UNIVERSITAT POLITÈCNICA DE CATALUNYA

Programa de Doctorat:

AUTOMÀTICA, ROBÒTICA I VISIÓ

Tesi Doctoral

# Understanding Human-Centric Images: From Geometry to Fashion

**Edgar Simo Serra**

Directors:
Francesc Moreno Noguer
Carme Torras

May 2015

## Abstract

Understanding humans from photographs has always been a fundamental goal of computer vision. Early works focused on simple tasks such as detecting the location of individuals by means of bounding boxes. As the field progressed, harder and more higher level tasks have been undertaken. For example, from human detection came the 2D and 3D human pose estimation in which the task consisted of identifying the location in the image or space of all different body parts, e.g., head, torso, knees, arms, etc. Human attributes also became a great source of interest as they allow recognizing individuals and other properties such as gender or age. Later, the attention turned to the recognition of the action being performed. This, in general, relies on the previous works on pose estimation and attribute classification. Currently, even higher level tasks are being conducted such as predicting the motivations of human behaviour or identifying the fashionability of an individual from a photograph.

In this thesis we have developed a hierarchy of tools that cover all these range of problems, from low level feature point descriptors to high level fashion-aware conditional random fields models, all with the objective of understanding humans from monocular RGB images. In order to build these high level models it is paramount to have a battery of robust and reliable low and mid level cues. Along these lines, we have proposed two low-level keypoint descriptors: one based on the theory of the heat diffusion on images, and the other that uses a convolutional neural network to learn discriminative image patch representations. We also introduce distinct low-level generative models for representing human pose: in particular we present a discrete model based on a directed acyclic graph and a continuous model that consists of poses clustered on a Riemannian manifold. As mid level cues we propose two 3D human pose estimation algorithms: one that estimates the 3D pose given a noisy 2D estimation, and an approach that simultaneously estimates both the 2D and 3D pose. Finally, we formulate higher level models built upon low and mid level cues for understanding humans from single images. Concretely, we focus on two different tasks in the context of fashion: semantic segmentation of clothing, and predicting the fashionability from images with metadata to ultimately provide fashion advice to the user.

In summary, to robustly extract knowledge from images with the presence of humans it is necessary to build high level models that integrate low and mid level cues. In general, using and understanding strong features is critical for obtaining reliable performance. The main contribution of this thesis is in proposing a variety of low, mid and high level algorithms for human-centric images that can be integrated into higher level models for comprehending humans from photographs, as well as tackling novel fashion-oriented problems.

# Acknowledgements

I would like to firstly thank both my advisors Francesc and Carme, to whom I owe this opportunity. I would also like to thank everyone who supported me during this long journey, and in particular, Rasputin and Nuryev.

This work has been partly sponsored by the following:

- A four-year PhD scholarship (BES-2012-054905).
- A four-month scholarship to visit the Toyota Technological Institute at Chicago (EEBB-I-14-08177).
- A four-month scholarship to visit the Nakamura Laboratory in Japan (EEBB-I-15-09283).
- Projects CINNOVA (201150E088), PAU+ (DPI2011-27510), ViSEN (ERA-Net Chistera PCIN-2013-047), and Intellact (FP7-ICT2009-6-269959).
- My own stubbornness.

*In loving memory of Misi*

# Contents

# List of Figures

# List of Tables

# Chapter 1

# Introduction

> I may not have gone where I intended to go, but I think I have ended up where I intended to be.
>
> *Douglas Adams*

Computer vision is a relatively new field with roughly half a century of tradition. It is a well known anecdote that it originally started out as a summer project for a first year undergraduate student in 1966, whose task was to “connect a television camera to a computer and get the machine to describe what it sees.” Had computer vision not been grossly underestimated and the project succeeded, this thesis would not have been possible and the author would be likely enjoying a long drink at the beach. However, seeing that this thesis is indeed finally completed, we come to the conclusion the project was not able to complete its ambitious task. Not only that, currently computer vision is a thriving field that is getting closer and closer to solving the 1966 summer problem.

In the beginning, computer vision was seen as more of a mathematical problem, which was additionally limited by the computational resources of the time. During that period, many of the tools and basic approaches we still use today were developed. Only in the last decade has the technology advanced sufficiently to tackle lofty computer vision problems and obtain reliable results that are making it into real world applications everywhere. One may argue that this started with the SIFT descriptor (Lowe, 2004), the most cited paper in computer vision[1], which eventually allowed the appearance of other notable works such as the deformable parts model (Felzenszwalb et al., 2008) for object detection. These models began to obtain significant results in identifying objects in natural photographs.

With the progress of time, datasets have increased in size and machine learning has played a larger and larger role in computer vision, to the point that it has now become an indispensable tool. One of the most important recent breakthroughs has been the use of Convolutional Neural Networks (CNN) for classification (Krizhevsky et al., 2012) which has outperformed existing approaches by a large margin. This has created an important resurgence of Artificial Neural Networks (ANN) in computer vision, as they are the basis for these techniques. In compliance with the current trends we rely heavily on machine learning throughout this thesis.

[1]Indeed, as it is common among computer vision researchers, we compulsively track the number of citations the SIFT paper has, which at the time of this writing is of 28,407 citations.

One particular area of computer vision is the understanding of human-centric images. This encompasses many different problems such as detection of individuals, 2D/3D pose estimation, attribute prediction, gaze prediction, clothing segmentation, etc. In particular, tasks such as face detection are already a reality and omnipresent for almost all types of digital cameras, while other problems such as assessing image memorability are still in their incipiency. In this thesis we focus on 3D human pose estimation, clothes segmentation, and predicting fashionability. For this purpose we have developed generative models for 3D human pose and feature point descriptors, which by themselves are unable to perform these tasks, but play a fundamental role in the algorithms that are able to do so.

Although our ultimate objective is higher level comprehension of human-centric images and problems such as the estimation of fashionability, this is unable to be performed without leveraging robust low level cues. Therefore it is critical to have a strong grasp of mid and low level algorithms. For this purpose we exploit a swath of existing features while supplementing them with our own. We show that our features are able to complement the existing ones while palliating their deficiencies.

## 1.1 Contributions

We organize the contributions into four groups: feature point descriptors, generative 3D human pose models, 3D human pose estimation, and probabilistically modelling fashion.

1. **Feature point descriptors.** We propose two different feature point descriptors: one based on heat diffusion and another based on Convolutional Neural Networks (CNN). In the first approach we model the image patch as a 3D surface. Afterwards we calculate the heat diffusion for logarithmically sampled time intervals and perform a Fast Fourier Transform on the resulting heat maps. We show that this approach is robust to both deformation and illumination changes. This descriptor, which we call *DaLI*, was published in (Simo-Serra et al., 2015b). The second work consists of using CNN to learn discriminative representations of image patches. This is done by using a Siamese CNN architecture, trained with pairs of corresponding patches. We propose learning with a sampling-based approach that in combination with heavy mining, which we denote "fracking", is able to significantly outperform hand-crafted features. This work is currently under submission (Simo-Serra et al., 2014c).
2. **Generative 3D human pose models.** We present two different generative models for parameterizing the 3D human pose: a Directed Acyclic Graph (DAG) and a mixture model for Riemannian manifolds. The DAG model uses discrete 3D poses obtained from clustering and learns the joint distribution of the resulting 3D poses and a latent space. By not having any loops, it is possible to efficiently map from the latent pose space to the 3D space and vice versa. This work was published as part of (Simo-Serra et al., 2012). For the second approach we model the 3D human pose by considering that it can be represented as data on a Riemannian manifold. This model is shown to preserve physical properties, such as distances between joints, much better than competing approaches. Additionally, we show that this scales well to large datasets and allows real-time sampling from the model. This work was published in (Simo-Serra et al., 2014b), with an extension for estimating velocities published in (Simo-Serra et al., 2015c).

3. **3D human pose estimation.** We propose two different algorithms for 3D human pose estimation from single monocular images. In the first work we assume we have a noisy estimation of the 2D human pose and use a linear formulation that allows us to project forward these estimations to the 3D space. We then use the distances between the 3D joints to estimate the anthropomorphism of the hypotheses and obtain a final solution. This work was published in (Simo-Serra et al., 2012). The second approach performs the 2D and 3D human pose estimation jointly in a single probabilistic framework. This is done by extending the pictorial structures Bayesian formulation (Felzenszwalb and Huttenlocher, 2005) to 3D, which naturally leads to a hybrid generative-discriminative model. To estimate the 3D pose we sample from a generative 3D pose model and then evaluate the hypotheses using a bank of discriminative 2D part detectors. We perform this alternate optimization until convergence. This work was published in (Simo-Serra et al., 2013).

4. **Probabilistically modelling fashion.** We finally tackle two different problems in the context of fashion: semantic segmentation of clothing, and predicting the fashionability of people in images. In the first problem we attempt to do fine-grained recognition of garments in an image. We propose a Conditional Random Fields (CRF) model that labels the different superpixels in the image. This is done by leveraging a set of strong unary potentials in combination with flexible pairwise potentials that are able to significantly outperform the state-of-the-art. This work was published in (Simo-Serra et al., 2014a). For the second problem of estimating fashionability, we have created a novel large dataset called *Fashion144k* generated by crawling posts from the largest online social fashion website. We aggregate votes or “likes” as a proxy for the fashionability of the posts and propose a large assortment of features that we compress using a deep Neural Network. Finally we interlace the features using a CRF model. The resulting model is then able to naturally learn different types of users, outfits, and settings, which are used to predict the fashionability of a post, and even give fashion advice. This work was published in (Simo-Serra et al., 2015a). Both these last two works are the fruition of collaborating with Prof. Raquel Urtasun and Prof. Sanja Fidler, from the University of Toronto.

Our focus is on high performance, and when necessary, fast models. We publish the code and the datasets when possible[2] in order to ensure our contributions are beneficial to the community. We hope this will encourage other researchers to use and compare against our approaches.

## Publications

The following is a list of the publications derived from this thesis:

[2] http://www.iri.upc.edu/people/esimo/

**Simo-Serra, E.**, Ramisa, A., Alenyà, G., Torras, C., and Moreno-Noguer, F. Single Image 3D Human Pose Estimation from Noisy Observations. In *IEEE Conference on Computer Vision and Pattern Recognition*, 2012.

**Simo-Serra, E.**, Quattoni, A., Torras, C., and Moreno-Noguer, F. A Joint Model for 2D and 3D Pose Estimation from a Single Image. In *IEEE Conference on Computer Vision and Pattern Recognition*, 2013.

**Simo-Serra, E.**, Torras, C., and Moreno-Noguer, F. Geodesic Finite Mixture Models. In *British Machine Vision Conference*, 2014.

**Simo-Serra, E.**, Fidler, S., Moreno-Noguer, F., and Urtasun, R. A High Performance CRF Model for Clothes Parsing. In *Asian Conference on Computer Vision*, 2014.

**Simo-Serra, E.**, Torras, C., and Moreno-Noguer, F. DaLI: Deformation and Light Invariant Descriptor. *International Journal of Computer Vision*, 1:1–1, 2015.

**Simo-Serra, E.**, Torras, C., and Moreno-Noguer, F. Lie Algebra-Based Kinematic Prior for 3D Human Pose Tracking. In *International Conference on Machine Vision and Applications*, 2015.

**Simo-Serra, E.**, Fidler, S., Moreno-Noguer, F., and Urtasun, R. Neuroaesthetics in Fashion: Modeling the Perception of Fashionability. In *IEEE Conference on Computer Vision and Pattern Recognition*, 2015.

Additionally a publication is still under submission:

**Simo-Serra, E.**, Trulls, E., Ferraz, L., Kokkinos, I., and Moreno-Noguer, F. Fracking Deep Convolutional Image Descriptors. arXiv preprint arXiv:1412.6537, 2014.

## 1.2 Thesis Overview

We group the work done conceptually into four major chapters as done for our contributions and then organize them from low to high level. Each of these chapters has an introductory section, several sections explaining different techniques, each corresponding roughly to a single publication, and a summary of the work. We additionally include a chapter to give an overview of our efforts as well as establishing some groundwork for some techniques we will use throughout the thesis. Finally the last chapter provides a general wrap-up and concludes the thesis.

**Chapter 2: Overview.** This introductory chapter gives a high level overview of the current state of the art of computer vision for human-centric images. We also introduce basic concepts of tools that we will use throughout the thesis.

**Chapter 3: Feature Point Descriptors.** This chapter focuses on our work on feature point descriptors. It gives a short introduction and then has two major sections that cover our 2015 IJCV and a paper under submission respectively.

**Chapter 4: Generative Models for 3D Human Pose.** This chapter focuses on the three different 3D human pose models that we use in the rest of the thesis, especially in Chapter 5. We first give an introduction followed by an in-depth review of models used in the past decade. The following two sections refer to the models used in our 2012 and 2013 CVPR papers. Afterward, we describe the approach of our 2014 BMVC paper in detail. We conclude with a summary of the different models.

**Chapter 5: 3D Human Pose Estimation.** This chapter focuses on our two approaches for estimating 3D human pose from single images. We first overview the approaches and elaborate on the related work in the field. The following two sections are devoted to our 2012 and 2013 CVPR papers, respectively. Lastly, we summarize the results.

**Chapter 6: Probabilistically Modelling Fashion.** In this chapter we look at the problem of modelling fashion. We start with the motivation of this problem and its impact in society before going into details of our 2014 ACCV and 2015 CVPR papers, respectively. The results are summarized at the end.

**Chapter 7: Conclusions.** The last chapter gives a high-level discussion of our contributions, how our work stands in the field, and current developments. We also pose open questions and sketch directions for future research.

# Chapter 2

# Overview

Scientific progress goes "boink"?

*Hobbes*

In this thesis we attempt to tackle very challenging high level computer vision problems. In order to do this, it is necessary to rely on robust low and mid level cues. In this chapter we will explain and give notions of the different cues and algorithms that can form part of these high level computer vision models. Of course the concepts of high and low are relative. In this chapter, and by extension the rest of this thesis, we shall consider as high level models those that perform tasks such as the ones we will present in Chapter 6, i.e., semantic segmentation of clothing and predicting fashionability from an image. Mid level cues will include pose estimation models such as those presented in Chapter 5 and foreground-background segmentation algorithms. Finally, low level cues will refer to features such as the prior distributions we describe in Chapter 4 or the feature point descriptors we will present in the next chapter.

We shall additionally discuss several machine learning models used throughout this thesis. In particular we will focus on logistic regression classifiers, Support Vector Machines (SVM), deep networks, and Conditional Random Fields (CRF). We shall formulate the different models, explain how they can be learnt, and briefly state some usage cases and applications of them.

This overview is not meant to be an exhaustive list of all the different cues and machine learning models commonly used in computer vision. Instead, it is meant to be a rough overview with a focus on the tasks considered in this thesis. A list of the different cues we will discuss and their relationship to this thesis can be seen in Table 2.1.

This chapter is divided into four sections that address low, mid, high level cues, and machine learning models, respectively.

## 2.1 Low Level Cues

We understand low level cues as simple features or priors that do not encode very prolific information. They are meant to be "austere" and fast to compute, and are always integrated into some larger model as they, by themselves, are unable to complete any task. However, this does not mean they are useless. For certain problems, just by combining and leveraging different weak low level cues it is possible to obtain significant performance. As an example we refer to the approach of (Krähenbühl and Koltun, 2011) that performs semantic segmentation with good results by only employing simple color features with some smoothing terms in a Conditional Random Fields framework.

| Level | Feature | Type | C3 | C4 | C5 | C6 |
|---|---|---|---|---|---|---|
| Low | Colour | Maps | ✓ | | | ✓ |
| | Gabor Filter | Maps | | | | ✓ |
| | Edge Detector | Maps | | | | |
| | Region Histogram | Regions | | | | ✓ |
| | Descriptors | Sparse, Maps | ✓ | | ✓ | ✓ |
| | PDF Prior | Likelihood Prior | | ✓ | ✓ | ✓ |
| Mid | Detectors | Maps, Bounding Boxes | | | ✓ | ✓ |
| | Foreground Segmentation | Maps | | | | ✓ |
| | Saliency | Maps | | | | |
| | Order 2 Pooling | Regions | | | | ✓ |
| | GIST | Image | | | | |
| | CNN | Regions, Image | | | | ✓ |
| | Pose | Spatial Prior | | | ✓ | ✓ |
| | Attributes | Generic | | | | ✓ |
| | Bag-of-Words | Regions, Image | | | | ✓ |
| High | Semantic Segmentation | Maps | | | | ✓ |

**Table 2.1: Different types of computer vision cues.** Overview of some commonly used cues in computer vision and their relationship with this thesis. We categorize them into low, mid and high level cues and based on their types. “Maps” generate 2D or 3D matrices where a value or vector corresponds to each pixel, respectively. “Regions” correspond to larger areas such as superpixels. “Image” type cues provide a single vector that corresponds to the entire image. In the right four columns we indicate to which chapter the cue relates to.

### Image Features

There are a variety of potential low level image features. The most simple one is to directly use the RGB color values. This is most commonly used in Convolutional Neural Networks (Krizhevsky et al., 2012) after normalizing the values, that is, subtracting the mean and dividing by the standard deviation of the pixel values. It is also possible to convert the RGB values to other color spaces such as HSV, YUV or CIE L*a*b. In general, it is not clear which color space is the best for a specific application and thus it requires experimental validation.

Since individual pixels do not contain much information, it is standard to instead group them into different regions which conceptually correspond to parts of the same object in the image (Achanta et al., 2012). This grouping is known as pooling and has two major effects: it first allows a much more efficient representation of the image as it is now reduced to a set of non-overlapping regions, and secondly, each region contains more information and thus becomes increasingly discriminative. One of the most popular algorithms is the gPb superpixel approach (P.Arbelaez et al., 2011). When working with these larger image regions that no longer consist of a single pixel but rather a group of pixels, it is common to compute histograms of their color values. This gives a compact representation of the color of the image patch. Furthermore, these histograms are normalized so that the total sum of all their elements is 1, allowing the different image regions to be directly compared regardless of the number of pixels they have.

Instead of trying to encode color information, it is also standard to try to directly encode textural information. One of the most common representations is using 2D Gabor filters, which are Gaussian kernel functions modulated by a sinusoidal plane waves and can be written as such:

$$
\begin{aligned}
g(x, y; f, \theta, \phi, \gamma, ) &= \frac{f^2}{\pi\gamma\eta} \exp\left(-\frac{x'^2 + \gamma^2 y'^2}{2\sigma^2}\right) \exp\left(i2\pi f x' + \phi\right) \\
x' &= x\cos\theta + y\sin\theta \\
y' &= -x\sin\theta + y\cos\theta \quad ,
\end{aligned}
\tag{2.1}
$$

where $f$ is the frequency of the sinusoidal factor, $\theta$ represents the orientation, $\phi$ is the phase offset, $\sigma$ is the standard deviation of the Gaussian envelope, and $\gamma$ is the spatial aspect ratio. By computing the responses of a bank of Gabor filters with different orientations and frequencies it is possible to get a local representation of the image texture. It is also standard in this case to calculate the normalized histogram of the responses as a texture representation of an image region or superpixel as described before.

Another approach is to detect edges in the image. This is commonly done by first blurring the image to reduce the image noise, before proceeding to convolve the image with simple Sobel filters. This filters are designed to detect lines of different orientation, e.g., horizontal and vertical lines. The output map resulting from convolving the image with the Sobel filters can then be used as an indication of where edges are in the image. This information is generally useful when attempting to distinguish contours of objects.

A slightly more advanced approach consists in using feature point descriptors. These are generally compact vector representations of small image patches. The most well known approach is the SIFT descriptor (Lowe, 2004). This descriptor first computes the gradient of a grayscale image, and then uses a rectangular grid in which all the sections build a histogram of the gradient values. These values are then weighted by a Gaussian centered on the patch. All the histograms are finally concatenated into a single vector which is normalized such that the sum of all the elements is one. As SIFT was originally designed to be used sparsely, there have been several alternatives proposed to be calculated densely on the image. The most popular one is the Histogram of Oriented Gradients (HOG) descriptor (Dalal and Triggs, 2005) shown in Fig. 2.1(e), while a more efficient modern variant is the DAISY descriptor (Tola et al., 2010), which uses overlapping circles instead of a rectangular grid.

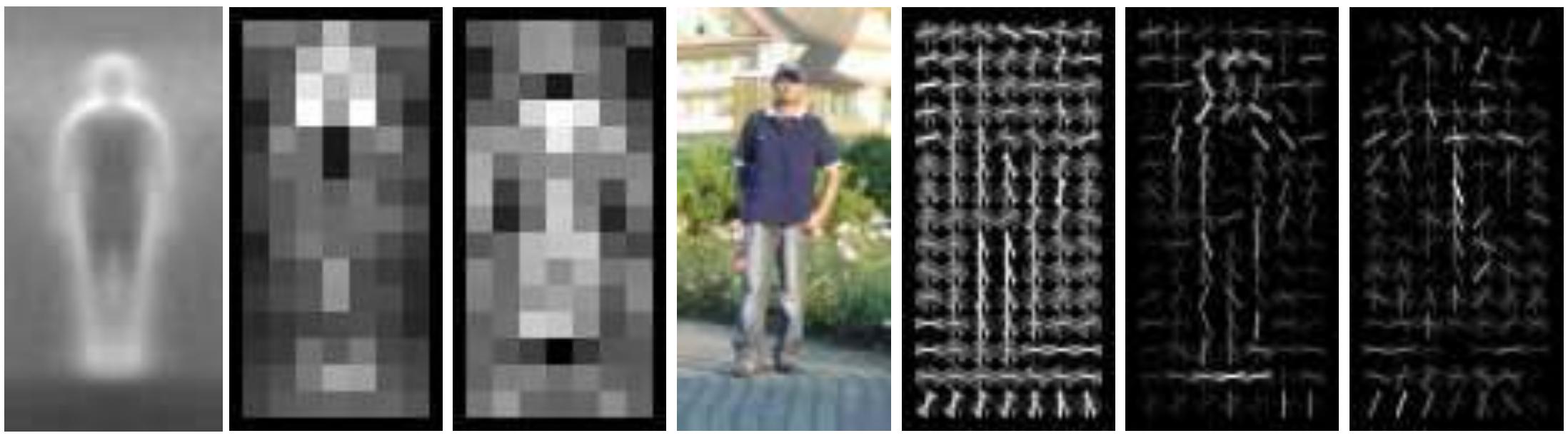

(a) (b) (c) (d) (e) (f) (g)

**Figure 2.1: Human detection with HOG descriptors.** The HOG detectors cue mainly on silhouette contours. *(a)* The average gradient image over the training data. *(b)* Each "pixel" shows the maximum positive SVM weight in the block centered on the pixel. *(c)* Likewise for the negative SVM weights. *(d)* A test image. *(e)* Its computed HOG descriptors. *(f,g)* The HOG descriptors weighted by the positive and negative SVM weights, respectively. Figure reproduced from (Dalal and Triggs, 2005).

### Simple Priors

It is also standard to use simple priors in addition to image features in computer vision. A very common prior is to use the location of a pixel within an image. This is based on the knowledge that for images taken by humans, in general, the subject is well focused and centered on the image. By using the position within the image as a feature it is possible to capture this relationship. For image patches either a histogram is calculated or the mean position for all the pixels in the patch is used.

When dealing with any sort of variable, there are usually some values that are more likely than others. An example would be the orientation of humans in images. In general they will be standing upright; it is less likely that they will be horizontal. In order to capture the prior distribution it is common to use mixture models to estimate the Probability Density Function (PDF) of the variable. This will help a model take into account which states or values are more likely.

## 2.2 Mid Level Cues

The cues we will discuss in this section are more elaborated than the low level ones, being generally the result of some more complex algorithm. Yet, they are able to capture concepts which are more useful when performing higher level tasks. The downside is that they are slower to compute.

### Image Features

Object detectors are often used as image feature. The most simple detectors consist in using a template of the object to be detected and convolving it with the image. In general, this is not done directly on the raw pixel values but on extracted descriptors, and the response correlates with the presence of the object. An example is the case of human detection (Dalal and Triggs, 2005) which was performed by convolving a template of Histogram of Oriented Gradients (HOG) descriptors at different scales. A representation of the approach can be seen in Fig. 2.1. As a feature it can be either used the output of the algorithm, in this case a bounding box of the object, or the associated likelihood map, which gives a spatial prior of where the objects is most likely to be in the image.

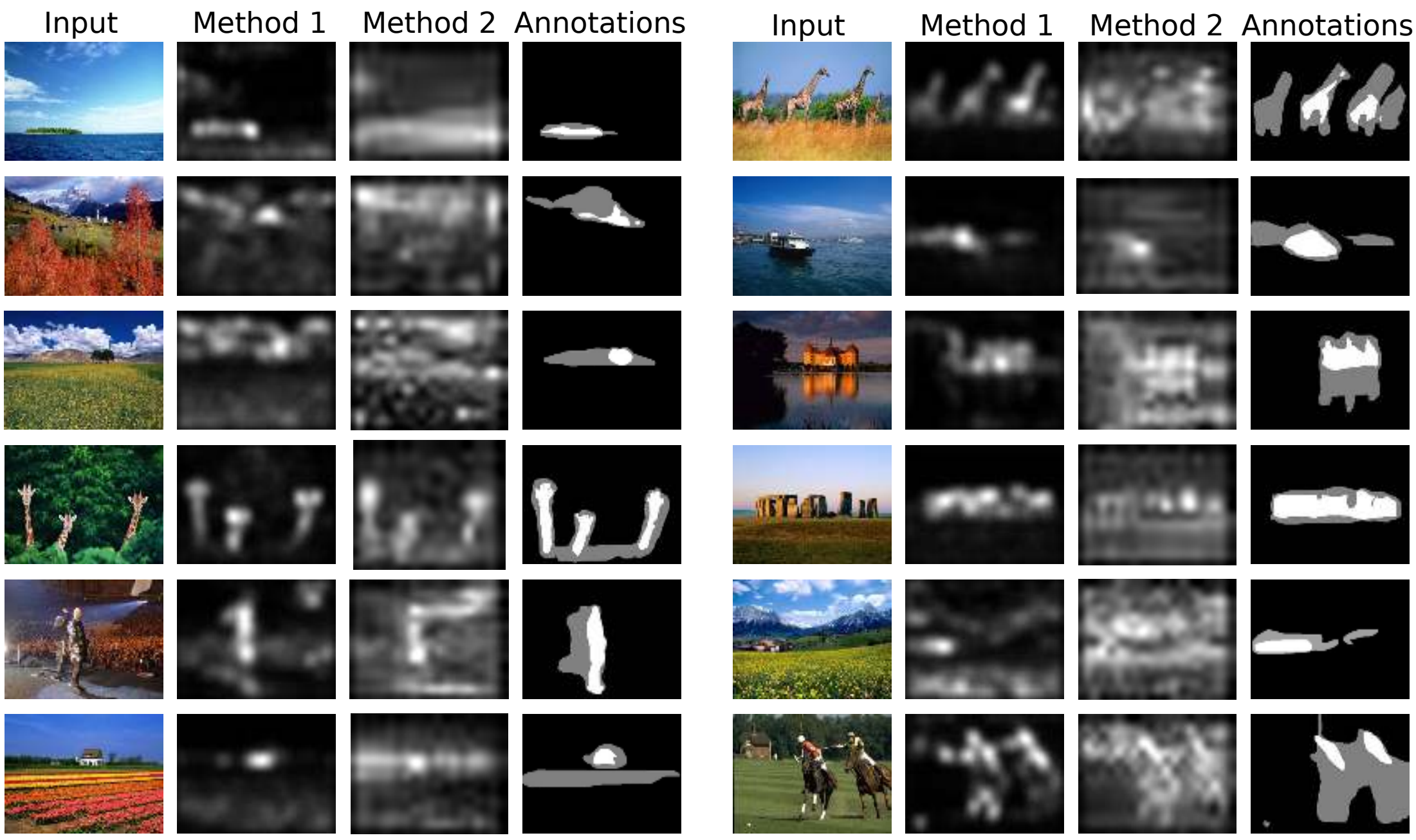

**Figure 2.2: Examples of saliency detection.** For input images we show the result of applying Method 1 (Hou and Zhang, 2007) and Method 2 (Itti and Koch, 2000). In order to obtain a ground truth of the saliency, all the images were labelled by four annotators. The black area of the annotations corresponds to non-salient regions, the grey area was a region identified as salient by at least one of four annotators, and the white area is the region selected by all four annotators. Figure reproduced from (Hou and Zhang, 2007).

Another widely used approach is to run a segmentation algorithm to first attempt to distinguish between what is likely to be the central object and what is likely to be the background in the image. One possible algorithm that can be used with this purpose is a Conditional Random Fields (CRF) model in which each pixel can have a state of being either foreground or background. Then, by using Gaussian prior on the color values for foreground and background, in addition with simple potentials between neighbouring pixels, it is possible to obtain fast segmentation results (Krähenbühl and Koltun, 2011). It is also possible to use more accurate and robust algorithms, although at a higher computational cost (Carreira and Sminchisescu, 2012).

An interesting feature proposed recently consists of trying to identify salient regions of the image. This is inspired by how the human visual cortex is able to quickly discern interesting objects to focus attention on. Saliency maps are in particular useful to descry the predominant regions of an image in order to focus on them (Todt and Torras, 2004). One of the more standardized approaches consists of extracting the spectral residue of an image and then constructing the saliency map in the spatial domain (Hou and Zhang, 2007). Several saliency maps are depicted in Fig. 2.2.

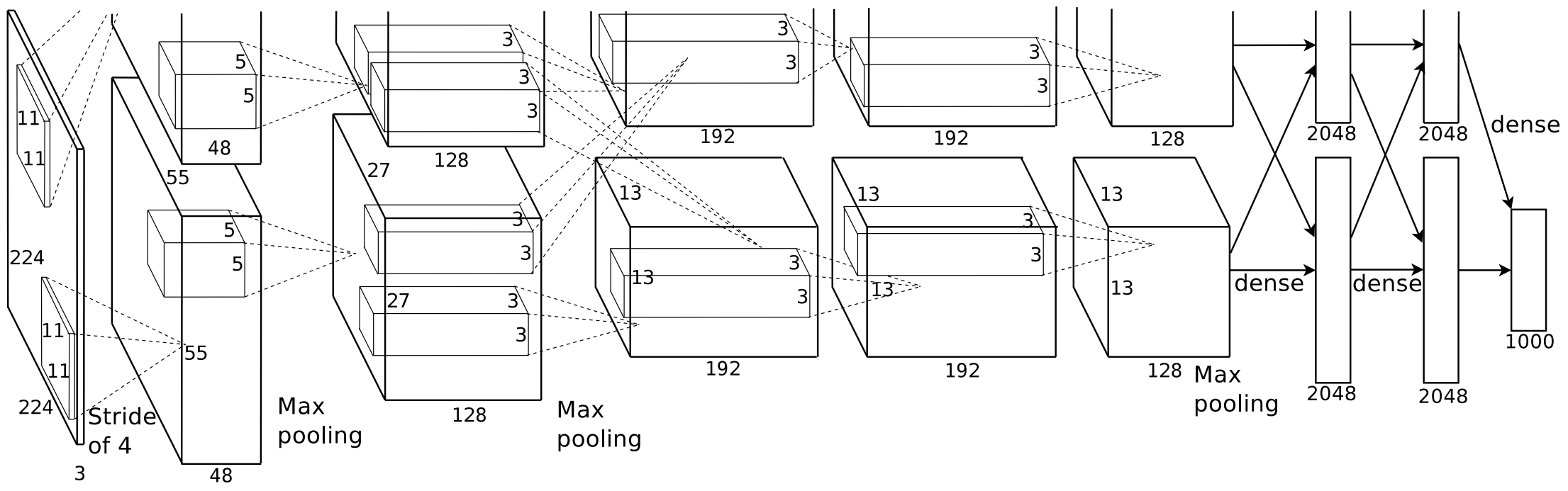

**Figure 2.3: Architecture of a convolutional neural network.** This network takes 224x224 RGB images as input, and outputs a 1000-dimensional vector that corresponds to the probability of the image belonging to one of 1000 different classes. It has 5 convolutional layers and 3 densely connected layers. This network was originally designed for the ILSVRC2012 challenge for a 1000 class classification challenge, where it obtained the first place. Figure reproduced from (Krizhevsky et al., 2012).

Instead of computing simple histograms of basic features, it is also possible to use higher order pooling. For example, (Carreira et al., 2012) performs second order pooling of SIFT descriptors with impressive results. In this approach the outer product of all descriptors extracted from an image region is computed. Then, either the max or the average of the resulting vector is used as a cue. More specifically, given a set of $n$ descriptors $\mathbf{x}_1, \ldots, \mathbf{x}_n$ of a region, the pooling is simply

$$G_{avg}(\mathbf{x}_1, \ldots, \mathbf{x}_n) = \frac{1}{n} \sum_i \mathbf{x}_i^{\mathrm{T}} \cdot \mathbf{x}_i \ , \tag{2.2}$$

for the average pooling case, and

$$G_{max}(\mathbf{x}_1, \ldots, \mathbf{x}_n) = \max_i \mathbf{x}_i^{\mathrm{T}} \cdot \mathbf{x}_i \ , \tag{2.3}$$

for the max pooling case. This approach yields stronger statistics of the region, allowing simple classifiers using these features to be much more discriminative than instead relying on more complex classifiers with simple features.

For full images, calculating a global descriptor which represents the entire image instead of a local area is also a possibility. One of the most popular global descriptors of this type is the GIST descriptor (Oliva and Torralba, 2001). This descriptor uses spectral and coarsely localized information in order to approximate a set of perceptual dimensions of the image such as naturalness, openness, etc. This results in a 544-dimension descriptor that is what is considered a "hand-crafted" feature, i.e., it is manually designed instead of beingautomatically learnt by a computer algorithm. Recently, learning deep Convolutional Neural Networks (CNN) to extract features has been used to obtain similar features with great results (Girshick et al., 2014). The networks to extract these features are usually trained on very large datasets for the task of image classification. After being trained, one or more of the final layers of the network are removed and the output of the new last layer is used as a mid level representation of the image. One of the most popular networks to extract these features is the "AlexNet" from (Krizhevsky et al., 2012), shown in Fig. 2.3. In this particular case the second-to-last 4096-dimensional layer is used as a global representation of the image. These features have been found to generalize very well to tasks other than the one originally trained for (Donahue et al., 2013). See (Rubio et al., 2015) for an example of using these features in a robotics application.

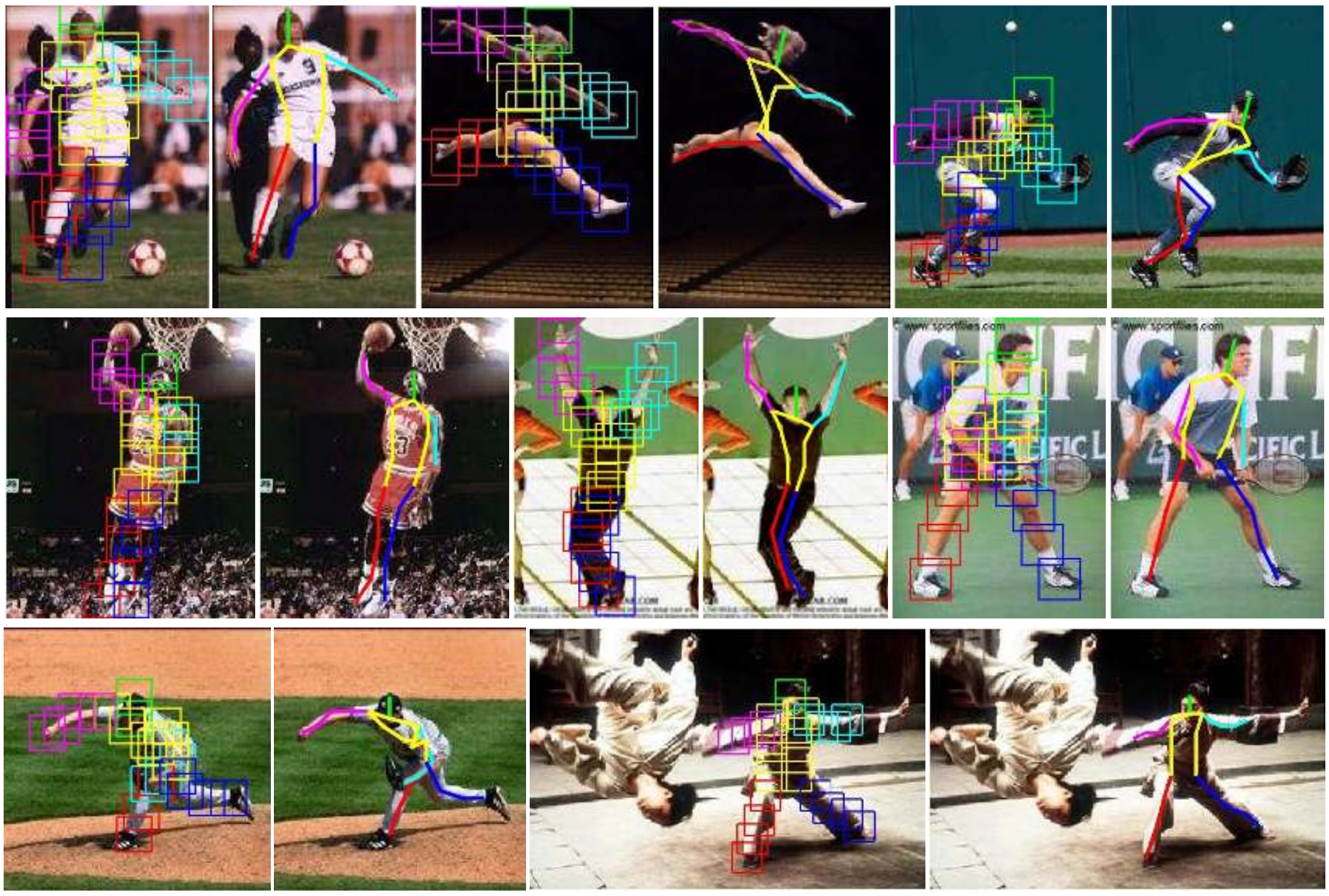

**Figure 2.4: 2D human pose estimation.** Some examples of 2D human pose estimations on different images. For each image pair, the bounding boxes of all individual parts are shown on the left, and the resulting joint skeleton is shown on the right. Figure reproduced from (Yang and Ramanan, 2011).

### Prior Models

When dealing with human-centric image tasks, having an estimation of the pose can be a very useful feature. In general a fast state-of-the-art algorithm such as (Yang and Ramanan, 2011) is used to estimate the 2D joints of all the individuals in an image. This model relies on a mixture of HOG templates for detecting different body parts. Then, it models co-occurrence and spatial relationships with a tree structure that allows efficient inference with dynamic programming. Once the pose is estimated, the location of the joints within the image can be used as a spatial prior. Furthermore, the relative pose of the human can also be used as a feature when performing tasks such as action recognition. Some examples of 2D human pose estimation can be seen in Fig. 2.4.

Another useful source of information consists in computing attributes representing some mid level information (Farhadi et al., 2009). For example, in the case of face recognition, this kind of information would consist in whether or not a person wears glasses and her/his gender or ethnicity. In general, these attributes are obtained by training discriminative models on annotated datasets.

### Bag-of-Words

A widely used approach to obtain features for regions or entire images is the bag-of-words model. This model was originally designed for extracting features from text, in which the bag-of-words is was just a dictionary of words. For a given input text, the number of occurrences that appear in the text of each word in the dictionary is counted. This is a simple way of obtaining a sparse representation that can then be used as features when training classifiers or other algorithms.

This approach is not limited to discrete features such as words in text, but can also be used for continuous features such as Gabor filter histograms or feature point descriptors. In this case, instead of using a dictionary of words, a codebook made of a list of representative features is constructed. A query feature is then made discrete by assigning it to the most similar entry in the codebook. This allows for a more global representation to be learned for either regions or whole images. The downside of this representation is that it does not take into account the spatial layout of the input data.

## 2.3 High Level Cues

Usually, high level cues represent the ultimate goal to be achieved. However, they can also be used as features in certain applications. We will briefly mention the possibility of using semantic segmentation algorithms as features for other models. Semantic segmentation differs from foreground segmentation in the sense there are more than two classes (foreground and background) and thus it is no longer possible to use efficient algorithms such as graphcuts (Boykov et al., 2001) to solve the problem. The output of a semantic segmentation algorithm is a label for all the pixels in the image and is highly related to the detection task, that is, regions in the image that belong the same class can be considered a detection of that particular object. It is not uncommon for these algorithms to use detectors as features.

## 2.4 Machine Learning Models

There is no doubt that machine learning plays a fundamental role in modern computer vision. From classifying objects in images to reasoning about the scene, it allows exploiting large sets of data to generate models that are capable of making predictions or taking decisions. There are several approaches to machine learning. In this section we shall consider the supervised learning problem which consists of designing models that are able to predict labels from input features. These models have a number of parameters which instead of being set manually, are algorithmically chosen such that they minimize the divergence between the model predictions and the known labels for annotated data.

In this section we present four different commonly used supervised models: logistic regression, support vector machines, deep networks, and conditional random fields. For each of the models we formulate the prediction rules and the learning function that is optimized when learning the parameters of the model. We will also give some notions on how they can be used and learned in practice, along with several examples.

### Logistic Regression

The logistic regression is a linear model that despite its name is used for classification and not regression. In particular, it outputs a single $[0, 1]$ value that is interpretable as a probability. The model parameters are a vector of weights $\mathbf{w}^{\mathrm{T}} = [w_1, \ldots, w_n]$ of the same length as the input $\mathbf{x}^{\mathrm{T}} = [x_1, \ldots, x_n]$, which may or may not include a bias term. The probability of $\mathbf{x}$ belonging to class $y = +1$ is written as:

$$f_{lr}(\mathbf{x}) = \frac{1}{1 + e^{-\mathbf{w}^{\mathrm{T}}\mathbf{x}}} \ . \tag{2.4}$$

The model parameters are optimized by minimizing the negative log-likelihood of the predictions with an additional regularization term:

$$\min_{\mathbf{w}} \frac{1}{2}\mathbf{w}^{\mathrm{T}}\mathbf{w} + C \sum_{(\mathbf{x},y)} \log\left(1 + e^{-y\mathbf{w}^{\mathrm{T}}\mathbf{x}}\right) \ , \tag{2.5}$$

where $C > 0$ is the regularization parameter, $y = \{-1, +1\}$ is the label of a particular training sample, and $\mathbf{x}$ are the features of the same sample. This optimization can be solved by using gradient-based methods. In particular we use the trust region Newton method implementation of LIBLINEAR (Fan et al., 2008).

We note that the standard formulation is for the two-class case. In order to generalize to the $m$-class case it is common to use a one-vs-all approach in which for each of the possible $m$ classes, a logistic regression is trained to predict only that particular class, using all the other classes as negatives. By concatenating all the individual logistic regression outputs a $m$-dimensional vector is obtained. The class with the largest value will be the prediction for the sample.

### Support Vector Machines

In contrast to logistic regression, Support Vector Machines (SVM) are non-probabilistic models. While they can be linear, usually non-linear variants are used. The decision function, or predicted class of $\mathbf{x}$ is:

$$f_{svm}(\mathbf{x}) = \operatorname{sgn}\left(\mathbf{w}^{\mathrm{T}}\phi(\mathbf{x}) + b\right) \ , \tag{2.6}$$

where $\phi(\mathbf{x})$ maps $\mathbf{x}$ into a high-dimension space, and $b$ is a bias, which is made explicit in this case.

The model parameters $\mathbf{w}$ and $b$ are found by maximizing the margin or the cleanest possible split between the training examples of different classes in the dataset. This is done by introducing slack variables $\xi_i$, which measure the degree of misclassification of the data sample $\mathbf{x}_i$. We consider two possible labels for each sample $y_i = \{-1, +1\}$. Thus the parameters are optimized by:

$$\begin{aligned} \min_{\mathbf{w},b,\boldsymbol{\xi}} \quad & \frac{1}{2}\mathbf{w}^{\mathrm{T}}\mathbf{w} + C\sum_{i=1}^{N} \xi_i \\ \text{subject to} \quad & y_i\left(\mathbf{w}^{\mathrm{T}}\phi(\mathbf{x}_i) + b\right) \geq 1 - \xi_i \ , \\ & \xi_i \geq 0, \quad i = 1, \ldots, N \end{aligned} \tag{2.7}$$

where $C > 0$ is the regularization parameter, and $N$ is the number of training samples. Due to the high dimensionality of $\mathbf{w}$ usually the dual problem is optimized:

$$\begin{aligned} \min_{\boldsymbol{\alpha}} \quad & \frac{1}{2}\boldsymbol{\alpha}^{\mathrm{T}}\mathbf{Q}\boldsymbol{\alpha} - \mathbf{e}^{\mathrm{T}}\boldsymbol{\alpha} \\ \text{subject to} \quad & \mathbf{y}^{\mathrm{T}}\boldsymbol{\alpha} = 0 \\ & 0 \leq \alpha_i \leq C, \quad i = 1, \ldots, N \end{aligned} \tag{2.8}$$

where $\mathbf{e} = [1, \ldots, 1]^{\mathrm{T}}$ is a vector of all ones, $\mathbf{Q}$ is an $N \times N$ positive semidefinite matrix, $Q_{ij} = y_i y_j K(\mathbf{x}_i, \mathbf{x}_j)$, and $K(\mathbf{x}_i, \mathbf{x}_j) = \phi(\mathbf{x}_i)^{\mathrm{T}}\phi(\mathbf{x}_j)$ is the kernel function.

After the optimization process, the optimal $\mathbf{w}$ satisfies:

$$\mathbf{w} = \sum_{i=1}^{N} y_i \alpha_i \phi(\mathbf{x}_i) \ , \tag{2.9}$$

which when combined with Eq. (2.6) yields an equivalent decision function for an input $\mathbf{x}$ based on the dual:

$$f_{svm}(\mathbf{x}) = \operatorname{sgn}\left(\sum_{i=1}^{N} y_i \alpha_i K(\mathbf{x}_i, \mathbf{x}) + b\right) \ . \tag{2.10}$$

We notice that in this case we have a new hyperparameter which is the choice of kernel function $K(\mathbf{x}_i, \mathbf{x})$. While there are many different options available, we consider the widely used Radial Basis Function (RBF) kernel defined by:

$$K(\mathbf{x}_i, \mathbf{x}) = \exp\left(-\gamma|\mathbf{x}_i - \mathbf{x}|^2\right) \ , \tag{2.11}$$

where $\gamma$ is the smoothness parameter of the kernel.

The optimization of the parameters of a SVM is a convex quadratic programming problem in both the primal (Eq. (2.7)) and the dual (Eq. (2.8)). We use the heuristic-based approach of LIBSVM (Chang and Lin, 2011) to optimize the dual.

### Deep Networks

Artificial Neural Networks (ANN) have a long tradition in computer science. However, only recently has their usage exploded in the field of computer vision. In particular the networks used, broadly coined "deep networks", have two important properties: they use convolutional layers to lower the number of parameters, and they have many layers, hence the name "deep". This recent widespread usage has been instigated by the very good performance obtained in specific computer vision tasks. This is due to many small improvements in combination with a significant increase in the available computational power. Up until now the training of networks with over 50 million parameters has been infeasible in a reasonable amount of time. Despite many minor improvements necessary for the improved performance of these networks, the underlying mathematics and formulation remains unchanged from several decades back.

An ANN is a directed acyclic graphical model, in which the nodes are called "neurons". The standard feed-forward network we shall consider is a network consisting of various layers. Each layer is only connected to both the previous layer and the next layer. We can then write the output of a layer $l$ as:

$$\mathbf{x}^l = \sigma\left((\mathbf{w}^l)^{\mathrm{T}}\mathbf{x}^{l-1} + b^l\right) \ , \tag{2.12}$$

where $\mathbf{w}^l$ is the weight vector of the layer, $b^l$ is the bias term, $\mathbf{x}^{l-1}$ is the output of the neurons in the previous layer, and $\sigma(\cdot)$ is a non-linear activation function. Commonly the hyperbolic tangent or rectified linear unit (ReLU) activation function is used. For more than one layer the chain rule is applied, i.e., the output of a layer is used as the input for the next layer.

In general these networks have a large number of parameters or weights $\mathbf{w}$. In order to learn these weights a technique called back-propagation is used. We assume we have a loss function $\Delta(\mathbf{y}, \mathbf{y}^*)$ where $\mathbf{y}$ is the prediction of the network and $\mathbf{y}^*$ is the ground truth or true label we want to predict. Back-propagation consists of computing $\frac{\partial \Delta}{\partial \mathbf{w}}$. This is done by first computing the error of the last layer $L$ as:

$$\delta^L = \frac{\partial \Delta}{\partial \mathbf{y}} \odot \sigma' \left( (\mathbf{w}^L)^{\mathrm{T}} \mathbf{x}^{L-1} + b^L \right) \ , \tag{2.13}$$

where $\mathbf{y} = \mathbf{x}^L$ is the output of the last layer, $\odot$ is the Hadamard product or element-wise product of two vectors, and $\sigma'$ in the derivative of the activation function. The errors of the other layers can then be written as a function of the next layer as:

$$\delta^l = \left( \left( \mathbf{w}^{l+1} \right)^{\mathrm{T}} \delta^{l+1} \right) \odot \sigma' \left( (\mathbf{w}^l)^{\mathrm{T}} \mathbf{x}^{l-1} + b^l \right) \tag{2.14}$$

Finally the derivatives for the weight $k$ of the neuron $j$ in the layer $l$ can be computed as:

$$\frac{\partial \Delta}{\partial \mathbf{w}^l_{jk}} = \mathbf{x}^{l-1}_k \delta^l_j \ , \tag{2.15}$$

and the bias for the layer $l$ becomes:

$$\frac{\partial \Delta}{\partial b^l} = \delta^l \ . \tag{2.16}$$

When learning the network, the features are propagated through the network for each sample using Eq. (2.12) in what is called a forward pass. Afterwards the loss $\Delta(\mathbf{y}, \mathbf{y}^*)$ is computed for that given sample's true label $\mathbf{y}^*$ and the output of the network $\mathbf{y}$, and is used to then perform a backwards pass. This consists of propagating the error backwards through the network by first using Eq. (2.13) for the top layer and then Eq. (2.14) for the remaining layers. Finally the partial derivatives of all the weights with respect to the loss are computed and used to update the weights and bias terms:

$$\mathbf{w}^{i+1} = \mathbf{w}^i - \lambda \frac{\partial \Delta}{\partial \mathbf{w}^i} \tag{2.17}$$

where $\lambda$ is the learning rate hyperparameter which controls the rate at which the weights are changed. This is done iteratively until some convergence criterion is met.

The usual approach to optimize the network is to use stochastic gradient descent, which is a variant of gradient descent in which only a subset of samples are used at each iteration to provide an estimate of $\frac{\partial \Delta}{\partial \mathbf{w}^i}$. This approach, in general, converges faster than standard gradient descent, which is fundamental for networks that can have over 50 million parameters.

In computer vision, instead of the fully connected network previously described, it is common to use what are known as Convolutional Neural Networks (CNN). The main difference here is that the output of a layer is obtained by convolving a filter, that is, the weights $\mathbf{w}$ are not independent for each output neuron but shared for all the neurons in the layer. In this way 2D spatial information is conserved. In general instead of having a single 2D output feature map, more than one filter is used. While the number of parameters decreases when using convolutional layers, the number of calculations necessary increases. For this purpose it is common to also use subsampling layers in which a convolution operator is used to decrease the size of the feature map. An example of a convolutional neural network can be seen in Fig. 2.5.

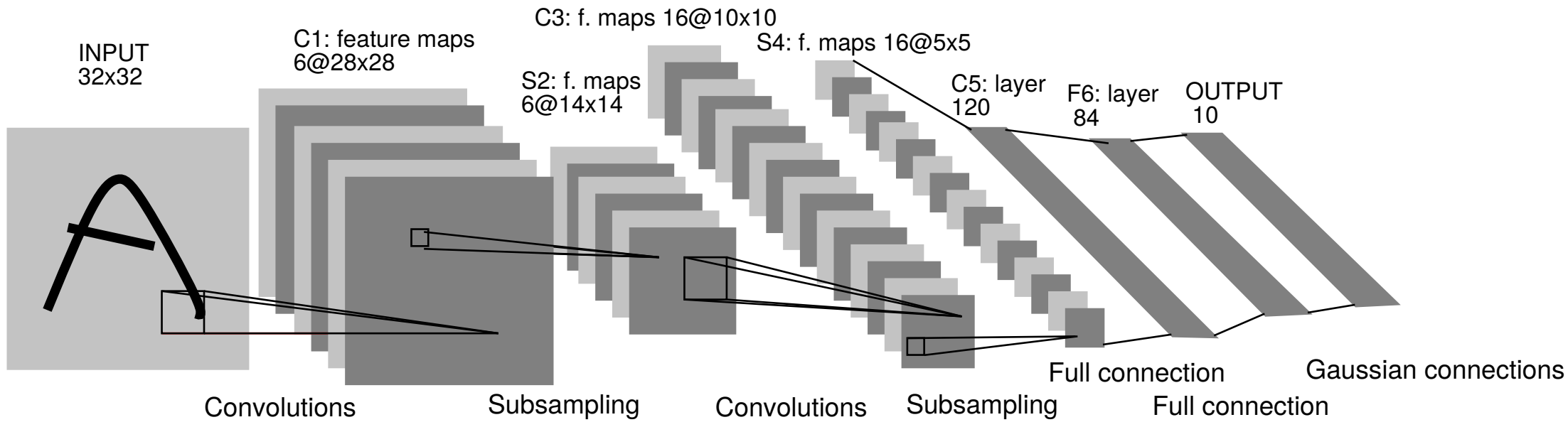

**Figure 2.5: Convolutional Neural Network (CNN) architecture.** The convolutional layers consist of several filters that are convolved with the input. By performing convolutions, fewer parameters are needed as they are shared when generating the output feature maps. In order to improve efficiency, subsampling layers are used in which another convolution operator is applied in order to reduce the size of the feature maps. After the convolutional layers, fully connected layers are used in which spatial information is lost. Figure reproduced from (Lecun et al., 1998).

In particular the CNN models used in computer vision tend to have many layers, and parameters in the order of tens of millions. In order to learn these models a large amount of data is necessary. To avoid overfitting the model to the training data many different techniques are used. The most common method consists of data augmentation. In this case a set of synthetic deformations are applied to the image, e.g., cropping, rotating, flipping horizontally, etc., in order to increase the number of training samples. This has been shown to help the network to generalize better.

As the specifics of each deep network are highly dependent on the task, we shall defer the explanation of details to the chapters in which they are used.

### Conditional Random Fields

When it comes to probabilistic models, one of the most important ones in computer vision is the Conditional Random Fields (CRF) model (Lafferty et al., 2001). These are a class of models used for structured prediction, that is, modelling output data that has a specific structure which takes the context into account. An example would be semantic segmentation of clothing in which although the pixels are being labelled, if there are pixels that belong to say the "boots" class, then there should be no pixels belonging to for example "heels", "sneakers" or "pumps" classes. In particular they are discriminative undirected probabilistic graphical models which encode known relationships between observations and construct consistent interpretations.

As indicated by its name, a CRF is modelling the conditional distribution $p(\mathbf{y}|\mathbf{x})$ of a random variable over the corresponding sequence labels $\mathbf{y}$, globally conditioned on a random variable over sequences to be labeled or observations $\mathbf{x}$. Note that $p(\mathbf{x})$ is not explicitly modelled as it is observed. Additionally the components of $\mathbf{y}$ are considered to come from a finite set of discrete states, however, $\mathbf{x}$ can come from a continuous distribution.

**Definition 2.4.1** *Let $G$ be a factor graph over $\mathbf{y}$. Then $p(\mathbf{y}|\mathbf{x})$ is a Conditional Random Field (CRF) if for any fixed $\mathbf{x}$, the distribution $p(\mathbf{y}|\mathbf{x})$ factorizes according to $G$.*

If $F = \{\psi_A\}$ is the set of factors in $G$, and each factor takes the exponential family form, then the conditional distribution can be written as,

$$p(\mathbf{y}|\mathbf{x}) = \frac{1}{Z(\mathbf{x})} \prod_{\psi_A \in G} \exp\left(\mathbf{w}_A^{\mathrm{T}} \mathbf{f}_A(\mathbf{x}_A, \mathbf{y}_A)\right) , \tag{2.18}$$

where $Z(x) = \sum_{\mathbf{y}} \prod_{\psi_A \in G} \exp\left(\mathbf{w}_A^{\mathrm{T}} \mathbf{f}_A(\mathbf{x}_A, \mathbf{y}_A)\right)$ is the partition function which ensures this is a probability, and $\mathbf{w}_A$ and $\mathbf{f}_A(\mathbf{x}_A, \mathbf{y}_A)$ are the weights and feature functions for the factor $\psi_A$, respectively. Note that weights can be shared among different factors; it is not uncommon for templates to be used for many different cliques in the graph. Furthermore, it is also typical to write the factors as potential functions, i.e., $\phi(\mathbf{y}) = \mathbf{w}^{\mathrm{T}}\mathbf{f}(\mathbf{x}, \mathbf{y})$, where $\mathbf{x}$ is dropped for notation simplicity.

In order to perform inference we can compute the Maximum A Posterior (MAP) estimate which consists of:

$$\mathbf{y}^* = \arg\max_{\mathbf{y}} \, p(\mathbf{y}|\mathbf{x}) = \arg\max_{\mathbf{y}} \sum_{\psi_A \in G} \mathbf{w}_A^T \mathbf{f}_A(\mathbf{x}_A, \mathbf{y}_A) \, . \tag{2.19}$$

Exact inference therefore consists of evaluating all the possible assignments of $\mathbf{y}$. For particular cases such as tree structures it is possible to efficiently compute the exact marginals. However, the general case is an NP-hard problem making it infeasible to evaluate all the states. Instead, it is usually solved by using approximation algorithms. We shall use a message passing algorithm called distributed convex belief propagation (Schwing et al., 2011) to perform inference. It belongs to the set of LP-relaxation approaches and has convergence guarantees, unlike other algorithms such as loopy belief propagation.

In order to perform learning we first formulate the conditional log-likelihood:

$$\sum_{\psi_A \in G} \mathbf{w}_A^T \mathbf{f}_A(\mathbf{x}_A, \mathbf{y}_A) - \log Z(\mathbf{x}) \, . \tag{2.20}$$

The learning problem can then be posed as minimizing the negative log-likelihood with an additional regularization term

$$\min_{\mathbf{w}} \; C\mathbf{w}^T\mathbf{w} + \sum_{(\mathbf{x},\mathbf{y})} \left( \log Z(\mathbf{x}) - \sum_{\psi_A \in G} \mathbf{w}_A^T \mathbf{f}_A(\mathbf{x}_A, \mathbf{y}_A) \right) , \tag{2.21}$$

where $C$ is once again the regularization parameter.

To learn the weights we will consider the primal-dual method of (Hazan and Urtasun, 2010), which is a structured prediction framework. It is based on message passing and has the characteristic that it has guaranteed convergence. It has been shown to be more efficient than other structured prediction learning algorithms.

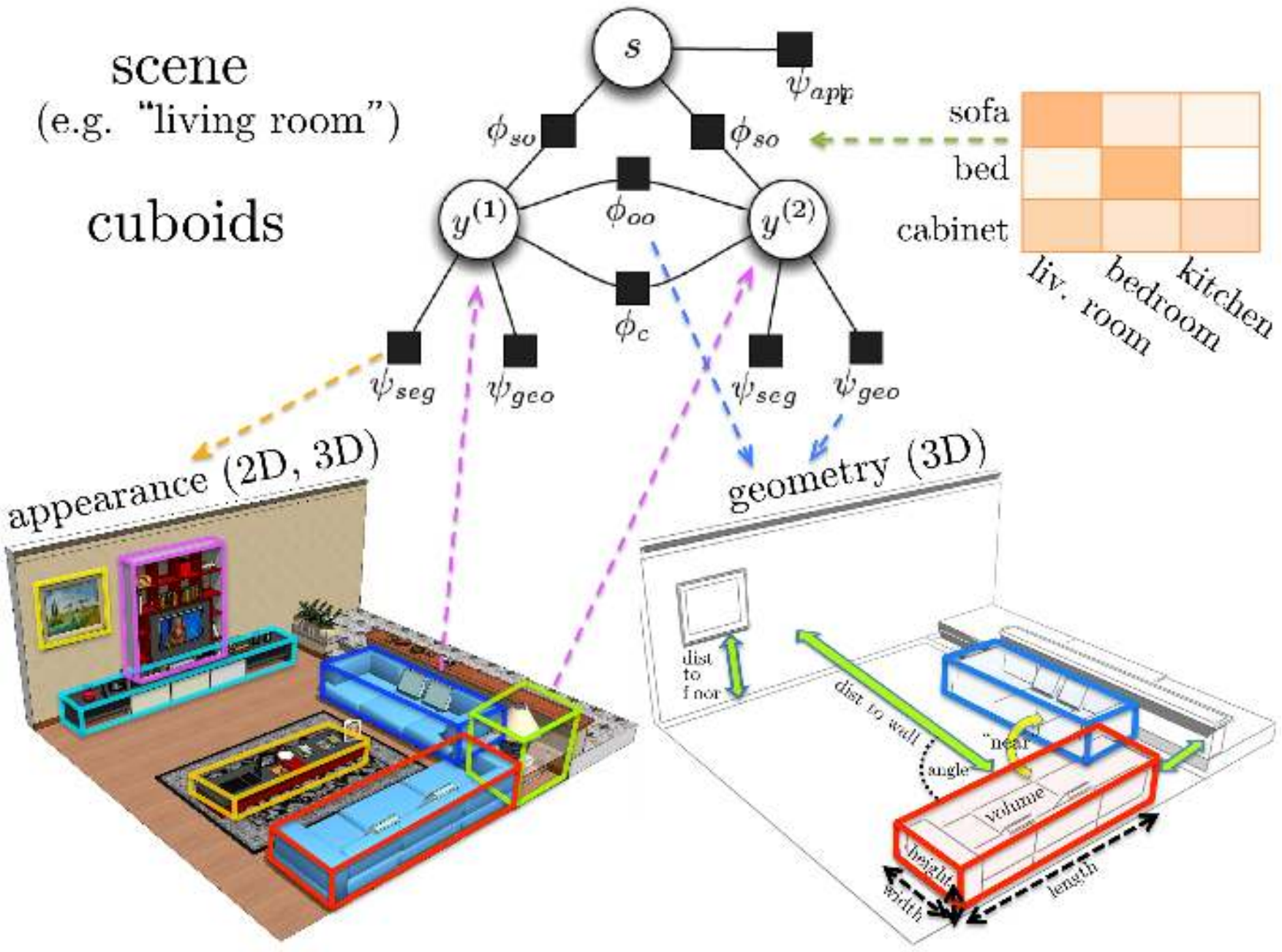

**Figure 2.6: Example of a Conditional Random Field (CRF) model.** This CRF attempts to localize the different objects in the scene using appearance and geometry cues. Additionally, relative geometry of the different objects is considered. Furthermore it is simultaneously performing scene classification using the context provided from the different objects in the scene. Figure reproduced from (Lin et al., 2013).

The main advantage of CRF models is the flexibility in defining the graph $G$ and design of the different feature functions $f_A(\mathbf{x}, \mathbf{y})$. Both must be heavily tailored to the application to capture the context and structure of the problem. As an example we describe the problem of joint 3D pose estimation of objects and scene detection from (Lin et al., 2013) shown in Fig. 2.6. In this particular case, there are two types of random variables $\mathbf{y}$: the scene variable $s$, which represents the type of scene; and the object variables $y^{(i)}$, which encode the 3D location of the object in the scene. Appearance factors are defined on all the nodes. The objects additionally have geometrical features, and there are factors capturing the geometric relationship between them. Finally there are factors between the scene and object nodes that capture the co-occurrences between them.

# Chapter 3

# Feature Point Descriptors

Representing a small part of an image, usually referred to as an image patch, as a compact vector allows performing many different useful tasks, e.g., finding the relative pose between two images or locating objects in images. Feature point descriptors are a way of representing these patches. It is well known that images contain a large amount of redundant information, by finding a smaller description of these patches it is possible to compare them in a more discriminative and efficient manner.

In this chapter we introduce two different descriptors developed as part of this thesis: the Deformation and Illumination Invariant (DaLI) descriptor and Convolutional Neural Network (CNN) based descriptors. We will discuss the implementation of both and summarize their strengths and possible usages.

## 3.1 Introduction

Feature point descriptors, i.e. the invariant and discriminative representation of local image patches, is a major research topic in computer vision. The field reached maturity with SIFT (Lowe, 2004), and has since become the cornerstone of a wide range of applications in recognition and registration. Descriptors are usually identified by their invariant properties. For example a descriptor that is invariant to rotation will be able to recognize patches independent of their orientation. On the other hand an illumination invariant descriptor would be useful facial recognition, but it would not be able to distinguish between night and day.

We show a summary of the aforementioned ubiquitous SIFT (Lowe, 2004) descriptor which initiated the widespread usage of descriptors in Fig. 3.1. By convolving the image patch with Gaussians to approximate the image gradient, computing histograms in different spatial bins, and then normalizing the resulting vector, Lowe was able to create a strong representation of a local image patch. The resulting descriptor is invariant to uniform scaling and orientation, while additionally being partially invariant to affine distortion and illumination changes.

While the most commonly used descriptors rely on convolving with Gaussians, we have focused on developing alternative descriptors that allow for more expressive rep-resentations of local image patches. In particular we have developed a descriptor based on embedding an image as a 3D mesh and then simulating the heat diffusion along the surface. We show that this representation is robust to non-rigid deformations. Furthermore, by performing a logarithmic sampling of the diffusion of heat at different time intervals and then computing the Fast Fourier Transform, we can make the descriptor also robust to illumination changes. We have created a deformation and illumination dataset in order to evaluate this descriptor and show it outperforms all other descriptors for this task.

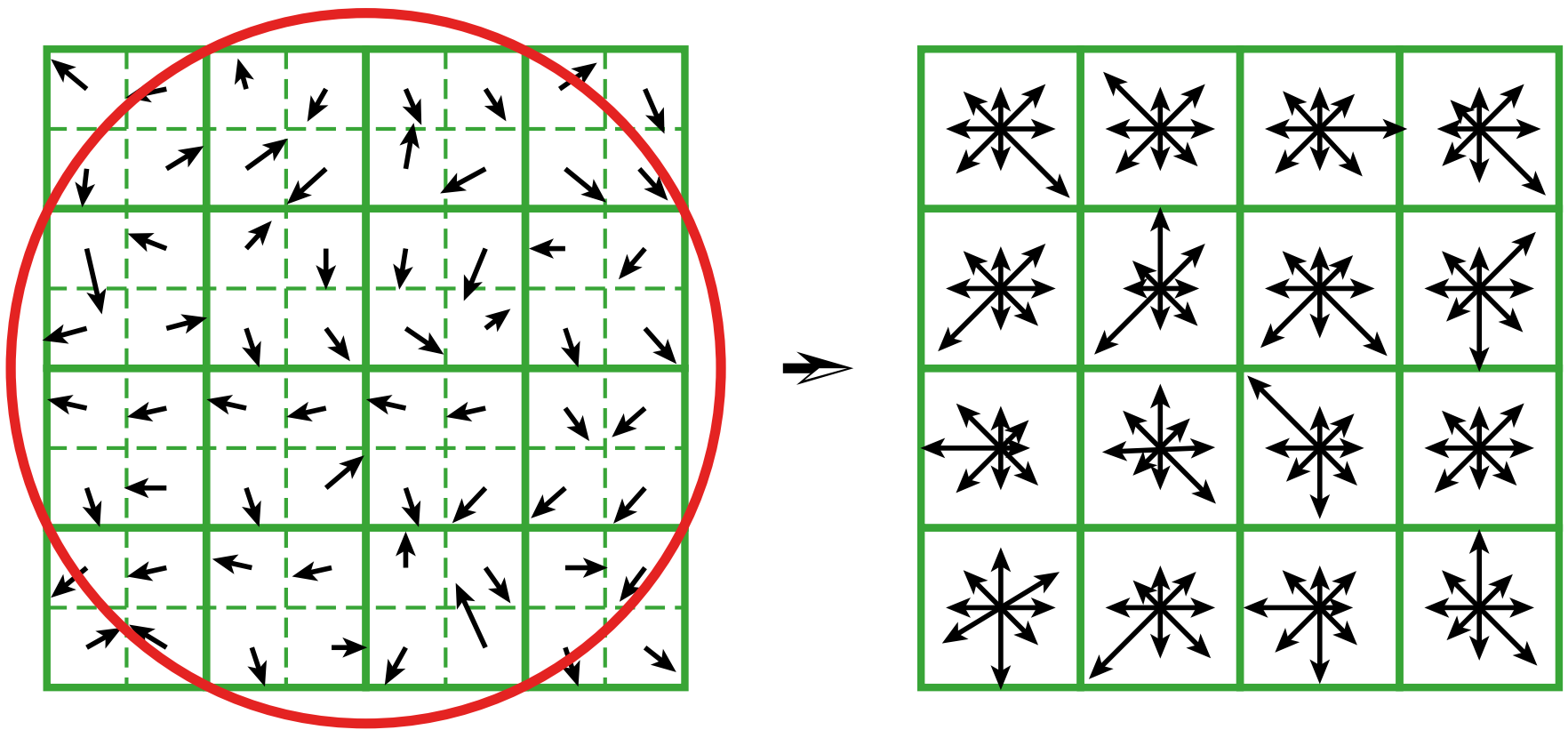

**Figure 3.1: SIFT descriptor.** The well known SIFT descriptor consists of 3D histograms of image gradients over spatial coordinates and gradient orientations. *Left:* The image gradients are weighted by a Gaussian window, indicated by the red circle. The length of the arrows corresponds to the sum of gradient magnitudes on a given direction. *Right:* The gradients in each of the $4 \times 4$ spatial blocks are collected in histograms with 8 bins each to form the final 128D SIFT descriptor. Released by Indif under CC-BY-SA-3.0, edited by the author.

Our second line of work consists of instead of hand-crafting these features, attempting to learn them using Convolutional Neural Networks (CNN). As it is not possible to learn these networks directly such as done for classification problems, we propose learning the network using a Siamese architecture. This consists of considering two image patches and whether or not they should correspond to the same point simultaneously. Both image patches are propagated forward through the network giving two different descriptors. Then we calculate the $L_2$ distance between both descriptors and apply a loss function that is meant to minimize the distance for two patches corresponding to the same object and maximize the distance for two patches corresponding to different objects. Afterwards, the error gradients are propagated backwards through the network for each patch. We use a dataset of patches extracted from Structure from Motion (Winder et al., 2009) for training, validation and testing. We will show that our sampling scheme in conjunction with large amounts of mining of samples is able to obtain a very large increase of performance over SIFT.

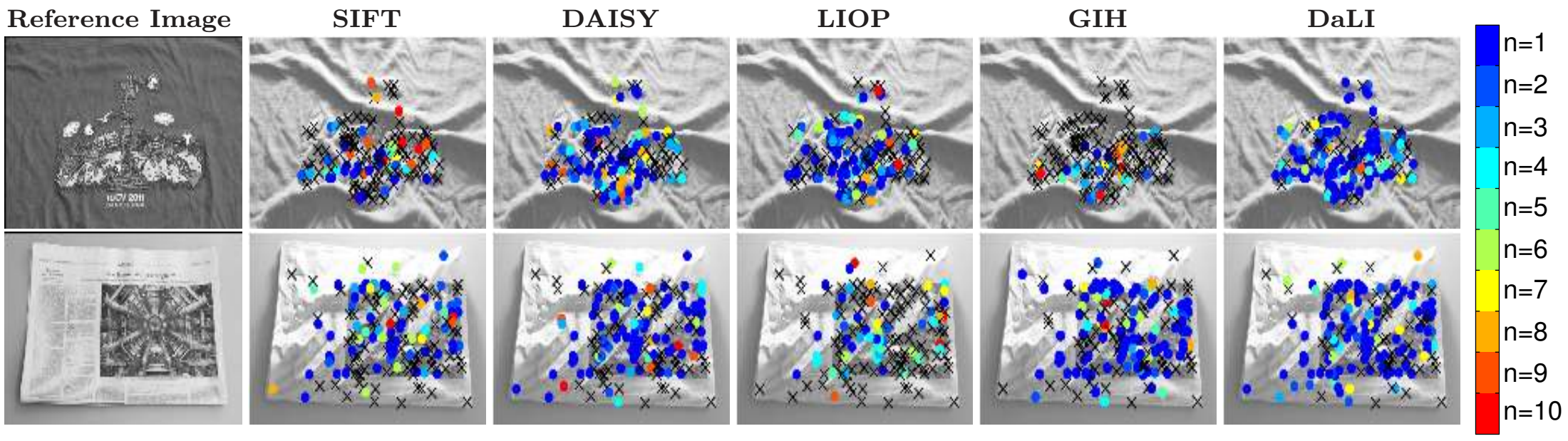

**Figure 3.2: Comparing DaLI against SIFT (Lowe, 2004), DAISY (Tola et al., 2010), LIOP (Wang et al., 2011) and GIH (Ling and Jacobs, 2005).** Input images correspond to different appearances of the object shown in the reference images, under the effect of non-rigid deformations and severe changes of illumination. Coloured circles indicate if the match has been correctly found among the first $n$ top candidates, where $n \leq 10$ is parameterized by the legend on the right. A feature is considered as mismatched when $n > 10$ and we indicate this with a cross. Note that the DaLI descriptor yields a significantly larger number of correct matches.

## 3.2 Deformation and Illumination Invariant (DaLI) Descriptor

Building invariant feature point descriptors is a central topic in computer vision with a wide range of applications such as object recognition, image retrieval and 3D reconstruction. Over the last decade, great success has been achieved in designing descriptors invariant to certain types of geometric and photometric transformations. For instance, the SIFT descriptor (Lowe, 2004) and many of its variants (Bay et al., 2006; Ke and Sukthankar, 2004; Mikolajczyk and Schmid, 2005; Morel and Yu, 2009; Tola et al., 2010) have been proven to be robust to affine deformations of both spatial and intensity domains. In addition, affine deformations can effectively approximate, at least on a local scale, other image transformations including perspective and viewpoint changes. However, as shown in Fig. 3.2, this approximation is no longer valid for arbitrary deformations occurring when viewing an object that deforms non-rigidly.

In order to match points of interest under non-rigid image transformations, recent approaches propose optimizing complex objective functions that enforce global consistency in the spatial layout of all matches (Cheng et al., 2008; Cho et al., 2009; Leordeanu and Hebert, 2005; Sanchez et al., 2010; Serradell et al., 2012; Torresani et al., 2008). Yet, none of these approaches explicitly builds a descriptor that goes beyond invariance to affine transformations. An interesting exception is (Ling and Jacobs, 2005), that proposes embedding the image in a 3D surface and using a Geodesic Intensity Histogram (GIH) as a feature point descriptor. However, while this approach is robust to non-rigid deformations, its performance drops under light changes. This is because a GIH considers deformations as one-to-one image mappings where image pixels only change their position but not the magnitude of their intensities.

To overcome the inherent limitation of using geodesic distances, we propose a novel descriptor based on the Heat Kernel Signature (HKS) recently introduced for non-rigid 3D shape recognition (Gębal et al., 2009; Rustamov, 2007; Sun et al., 2009), and which besides invariance to deformation, has been demonstrated to be robust to global

**DaLI Descriptor Slices at Frequencies** $w = \{0, 1, 2, 3, 4, 5\}$

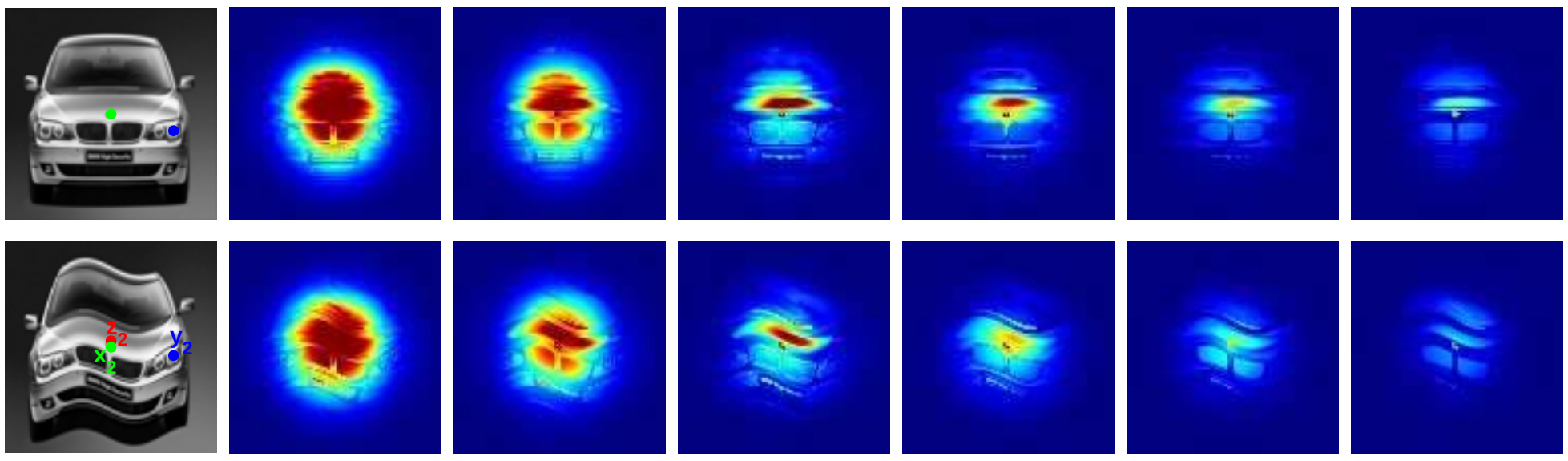

**Figure 3.3: Visualization of a DaLI descriptor.** The central idea is to embed image patches on 3D surfaces and describe them based on heat diffusion processes. The heat diffusion is represented as a stack of images in the frequency domain. The images show various slices of our descriptor for two different patches.

isotropic (Bronstein and Kokkinos, 2010) and even affine scalings (Raviv et al., 2011). In general, the HKS is particularly interesting in our context of images embedded on 3D surfaces, because illumination changes produce variations on the intensity dimension that can be seen as local anisotropic scalings, for which (Bronstein and Kokkinos, 2010) still shows a good resilience.

Our main contribution is thus using the tools of diffusion geometry to build a descriptor for 2D image patches that is invariant to both non-rigid deformations and photometric changes. An example of two descriptors are shown in Fig. 3.3. To construct our descriptor we consider an image patch $P$ surrounding a point of interest, as a surface in the $(u, v, \beta I(\mathbf{u}))$ space, where $(u, v)$ are the spatial coordinates, $I(\mathbf{u})$ is the intensity value at $(u, v)$, and $\beta$ is a parameter which is set to a large value to favor anisotropic diffusion and retain the gradient magnitude information. Drawing inspiration from the HKS (Gębal et al., 2009; Sun et al., 2009), we then describe each patch in terms of the heat it dissipates onto its neighborhood over time. To increase robustness against 2D and intensity noise, we use multiple such descriptors in the neighborhood of a point, and weigh them by a Gaussian kernel. As shown in Fig. 3.2, the resulting descriptor (which we call DaLI, for Deformation and Light Invariant) outperforms state-of-the-art descriptors in matching points of interest between images that have undergone non-rigid deformations and photometric changes.

We propose alternatives to both alleviate the high cost of the heat kernel computation and to reduce the dimensionality of the descriptor. In particular we investigate topologies with varying vertex densities. This allows reducing the effective size of the underlying mesh, and hence to speed up the DaLI computation time by a factor of over 4 with respect to a simple square mesh grid. In addition, we have also compacted the size of the final descriptor by a factor of 50× using a Principal Component Analysis (PCA) for dimensionality reduction. As a result, the descriptor we propose here can be computed and matched much faster when compared to (Moreno-Noguer, 2011), while preserving the discriminative power.

For evaluation, we acquired a challenging dataset that contains 192 pairs of real images, manually annotated, of diverse materials under different degrees of deformation and being illuminated by radically different illumination conditions. Fig. 3.2-left shows two samples of our dataset. We believe this is the first deformation and illumination dataset for evaluating image descriptors using real-world objects, and have made the dataset along with the code of the DaLI descriptor publicly available[1].

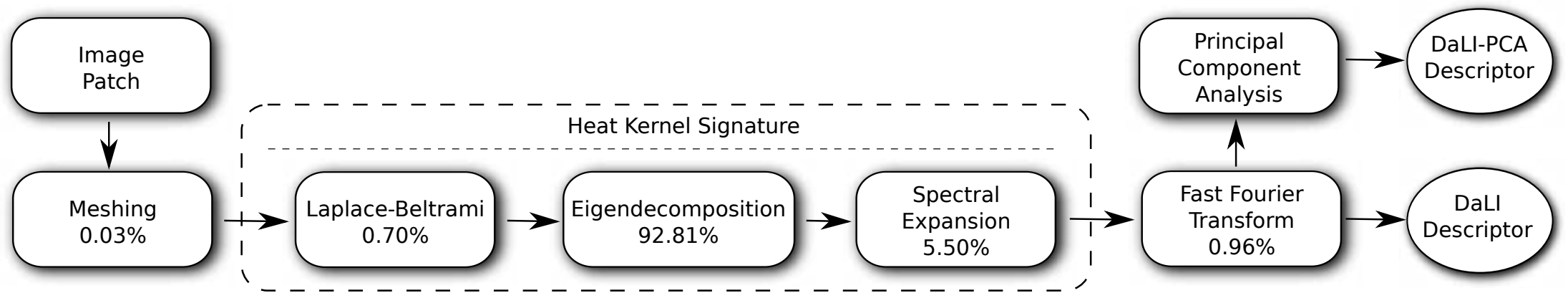

**Figure 3.4: Flowchart of the algorithm used to calculate the DaLI and DaLI-PCA descriptors.** The percentages below each of the steps indicate the total amount of the contribution of that step to the computation time. Observe that 99% of the computation time corresponds to the Heat Kernel Signature calculation and specifically almost entirely to the eigendecomposition of the Laplace-Beltrami operator.

## Related Work

The SIFT descriptor (Lowe, 2004) has become the main reference among feature point descriptors, showing great success in capturing local affine deformations including scaling, rotation, viewpoint change and certain lighting changes. Since it is relatively slow to compute, most of the subsequent works have focused on developing faster descriptors (Bay et al., 2006; Calonder et al., 2012; Ke and Sukthankar, 2004; Mikolajczyk and Schmid, 2005; Tola et al., 2010). Scale and rotation invariance has also been demonstrated in (Kokkinos et al., 2012) using a combination of logarithmic sampling and multi-scale signal processing, although that requires large image patches which make the resulting descriptor more sensitive to other deformations. Indeed, as discussed in (Vedaldi and Soatto, 2005), little effort has been devoted to building descriptors robust to more general deformations.

The limitations of the affine-invariant descriptors when solving correspondences between images of objects that have undergone non-rigid deformations are compensated by enforcing global consistency, both spatial and photometric, among all features (Belongie et al., 2002; Berg et al., 2005; Cheng et al., 2008; Cho et al., 2009; Leordeanu and Hebert, 2005; Sanchez et al., 2010; Serradell et al., 2012; Torresani et al., 2008), or introducing segmentation information within the descriptor itself (Trulls et al., 2013, 2014). In any event, none of these methods specifically handles the non-rigid nature of the problem, and they rely on solving complex optimization functions for establishing matches.

An alternative approach is to directly build a deformation invariant descriptor. With that purpose, recent approaches in two-dimensional shape analysis have proposed using different types of intrinsic geometry. For example, (Bronstein et al., 2007; Ling and Jacobs, 2007) define metrics based on the inner-distance, and (Ling et al., 2010) proposes using geodesic distances. However, all these methods require the shapes to be segmented out from the background and represented by binary images, which is difficult to do in practice. In (Ling and Jacobs, 2005), it was shown that geodesic distances, in combination with an appropriate 3D embedding of the image, were adequate to achieve deformation invariance in intensity images. Nonetheless, this method assumes that pixels only change their image locations and not their intensities and, as shown in Fig. 3.2, is prone to failure under illumination changes.

[1] http://www.iri.upc.edu/people/esimo/research/dali/

There have also been efforts to build illumination invariant descriptors. Such works consider strategies based on intensity ordering and spatial sub-division (Fan et al., 2012; Gupta and Mittal, 2007, 2008; Gupta et al., 2010; Heikkilä et al., 2009; Tang et al., 2009; Wang et al., 2011). While these approaches are invariant to monotonically increasing intensity changes, their success rapidly falls when dealing with photometric artifacts produced by complex surface reflectance or strong shadows.

The DaLI descriptor we propose can simultaneously handle such relatively complex photometric and spatial warps. Following (Ling and Jacobs, 2005), we represent the images as 2D surfaces embedded in the 3D space. This is in fact a common practice, although it has been mostly employed for low level vision tasks such as image denoising (Sochen et al., 1998; Yezzi, 1998) or segmentation (Yanowitz and Bruckstein., 1989). The fundamental difference between our approach and (Ling and Jacobs, 2005) is that we then describe each feature point on the embedded surface considering the heat diffusion over time (Gębal et al., 2009; Lévy, 2006; Sun et al., 2009) instead of using a Geodesic Intensity Histogram. As we will show in the results section this yields substantially improved robustness, especially to illumination changes. Heat diffusion theory has been used by several approaches for the analysis of 3D textured (Kovnatsky et al., 2011) and non-textured shapes (de Goes et al., 2008; Lévy, 2006; Reuter et al., 2006; Rustamov, 2007), but to the best of our knowledge, it has not been used before to describe patches in intensity images.

One of the main limitations of the methods based on the heat diffusion theory is the high complexity cost they require. The bottleneck of their computation lies on an eigendecomposition of a $n_v \times n_v$ Laplacian matrix (see Fig. 3.5), where $n_v$ is the number of vertices of the underlying mesh. This has been addressed by propagating the eigenvectors across different mesh resolutions (Shi et al., 2006; Wesseling, 2004) or using matrix exponential approximations (Vaxman et al., 2010). In this work, an annular multi-resolution grid will be used to improve the efficiency of the DaLI computation. Additionally, PCA will be used to reduce the dimensionality of the original DaLI descriptor (Moreno-Noguer, 2011), hence speeding up the matching process as well.

## Deformation and Light Invariant Descriptor

Our approach is inspired by current methods (Gębal et al., 2009; Sun et al., 2009) that suggest using diffusion geometry for 3D shape recognition. In this section we show how this theory can be adapted to describe 2D local patches of images that undergo non-rigid deformations and photometric changes. A general overview of the different steps needed to compute the DaLI and DaLI-PCA descriptors can be seen in Fig. 3.4 and are explained more in detail below.

### Invariance to Non-Rigid Deformations

Let us assume we want to describe a 2D image patch $P$, of size $S_P \times S_P$ and centered on a point of interest $\mathbf{p}$. In order to apply the diffusion geometry theory to intensity patches we regard them as 2D surfaces embedded in 3D space (Fig. 3.5 bottom-left). More formally, let $f : P \to \mathcal{M}$ be the mapping of the patch $P$ to a 3D Riemannian

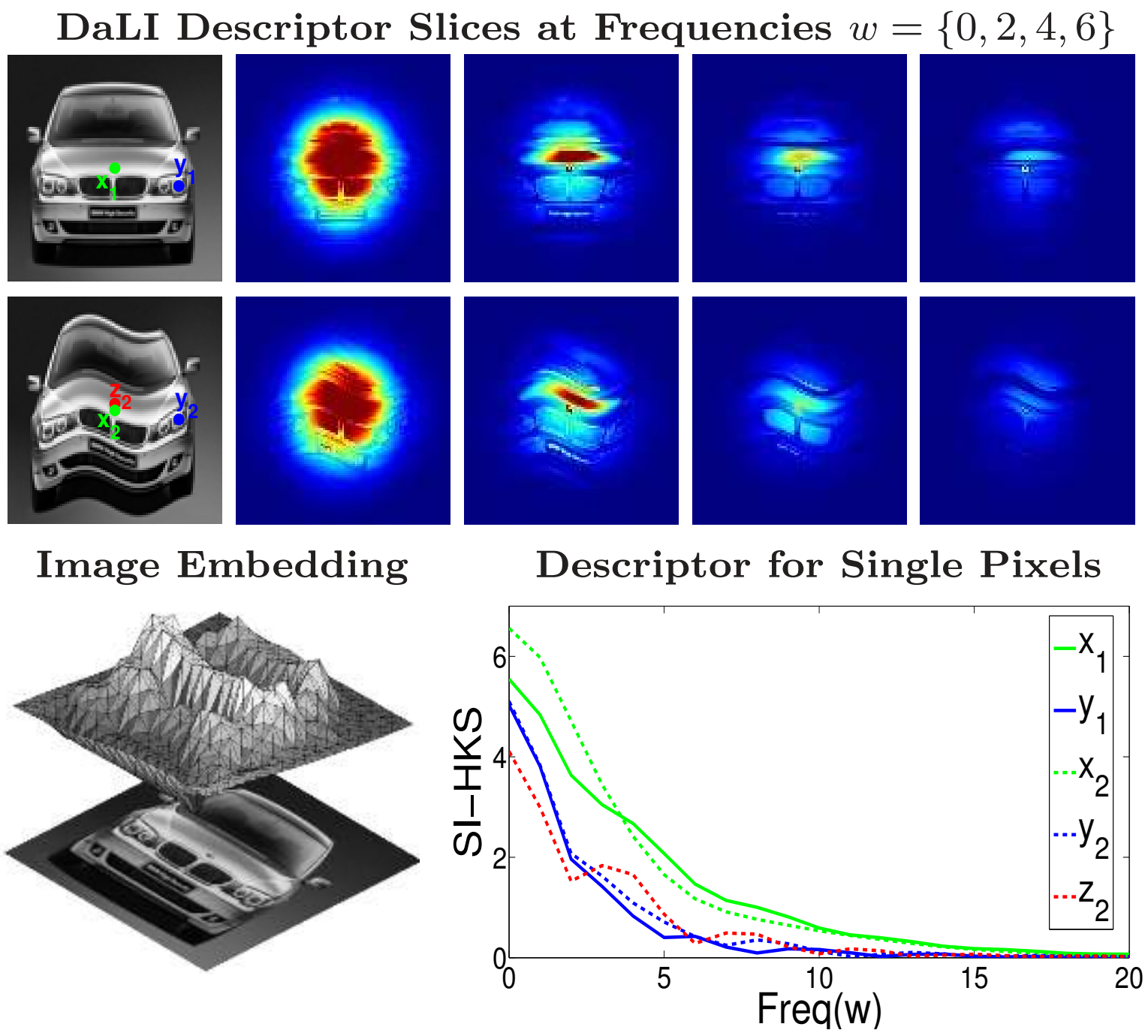

**Figure 3.5: DaLI descriptor.** Our central idea is to embed image patches in 3D surfaces and describe them based on heat diffusion processes. We represent the heat diffusion as a stack of images in the frequency domain. The top images show various slices of our descriptor for two different patches. The bottom-right graph depicts the value of the descriptor for the pixels marked by colour circles in the upper images. Note that corresponding pixels have very similar signatures. However, the signature may significantly change from one pixel to its immediate neighbor. For instance, $\mathbf{z}_2$ is at one pixel distance from $\mathbf{x}_2$, but their signatures are rather different. As a consequence, using the signature of a single point as a descriptor is very sensitive to 2D noise in the feature detection process. We address this by simultaneously considering the signature of all the pixels within the patch, weighted by a Gaussian function of the distance to the center of the patch.

manifold $\mathcal{M}$. We explicitly define this mapping by:

$$f : \mathbf{u} \rightarrow (u, v, \beta I(\mathbf{u})) \quad \forall \mathbf{u} \in P \;, \tag{3.1}$$

where $I(\mathbf{u})$ is the pixel intensity at $\mathbf{u} = (u, v)^\top$, and $\beta$ is a parameter that, as we will discuss later, controls the amount of gradient magnitude preserved in the descriptor.

Several recent methods (Gębal et al., 2009; Lévy, 2006; Reuter et al., 2006; Rustamov, 2007; Sun et al., 2009) have used the heat diffusion geometry for capturing the local properties of 3D surfaces and performing shape recognition. Similarly, we describe each patch $P$ based on the heat diffusion equation over the manifold $\mathcal{M}$:

$$\left( \triangle_{\mathcal{M}} + \frac{\partial}{\partial t} \right) h(\mathbf{u}, t) = 0 \;,$$

where $\triangle_{\mathcal{M}}$ is the *Laplace-Beltrami operator*, a generalization of the Laplacian to non-Euclidean spaces, and $h(\mathbf{u}, t)$ is the amount of heat on the surface point $\mathbf{u}$ at time $t$.

The solution $k(\mathbf{u}, \mathbf{v}, t)$ of the heat equation with an initial heat distribution $h_o(\mathbf{u}, t) = \delta(\mathbf{u} - \mathbf{v})$ is called the *heat kernel*, and represents the amount of heat that is diffused between points $\mathbf{u}$ and $\mathbf{v}$ at time $t$, considering a unit heat source at $\mathbf{u}$ at time $t = 0$. For a compact manifold $\mathcal{M}$, the heat kernel can be expressed by following spectral expansion (Chavel, 1984; Reuter et al., 2006):

$$k(\mathbf{u}, \mathbf{v}, t) = \sum_{i=0}^{\infty} e^{-\lambda_i t} \boldsymbol{\phi}_i(\mathbf{u}) \boldsymbol{\phi}_i(\mathbf{v}) \ , \tag{3.2}$$

where $\{\lambda_i\}$ and $\{\boldsymbol{\phi}_i\}$ are the eigenvalues and eigenfunctions of $\triangle_{\mathcal{M}}$, and $\boldsymbol{\phi}_i(\mathbf{u})$ is the value of the eigenfunction $\boldsymbol{\phi}_i$ at the point $\mathbf{u}$. Based on this expansion, (Sun et al., 2009) proposes describing a point $\mathbf{p}$ on $\mathcal{M}$ using the Heat Kernel Signature

$$\text{HKS}(\mathbf{p}, t) = k(\mathbf{p}, \mathbf{p}, t) = \sum_{i=0}^{\infty} e^{-\lambda_i t} \boldsymbol{\phi}_i^2(\mathbf{p}) \ , \tag{3.3}$$

which is shown to be isometrically-invariant, and adequate for capturing both the local properties of the shape around $\mathbf{p}$ (when $t \to 0$) and the global structure of $\mathcal{M}$ (when $t \to \infty$).

However, while on smooth surfaces the HKS of neighboring points are expected to be very similar, when dealing with the wrinkled shapes that may result from embedding image patches, the heat kernel turns to be highly unstable along the spatial domain (Fig. 3.5 bottom-right). This makes the HKS particularly sensitive to noise in the 2D location of the keypoints. To handle this situation, we build the descriptor of a point $\mathbf{p}$ by concatenating the HKS of all points $\mathbf{u}$ within the patch $P$, properly weighted by a Gaussian function of the distance to the center of the patch. We therefore define the following Deformation Invariant (DI) descriptor:

$$\text{DI}(\mathbf{p}, t) = [\text{HKS}(\mathbf{u}, t) \cdot G(\mathbf{u}; \mathbf{p}, \sigma)]_{\forall \mathbf{u} \in P} \ , \tag{3.4}$$

where $G(\mathbf{u}; \mathbf{p}, \sigma)$ is a 2D Gaussian function centered on $\mathbf{p}$ having a standard deviation $\sigma$, evaluated at $\mathbf{u}$. Note that for a specific time instance $t$, $\text{DI}(\mathbf{p}, t)$ is a $S_P \times S_P$ array.

The price we pay for achieving robustness to 2D noise is an increase of the descriptor size. That is, if $\text{HKS}(\mathbf{p}, t)$ is a function defined on the temporal domain $\mathbb{R}^+$ discretized into $n_t$ equidistant intervals, the complete DI descriptor $\text{DI}(\mathbf{p}) = [\text{DI}(\mathbf{p}, t_1), \ldots, \text{DI}(\mathbf{p}, t_{n_t})]$ will be defined on $S_P \times S_P \times n_t$, the product of the spatial and temporal domains. However, note that for our purposes this is still feasible, because we do not need to compute a descriptor for every pixel of the image, but just for a few hundreds of points of interest. Furthermore, as we will next discuss, the descriptor may be highly compacted if we represent it in frequency domain instead of time domain, and even further compacted by using dimensionality reduction techniques such as Principal Component Analysis (PCA).

### Invariance to Illumination Changes

An inherent limitation of the descriptor introduced in Eq. (3.4) is that it is not illumination invariant. This is because light changes scale the manifold $\mathcal{M}$ along the intensity axis, and the HKS is sensitive to scaling. It can be shown that an isotropic scaling of the manifold $\mathcal{M}$ by a factor $\alpha$, scales the eigenvectors and eigenvalues of Eq. (3.2) by factors $1/\alpha$ and $1/\alpha^2$, respectively (Reuter et al., 2006). The HKS of a point $\alpha\mathbf{p} \in \alpha\mathcal{M}$ can then be written as

$$\mathrm{HKS}(\alpha\mathbf{p}, t) = \sum_{i=0}^{\infty} e^{-\frac{\lambda_i}{\alpha^2} t} \frac{\phi_i^2(\mathbf{p})}{\alpha^2} = \frac{1}{\alpha^2} \mathrm{HKS}(\mathbf{p}, \frac{t}{\alpha^2}) \,, \tag{3.5}$$

which is an amplitude and time scaled version of the original HKS.

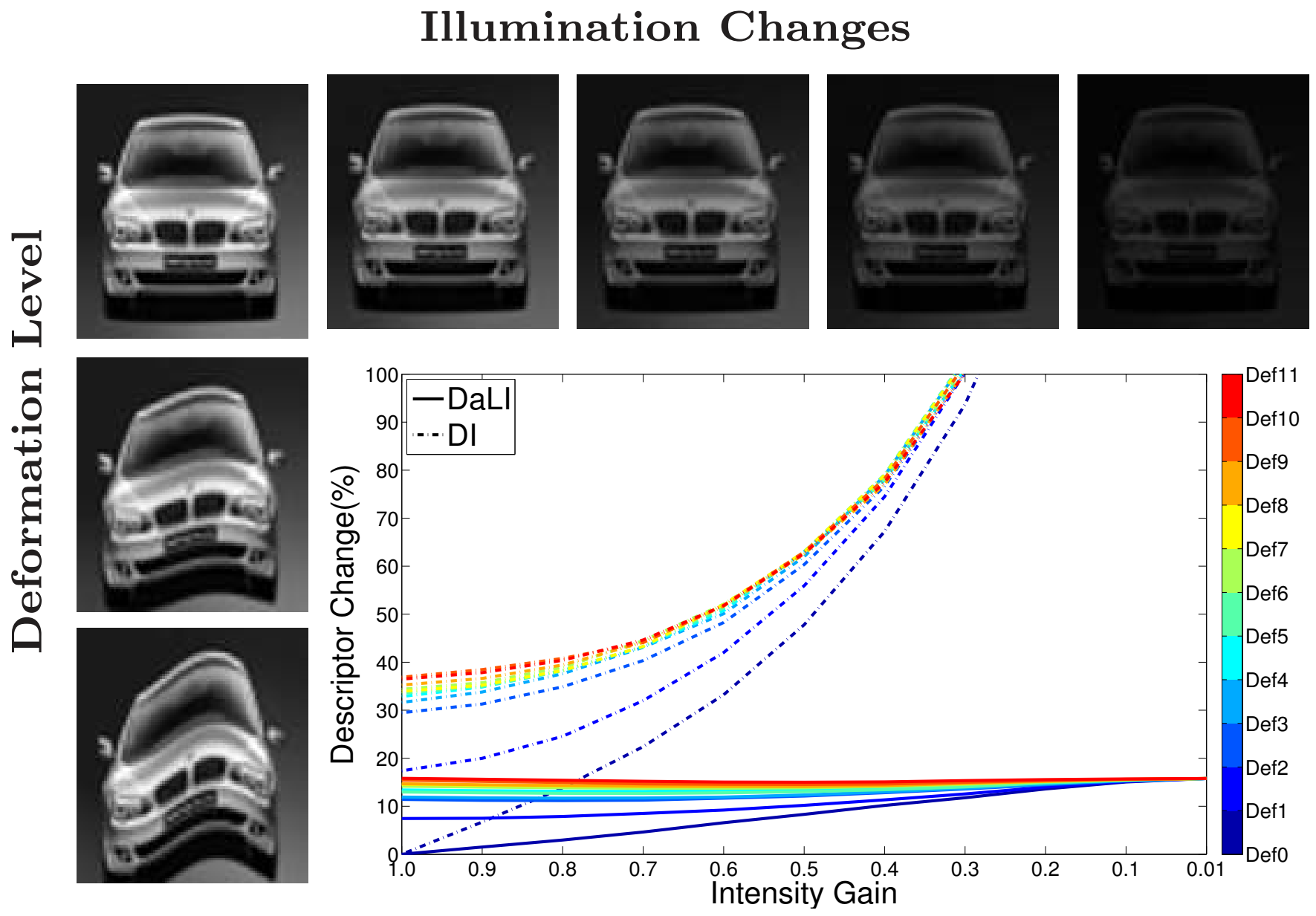

**Figure 3.6: Invariance of the DaLI and DI descriptors to non-rigid deformations and illumination changes.** *Top row and left column images:* Different degrees of deformation and light changes applied on the top left reference patch $P_0$. Deformations are applied according to a function $\mathrm{Def}(\cdot) \in \{\mathrm{Def0}, \ldots, \mathrm{Def11}\}$, where Def11 corresponds to the maximal deformation. Light changes are produced by scaling the intensity of $P_0$ by a gain $g \in [0, 1]$. *Bottom Graph:* Given a deformation $\mathrm{Def}(\cdot)$ and a gain factor $g$, we compute the percentage of change of the DI descriptor by $\|\mathrm{DI}(P_0) - \mathrm{DI}(\mathrm{Def}(gP_0))\| / \|\mathrm{DI}(P_0)\|$. The percentage of change for DaLI is computed in a similar way. Observe that DaLI is much less sensitive than DI, particularly to illumination changes.

Nonetheless, under isotropic scalings, several alternatives have been proposed to remove the dependence of the HKS on the scale parameter $\alpha$. For instance, (Reuter et al., 2006) suggests normalizing the eigenvalues in Eq. (3.2). In this work we followed (Bronstein and Kokkinos, 2010), that applies three consecutive transformations on the HKS. First, the time-dimension is logarithmically sampled, which turns the time scaling into a time-shift, that is, the right-hand side of Eq. (3.5) begets $\alpha^{-2}\mathrm{HKS}(\mathbf{p}, -2\log\alpha + \log t)$. Second, the amplitude scaling factor is removed by taking logarithm and derivative w.r.t. $\log t$. The Heat Kernel then becomes $\frac{\partial}{\partial \log t} \log \mathrm{HKS}(\mathbf{p}, -2\log\alpha + \log t)$. The time-shift term $-2\log\alpha$ is finally removed using the magnitude of the Fourier transform, which yields $\mathrm{SI\text{-}HKS}(\mathbf{p}, w)$, a scale invariant version of the original HKS in the frequency domain. In addition, since most of the signal information is concentrated in the low-frequency components, the size of the descriptor can be highly reduced com-pared to that of HKS$(\mathbf{p}, t)$ by eliminating the high-frequency components past a certain frequency threshold $w_{max}$.

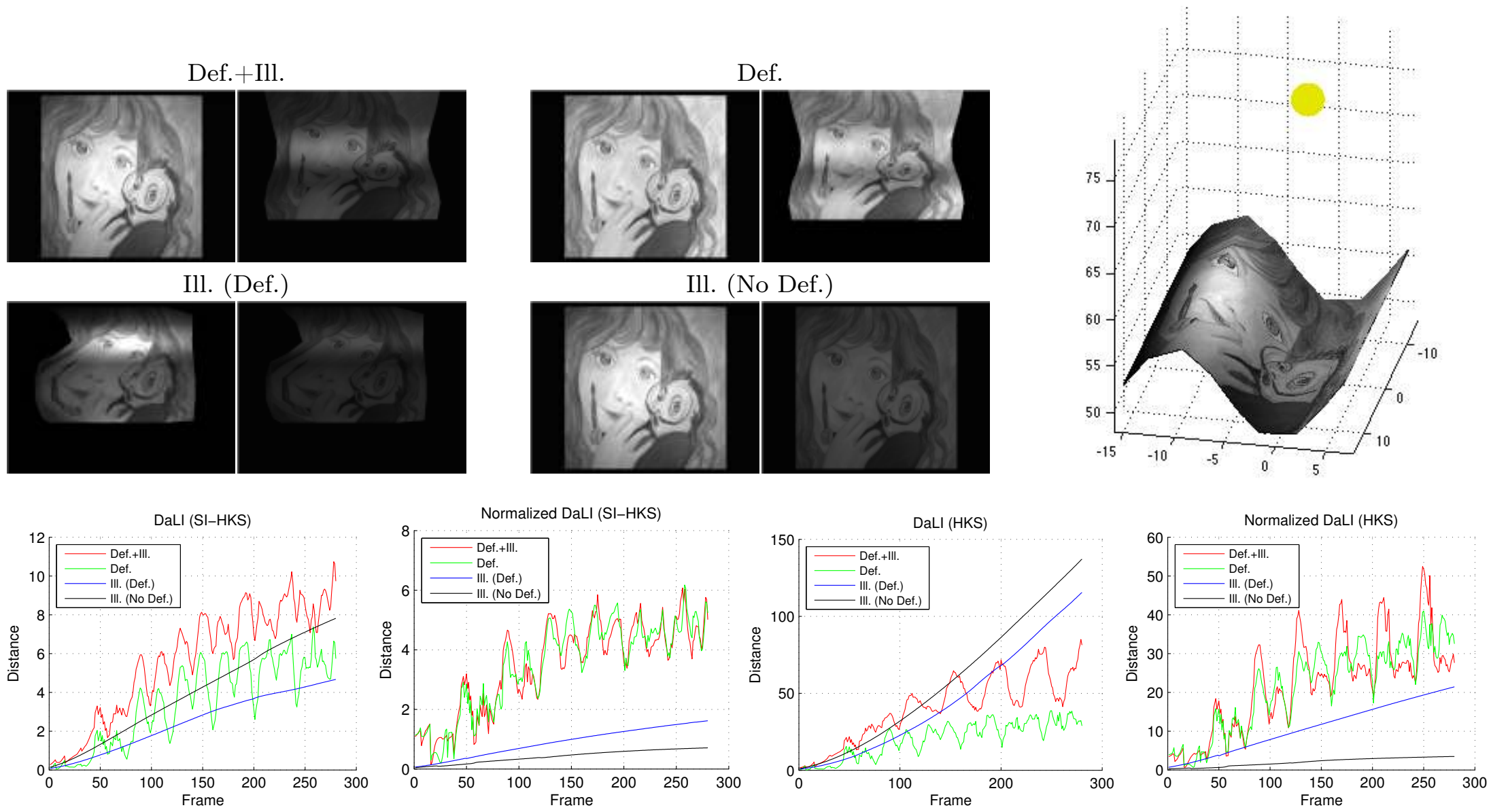

**Figure 3.7: Evaluation of the descriptor robustness on synthetic sequences.** In the top-left we show two sample images (the reference image and one specific frame) from all four different scenarios we consider. In the top-right we show an 3D view of the rendering process, with the light position placed near the mesh and producing patterns of different brightness on top of the surface. The bottom row depicts the descriptor distance between every input frame and the reference image for different descriptor variants.

As we will show in the results section, another advantage of the SI-HKS signature is that although it is specifically designed to remove the dependence of the HKS on isotropic scalings, it is quite resilient to anisotropic transformations, such as those produced by photometric changes that only affect the intensity dimension of the manifold $\mathcal{M}$. Thus, we will use this signature to define our Deformation and Light Invariant (DaLI) descriptor:

$$\text{DaLI}(\mathbf{p}, w) = [\text{SI-HKS}(\mathbf{u}, w) \cdot G(\mathbf{u}; \mathbf{p}, \sigma)]_{\forall \mathbf{u} \in P} \ .$$

Again, the full DaLI$(\mathbf{p})$ descriptor is defined as a concatenation of $w_{max}$ slices in the frequency domain, each of size $S_P \times S_P$.

Fig. 3.5-top shows several DaLI slices at different frequencies for a patch and a deformed version of it. As said above, observe that most of the signal is concentrated in the low frequency components. In Fig. 3.6 we compare the sensitivity of the DI and DaLI descriptors to deformation and light changes, simulated here by a uniform scaling of the intensity channel. Note that DaLI, in contrast to DI, remains almost invariant to light changes, and it also shows a better performance under deformations. In the results section, we will show that this invariance is also accompanied by a high discriminability, yielding significantly better results in keypoint matching than existing approaches.

In order to get deeper insight about the properties of the DaLI descriptor, we have further evaluated the HKS and SI-HKS descriptor variants on a synthetic experiment, in which we have rendered various sequences of images of a textured 3D wave-like mesh under different degrees of deformation and varying illumination conditions. The surface's reflectance is assumed to be Lambertian and the light source is moved near the surface, producing lighting patterns that combine both shading and the effects of the inverse-square falloff law.

We have analyzed four particular situations: *Def.+Ill.*, varying both deformation and the light source position; *Def.*, varying deformation and keeping the light source at infinity; *Ill. (Def.)*, starting with a largely deformed state which is kept constant along the sequence and varying the light source position; and *Ill. (No Def.)*, varying the light source position while keeping the surface flat. The mesh deformation in the first two sequences, corresponds to a sinusoidal warp, in which the amplitude of the deformation increases with the frame number. The varying lighting conditions in all experiments except the second, are produced by smoothly moving the light source on a hemisphere very close to the surface. Two frames from each of these sequences are shown in the top-left of Fig. 3.7.

For the evaluation, we computed the $L_2$ norm between pairs of descriptors at the center of the first and $n$-th frames of the sequence. The results are depicted in Fig. 3.7-bottom. When computing the distances, we consider two situations: normalizing the intensity of the input images so that the pixels follow a distribution $\mathcal{N}(0,1)$, and directly using the input image intensities. The most interesting outcome of this experiment is how the non-normalized SI-HKS descriptor has comparable distances for all the scenarios. On the other hand, the normalized versions (SI-HKS and HKS) seem to distinguish largely between whether there is or is not deformation. It is also worth noting that this normalization creates some instability at the earlier frames while the non-normalized SI-HKS descriptor starts at nearly 0 error and increases smoothly for all scenarios. Note also the low performance of the non-normalized HKS descriptor under illumination changes as seen by the exponential curves for the illumination scenarios *Ill. (Def)* and *Ill. (No Def)*, and the large fluctuations for both the deformation *Def* and the illumination changing scenario *Def. + Ill*. This indicates the importance of the logarithmic sampling and Fourier transform process we apply to make HKS illumination invariant.

### Handling In-Plane Rotation

Although DaLI tolerates certain amounts of in-plane rotation, it is not designed for this purpose. This is because with the aim of increasing robustness to 2D noise, we built the descriptor using all the pixels within the patch, and their spatial relations have been retained. Thus, if the patch is rotated, the descriptor will also be rotated.

In order to handle this situation, during the matching process we will consider several rotated copies of the descriptors. Therefore, given $\mathrm{DaLI}(\mathbf{p}_1)$ and $\mathrm{DaLI}(\mathbf{p}_2)$ we will compare them based on the following metric

$$\mathrm{d}(\mathbf{p}_1, \mathbf{p}_2) = \arg\min_{\theta_i} \|\mathrm{R}_{\theta_i}(\mathrm{DaLI}(\mathbf{p}_1)) - \mathrm{DaLI}(\mathbf{p}_2)\|$$

where $\|\cdot\|$ denotes the $L_2$ norm and $\mathrm{R}_{\theta_i}(\mathrm{DaLI}(\mathbf{p}))$ rotates $\mathrm{DaLI}(\mathbf{p})$ by an angle $\theta_i$. This parameter is chosen among a discrete set of values $\boldsymbol{\theta}$.

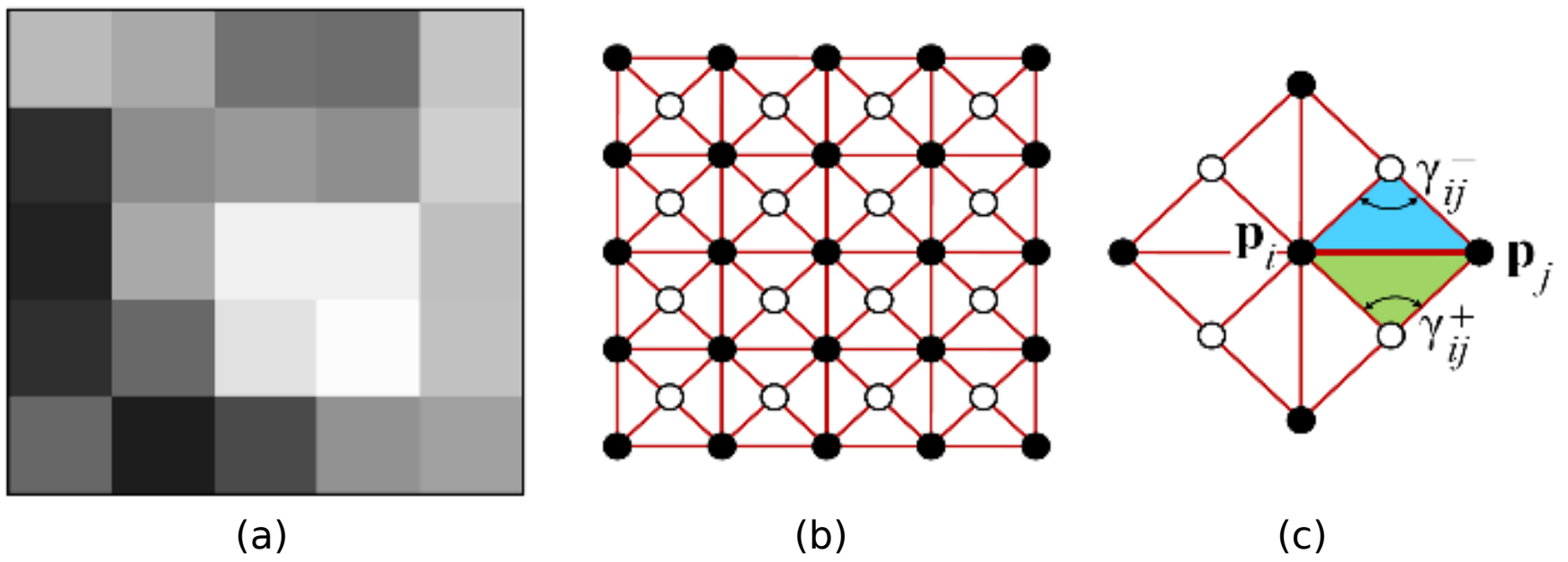

**Figure 3.8: Representation of an image patch.** (a) Image patch. (b) Representation of the patch as a triangular mesh. For clarity of presentation we only depict the $(u, v)$ dimension of the mesh. Note that besides the vertices placed on the center of the pixels (filled circles) we have introduced additional intra-pixel vertices (empty circles), that provide finer heat diffusion results and higher tolerance to in-plane rotations. (c) Definition of the angles used to compute the discrete Laplace-Beltrami operator.

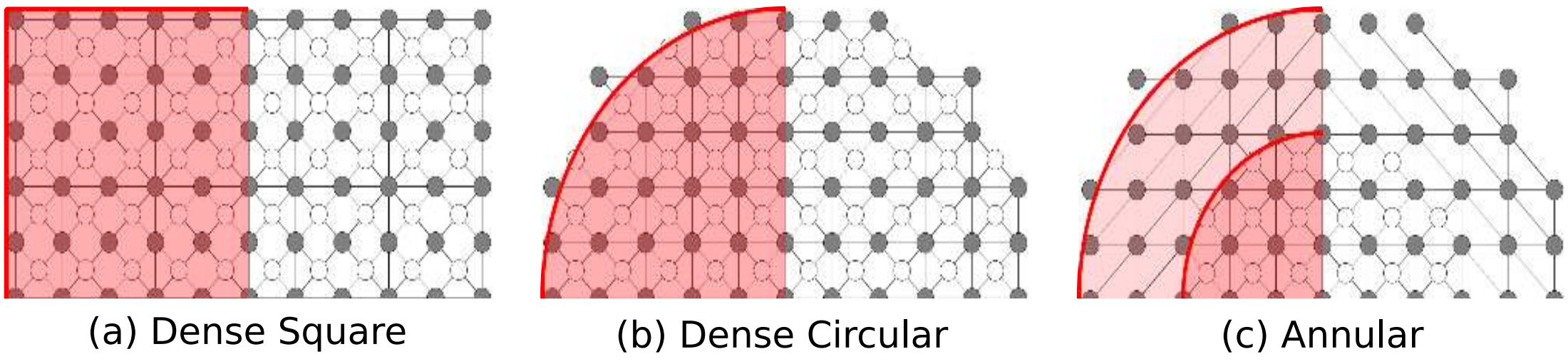

**Figure 3.9: Different mesh triangulations.** Upper half of three different triangulations of a $11 \times 11$ image patch. The shading on the left half of the mesh indicates the density of the meshing. Dark red shading indicates high density and lighter red shading corresponds to low density. (a) Dense Square Mesh, with the same topology as in Fig. 3.8. By using circular meshes (b, c), we reduce the number of vertices and thus, the computation time of the heat kernel. In the case of the annular mesh (c), a further reduction of the number of nodes is achieved by having a variable resolution of the mesh that is more dense at the center. The edges of the annular mesh preserve symmetry around the central point in order to favor uniform heat diffusion.

This rotation handling will not be necessary when using Principal Component Analysis to compress the descriptor size as we describe in Section 3.2.

### Implementation Details

We next describe a number of important details to be considered for the implementation of the DaLI descriptor.

**Geometry of the embedding.** For the numerical computation of the heat diffusion, it is necessary to discretize the surface. We therefore represent the manifold $\mathcal{M}$ on which the image patch is embedded using a triangulated mesh. Fig. 3.8(b) shows the underlying structured 8-neighbour representation we use. Although it requires introducing additional *virtual* vertices between the pixels, its symmetry with respect to both axes provides robustness to small amounts of rotation, and more uniform diffusions than other configurations.

| Mesh Type | # Pixels | # Vertices ($n_v$) | # Faces ($n_f$) | Time (s) |
|---|---|---|---|---|
| Dense Square | 1681 | 3281 | 6400 | 1.988 |
| Dense Circular | 1345 | 2653 | 5144 | 1.509 |
| Annular | 1345 | 1661 | 3204 | 0.460 |

**Table 3.1: DaLI computation time and mesh complexity for different triangulations.** We consider a circular patch with outer radius $S = 20$, and inner radius $S_o = 10$ (for the Annular mesh).

As seen in Fig. 3.4, nearly all the computation time of the DaLI descriptor is spent calculating the Laplace-Beltrami eigenfunctions of the triangulated mesh. In the following subsection we will show that this computation turns to have a cubic cost on the number of vertices of the mesh, hence, important speed gains can be achieved by lowering this number. For this purpose we further considered a circular mesh (Fig. 3.9(b)), and a mesh with a variable density, like the one depicted in Fig. 3.9(c), where a lower resolution annulus is used for the pixels further away from the center.

By using an annular mesh with an inner radius $S_o = S/2$, where $S$ is the size of the outer radius, we were able to speed up the computation of the DaLI descriptor by a factor of four compared to the Dense Squared configuration (see Table 3.1). Most importantly, this increase in speed did not result in poorer recognition rates.

Another important variable of our design is the magnitude of the parameter $\beta$ in Eq. (3.1), that controls the importance of the intensity coordinate with respect to the $(u, v)$ coordinates. In particular, as shown in Fig. 3.10, large values of $\beta$ allow our descriptor to preserve edge information. This is a remarkable feature of the DaLI descriptor, because besides being deformation and illumination invariant, edge information is useful to discriminate among different patches.

**Discretization of the Laplace-Beltrami operator** In order to approximate the Laplace-Beltrami eigenfunctions on the triangular mesh we use the cotangent scheme described in (Pinkall and Polthier, 1993). We next detail the main steps.

Let $\{\mathbf{p}_1, \ldots, \mathbf{p}_{n_v}\}$ be the vertices of a triangular mesh, associated to an image patch embedded on a 3D manifold. We approximate the discrete Laplacian by a $n_v \times n_v$ matrix $\mathbf{L} = \mathbf{A}^{-1}\mathbf{M}$ where $\mathbf{A}$ is a diagonal matrix in which $\mathbf{A}_{ii}$ is proportional to the area of all triangles sharing the vertex $\mathbf{p}_i$. $\mathbf{M}$ is a $n_v \times n_v$ sparse matrix computed by:

$$\mathbf{M}_{ij} = \begin{cases} \sum_k m_{ik} & \text{if } i = j \\ -m_{ij} & \text{if } \mathbf{p}_i \text{ and } \mathbf{p}_j \text{ are adjacent} \\ 0 & \text{otherwise} \end{cases}$$

where $m_{ij} = \cot \gamma_{ij}^{+} + \cot \gamma_{ij}^{-}$, and $\gamma_{ij}^{+}$ and $\gamma_{ij}^{-}$ are the two opposite angles depicted in Fig. 3.8(c), and the subscript '$k$' refers to all neighboring vertices of $\mathbf{p}_i$.

The eigenvectors and eigenvalues of the discrete La-place-Beltrami operator can then be computed from the solution of the generalized eigenproblem $\mathbf{M}\mathbf{\Phi} = \mathbf{\Lambda}\mathbf{A}\mathbf{\Phi}$, where $\mathbf{\Lambda}$ is a diagonal matrix with the eigenvalues $\{\lambda_i\}$ and the columns of $\mathbf{\Phi}$ correspond to the eigenvectors $\{\boldsymbol{\phi}_i\}$ in Eq. (3.2).

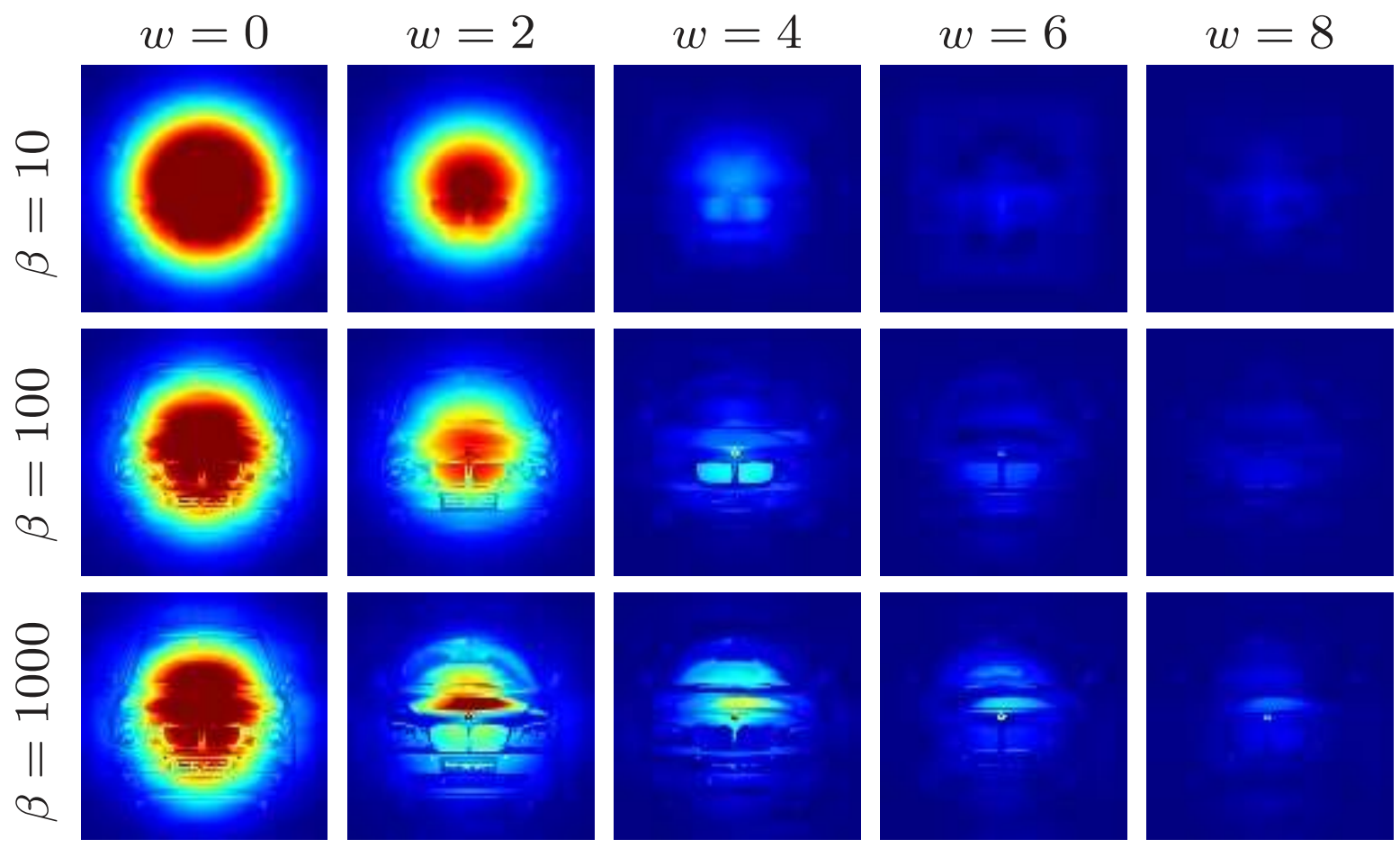

**Figure 3.10: Preserving edge information.** Larger values of the parameter $\beta$ in Eq. (3.1) allow the descriptor to retain edge information. Each row depicts the DaLI descriptor at frequencies $w = \{0, 2, 4, 6, 8\}$ for a different value of $\beta$ computed on the car image from Fig. 3.5. Observe that for low values of $\beta$ there is blurring on the higher frequencies of the descriptor.

Note that the computational cost of the eigendecomposition is cubic in the size of $\mathbf{M}$, i.e., $\mathcal{O}(n_v^3)$. As discussed in the previous subsection, we mitigate this cost by choosing mesh topologies where the number of vertices is reduced. In addition, since the eigenvectors $\boldsymbol{\phi}_i$ with smallest eigenvalues have the most importance when calculating the HKS from Eq. (3.3), we can approximate the actual value by only using a subset formed by the $n_\lambda$ eigenvectors with smallest eigenvalues. Both these strategies allow the HKS calculation to be tractable in terms of memory and computation time.

Finally, Table 3.2 summarizes all the parameters that control the shape and size of the DaLI descriptor. The way we set their default values, shown between the parentheses, will be discussed in Section 3.2.

### Deformation and Varying Illumination Dataset

In order to properly evaluate the deformation and illumination invariant properties of the DaLI descriptor and compare it against other state-of-the-art descriptors, we have collected and manually annotated a new dataset of deformable objects under varying illumination conditions. The dataset consists of twelve objects of different materials with four deformation levels and four illumination conditions each, for a total of 192 unique images. All images have a resolution of $640 \times 480$ pixels and are grayscale.

The types of objects in the dataset are 4 shirts, 4 newspapers, 2 bags, 1 pillowcase and 1 backpack. They were chosen in order to evaluate all methods against as many different types of deformation as possible. The objects can be seen in the top of Fig. 3.11.

| Symbol | Parameter Description. (Default Value) |
|---|---|
| $S$ | Outer radius of the annulus. (20) |
| $S_o$ | Inner radius of the annulus. (10) |
| $\beta$ | Magnitude of the embedding. (500) |
| $\sigma$ | Standard deviation of Gaussian weighting. ($\frac{S}{2}$) |
| $n_\lambda$ | # of eigenvectors of the Laplace-Beltrami operator. (100) |
| $n_t$ | # of intervals in the temporal domain. (100) |
| $w_{max}$ | # of frequency components used. (10) |
| $\theta_i$ | Rotation angles for descriptor comparison. ($\{-5, 0, +5\}$) |
| $n_v$ | # of mesh vertices.(1661) |
| $n_f$ | # of triangular faces in the mesh. (3204) |
| $n_{pca}$ | # of PCA components for the DaLI-PCA. (256) |

**Table 3.2: DaLI Parameters.** We show the parameters of the DaLI descriptor with the default values in parenthesis.

### Deformation and Illumination Conditions

The pipeline to acquire the images of each object consisted of, while keeping the deformation constant, changing the illumination before proceeding to the next deformation level. All images were taken in laboratory conditions in order to fully control the settings for a suitable evaluation.

The reference image was acquired from an initial configuration where the object was straightened out as much as possible. While deformations are fairly subjective, as they were done incrementally over the previous deformation level, they are representative of increasing levels of deformation. Different deformation levels of an object with the same illumination conditions are shown in the middle-left of Fig. 3.11.

The illumination changes were produced by using two high power focus lamps. The first one was placed vertically over the object, at a sufficient distance to guarantee a uniform global illumination of the object's surface. The second lamp was placed at a small elevation angle and close to the object, in order to produce harsh shadows and local illumination artifacts. By alternating the states of these lamps, four different illumination levels are achieved: no illumination, global illumination, global with local illumination, and local illumination. The different illumination conditions for constant deformation levels can be seen in the bottom-left of Fig. 3.11. Note that even with moderate deformations, the presence of the local illumination causes severe appearance changes.

### Manual Annotations

To build the ground truth annotations, we initially detected interest points in all images using a multi-scale Difference of Gaussians filter (Lowe, 2004). This yielded ap-proximately between $500 - 600$ feature points per image, each consisting of a 2D image coordinate and its associated scale.

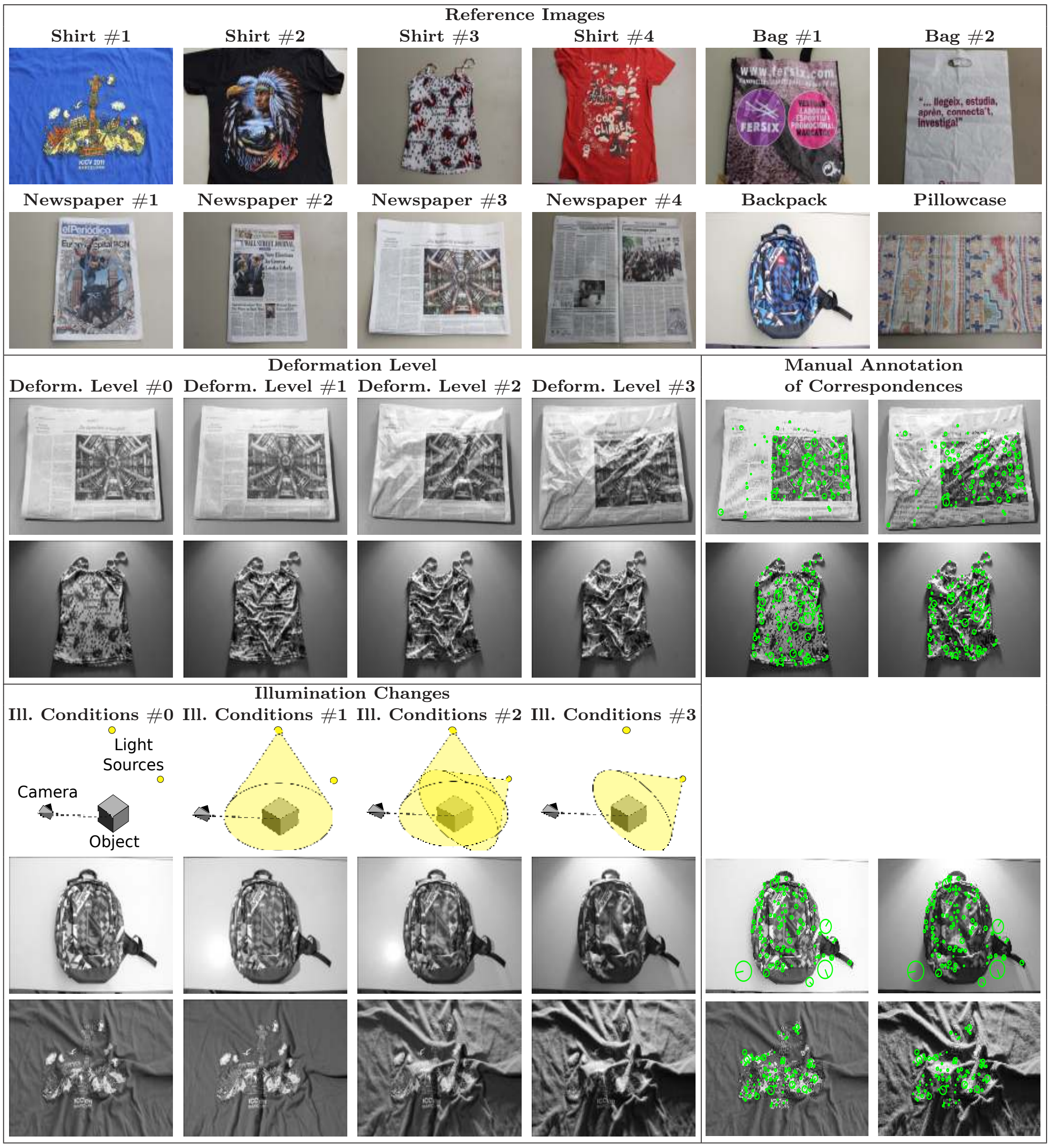

**Figure 3.11: Deformable and varying illumination dataset.** *Top:* Reference images of the twelve objects in the dataset. Each object has four deformation levels and four illumination levels yielding a total of 16 unique images per object. *Middle-left:* Sample series of images with increasing deformation levels, and constant illumination. *Bottom-left:* Sample images of the different illumination conditions taken for a deformation level of each object. The illumination conditions #0, #1, #2 and #3 correspond to no illumination, global illumination, global+local illumination, and local illumination, respectively. *Middle-right and bottom-right:* Examples of feature points matched across image pairs. The first column corresponds to the reference image for the object. These feature points are detected using Differences of Gaussians (DoG) and are matched by manual annotation. Each feature point consists of image coordinates, scale coordinates and orientation.

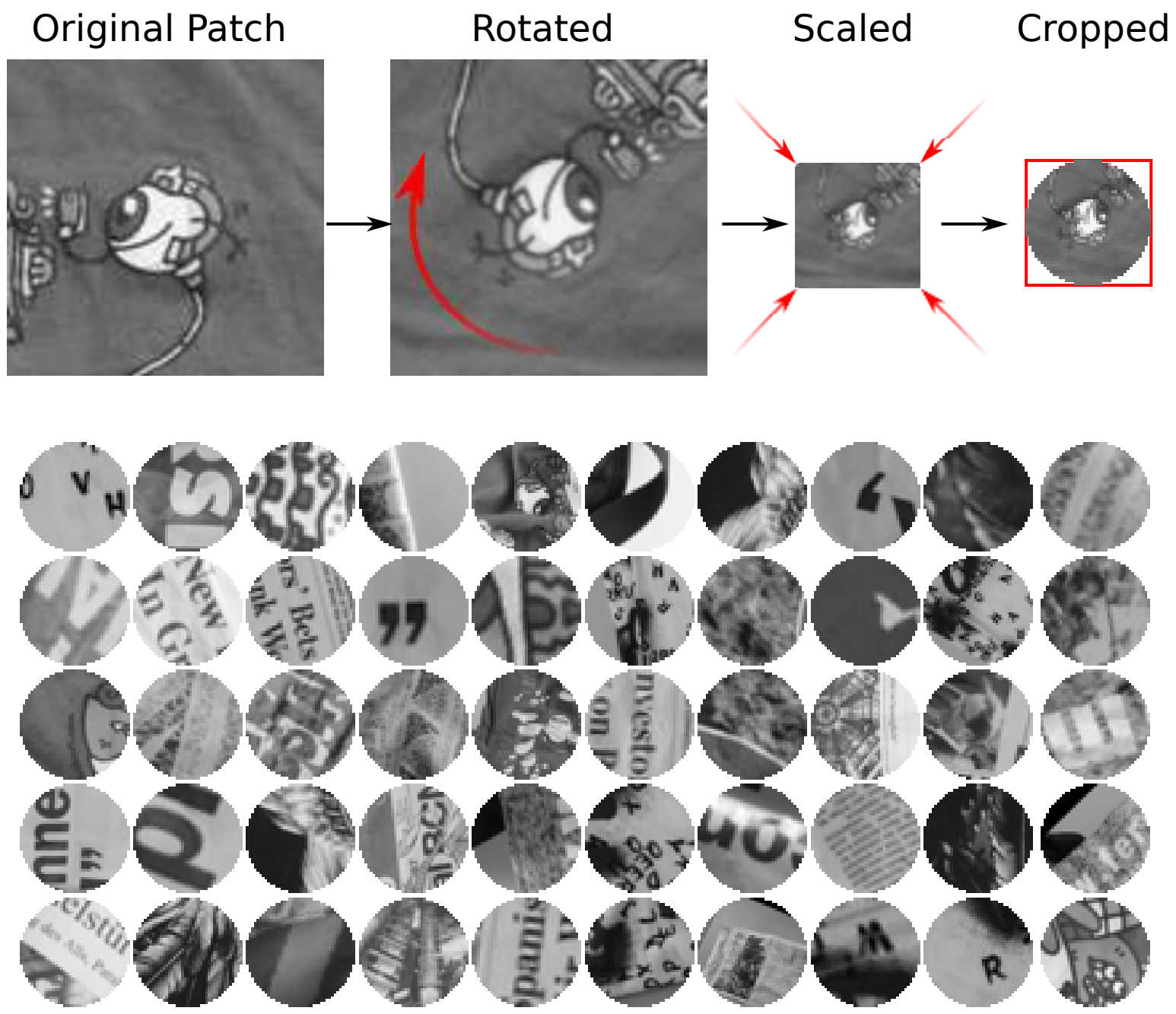

**Figure 3.12: Extracting patches from points of interest.** *Top:* Outline of the process used to obtain patches for evaluating image descriptors. For each feature point, we initially extract a square patch centered on the feature, and whose size is proportional to the scale factor of the interest point. The patch is then rotated according to the orientation of the feature point, and finally scaled to a constant size and cropped to be in a circular shape. *Bottom:* Sample patches from the dataset, already rotated and scaled to a constant size in order to make them rotation and scale invariant.

These feature points were then manually matched for each deformation level against the undeformed reference image, resulting in three pairs of matched feature points. All matches were done with top-light illumination conditions (Ill. Conditions #1, Fig. 3.11) to facilitate the annotation task and maximize the number of repeated features between each pair of images. The matching process yielded between 100 and 200 point correspondences for each pair of reference and deformed images. The same feature points are used for all illumination conditions for each deformation level. The middle-right images of Fig. 3.11 show a few samples of our annotation. Note that the matched points are generally not near the borders of the image to avoid having to clip when extracting image patches.

As we will discuss in the experimental section, in this work we seek to compare the robustness of the DaLI and other descriptors to only deformation and light changes. Yet, although the objects in the dataset are not globally rotated, the deformations do produce local rotations. In order to compensate for this we use the SIFT descriptor as done in (Mikolajczyk and Schmid, 2005) to compute the orientation of each feature point, and align all corresponding features. When a feature point has more than one dominant orientation, we consider each of them to augment the set of correspondences.

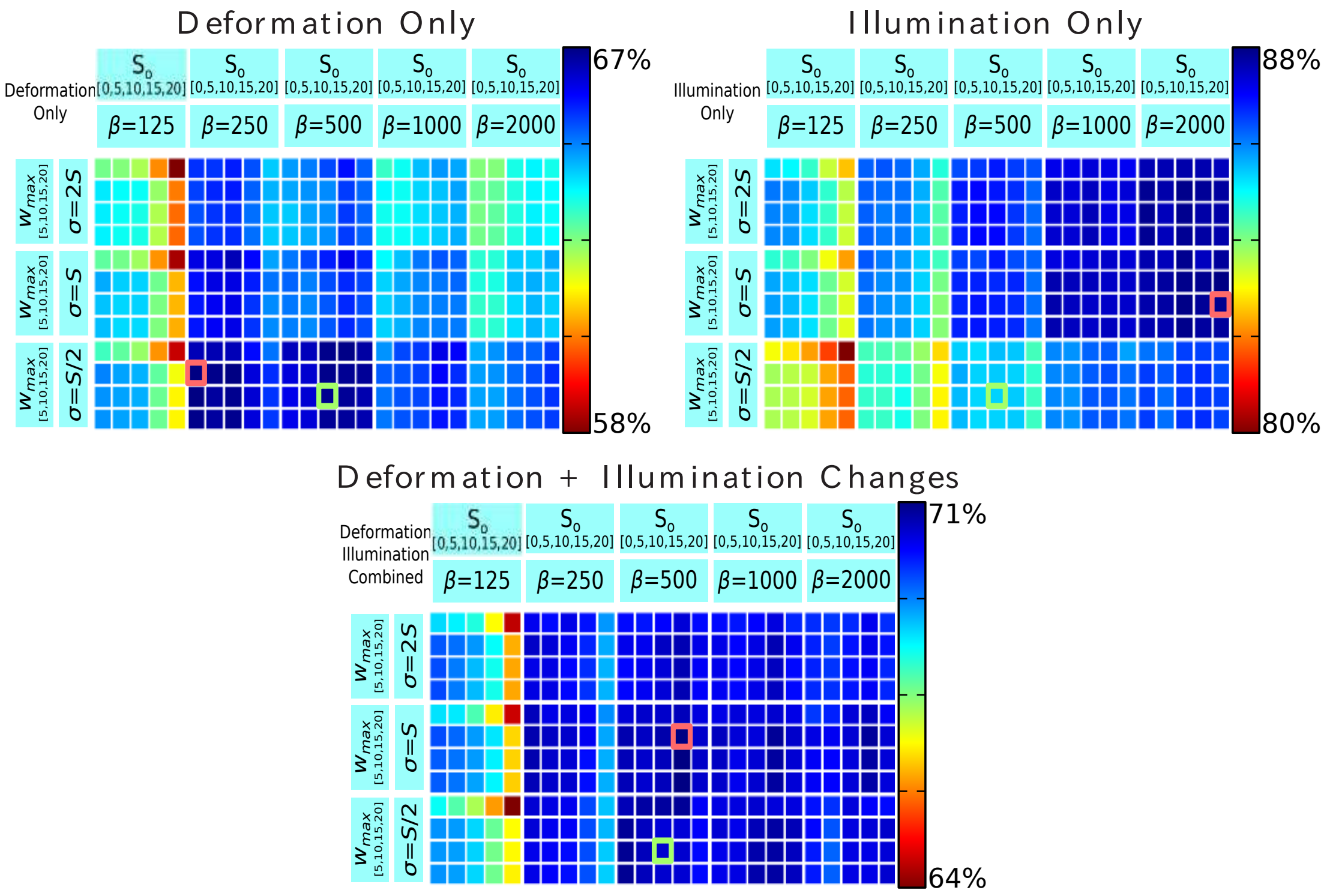

**Figure 3.13: DaLI performance for different values of the parameters $S_o$, $\beta$, $\sigma$ and $w_{max}$.** We compute and average the matching rate for the Shirt #1 and the Newspaper #1 objects in the dataset using $S_o \in \{0, 5, 10, 15, 20\}$ pixels, $\beta \in \{125, 250, 500, 1000, 2000\}$, $\sigma \in \{\frac{S}{2}, S, 2S\}$ and $w_{max} \in \{5, 10, 15, 20\}$ for three scenarios: both deformation and illumination changes, only deformation changes, and only illumination changes. The graphs depict the results of this 4D parameter exploration, where the colour of each square represents the percentage of correctly matched points for a specific combination of the parameters. In order to visualize the differences, we scale the values separately for each scenario. The best parameters for each scenario are marked in red and can be seen to vary greatly amongst themselves. We use a compromise, and for all the experiments in this section we set these parameters (highlighted in green) to $\beta = 500$, $S_o = 10$, $\sigma = \frac{S}{2}$ and $w_{max} = 10$.

### Evaluation Criteria

In order to perform fair comparisons, we have developed a framework to evaluate local image descriptors on even grounds. This is done by converting each feature point into a small image patch which is then used to compute descriptors. This allows the evaluation of the exact same set of patches for different descriptors.

For each feature point we initially extract a square patch around it, with a size proportional to the feature point's scale. In the results section we will discuss the value of the proportionality constant we use. The patch is then rotated by the feature point's orientation using bilinear interpolation, and scaled to a constant size, which we have set to $41 \times 41$ pixels as done in (Mikolajczyk and Schmid, 2005). Finally, the patch is cropped to a circular shape. This results in a scale and rotation invariant circular image patch with a diameter of 41 pixels. The steps for extracting the patches are outlined in the top of Fig. 3.12, and the bottom of the figure shows a few examples of patches from the dataset.

Given these "normalized patches" we then assess the performance of the descriptors as follows. For each pair of reference/deformed images, we extract the descriptors of all feature points in both images. We then compute the $L_2$ distance between all descriptors from the reference and the deformed image. This gives a distance matrix, which is rectangular instead of square due to the creation of additional feature points when there are multiple dominant orientations. Patches that have different orientations but share the same location are treated as a unique patch. As evaluation metric we use a descriptor-independent detection rate, which is defined for the $n$ top matches as:

$$\text{Detection Rate}(n) = \frac{100 \cdot N_c(n)}{N} \,, \tag{3.6}$$

where $N_c(n)$ is the number of feature points from the reference image that have the correct match among the top $n$ candidates in the deformed image, and $N$ is the total number of feature points in the reference image.

For the experimental results we will discuss in the following section, we consider three different evaluation scenarios: deformation and illumination, only deformation, and only illumination. In the first case we compare all combinations of deformation and illumination with respect to the reference image which has no additional illumination (ill. conditions #0) and no deformation (deform. level #0). This represents a total of 15 comparisons for each object. In the second case we consider only varying levels of deformation for each illumination condition, which yields 12 different comparisons per object (three comparisons per illumination level). When only considering illumination, each deformation level is compared to all illumination conditions. Again, this gives rise to 12 comparisons per object (three comparisons per deformation level).

## Results

We next present the experimental results, in which we discuss the following main issues: an optimization of the descriptor parameters, a PCA-based strategy for compressing the descriptor representation, and the actual comparison of DaLI against other state-of-the-art descriptors, for matching points of interest in the proposed dataset. Finally, we analyze specific aspects such as the performance of all descriptors in terms of their size, the benefits of normalizing the intensity of input images, and a real application in which the descriptors are compared when matching points of interest in real sequences of a deforming cloth and a bending paper.

### Choosing Descriptor's Parameters

We next study the influence and set the values of the DaLI parameters of Table 3.2. As the size $S_P$ of the patch is fixed, causing the descriptor radius $S$ to be also fixed, we will look at finding the appropriate value of other parameters, namely the magnitude $\beta$ of the embedding, the degree $\sigma$ of smoothing within the patch, the inner radius of the annulus $S_o$ and the dimensionality $w_{max}$ of the descriptor in the frequency domain. In order to find their optimal values, we used two objects in the dataset (Shirt #1 and Newspaper #1), and computed matching rates of their feature points for a wide range of values for each of these parameters.

It is worth to point out that the number of eigenvectors $n_\lambda$ of the Laplace-Beltrami operator was set to 100 in all cases. Note that this value represents a very small portion of all potential eigenvectors, in the order of two thousands (equal to the number of vertices $n_v$). Using a lesser number of them would eventually deteriorate the results, while not providing a significant gain in efficiency, and using more of them, almost did not improve the performance. Similarly, the number $n_t$ of intervals in which the temporal domain is split is set to 100. Again, this parameter had almost no influence, neither in the performance of the descriptor nor in its computation time.

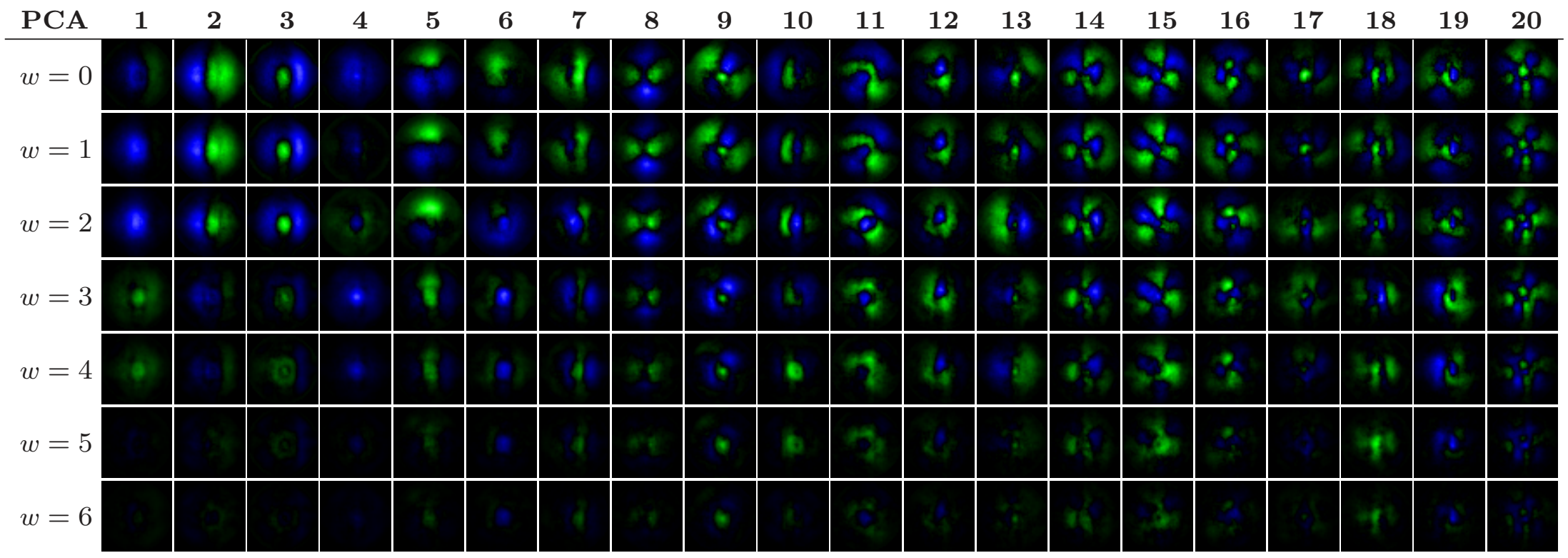

**Figure 3.14: The first 7 frequencies of the first 20 components of the PCA basis.** These are computed from images of two objects from the dataset (Shirt #1 and Newspaper #1). Each vector is normalized for visualization purposes. Positive values are displayed in green while negative values are displayed in blue. Most of the components do not contain much information at frequencies $w > 6$ and thus they are not displayed, although they are considered in the DaLI-PCA descriptor.

Figure 3.13 depicts the results of the parameter sweeping experiment. We display the rates for three scenarios: when considering both deformation and illumination changes, only deformation changes, and only illumination changes. The most influential parameters are the weighting factor $\sigma$ and to a lesser extent the magnitude of the embedding $\beta$. We see that for a wide range of parameters, the results obtained are very similar when considering both illumination and deformation, however, there is a balance to be struck between both deformation and illumination invariance. By increasing deformation invariance, illumination invariance is reduced and vice-versa. Finally we use a compromise, and the parameters we choose for all the rest of experiments are $\beta = 500$, $S_o = 10$, $\sigma = \frac{S}{2}$ and $w_{max} = 10$, besides the $n_\lambda = 100$ and $n_t = 100$ we mentioned earlier.

### Compression with PCA

The DaLI descriptor has the downside of having a very high dimensionality, as its size is proportional to the product of the number of vertices $n_v$ used to represent the patch and the number of frequency components $w_{max}$. For instance, using patches with a diameter of 41 pixels and considering the first 10 frequency slices, results in a 13,450-dimensional descriptor (1345 pixels by 10 frequency slices), requiring thus large amounts of memory and yielding slow comparisons. However, since the descriptor is largely redundant, it can be compacted using dimensionality reduction techniques such as (C. Strecha and Fua, 2012; Cai et al., 2011; Philbin et al., 2010).

As a simple proof of concept, we have used Principal Component Analysis for performing such compression. The PCA covariance matrix is estimated on 10,436 DaLI descriptors extracted from images of the Shirt #1 and Newspaper #1. The $n_{pca} \ll n_v \cdot w_{max}$ largest eigenvectors are then used for compressing an incoming full-size DaLI descriptor. The resulting compacted descriptor, which we call DaLI-PCA, can be efficiently compared with other descriptors using the Euclidean distance. Fig. 3.14 shows the first 7 frequencies of the first 20 vectors of the PCA-basis. It is interesting to note that most of the information can be seen to be in the lower frequencies. This can be considered an experimental justification for the frequency cut off applied with the $w_{max}$ parameter, which we have previously set to 10.

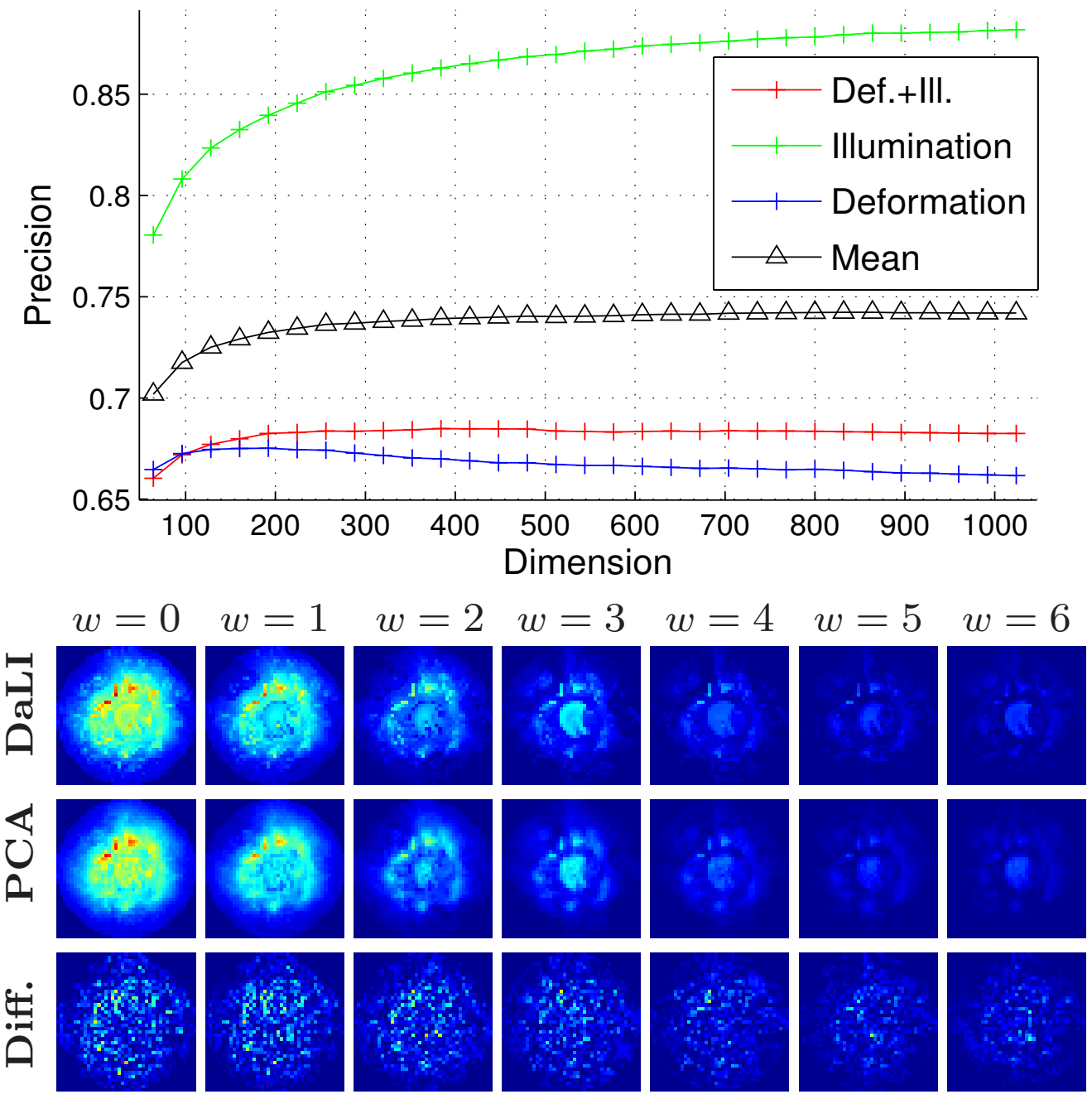

**Figure 3.15: PCA approximation of the DaLI descriptor.** *Top:* DaLI-PCA performance for different compression levels. Note that the overall mean precision does not vary much for $n_{pca} > 256$ components. *Bottom:* Comparison of an original DaLI descriptor with its compressed DaLI-PCA version obtained using 256 PCA components. For visualization purposes the values are normalized and the difference shown in the third row is scaled by $5\times$.

In order to choose the appropriate dimension $n_{pca}$ of the PCA-basis, we have used our dataset to evaluate the matching rate of DaLI-PCA descriptors for different compression levels. The results are summarized in Fig. 3.15-top, and show that using fewer dimensions favors deformation invariance (actually, PCA can be understood as a smoothing that undoes some of the harm of deformations) while using more dimensions favors illumination invariance. The response to joint deformation and illumination changes does not improve after using between $200 - 300$ components, and this has been the criterion we used to set $n_{pca} = 256$ for the rest of the experiments in this section. In Fig. 3.15-bottom we compare the frequency slices for an arbitrary DaLI descriptor and its approximation with 256 PCA-modes. Observe that the differences are almost negligible.

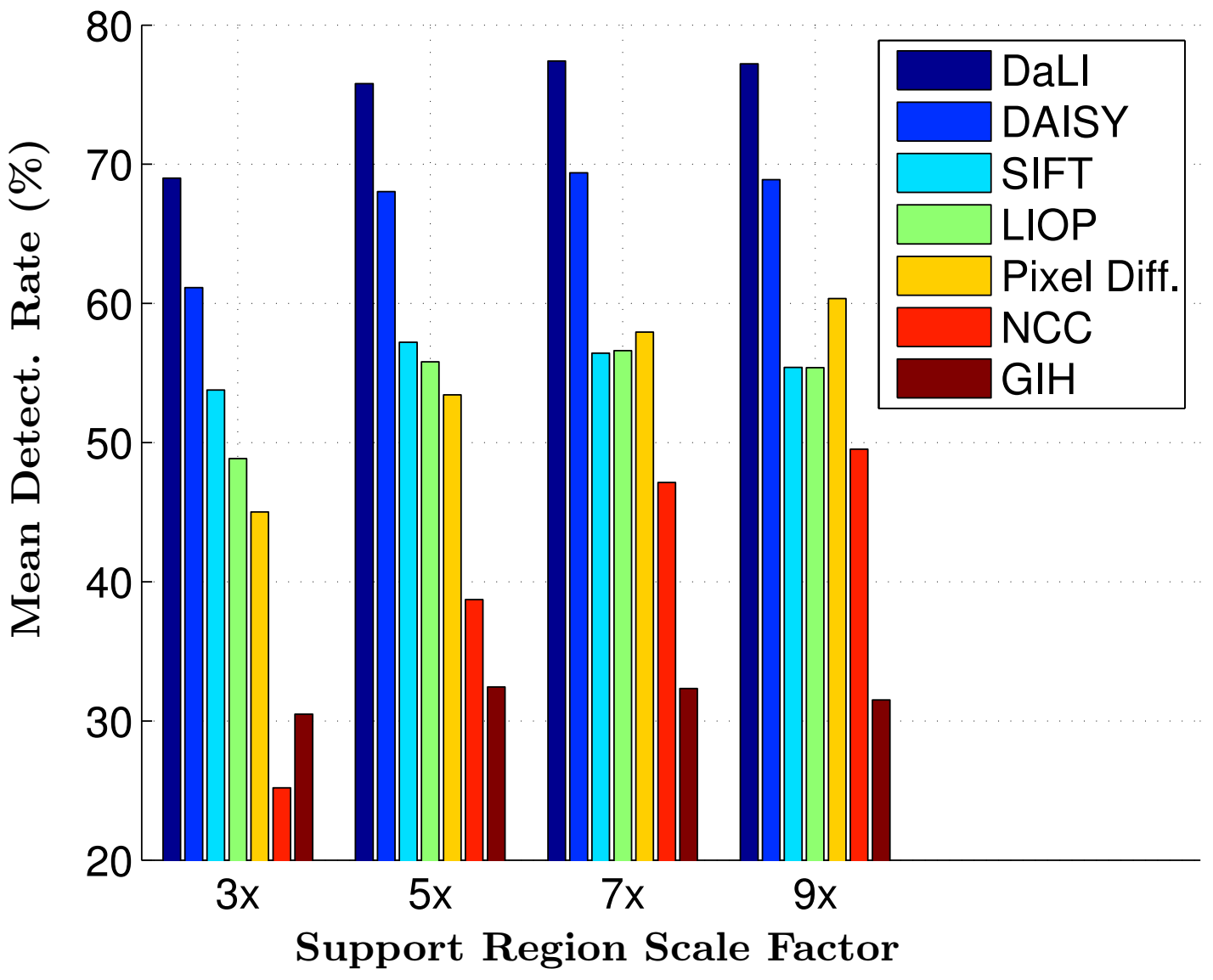

**Figure 3.16: Mean detection rates obtained by scaling regions of interest with different factors.** While a 3× scale factor does lower the overall performance, the difference between a 5×, 7× or 9× scale factor is minimum for descriptors other than weighted pixel differences (Pix. Diff.) or normalized cross covariance (NCC), which do improve as interest regions increase in size. The results of the graph correspond to the average of the mean detection rates with Deformation+Illumination changes, Illumination-only changes and Deformation-only changes.

### Comparison with Other Approaches

We compare the performance of our descriptors (both DaLI and DaLI-PCA) to that of SIFT (Lowe, 2004), DAISY (Tola et al., 2010), LIOP (Wang et al., 2011), GIH (Ling and Jacobs, 2005), Normalized Cross Correlation (NCC) and Gaussian-weighted Pixel Difference. SIFT and DAISY are both descriptors based on Differences of Gaussians (DoG) and spatial binning which have been shown to be robust to affine deformations and to certain amount of illumination changes. LIOP is a recently proposed descriptor based on intensity ordering making it fully invariant to monotonic illumination changes. GIH is a descriptor specifically designed to handle non-rigid image deformations, but as pointed out previously, it assumes these deformations are the result of changing the position of the pixels within the image and not their intensity. NCC is a standard region-based metric known to possess illumination-invariant properties. Finally, we compare against a Gaussian-weighted pixel difference using the same convolution scheme as used for the DaLI descriptor. Standard parameters suggested in the original papers are used for all descriptors except for the LIOP descriptor in which using a larger number of neighboring sample points (8 instead of 4 neighbors) results in a higher performance at the cost of a larger descriptor (241,920 instead of 144 dimensions). The LIOP and SIFT implementations are provided by VLfeat (Vedaldi and Fulkerson, 2008). We use the authors' implementation of DAISY and GIH.

The evaluation is done on the dataset presented in Section 3.2. All the descriptors are therefore tested on exactly the same image patches in order to exclusively judge the capacity of local feature representation. Yet, as mentioned in Sec. 3.2, the dataset still requires setting the scale factor to use for the points of interest. This value corresponds to the relative size of each image patch with respect to the scale value obtained from the DoG feature point detector. For this purpose, we evaluated the response of all descriptors for scale factors of 3×, 5×, 7× and 9×. The results are shown in Fig. 3.16. Although the SIFT implementation uses a default value of 3×, we have observed that the performance of all descriptors improves by increasing the patch size. Note that this does not result in a higher computational cost, as the final size of the patch is normalized to a circular shape with a diameter of 41 pixels. The maximum global response for all descriptors is achieved when using a $7\times$ scale factor, which is the value we use for all the experiments reported below.

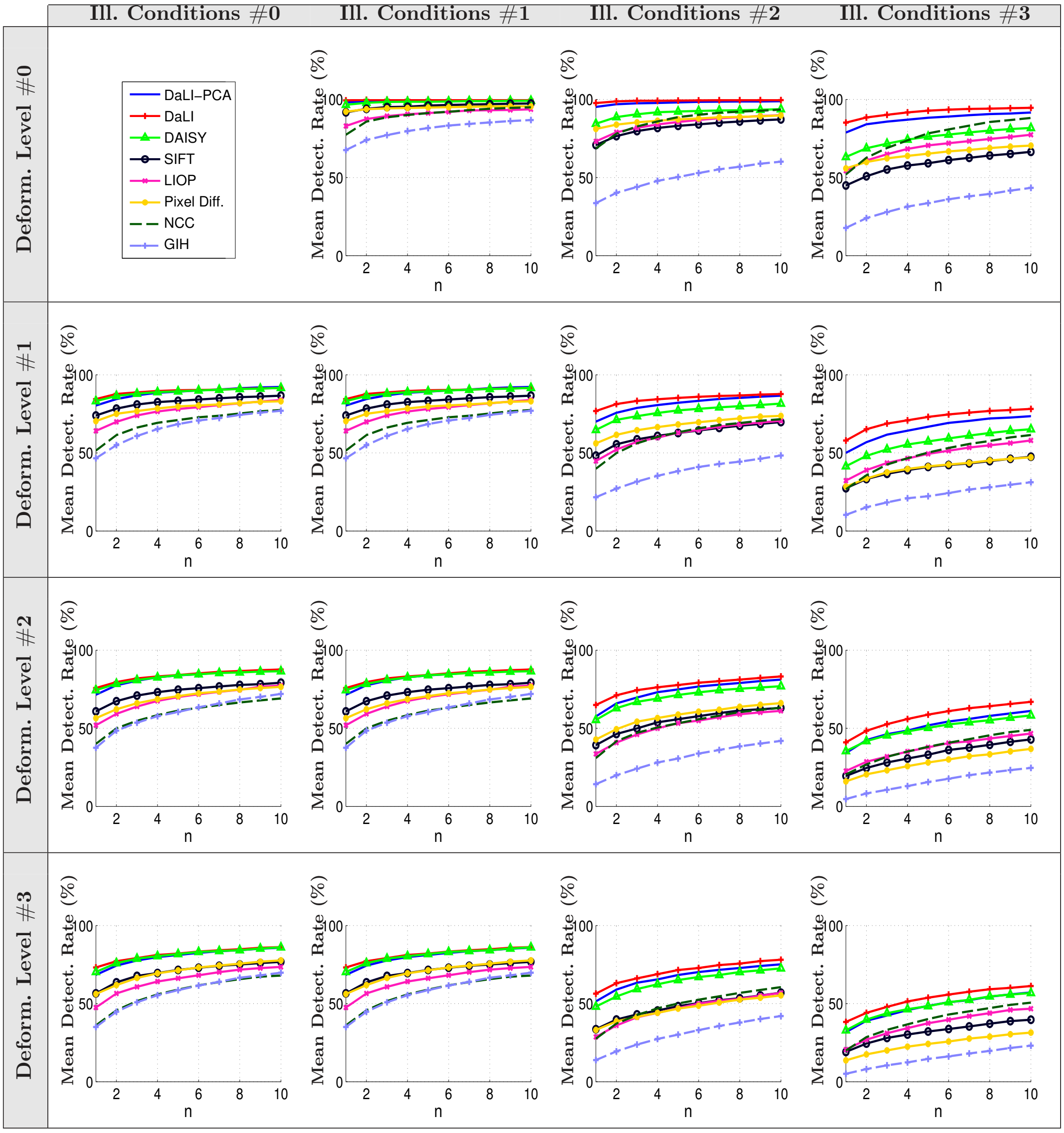

**Figure 3.17: Detection rate when simultaneously varying deformation level and illumination conditions.** Each graph represents the average of the mean detection rate between the reference image (ill. conditions #0 and deform. level #0) and all images in the dataset under specific light and deformation conditions.

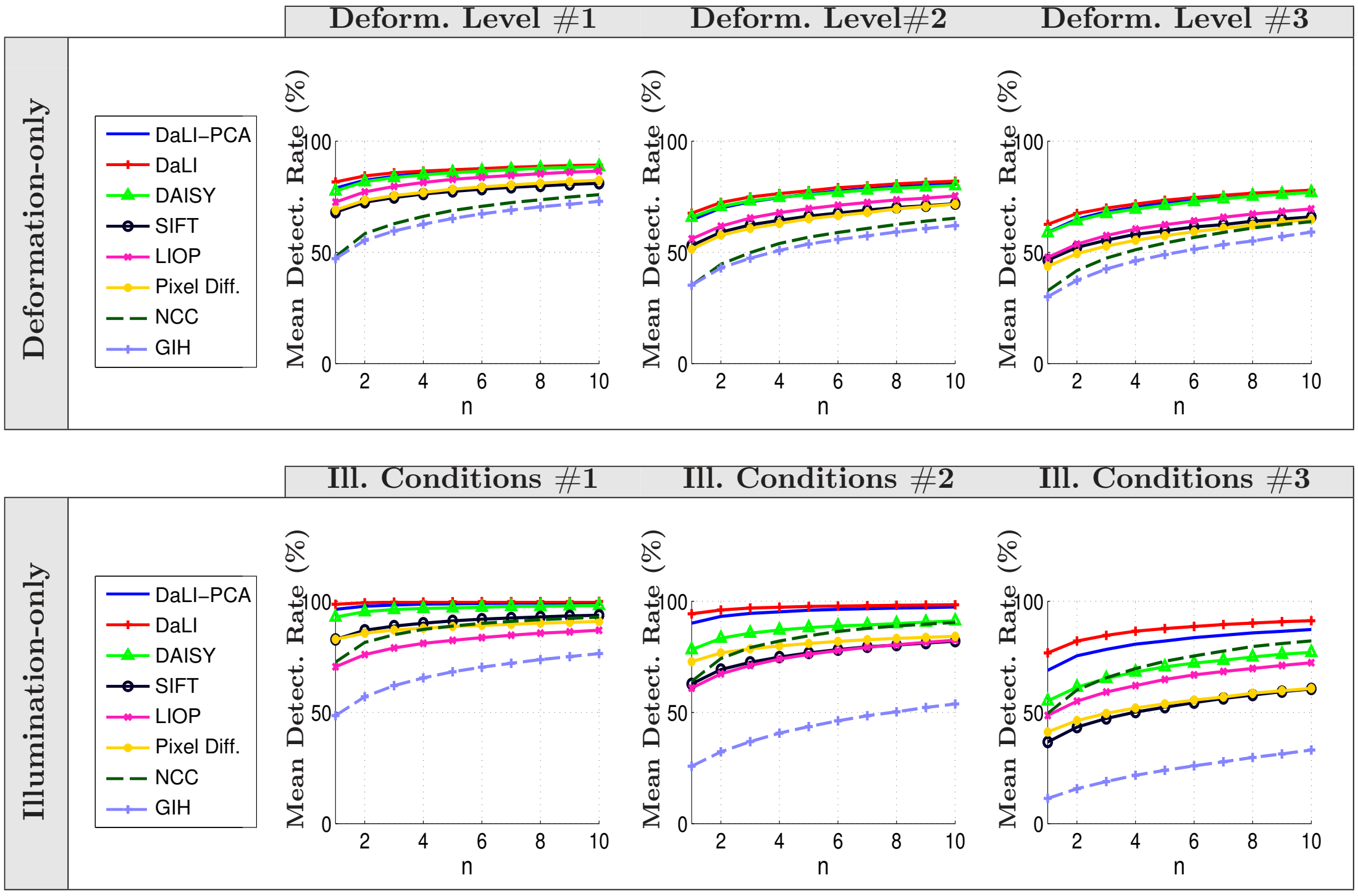

**Figure 3.18: DaLI descriptor results for varying only deformation or illumination.** *Top:* Results when varying only the deformation while keeping the illumination conditions constant. It can be seen that both DaLI and DAISY largely outperform the rest of descriptors. *Bottom:* Results of varying only the illumination conditions while keeping the deformation level constant. Note that only DaLI remains robust to illumination changes. The performance of DAISY falls roughly a 20% compared to DaLI.

The results for concurrent deformation and illumination are summarized in Fig. 3.17. DaLI consistently outperforms all other descriptors, although the more favorable results are obtained under large illumination changes. The performance of DAISY is very similar to that of DaLI when images are not affected by illumination artifacts. In this situation, the detection rates of DAISY are approximately between $2-5\%$ below to those obtained by DaLI. However, when illumination artifacts become more severe, the performance of DAISY rapidly drops, yielding detection rates which are more than 20% below DaLI. SIFT, LIOP, and Pixel Difference yield similar results, with SIFT being better at weak illumination changes and LIOP better at handling strong illumination changes. Yet, these three methods are one step behind DaLI and DAISY. NCC generally performs worse except in situations with large illumination changes, where it even outperforms DAISY. On the other hand, GIH performs quite poorly even when no light changes are considered. This reveals another limitation of this approach, in that it assumes the effect of deformations is to locally change the position of image pixels, while in real deformations some of the pixels may disappear due to occlusions. Although our approach does not explicitly address occlusions, we can partially handle them by weighing the contribution of the pixels within each patch, by a function decreasing with the distance to the center. Thus, most of the information of our descriptor is concentrated in a small region surrounding the point of interest, hence making it less sensitive to occlusions. The results also show that the compressed DaLI-PCA follows a similar pattern as DaLI, and specially outperforms DAISY under severe illumination conditions.

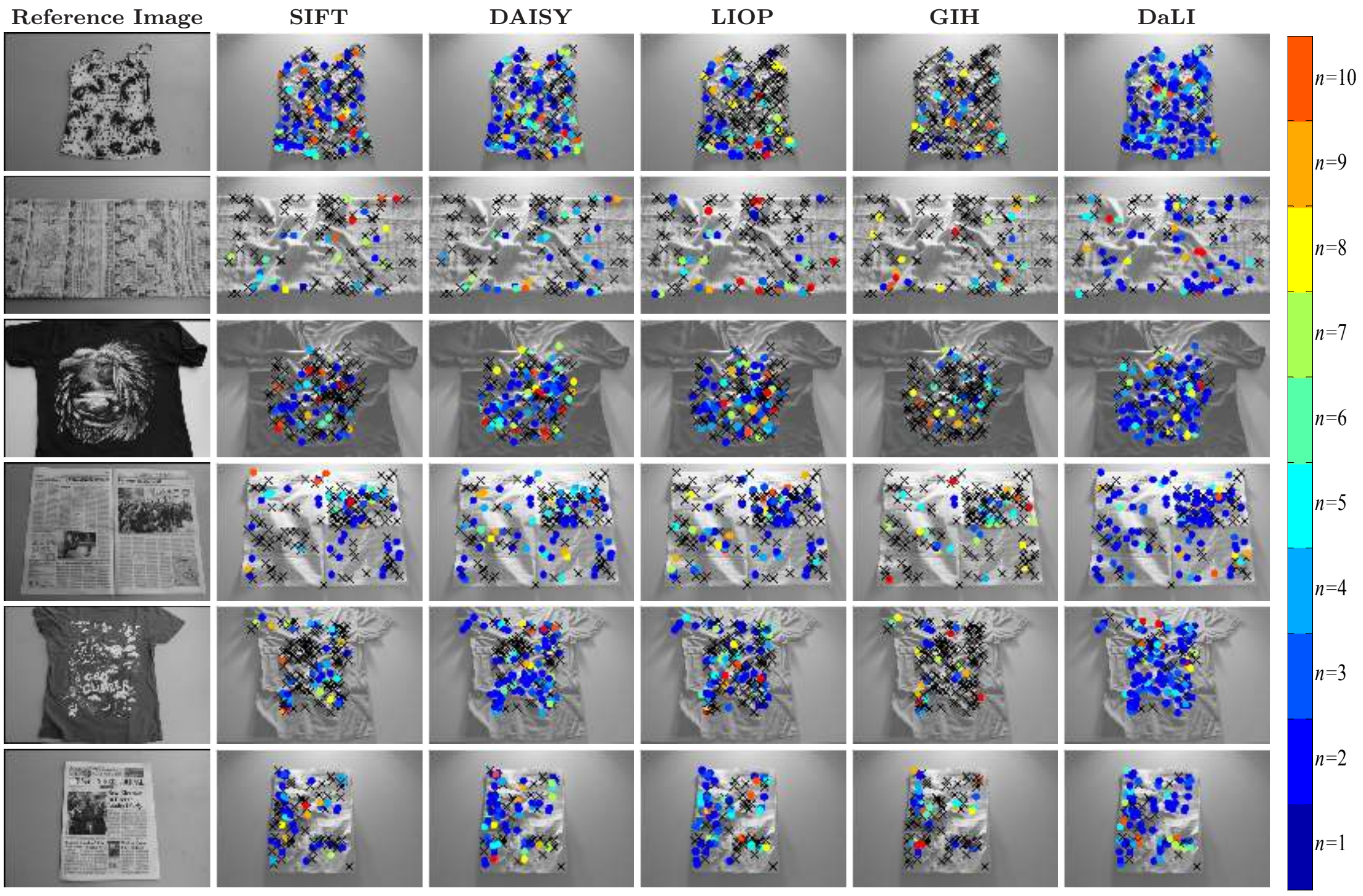

**Figure 3.19: Sample results from the dataset.** As in Fig. 3.2, the colour of the circles indicates the position $n$ of the correct match among the top candidates. If $n > 10$ we consider the point as unmatched and mark it with a cross.

In Fig. 3.18 we give stronger support to our arguments by independently evaluating deformations and illumination changes. These graphs confirm that under deformation-only changes, DaLI outperforms DaLI-PCA and DAISY by a small margin of roughly 3%. Next, SIFT, LIOP, and Pixel Difference yield similar results, roughly 20% below DaLI in absolute terms. GIH and NCC yield also similar results, although their performance is generally very poor. When only illumination changes are considered, both DaLI and DaLI-PCA significantly outperform other descriptors, by a margin larger than 20% when dealing with complex illumination artifacts. The only notable difference in this scenario is that the NCC descriptor outperforms SIFT and Pixel Difference. As GIH is not invariant to illumination changes, it obtains poor results. Similarly, since LIOP is designed to be invariant to monotonic lighting changes, it does not perform that well in real images that undergo complex illumination artifacts.

In summary, the experiments have shown that DaLI globally obtains the best performance. Its best relative response when compared with other descriptors is obtained when the deformations are mild and the light changes drastic. Some sample results on particular images taken from the dataset can be seen in Fig. 3.19. Additionally, numeric results for the best candidate ($n = 1$ in Eq. 3.6) under different conditions for all descriptors are shown in Table 3.3.

| Descriptor | Deformation | Illumination | Deformation+Illumination |
|---|---|---|---|
| DaLI-PCA | 67.425 | 85.122 | 68.368 |
| **DaLI** | **70.577** | **89.895** | **72.912** |
| DAISY | 67.373 | 75.402 | 66.197 |
| SIFT | 55.822 | 60.760 | 53.431 |
| LIOP | 58.763 | 60.014 | 52.176 |
| Pixel Diff. | 54.714 | 65.610 | 54.382 |
| NCC | 38.643 | 62.042 | 41.998 |
| GIH | 37.459 | 28.556 | 31.230 |

**Table 3.3: Evaluation results on the dataset for all descriptors.** Results are obtained by averaging the first match percentage values over all images being tested under all different conditions.

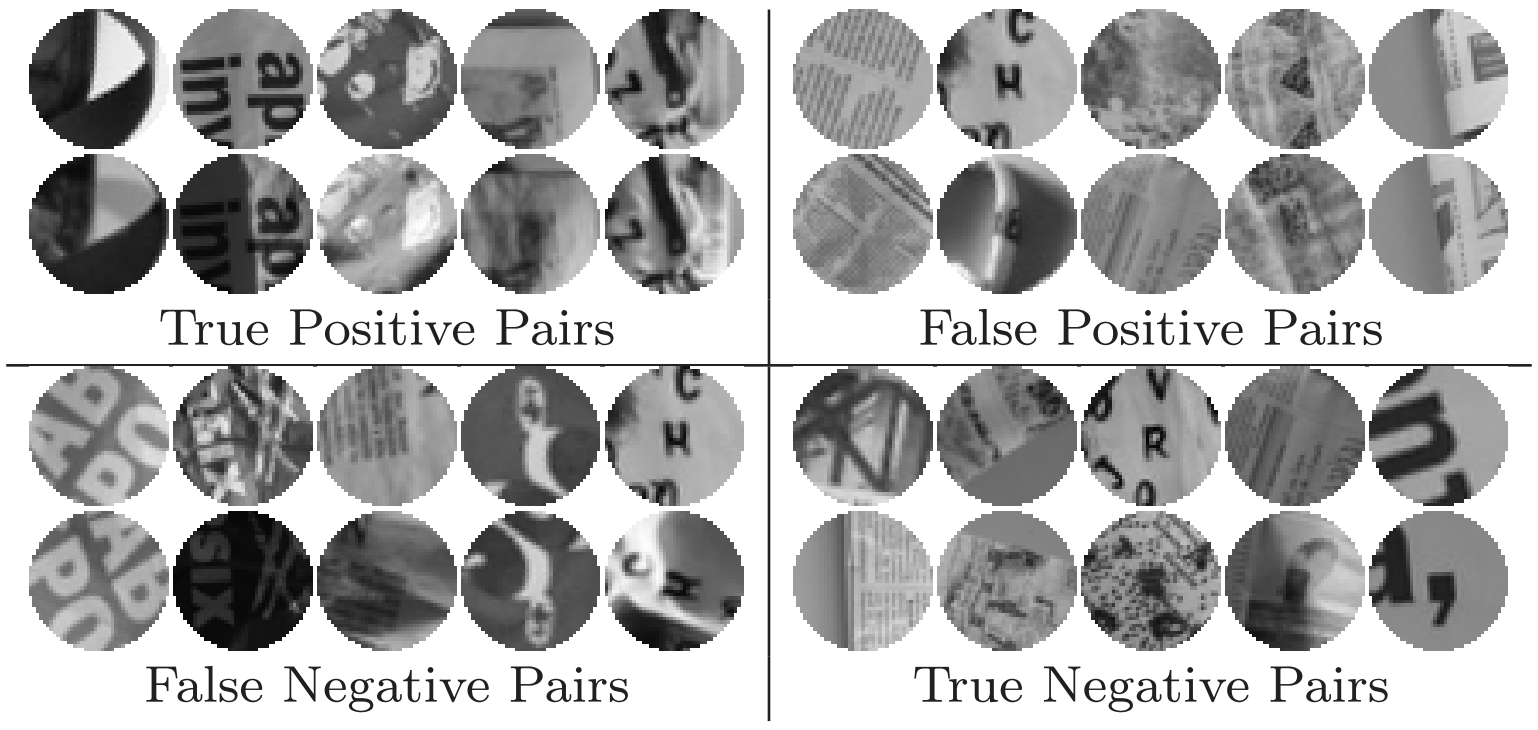

**Figure 3.20: True positive, false positive, false negative and true negative image patch pairs.** These were obtained using the DaLI descriptor on the dataset. Note that most of the false negatives are due to large orientation changes across feature points.

Finally, examples of particular patch matches are depicted in Fig. 3.20. The true positives pairs can be seen to be matched despite large changes. On the other hand, the false negatives seem largely generated by differences in orientations of the feature points: they correspond to the same patch, only rotated. The false positives share some similarity, although they are mainly from heavily deformed images.

### Descriptor Size Performance

Since larger descriptors may a priori have an unfair advantage, we next provide results of an additional experiment in which we compare descriptors having similar sizes. The LIOP we calculate in this case uses 4 neighbours instead of the 8 neighbours we considered before, which results in a smaller size, although also in a lower performance. GIH is originally 176-dimensional, thus the results are the same as in Table 3.3. NCC and Pixel Diff, are not considered for this experiment as their size is $41 \times 41 = 1681$.

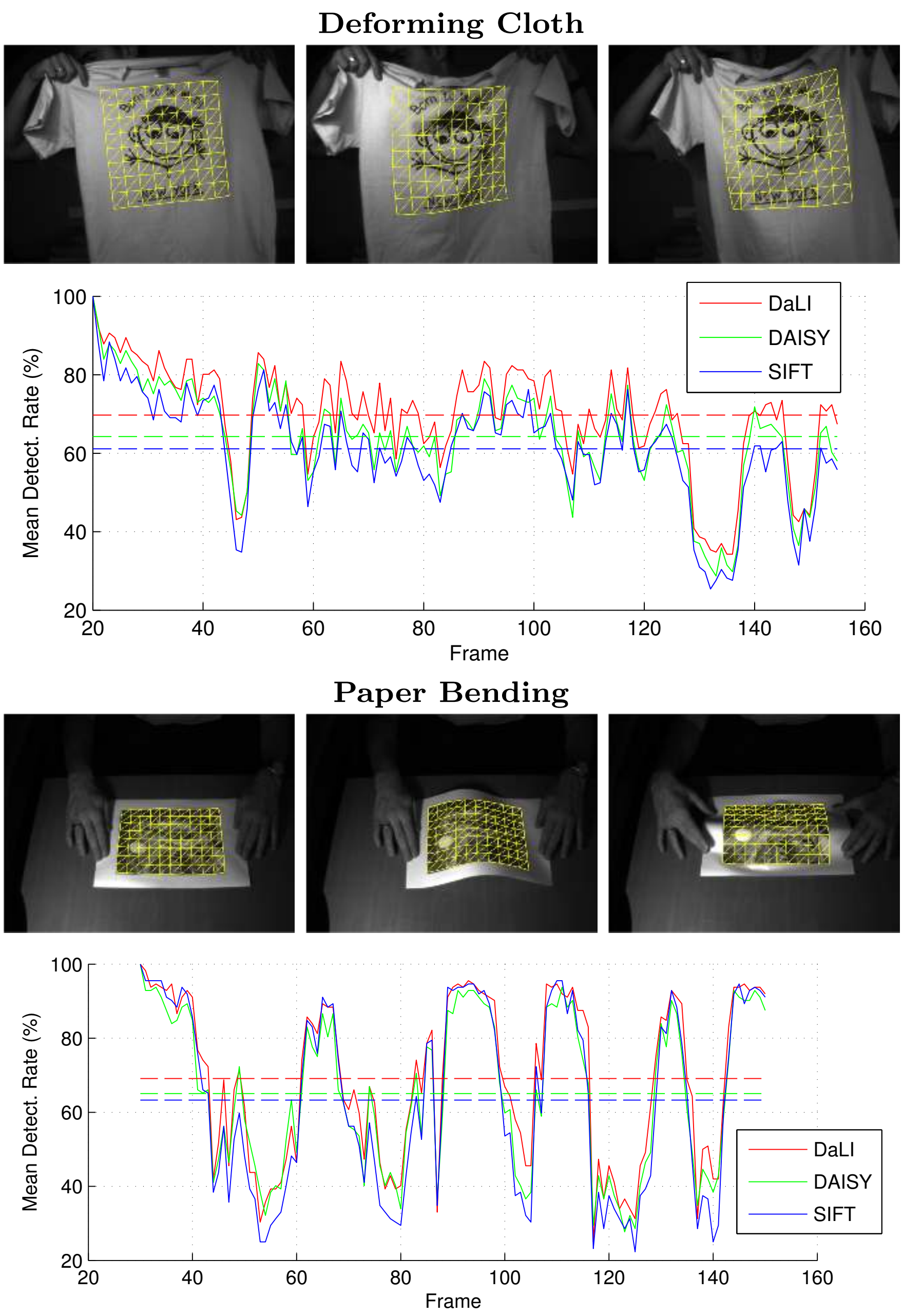

**Figure 3.21: Mean detection accuracy on two real world videos from (Moreno-Noguer and Fua, 2013).** For each sequence we show three example frames from each video in the top row. In the bottom row we plot the accuracy for each frame for three descriptors: DaLI, DAISY and SIFT. Additionally the mean for each descriptor is displayed as a dashed line.

Results are shown in Table 3.4. We can see that the 128-dimensional DaLI-PCA outperforms all other descriptors except the 256-dimensional DaLI-PCA. It is worth

| Descriptor | Size | Deformation | Illumination | Deformation+Illumination |
|---|---|---|---|---|
| DaLI-PCA | 128 | **67.45** | 82.34 | 67.71 |
| SIFT | 128 | 55.82 | 60.76 | 53.43 |
| LIOP | 144 | 54.01 | 44.89 | 44.45 |
| DAISY | 200 | 67.37 | 75.40 | 66.20 |
| GIH | 176 | 37.46 | 28.56 | 31.23 |
| **DaLI-PCA** | **256** | 67.43 | **85.12** | **68.37** |

**Table 3.4: Comparison of performance and descriptor size.**

noting the large performance gain obtained over the standard SIFT descriptor.

### Effects of Intensity Normalization

We next extend the analysis we introduced in Sect. 3.2 in which we evaluated SI-HKS and HKS with and without pre-normalizing the intensity of input images. We will also consider SIFT and DAISY, which have been the most competitive descriptors in previous experiments. Since SIFT/DAISY implementations require the pixels to be in a $[0, 1]$ range, we have normalized each image patch so that the pixels follow the distribution $\mathcal{N}(0.5, (2 \cdot 1.956)^{-1})$. This makes it so that on average 95% of the pixels will fall in $[0, 1]$. Pixels outside of this range are set to either 0 or 1.

We compare the DaLI descriptor (both its SI-HKS and HKS variants), DAISY and SIFT, with and without normalization. Results are shown in Table 3.5. We can see that for DAISY and SIFT, since they perform a final normalization stage, the results do not have any significant change. In the case of the DaLI descriptor, though, we observe that there is a rather significant performance increase when using the SI-HKS variant over the HKS one, even with patch normalization. This demonstrates again that the role of the Fourier Transforms applied in HKS to make it illumination invariant go far beyond a simple normalization. In addition, SI-HKS compresses the descriptor in the frequency domain and is one order of magnitude smaller than the HKS variant.

### Evaluation on Real World Sequences

This section describes additional experiments on two real world sequences of deforming objects, taken from (Moreno-Noguer and Fua, 2013). One consists of a T-Shirt being waved in front of a camera (Deforming Cloth) and the other consists of a piece of paper being bended in front of a camera (Paper Bending). We use points of interest computed with the Differences of Gaussians detector (DoG) and follow the same patch extraction approach as in the rest of the paper. The points of interest are calculated for the first frame in each sequence and then propagated using the provided 3D ground truth to the other frames. We use the same descriptor parameters as in the rest of the experiments, and seek to independently match the points of interest in the first frame to those of all the other frames.

| Descriptor | Normalization? | Deformation | Illumination | Deformation+ Illumination |
|---|---|---|---|---|
| **DaLI (SI-HKS)** | **No** | **70.58** | **89.90** | **72.91** |
| DaLI (SI-HKS) | Yes | 70.38 | 88.60 | 72.28 |
| DaLI (HKS) | No | 66.27 | 84.21 | 67.83 |
| DaLI (HKS) | Yes | 67.20 | 84.62 | 69.42 |
| DAISY | No | 67.37 | 75.40 | 66.20 |
| DAISY | Yes | 67.08 | 75.59 | 66.27 |
| SIFT | No | 55.82 | 60.76 | 53.43 |
| SIFT | Yes | 55.05 | 61.83 | 53.21 |

**Table 3.5: Effect of normalizing image patches for various descriptors.**

As we can observe in Fig. 3.21, DaLI outperforms both DAISY and SIFT[2] . We obtain a 5.5% improvement over DAISY on the Deforming Cloth sequence and a 4.1% improvement on the Paper Bending sequence. Note that these sequences do not have as complicated illumination artifacts as our dataset, an unfavorable situation for our descriptor. Yet, DaLI still consistently outperforms other approaches along the whole sequence.

## 3.3 Deep Architectures for Descriptors

While most descriptors use hand-crafted features (Lowe, 2004; Bay et al., 2006; Kokkinos et al., 2012; Trulls et al., 2013), including the DaLI descriptor (Simo-Serra et al., 2015b) presented in the previous section, there has recently been interest in using machine learning algorithms to learn descriptors from large databases.

In this section we draw inspiration on the recent success of Deep Convolutional Neural Networks on large-scale image classification problems (Krizhevsky et al., 2012; Szegedy et al., 2013) to build discriminative descriptors for local patches. Specifically, we propose an architecture based on a Siamese structure of two CNNs that share the parameters. We compute the $L_2$ norm on their output, i.e. the descriptors, and use a loss that enforces the norm to be small for corresponding patches and large otherwise. We demonstrate that this approach allows us to learn compact and discriminative representations.

To implement this approach we rely on the dataset of (Brown et al., 2011), which contains over 1.5M grayscale $64\times64$ image patches from different views of 500K different 3D points. With such large datasets it becomes intractable to exhaustively explore all corresponding and non-corresponding pairs. Random sampling is typically used; however, most correspondences are not useful and hinder the learning of a discriminant mapping. We address this issue with aggressive mining of "hard" positives and negatives and which proves fundamental in order to obtain discriminative learned descriptors. In particular, in some of the tests we obtain up to a 169% increase in performance with SIFT as a baseline.

[2]Again, we only compare against DAISY and SIFT, as these are the descriptors which have been more competitive in the experiments with the full dataset.

### Related Work

Recent developments in the design of local image descriptors are moving from carefully-engineered features (Lowe, 2004; Bay et al., 2006) to learning features from large volumes of data. This line of works includes unsupervised techniques based on hashing as well as supervised approaches using Linear Discriminant Analysis (Brown et al., 2011; Gong et al., 2012; C. Strecha and Fua, 2012), boosting (Trzcinski et al., 2012), and convex optimization (Simonyan et al., 2014).

We explore solutions based on deep convolutional networks (CNNs). CNNs have been used in computer vision for decades, but are currently experiencing a resurgence kickstarted by the accomplishments of (Krizhevsky et al., 2012) on large-scale image classification. The application of CNNs to the problem of descriptor learning has already been explored by some researchers (Jahrer et al., 2008; Osendorfer et al., 2013). These works are however preliminary, and many open questions remain regarding the practical application of CNNs for learning descriptors, such as the most adequate network architectures and application-dependent training schemes. We aim to provide a rigorous analysis of several of these topics. In particular, we use a Siamese network (Bromley et al., 1994) to train the models, and experiment with different network configurations inspired by the state-of-the-art in deep learning.

Additionally, we demonstrate that aggressive mining of both "hard" positive and negative matching pairs greatly enhances the learning process. Mining hard negatives is a well-known procedure in sliding-window detectors (Felzenszwalb et al., 2010), where the number of negative samples is virtually unlimited and yet most negatives are easily discriminated. Similar techniques have been applied to CNNs for object detection (Girshick et al., 2014; Szegedy et al., 2013).

### Learning Deep Descriptors

Given an intensity patch $\mathbf{x} \in \mathbb{R}^N$, the descriptor of $\mathbf{x}$ is a non-linear mapping $D(\mathbf{x})$ that is expected to be discriminative, i.e. descriptors for image patches corresponding to the same point should be similar, and dissimilar otherwise.

In the context of multiple-view geometry, descriptors are typically computed for salient points where scale and orientation can be reliably estimated, for invariance. Patches then capture local projections of 3D scenes. Let us consider that each image patch $\mathbf{x}_i$ has an index $p_i$ that uniquely identifies the 3D point which roughly projects onto the 2D patch, from a specific viewpoint. Therefore, taking the $\mathrm{L}_2$ norm as a similarity metric between descriptors, for an ideal descriptor we would wish that

$$d_D(\mathbf{x}_1, \mathbf{x}_2) = \|D(\mathbf{x}_1) - D(\mathbf{x}_2)\|_2 = \begin{cases} 0 & \text{if } p_1 = p_2 \\ \infty & \text{if } p_1 \neq p_2 \end{cases} . \tag{3.7}$$

We propose learning descriptors using a Siamese network (Bromley et al., 1994), i.e. optimizing the model for pairs of corresponding or non-corresponding patches, as shown in Fig. 3.22a. We propagate the patches through the model to extract the descriptors and compute their $\mathrm{L}_2$ norm, which is a standard similarity measure for

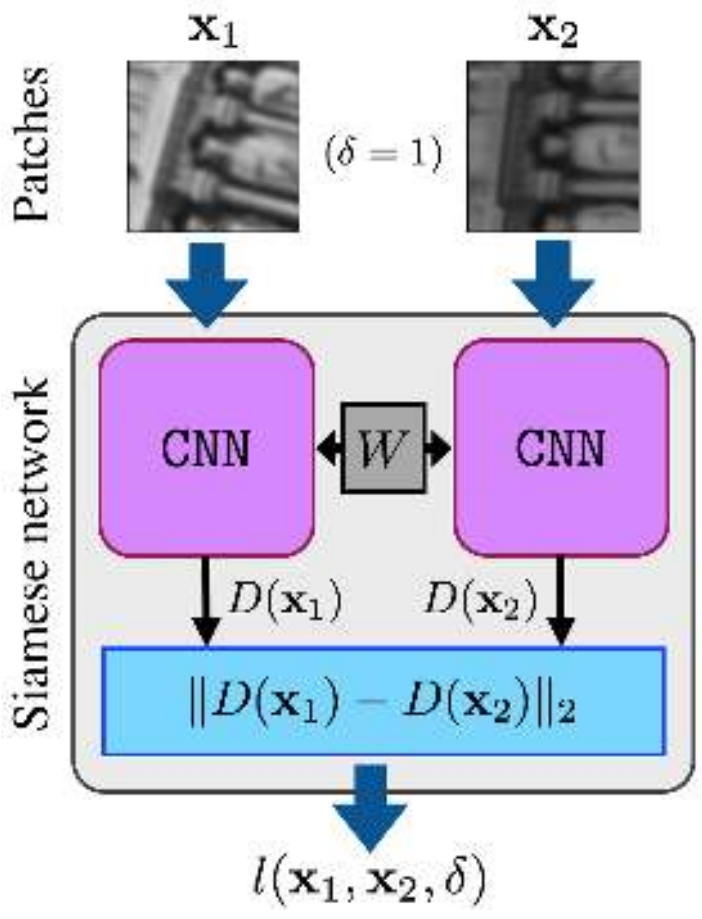

**(a)** Siamese network.

| Name | Layer 1 | Layer 2 | Layer 3 | Layer 4 |
|---|---|---|---|---|
| CNN3_NN1 | 32x7x7<br>x2 pool | 64x6x6<br>x3 pool | 128x5x5<br>x4 pool | 128<br>- |
| CNN3 | 32x7x7<br>x2 pool | 64x6x6<br>x3 pool | 128x5x5<br>x4 pool | -<br>- |
| CNN2a_NN1 | 32x5x5<br>x3 pool | 64x5x5<br>x4 pool | 128<br>- | -<br>- |
| CNN2b_NN1 | 32x9x9<br>x4 pool | 64x5x5<br>x5 pool | 128<br>- | -<br>- |
| CNN2 | 64x5x5<br>x4 pool | 128x5x5<br>x11 pool | -<br>- | -<br>- |
| CNN1_NN1 | 32x9x9<br>x14 pool | 128<br>- | -<br>- | -<br>- |

**(b)** Various convolutional neural network architectures.

**Figure 3.22: Overview of the network architectures used.** *Left:* Schematic of a Siamese network, where pairs of input patches are processed by two copies of the same CNN. *Right:* Different CNN configurations considered.

image descriptors. We then compute the loss function on this distance. Given a pair of patches $\mathbf{x}_1$ and $\mathbf{x}_2$ we define a loss function of the form

$$l(\mathbf{x}_1, \mathbf{x}_2, \delta) = \delta \cdot l_P(d_D(\mathbf{x}_1, \mathbf{x}_2)) + (1 - \delta) \cdot l_N(d_D(\mathbf{x}_1, \mathbf{x}_2)) \;, \tag{3.8}$$

where $\delta$ is the indicator function, which is 1 if $p_1 = p_2$, and 0 otherwise. $l_P$ and $l_N$ are the partial loss functions for patches corresponding to the same 3D point and to different points, respectively. When performing back-propagation, the gradients are independently accumulated for both descriptors, but jointly applied to the weights, as they are shared.

Although it would be ideal to optimize directly for Eq. (3.7), we relax it, using a margin $m$ for $l_N(\cdot)$. In particular, we consider the hinge embedding criterion (Mobahi et al., 2009)

$$l_P(d_D(\mathbf{x}_1, \mathbf{x}_2)) = d_D(\mathbf{x}_1, \mathbf{x}_2) \tag{3.9}$$

$$l_N(d_D(\mathbf{x}_1, \mathbf{x}_2)) = \max(0, m - d_D(\mathbf{x}_1, \mathbf{x}_2)) \;. \tag{3.10}$$

### Convolutional Neural Network Descriptors

When designing the structure of the CNN we are limited by the size of the input data, in our case 64×64 patches from the dataset of (Brown et al., 2011). Note that larger patches would allow us to consider deeper networks, and possibly more informative descriptors, but at the same time they would be also more susceptible to occlusions. We consider networks of up to three convolutional layers, followed by up to a single additional fully-connected layer. We target descriptors of size 128, the same as SIFT (Lowe, 2004); this value also constrains the architectures we can explore.

As usual, each convolutional layer consists four sub-layers: filter layer, non-linearity layer, pooling layer and normalization layer. Since sparser connectivity has been shown to improve performance while lowering parameters and increasing speed (Culurciello et al., 2013), except for the first layer, the filters are not densely connected to the previous layers. Instead, they are sparsely connected at random, so that the mean number of connections each input layer has is constant.

Regarding the non-linear layer, we use hyperbolic tangent (Tanh), as we found it performs better than Rectified Linear Units (ReLU). We use $L_2$ pooling for the pooling sublayers, which were shown to outperfom the more standard max pooling (Sermanet et al., 2012). Normalization has been shown to be important for deep networks (Jarrett et al., 2009) and fundamental for descriptors (Mikolajczyk and Schmid, 2005). We use subtractive normalization for a 5×5 neighbourhood with a Gaussian kernel. We will later justify these decisions empirically.

An overview of the architectures we consider is given in Fig. 3.22b. We choose a set of six networks, from 2 up to 4 layers. The architecture hyperparameters (number of layers and convolutional/pooling filter size) are chosen so that no padding is needed. We consider models with a final fully-connected layer as well as fully convolutional models, where the last sublayer is a pooling layer. Our implementation is based on Torch7 (Collobert et al., 2011).

### Stochastic Sampling Strategy and Mining

Our goal is to optimize the network parameters from an arbitrarily large set of training patches. Let us consider a dataset with $N$ patches and $M \leq N$ unique 3D patch indices, each with $n_i$ associated image patches. Then, the number of matching image patches or positives $N_P$ and the number of non-matching images patches or negatives $N_N$ in the dataset is

$$N_P = \sum_{i=1}^{M} \frac{n_i(n_i-1)}{2} \qquad \text{and} \qquad N_N = \sum_{i=1}^{M} n_i(N-n_i)\ . \tag{3.11}$$

In general both $N_P$ and $N_N$ are intractable to exhaustively iterate over. We approach the problem with random sampling. For gathering positives samples we can randomly choose a set of $B_P$ 3D point indices $\{p_1, \cdots, p_{B_P}\}$, and choose two patches with corresponding 3D point indices randomly. For negatives it is sufficient to choose $B_N$ random pairs with non-matching indices.

However, when the pool of negative samples is very large random sampling will produce many negatives with a very small loss, which do not contribute to the global loss, and thus stifle the learning process. Instead, we can iterate over non-corresponding patch pairs to search for "hard" negatives, i.e. with a high loss. In this manner it becomes feasible to train discriminative models faster while also increasing performance. This technique is commonplace in sliding-window classification.

Therefore, at each epoch we generate a set of $B_N$ randomly chosen patch pairs, and after forward-propagation through the network and computing their loss we keep only a subset of the $B_N^M$ "hardest" negatives, which are back-propagated through the network in order to update the weights. Additionally, the same procedure can be used over the positive samples, i.e. we can sample $B_P$ corresponding patch pairs and prune them down to the $B_P^M$ "hardest" positives. We show that the combination of aggressively mining positive and negative patch pairs allows us to greatly improve the discriminative capability of learned descriptors. Note that extracting descriptors with the learned models does not further require the Siamese network and does not incur the computational costs related to mining.

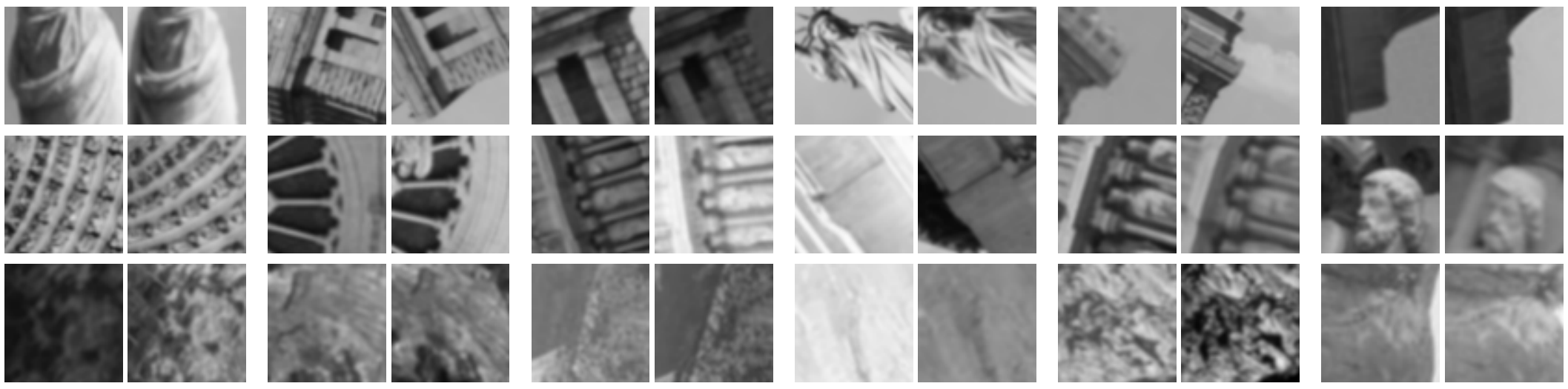

**Figure 3.23: Pairs of corresponding samples from the Multi-view Stereo Correspondence dataset.** Top row: 'Liberty' (LY). Middle row: 'Notre Dame' (ND). Bottom row: 'Yosemite' (YO).

### Learning

We normalize the dataset by subtraction of the mean of the training patches and division by their standard deviation. We then learn the weights by performing stochastic gradient descent. We use a learning rate that decreases by an order of magnitude every fixed number of iterations. Additionally, we use standard momentum in order to accelerate the learning process. We use a subset of the data for validation, and stop training when the metric we use to evaluate the learned models converges. Due to the exponentially large pool of positives and negatives available for training and the small number of parameters of the architectures, no techniques to cope with overfitting are used. The particulars of the learning procedure are detailed in the following section.

## Results

For evaluation we use the Multi-view Stereo Correspondence dataset (Brown et al., 2011), which consists of $64\times64$ grayscale image patches sampled from 3D reconstructions of the Statue of Liberty (LY), Notre Dame (ND) and Half Dome in Yosemite (YO). Patches are extracted using the Difference of Gaussians detector (Lowe, 2004), and determined as a valid correspondence if they are within 5 pixels in position, 0.25 octaves in scale and $\pi/8$ radians in angle. Fig. 3.23 shows some samples from each set, which contain significant changes in position, rotation and illumination conditions, and often exhibit very noticeable perspective changes.

We join the data from LY and YO to form a training set with over a million patches. Out of these we reserve a subset of 10,000 unique 3D points for validation (roughly 30,000 patches). The resulting training dataset contains 1,133,525 possible positive patch combinations and $1.117\times10^{12}$ possible negative combinations. This skew is common in correspondence problems such as stereo or structure from motion; we address it with aggressive mining.

A popular metric for classification systems is the Receiving Operator Characteristic (ROC), used e.g. in (Brown et al., 2011), which can be summarized by its Area Under the Curve (AUC). However, ROC curves can be misleading when the number of positive and negative samples are very different (Davis and Goadrich, 2006), and is already nearly saturated for the baseline descriptor SIFT. A richer metric is the Precision-Recall curve (PR). We benchmark our models with PR curves and their AUC. In particular, for each of the 10,000 unique points in the validation set we randomly sample two corresponding patches and 1,000 non-corresponding patches, and use them to compute the PR curve. We rely on the validation set for the LY+YO split to examine different configurations, network architectures and mining techniques.

| Architecture | Parameters | PR AUC |
|---|---|---|
| SIFT | — | 0.361 |
| CNN1_NN1 | 68,352 | 0.032 |
| CNN2 | 27,776 | 0.379 |
| CNN2a_NN1 | 145,088 | 0.370 |
| CNN2b_NN1 | 48,576 | 0.439 |
| CNN3_NN1 | 62,784 | 0.289 |
| CNN3 | 46,272 | **0.558** |

**Table 3.6: Effect of network architectures.** We look at the effects of network depth, and fully convolutional networks vs networks with a fully-connected layer. The PR AUC is calculated on the validation set for the top-performing iteration.

| Architecture | PR AUC |
|---|---|
| SIFT | 0.361 |
| CNN3 | **0.558** |
| CNN3 ReLU | 0.442 |
| CNN3 No Norm | 0.511 |
| CNN3 MaxPool | 0.420 |

**Table 3.7: Comparison of network hyperparameters.** We compare the fully convolutional CNN3 architecture with different hyperparameters settings such as Tanh and ReLU rectification layers, without normalization, and with max pooling instead of $L_2$ pooling. The best results are obtained for Tanh units, normalization, and $L_2$ pooling (i.e. 'CNN3'). The PR AUC is computed on the validation set for the top-performing iteration.

Finally, we evaluate the top-performing models over the test set. We follow the same procedure as for validation, compiling the results for 10,000 points with 1,000 non-corresponding matches each, now over 10 different folds. We run three different splits, for generalization: LY+YO (tested on ND), LY+ND (tested on YO), and YO+ND (tested on LY).

We will consider all hyperparameters to be the same unless otherwise mentioned, i.e. a learning rate of 0.01 that decreases ever $10,000$ iterations by a factor of 10. We consider negative mining with $B_N = 256$ and $B_N^M = 128$, and no positive mining; i.e. $B_P = B_P^M = 128$.

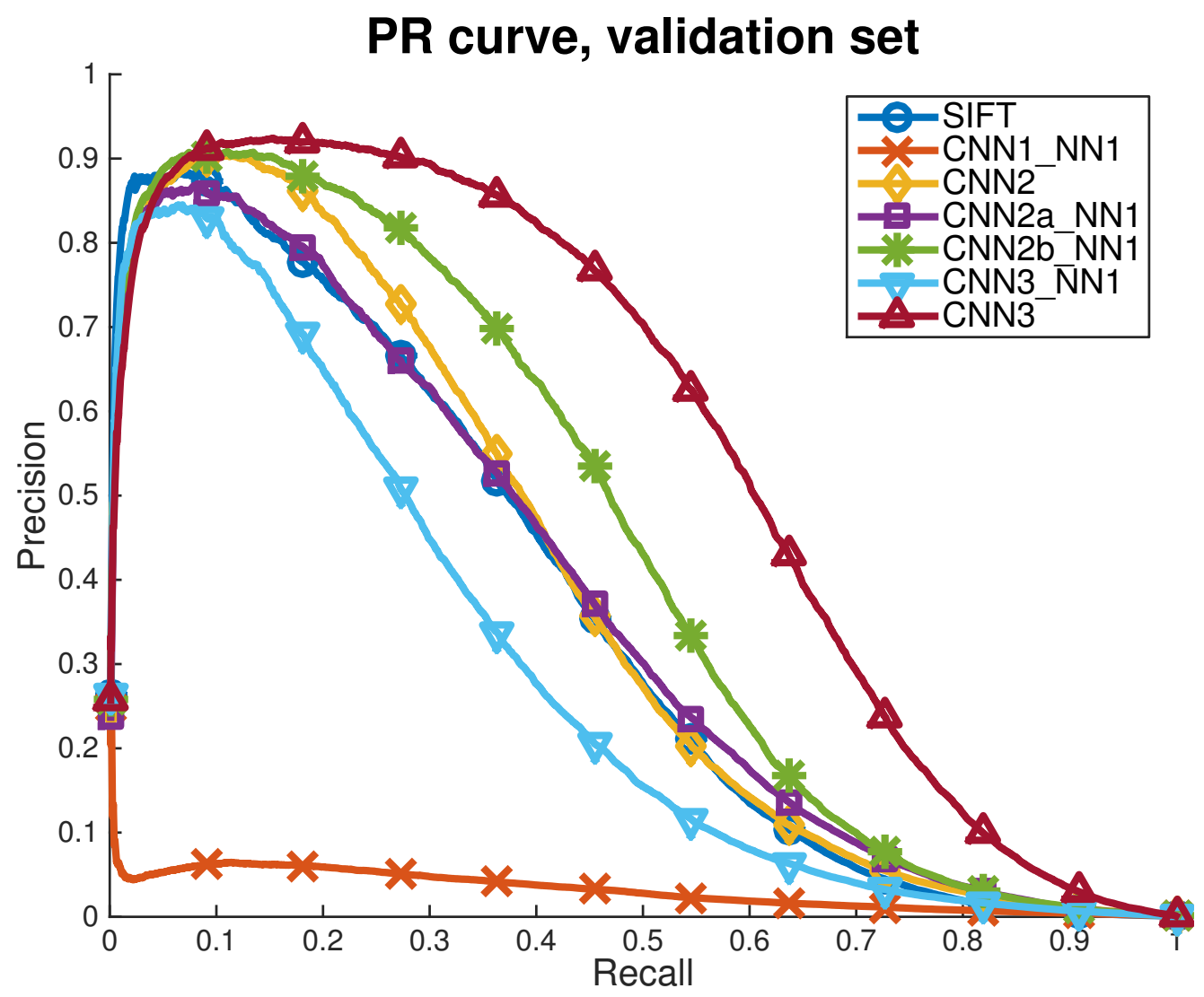

**Figure 3.24: Precision-recall curves for different network architectures.** We look at the effect of network depth (up to 3 CNN layers), and fully convolutional networks vs networks with final fully-connected layer (NN1). Fully-convolutional models outperform models with fully-connected neural network at the end.

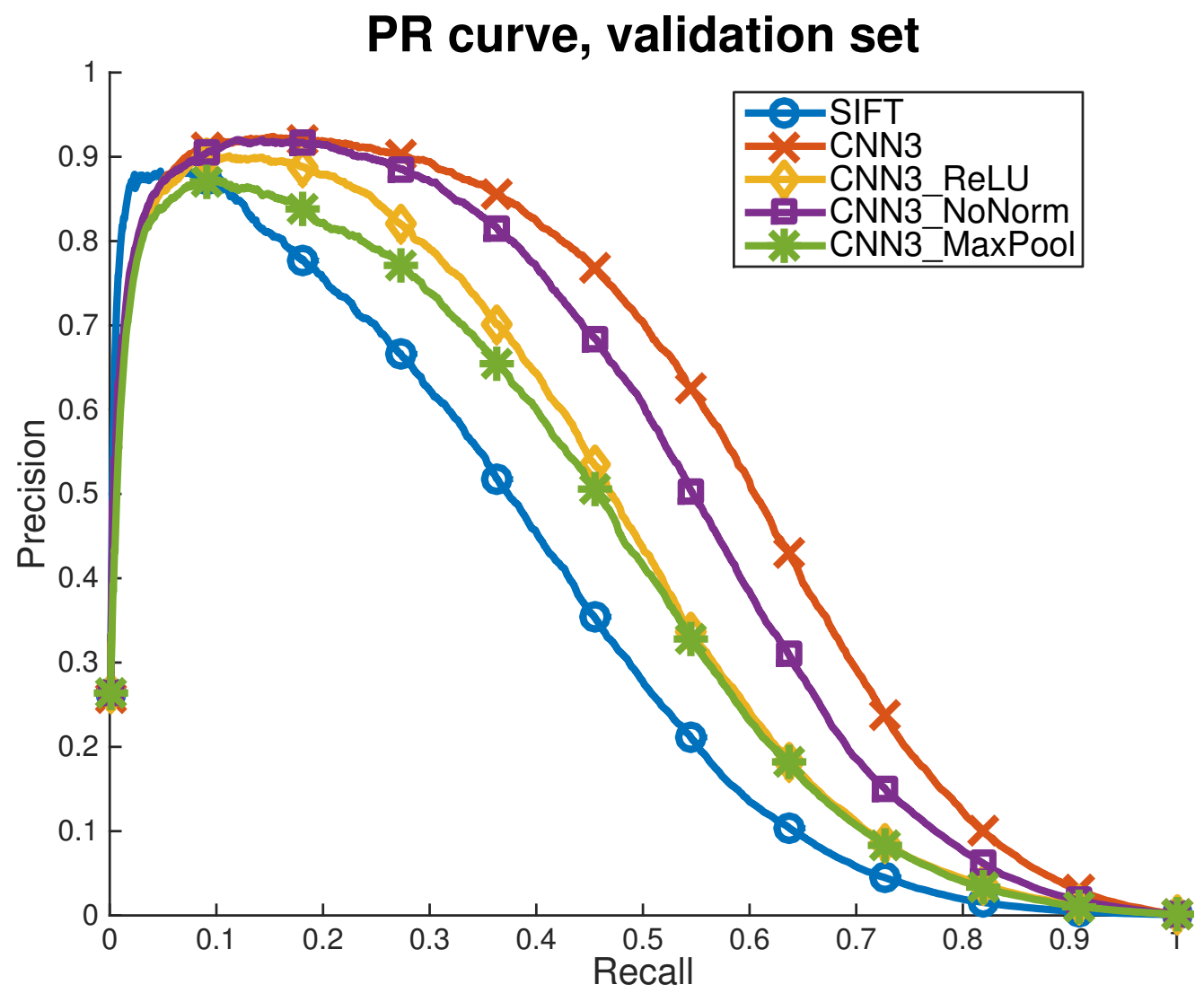

**Figure 3.25: Precision-recall curves for different network hyperparameters.** We analyze fully convolutional CNN3 models with Tanh and ReLU, without normalization, and with max pooling instead of $L_2$ pooling. The best results are obtained for Tanh units, normalization, and $L_2$ pooling ('CNN3' model).

### Depth and Fully Convolutional Architectures

The network depth is constrained by the size of the patch. We consider only up to 3 convolutional layers (CNN1-3). Additionally, we consider adding a single fully-connected layer at the end (NN1). Fully-connected layers increase the number of parameters by a large factor, which increases the difficulty of learning and can lead to overfitting. We show the results of the various architectures we evaluate in Table 3.6 and Fig. 3.24. Deeper networks outperform shallower ones, and architectures with a fully-connected layer at the end do worse than fully convolutional architectures. In the following experiments with consider only models with 3 convolutional layers.

| $R_P$ | $R_N$ | Cost | PR AUC |
|---|---|---|---|
| 1 | 1 | — | 0.366 |
| 1 | 2 | 20% | 0.558 |
| 2 | 2 | 35% | 0.596 |
| 4 | 4 | 48% | 0.703 |
| 8 | 8 | 67% | **0.746** |
| 16 | 16 | 80% | 0.538 |

**Table 3.8: Effect of mining samples on performance.** Mining factors indicate the samples considered ($B_P$, $B_N$), i.e. 1: 128, 2: 256, 4: 512, 8: 1024, and 16: 2048, of which 128 are used for training. Column 3 indicates the fraction of the computational cost spent mining.

| Architecture | Output | Parameters | PR AUC |
|---|---|---|---|
| SIFT | 128D | — | 0.361 |
| CNN3 | 128D | 46,272 | **0.596** |
| CNN3 Wide | 128D | 110,496 | 0.552 |
| CNN3_NN1 | 128D | 62,784 | 0.456 |
| CNN3_NN1 | 32D | 50,400 | 0.389 |

**Table 3.9: Effect of number of filters and fully-connected layer.** The best results are obtained with fully-convolutional networks with a fewer number of filters.

### Hidden Units Mapping, Normalization, and Pooling

It is generally accepted that Rectified Linear Units (ReLU) perform much better in classification tasks (Krizhevsky et al., 2012) than other non-linear functions. They are, however, ill-suited for regression tasks (such as the problem in hand), as they can only output positive values. We consider both the standard Tanh and ReLU. For the ReLU case we still use Tanh for the last layer. We also consider not using the normalization sublayer for each of the convolutional layers. Finally, we consider using max pooling rather than $L_2$ pooling. We show results for the fully-convolutional CNN3 architecture in Table 3.7 and Fig. 3.25. The best results are obtained with Tanh, normalization and $L_2$ pooling ('CNN3' in the table/plot). We will use this configuration in the following experiments.

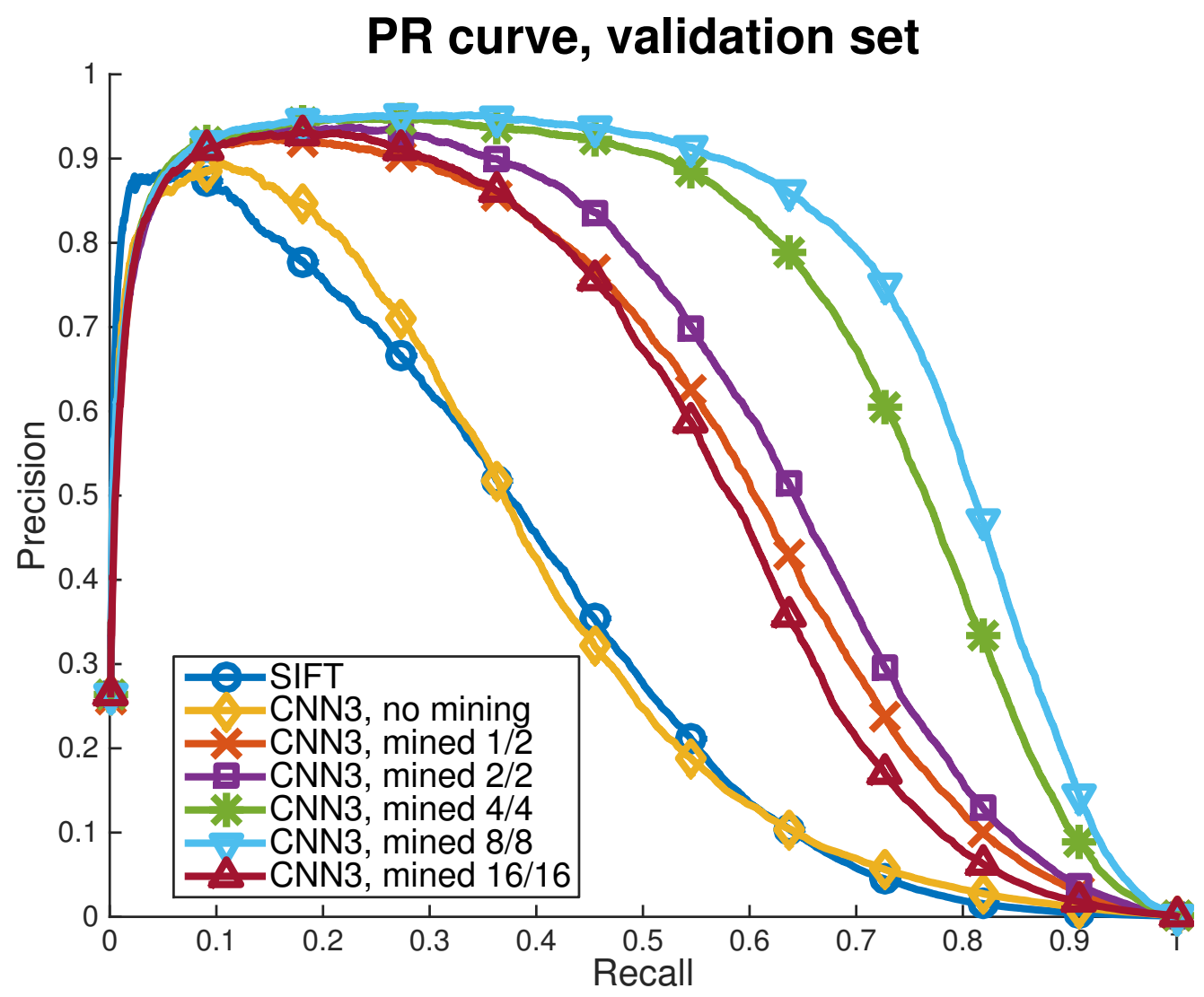

**Figure 3.26: Precision-recall curves for different levels of mining.** The best results are obtained with 8/8 factors, i.e. pools of 1024 samples filtered down to 128.

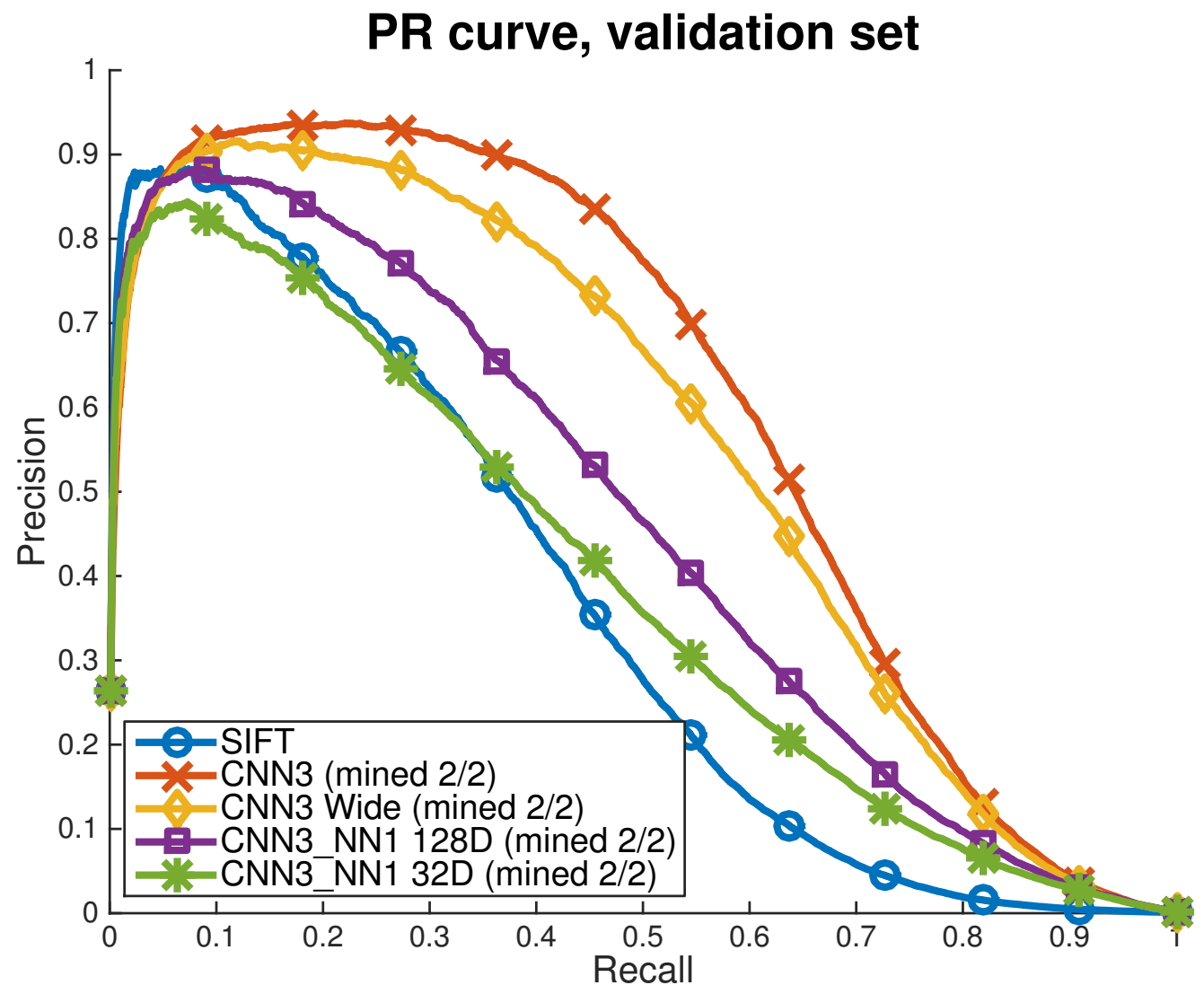

**Figure 3.27: Precision-recall curves for different number of filters and fully-connected layers.** We show several results for different number of filters and fully-connected layers for dimensionality reduction. The best results are obtained with fully-convolutional networks.

### Mining

We analyze the effect of both positive and negative mining by training different models in which a large, initial pool of $B_P$ positives and $B_N$ negatives are pruned to a smaller number of 'hard' positive and negative matches, used to update the parameters of the network. We keep this batch size for parameter learning constant, with $B_N^M = 128$ and $B_P^M = 128$, and increase the ratio of both negative mining $R_N = B_N/B_N^M$ and positive mining $R_P = B_P/B_P^M$. We keep all other parameters constant. We use the notation $R_P/R_N$, for brevity.

| Train | Test | SIFT | CNN3 mine-1/2 | CNN3 mine-2/2 | CNN3 mine-4/4 | CNN3 mine-8/8 | Improvement (Best vs SIFT) |
|---|---|---|---|---|---|---|---|
| LY+YOS | ND | 0.349 | 0.535 | 0.555 | 0.630 | **0.667** | 91.1% |
| LY+ND | YOS | 0.425 | 0.383 | 0.390 | 0.502 | **0.545** | 28.2% |
| YOS+ND | LY | 0.226 | 0.460 | 0.483 | 0.564 | **0.608** | 169.0% |

**Table 3.10: Precision-recall area under the curve for the generalized results over the three dataset splits.**

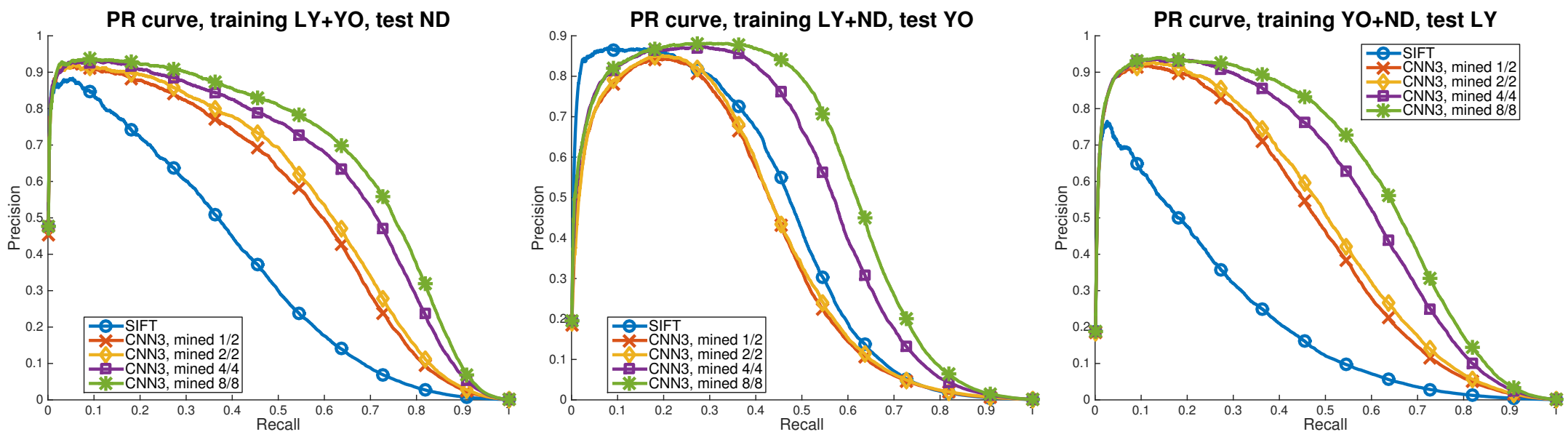

**Figure 3.28: Precision-recall curves for the generalized results over the three dataset splits.**

Large mining factors have a high computational cost, up to 80% of the total computational cost, which includes mining (i.e. forward propagation of all $B_P$ and $B_N$ samples) and learning (i.e. backpropagating the "hard" positive and negative samples). In order to speed up the learning process we initialize the CNN3 models with positive mining, i.e. 2/2, 4/4, 8/8 and 16/16, with an early iteration of a model trained only with negative mining (1/2).

Results are shown in Table 3.8 and Fig. 3.26. We see that for this particular problem aggressive mining is fundamental. This is likely due to the extremely large number of both negatives and positives in the dataset, in combination with models with a low number of parameters. We observe a drastic increase in performance up to 8/8 mining factors.

### Number of Filters and Descriptor Dimension

We analyze increasing the number of filters in the CNN3 model, and adding a fully-connected layer that can be used to decrease the dimensionality of the descriptor. We consider increasing the number of filters in layers 1 and 2 from 32 and 64 to 64 and 96, respectively. Additionally, we double the number of internal connections between layers. This more than doubles the number of parameters in this network. To analyze descriptor dimensions we consider the CNN3_NN1 model and change the number of outputs in the last fully-connected layer from 128 to 32. In this case we consider positive mining with $B_P = 256$ (i.e. 2/2). Results can be seen in Table 3.9 and Fig. 3.27. The best results are obtained with fewer filters and fully-convolutional networks. Additionally we notice that mining is also instrumental for models the NN1 layer (compare results with Table 3.6).

### Generalization

In this section we consider the three dataset splits. We train the best performing models, i.e. CNN3 with different mining ratios, on a combination of two sets, and test them on the remaining set. We select the training iteration that performs best over the validation set. The test datasets are very large (up to 633K patches) and we use the same procedure as for validation, evaluating 10,000 unique points, each with 1,000 random non-corresponding matches. We repeat this process over 10 folds, thus considering 100,000 sets of one corresponding patch vs 1,000 non-corresponding patches. We show results in terms of PR AUC in Table 3.10, and the corresponding PR curves are pictured in Fig. 3.28.

We report consistent improvements over the baseline, i.e. SIFT. The performance varies significantly from split to split; this is due to the nature of the different sets. 'Yosemite' contains mostly fronto-parallel translations with illumination changes and no occlusions (Fig. 3.23, row 3); SIFT performs well on this data. Our learned descriptors outperform SIFT on the high-recall regime (over 20% of the samples; see Fig. 3.28), and is 28% better overall in terms of PR AUC. The effect is much more dramatic on 'Notredame' and 'Liberty', which contains significant patch translation and rotation, as well as viewpoint changes around outcropping non-convex objects which result in occlusions (see Fig. 3.23, rows 1-2). Our learned descriptors outperform SIFT by 91% and 169% testing over ND and LY, respectively.

### Qualitative analysis

Fig. 3.29 shows samples of matches retrieved with our CNN3-mined-4/4 network, over the validation set for the first split. In this experiment the corresponding patches were ranked in the first position in 76.5% of cases; a remarkable result, considering that every true match had to be chosen from a pool of 1,000 false correspondences. The right-hand image shows cases where the ground truth match was not ranked first; notice that most of these patches exhibit significant changes of appearance. We include a failure case (highlighted in red), caused by a combination of large viewpoint and illumination changes; however, these misdetections are very uncommon.

## 3.4 Summary

In this chapter we have presented two vastly different approaches at tackling the same problem: discriminative representation of local image patches, i.e., feature point descriptors. Our first approach is more traditional in the sense it is a hand-crafted descriptor for the specific problem of deformation and illumination invariant feature point descriptors. For our second approach we propose a data-driven approach in which given enough data we are able to learn the descriptor. Both approaches are complementary and can be used in a variety of different applications.

On one hand, heat diffusion theory has been recently shown effective for 3D shape recognition tasks. We have proposed using these techniques to build DaLI, a feature point descriptor for 2D image patches, that is very robust to both non-rigid deformations and illumination changes. The advantages of our method with respect to the state-of-the-art have been demonstrated by extensively testing on a new deformation and varying illumination evaluation dataset. We have also shown that simple dimensionality reduction techniques such as PCA can be effectively used to reduce dimensionality while maintaining similar performance. This seems to give the intuition that further improvements can be obtained by using more advanced and powerful techniques such as LDAHash (C. Strecha and Fua, 2012). Work has also been done in optimizing the calculation speed by means of more complex meshing to reduce the cost of computing the eigenvectors of the Laplace-Beltrami operator.

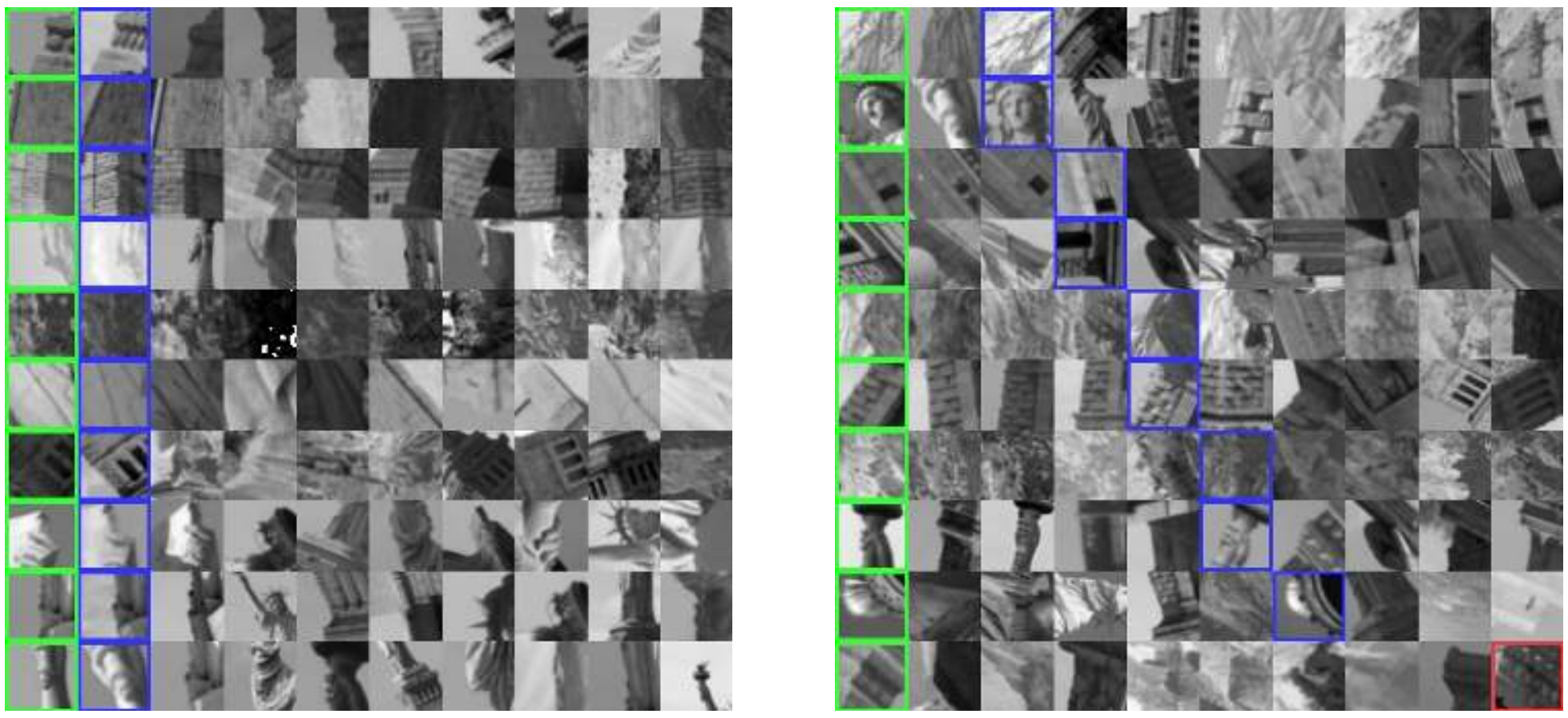

**Figure 3.29: Samples of matches retrieved with our descriptor.** Each row depicts the reference patch (in green) and the top matching candidates, sorted from left to right by decreasing similarity. The ground truth match is highlighted in blue. *Left:* Examples where the ground truth match is retrieved in the first position. *Right:* Examples in which the ground truth match is ranked at positions 2-6. The last row shows a failure case, highlighted in red, where the correct match is ranked 632/1000.

We have also shown that it is possible to learn feature point descriptors instead of manually defining them. This is a non-standard learning problem due to using small image patches and not being able directly optimize the model. Instead we propose learning the model with a Siamese network that uses two patches simultaneously to learn a deep convolutional network. Furthermore we propose a novel training scheme based on aggressive mining of both positive and negative correspondences that we have shown is critical to obtain high performance. We have performed extensive evaluation considering a wide range of architectures and hyperparameters to fully evaluate our method. Our results show that given the right training data we are able to obtain a 2.5x performance increase over the widely used SIFT descriptor (Lowe, 2004). While the results are preliminary, we believe that by creating a much larger and representative dataset it should be possible to learn a very general descriptor that could be used as a drop-in replacement for existing descriptors to improve performance.

Both presented approaches have been proven to be extremely competitive with the state-of-the-art and are complementary with existing and widely used difference of Gaussian based approaches. We have released the code for the evaluation and computation of the DaLI descriptor[3] to encourage other researchers to compare against and use our descriptor in other applications. While we have not at the time of this writing released the code for our deep descriptor framework, we intend to release it for the same reasons.

[3] http://www.iri.upc.edu/people/esimo/research/dali/

# Chapter 4

# Generative 3D Human Pose Models

In this chapter we give an overview of different models for 3D Human Pose estimation in order of increasing complexity and expressiveness. This is not meant to be a complete list and only mentions models used throughout the course of this thesis. In particular we shall discuss linear latent models, Directed Acyclic Graph (DAG) models, and we will give a slightly longer description of the Geodesic Finite Mixture Model (GFMM).

## 4.1 Introduction

There has been a long tradition of generative models in the task of 3D human pose estimation. This is due to the space of possible 3D poses being extremely large. Additionally, for 3D pose estimation there is no natural grid such as pixels used in 2D pose estimation. This makes it unnatural to define discriminative models such as the pictorial structures model (Felzenszwalb and Huttenlocher, 2005) widely used in the 2D for the 3D case. Furthermore, when tracking 3D poses, there is a strong spatial pose prior which encourages searching the solution space near the solution of the previous frame, which can be incorporated elegantly into generative models.

In this chapter we shall focus on three different approaches used in this thesis: linear latent models, Directed Acyclic Graph (DAG) models, and Geodesic Finite Mixture Models (GFMM) We will give a special focus to the latter model which is a novel model proposed in this thesis. For all these poses we consider the static scenario, i.e., we do not consider motion nor any other temporal information. Additionally all models make use of latent variables to model the pose.

Each model has a different set of strengths and weaknesses. We focus on four different model attributes: complexity, scalability, consistency, and whether it is modelling the Probability Density Function (PDF) of the data. Complexity roughly corresponds to the level of technical and algorithmic complexity of the model. As to be expected, linear models have very low complexity in both formulation and speed, however, they tend to be much weaker representations. As dataset sizes are getting larger and larger, scalability or the ability of the model to benefit from large amounts of data is becoming more and more important. Additionally for 3D human pose estimation, we know

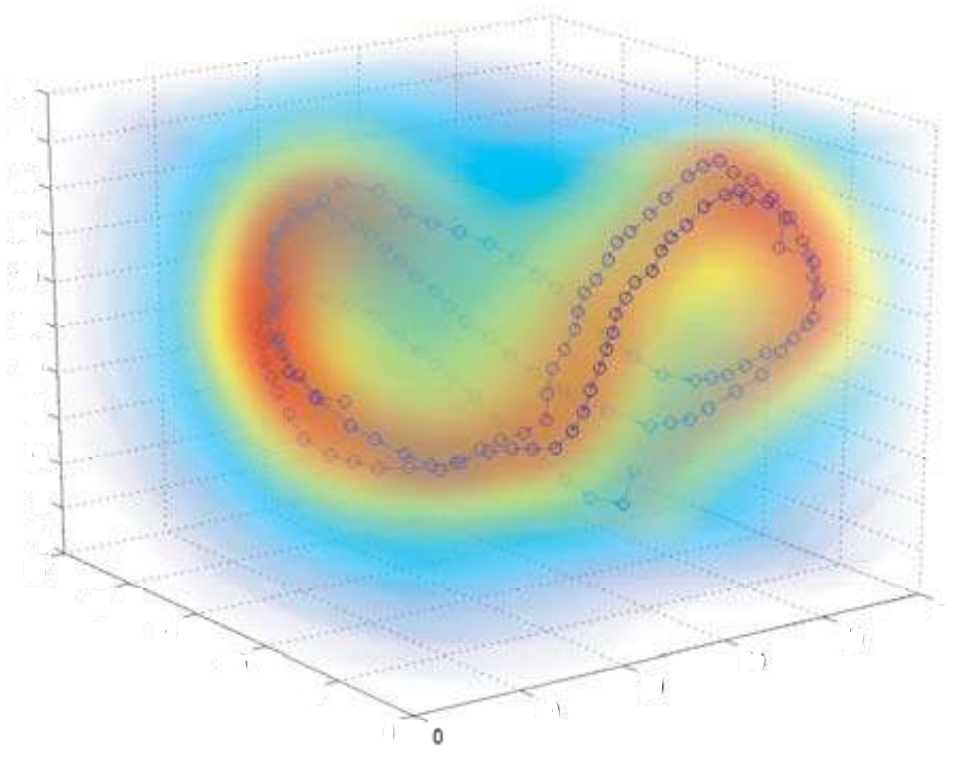

**(a)** Probability Density Function

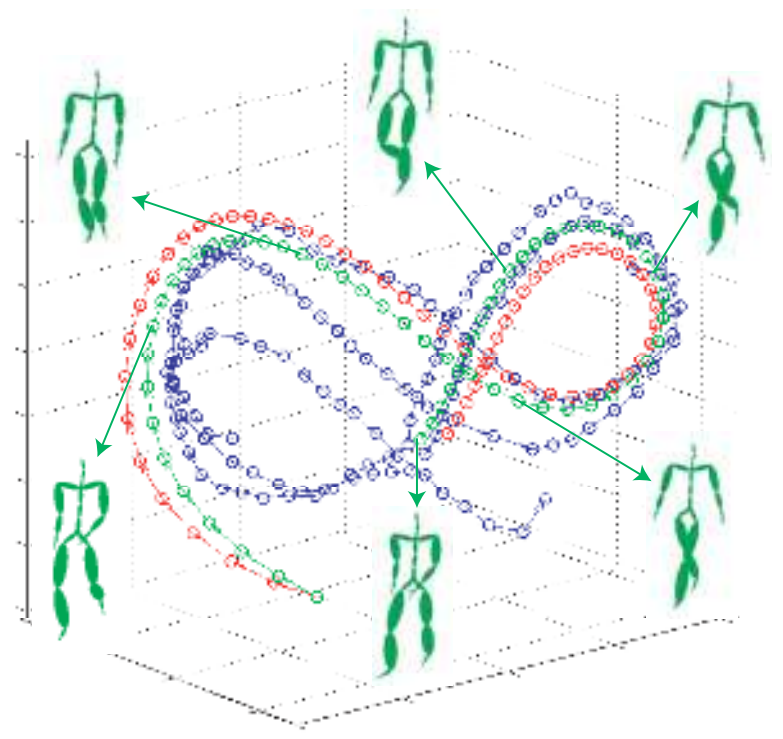

**(b)** Trajectory reconstruction

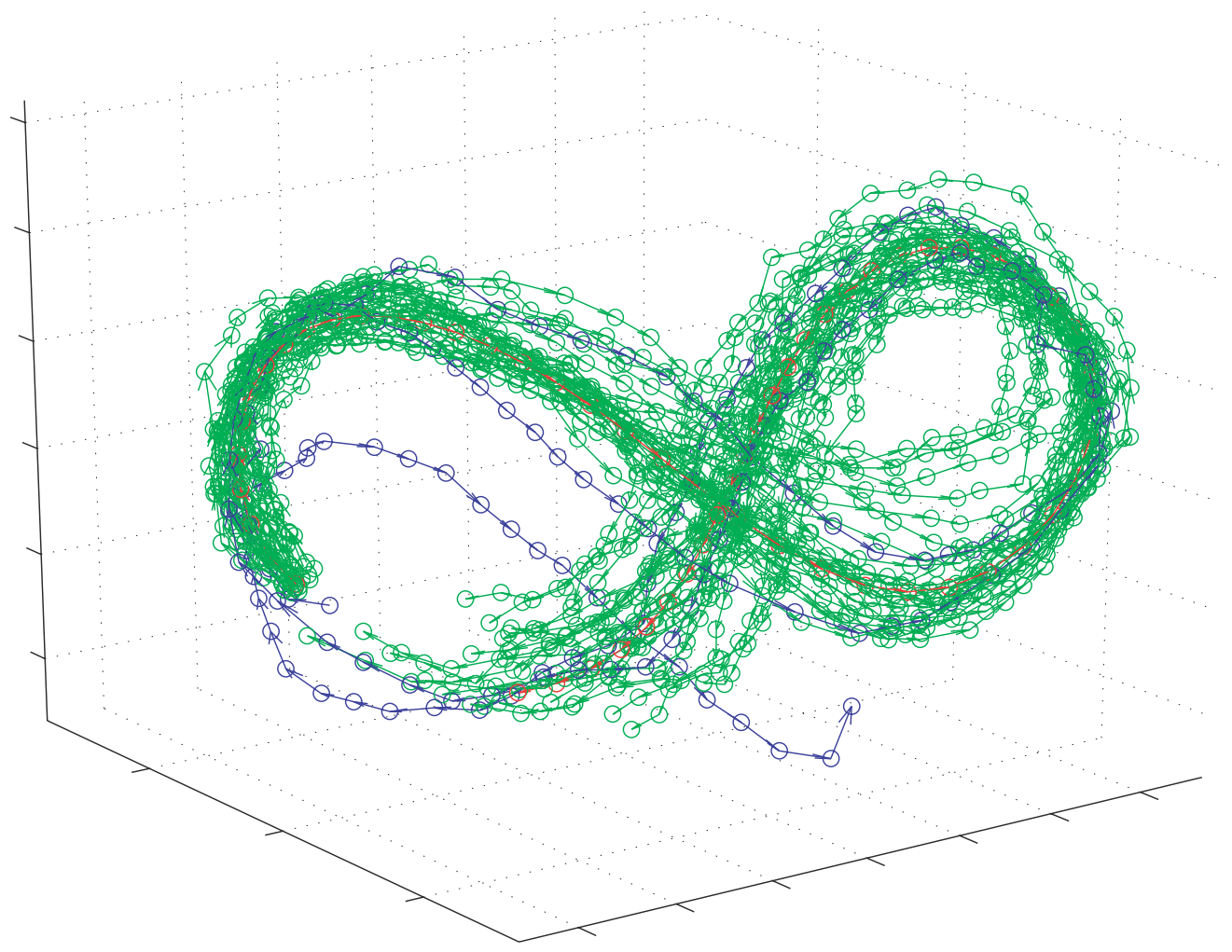

**(c)** Random trajectories drawn from the model the data lies on a manifold for a given individual. Consistency refers to whether or not an algorithm is able to generate poses only on this manifold, or if it can erroneously also generate outside the manifold and thus generate non-anthropomorphic poses. Finally, not all models attempt to represent the PDF of the data. In particular linear models are unable to, while more expressive models are able to. Although there are more model attributes that can be taken into account, we find that these four are the most representative of 3D pose models.

**Figure 4.1: Example of a GPDM with a 3-dimensional latent space.** Model are learnt from a walking sequence of 2.5 gaits on the CMU mocap dataset (mocap.cs.cmu.edu). Figures reproduced from (Wang et al., 2005).

For the entire chapter we shall write a 3D pose as the vector $\mathbf{x} = [\mathbf{p}_1^{\mathrm{T}}, \cdots, \mathbf{p}_{n_v}^{\mathrm{T}}]$, where $\mathbf{p}_i$ are the 3D positions of the skeleton joints. In general the objective is to model the distribution of these poses, that is $p(\mathbf{x})$.

## 4.2 Related Work

One of the more successful families of models for 3D human pose estimation have been the Gaussian Process family of models. Of which the most important is the Gaussian Process Latent Variable Model (GPLVM) (Lawrence, 2005), in which a continuous low-dimensional latent space is learnt for a single cycle of a motion. This has been extended to dynamics with the Gaussian Process Dynamic Model (GPDM) proposed by Wang et al. (Wang et al., 2005; Urtasun et al., 2006; Wang et al., 2008; A. Yao and Urtasun, 2011), which was also extended to include topological constraints (Urtasun et al., 2007). A visualization of a 3-dimensional GPDM is shown in Fig. 4.1. Hierarchical variants (hGPLVM) have also been used in a tracking by detection approach (Andriluka et al., 2010). However, Gaussian Processes do not scale well to large datasets due to their $\mathcal{O}(n^3)$ complexity for prediction. Sparse approximations do exist (Quiñonero-candela et al., 2005), but in general do not perform as well.

There have been other approaches such as learning Conditional Restricted Boltzmann Machines (CRBM) (Taylor et al., 2010). However, these methods have a very complex learning procedure that makes use of several approximations and thus it is not easy to train good models. Li et al. (Li et al., 2010) proposed the Globally Coordinated Mixture of Factor Analyzers (GCMFA) model which is similar to the GPLVM in the sense it is performing a strong non-linear dimensionality reduction. Yet, as in GPLVM models it does not scale well to large datasets. A more simple approach would be the Gaussian Mixture Model (GMM) that have been used extensively to encode the relationship between joints (Sigal et al., 2004; Daubney and Xie, 2011; Sigal et al., 2012). This is usually done in the angular space. However, as this space is not a vector space due to an inherent periodicity, they introduce heuristics to mitigate the effect it has on the GMM. On the other hand our proposed GFMM does not have this issue and additionally provides a much better approximation of the underlying manifold, without any notable computational slowdown.

Additionally recently there have been many development on statistics on Riemannian manifolds (Pennec, 2006) that have made their way into computer vision (Fletcher et al., 2004; Pennec et al., 2006; Davis et al., 2007), and in particular human motion (Sommer et al., 2010; Brubaker et al., 2012). The most popular model is the Principal Geodesic Analysis (PGA) model (Fletcher et al., 2004), in which the data is projected onto a linear space which is an approximation of the manifold. As this space is linear, another linear model, in particular a linear latent model, can be constructed in this approximated space. While this is a simple scalable and consistent approach, it is not modelling a probability density function.

Not all work has been done on joint-based models. As it is quite common to use silhouettes for 3D human pose tracking, shape-based models have been also proposed. Some of the more simple ones model the body with a set of cylinders which are fitted to a joint model (Sigal et al., 2004, 2012). However, there do exist approaches that directly work with 3D meshes obtained from laser scans. One of the more well-known approaches applies a Principal Component Analysis (PCA) to the 3D meshes (Anguelov et al., 2005; Hasler et al., 2009). An example of 3D scans is shown in Fig. 4.2. The downside of using silhouettes to match 3D mesh models is that it is usually dependent on background subtraction in order to get reliable silhouette estimates. On the other hand the joint based models we consider can be used with other types of image evidence as we will show in the next chapter.

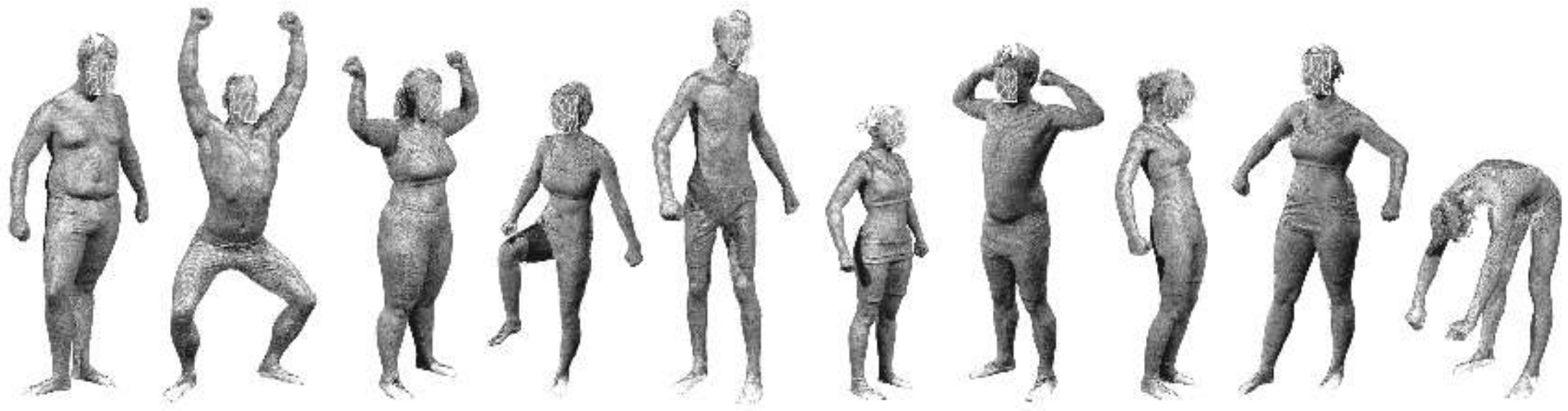

**Figure 4.2: Examples of 3D body shape scans.** These scans are used to obtain a set of deformation basis to represent body shape and 3D pose independently. Figure reproduced from (Hasler et al., 2009).

| Model | Complexity | Scales? | Consistent? | PDF? |
|---|---|---|---|---|
| GMM (Sigal et al., 2004) | Low | Yes | No | Yes |
| PGA (Fletcher et al., 2004) | Low | Yes | Yes | No |
| GPLVM (Lawrence, 2005) | Low | No | No | Yes |
| GPDM (Wang et al., 2005) | Medium | No | No | Yes |
| hGPLVM (Andriluka et al., 2010) | Medium | No | No | Yes |
| CRBM (Taylor et al., 2010) | High | Yes | No | Yes |
| GCMFA (Li et al., 2010) | High | No | No | Yes |
| PCA | Low | Yes | No | No |
| DAG | Medium | Yes | No | Yes |
| GFMM | Low | Yes | Yes | Yes |

**Table 4.1: Comparison of different generative pose models.** We analyze several commonly used human pose models. We consider the complexity of the models which we classify as low, medium or high. Models are considered to scale if they can handle well large amounts of data (∼100k samples) and to be consistent if they use geodesic distances instead of other metrics. Additionally we show whether or not a model is actually modelling the Probability Density Function (PDF) of the data. The bottom three correspond to the models used in this thesis.

We show an overview of the aforementioned models in Table 4.1. We focus on four model attributes: complexity, scalability, consistency, and whether or not the model represents the Probability Density Function (PDF) of the data. We also include the three models used in this thesis: latent linear model (PCA), Directed Acyclic Graph (DAG) and Geodesic Finite Mixture Model (GFMM). We can see that our models can all scale well to large data while having a medium to low complexity. Additionally the GFMM is consistent with the 3D human pose manifold.

## 4.3 Linear Latent Model

Latent linear models are well known in the field of 3D pose estimation and shape recovery (Moreno-Noguer et al., 2009, 2010; Moreno-Noguer and Porta, 2011; Moreno-Noguer and Fua, 2013). Their extremely low complexity makes them very attractive for many different applications. They consist in finding a linear transformation that projects data points to a, generally smaller, latent space. This is very useful as the true degrees of freedom of high dimensional data is usually much smaller than the dimensionality of the data. In general, there is no linear mapping between the pose manifold and the 3D pose, however, the coarse approximation can be considered sufficient when combined with additional constraints such as illumination (Moreno-Noguer and Fua, 2013).

One of the more simple yet widely used approaches consists in finding a linear latent space using Principal Component Analysis (Tenenbaum et al., 2000). This consists in finding a new linear basis for the data where the vectors are sorted by covariance. By then discarding the vectors with lower covariance it is possible to find a lower dimension linear subspace that conserves most of the data variability. In the context of deformable object, the vectors of the new basis can be interpreted as different deformation modes of the object. This approach is unsupervised, i.e., no labels are needed for the data. An alternative is Linear Discriminant Analysis, in which a projection that attempts to maximize the difference between the class of each sample is found. This is a supervised approach, that is, each sample has a label that we wish to predict.

### Deformation Modes

We first assume that the 3D pose can be represented as a linear combination of a mean 3D pose $\mathbf{x}_0$ and $n_m$ deformation models $\mathbf{Q} = [\mathbf{q}_1, \cdots, \mathbf{q}_{n_m}]$

$$\mathbf{x} = \mathbf{x}_0 + \sum_{i=1}^{n_m} \alpha_i \mathbf{q}_i = \mathbf{x}_0 + \mathbf{Q}\boldsymbol{\alpha} \,, \tag{4.1}$$

where $\boldsymbol{\alpha} = [\alpha_1, \ldots, \alpha_{n_m}]^\mathrm{T}$ are the unknown weights that define the current 3D pose and can be interpreted as parameters of the learnt latent space corresponding tho the pose $\mathbf{x}$.

The matrix $\mathbf{Q}$ can be found by simply calculating the eigenvalues of the covariance of the training data. Assuming we have $N$ poses $\mathbf{x}_i$ we can rewrite them in matrix form as

$$\mathbf{X} = \begin{bmatrix} (\mathbf{x}_1 - \mathbf{x}_0)^\mathrm{T} \\ (\mathbf{x}_2 - \mathbf{x}_0)^\mathrm{T} \\ \vdots \\ (\mathbf{x}_N - \mathbf{x}_0)^\mathrm{T} \end{bmatrix} \,, \tag{4.2}$$

where $\mathbf{x}_0 = \frac{1}{N} \sum_i \mathbf{x}_i$.

The covariance $\mathbf{C}$ of the training data can then be calculated by

$$\mathbf{C} = \frac{1}{N-1} \mathbf{X}^\mathrm{T} \mathbf{X} \,. \tag{4.3}$$

The eigenvectors $\mathbf{V}$ and eigenvalues $\boldsymbol{\lambda}$ can then be computed by

$$\mathbf{V}^{-1} \mathbf{C} \mathbf{V} = \mathrm{diag}(\boldsymbol{\lambda}) \,. \tag{4.4}$$

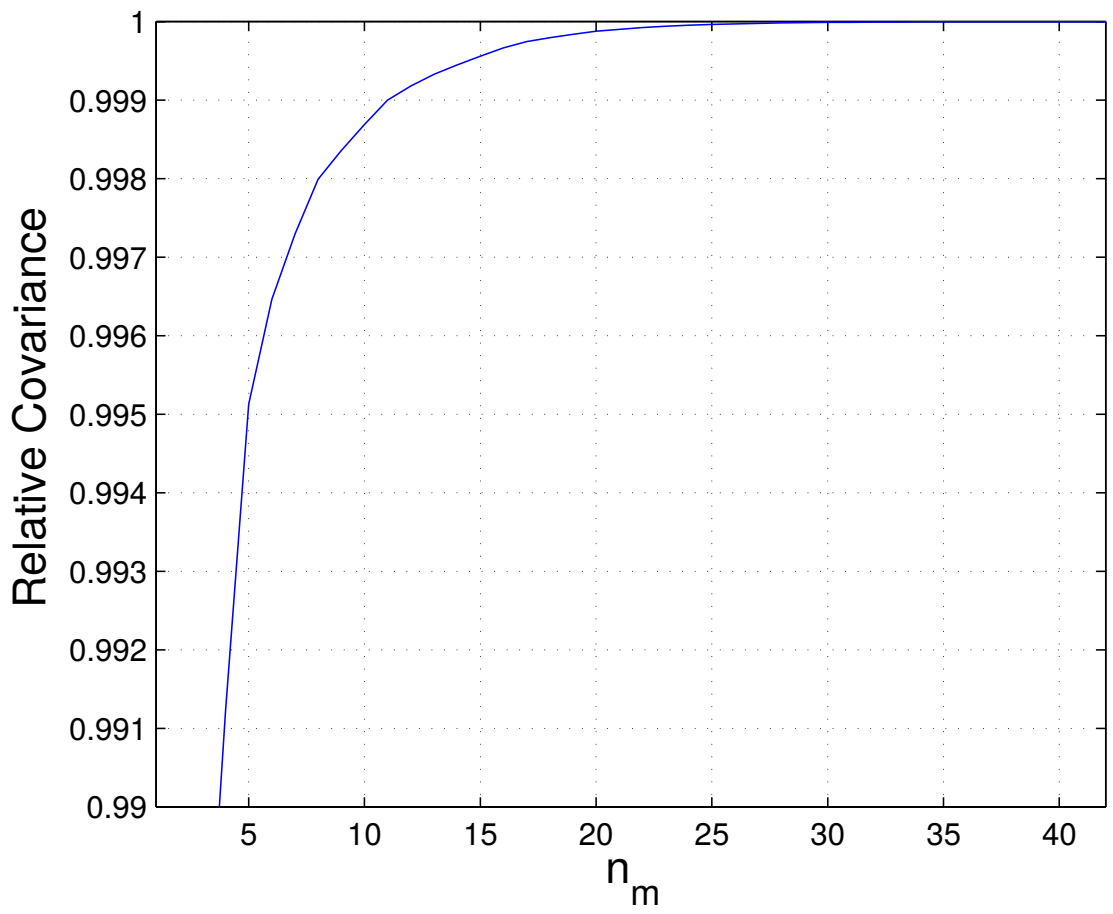

**Figure 4.3: Effect of $\sigma_{\text{threshold}}$ in linear latent models.** We plot how the parameter $\sigma_{\text{threshold}}$ affects the amount of covariance retained or relative covariance of the data for the HumanEva walking action. In particular with $n_m = 21$ we are able to retain 99.99% of the covariance.

Without loss of generality we can assume the eigenvectors $\mathbf{V}$ are sorted in descending order by their eigenvalues $\boldsymbol{\lambda}$.

Finally the matrix $\mathbf{Q}$ is just the first $n_m$ vectors of the eigenvectors $\mathbf{V}$. As a criteria for selecting the value of $n_m$, the percent of the total variance explained by those vectors can be used. Thus $n_m$ can be chosen by

$$\arg\min_{n_m} n_m \quad \text{subject to} \quad \sigma_{\text{threshold}} < \frac{\sum_{i=1}^{n_m} d_i}{\sum_{i=1}^{N} d_i} \,. \tag{4.5}$$

Thus instead of choosing the value of $n_m$ directly, it is chosen based on a more interpretable threshold $\sigma_{\text{threshold}}$, which is the only hyperparameter of the model. Typical values for $\sigma_{\text{threshold}}$ are around 0.99.

## Results

We plot the effect of $\sigma_{\text{threshold}}$ in Fig. 4.3. This is the only hyperparameter of the model and is critical for good performance as it serves as a lower-bound for 3D pose estimation error. From our experiments we found that setting $n_m = 21$, corresponding to $\sigma_{\text{threshold}} = 0.9999$, offers a good balance between dimensionality and low error. We also show a qualitative example of how the dimensionality affects the reconstruction quality in Fig. 4.4.

The main advantage of these linear models is they are very fast to compute and very easy to use. They are completely unsupervised and depend on a single hyperparameter. The main disadvantage is they only allow for linear relationships and are not truly modelling $p(\mathbf{x})$, i.e., they only model a mapping, not an actually Probability Density Function (PDF). Additionally were we to sample from the latent space $\boldsymbol{\alpha}$ there is no guarantee that the poses will be anthropomorphic or human-like. Thus if sampling additional criteria will have to be used to select only the anthropomorphic poses.

In Section 5.3 we shall show an application in which we benefit from the linear formulation. In particular we are able to combine the linear pose model with a linear camera projection model in order to project noisy 2D joint observations into the latent space. As the formulation is linear, there is an exact solution, which generates a set of 3D hypothesis that can be used to perform 3D pose estimation.

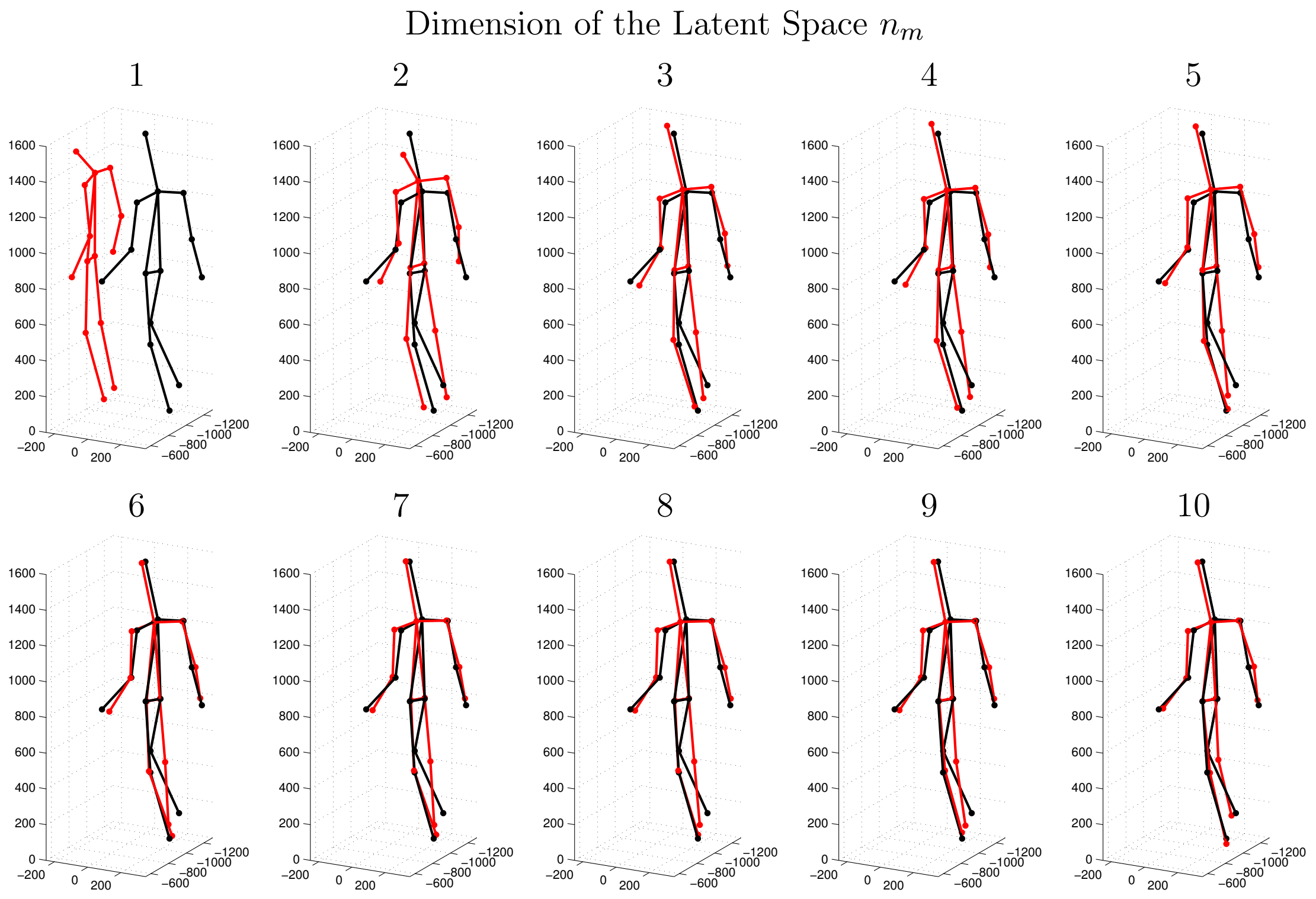

**Figure 4.4: Dimensionality of the Linear Latent Space.** We show the effects of the dimensionality of the linear latent space for reconstructing. We show the ground truth in black and the reconstructed pose for $n_m = \{1, \ldots, 10\}$. The first three dimensions serve to appropriately position the joints. Additional dimensions improve smaller details such as matching the hands and the feet.

## 4.4 Directed Acyclic Graphs

Using probabilistic graphical models is very common in computer vision. In particular star-shaped models (Felzenszwalb et al., 2008, 2010) and tree-shaped models (Yang and Ramanan, 2011) have been widely used as exact inference can be performed. In the case of general graphical models, inference is no longer exact due to the presence of loops and must be performed by message passing algorithms (Sigal et al., 2012; Andriluka et al., 2012). We instead opt to use Directed Acyclic Graphs (DAG), that can capture more complicated relationships in comparison to tree models while still having exact inference, as there no loops in the graphical model.

This model attempts to characterize $p(\mathbf{x})$ through a latent space. That is $p(\mathbf{x})$ is not modelled directly, instead it is modelled as $p(\mathbf{x}) = p(\mathbf{x}|\mathbf{h})p(\mathbf{h})$ where $\mathbf{h}$ is a latent space. In order to be tractable the poses are discretized using a clustering algorithm such as k-means into a set of base poses $\mathcal{X}^L$. Additionally we define the lower dimensional latent space as another discrete set $\mathcal{H}$. The problem can then be seen as learning a compression function $\phi(X^L) : \mathcal{X}^L \to \mathcal{H}$ that maps points in the high dimensional local 3D pose space to a lower dimensional space.

### Discrete Joint Model

In order to obtain the discrete pose set $\mathcal{X}^L$ the $\mathbf{x}$ poses in 3D are first transformed to a local reference such that they only represent local deformations. Then they are mapped from the $\mathbb{R}^{N\times 3}$ continuous pose space $\mathbf{x}$ to a discrete domain by doing vector quantization of groups of 3D joint positions. More specifically, the joints are grouped into five coarse parts: right arm ($ra$), left arm ($la$), right leg ($rl$), left leg ($ll$) and torso+head ($th$). Thus, every pose can be mapped to a discrete vector $T = [ra, la, rl, ll, th] \in \mathcal{K}^5$ where $ra, la, rl, ll, th$ are cluster indexes belonging to $\mathcal{K} = \{1, 2, .., k\}$ of the corresponding 3D joint positions.

We will now define a joint distribution over latent variables $H \in \mathcal{H} = \{1, .., n\}$ (where $n$ is the number of latent states), and observed variables $T$. The model is given by the following generative process:

- Sample a latent state $i$ according to $p(h_0)$
- For all the parts associated with arm and leg positions, sample discrete locations: $< ra, la, rl, ll >$ and states $< j, l, m, n >$ according to the conditional distributions: $p(h_{ra} = j, ra \mid i)$, $p(h_{la} = l, la \mid i)$, $p(h_{rl} = m, rl \mid i)$ and $p(h_{ll} = n, ll \mid i)$
- Sample a pair of latent states: $< q, w >$ (associated with the positions of the upper body and lower body joints) according to $p(h_u = q \mid h_{ra} = j, h_{la} = l)$ and $p(h_l = w \mid h_{rl} = m, h_{ll} = n)$
- Sample a discrete location $th$ and a state $r$ from $p\left(h_{(th)} = r, th \mid h_u = q, h_l = w\right)$

Given this generative model we define the probability of a discrete 3D position $T = [ra, la, rl, ll, th]$ as:

$$
\begin{aligned}
p(T) &= \sum_{\mathcal{H}} p(T, H) \\
&= \sum_{\mathcal{H}} p(h_0)\, p(h_{ra}, ra \mid h_0)\, p(h_{la}, la \mid h_0) \\
&\quad p(h_{rl}, rl \mid h_0)\, p(h_{ll}, ll \mid h_0)\, p(h_u \mid h_{ra}, h_{la}) \\
&\quad p(h_{ll}, ll \mid h_0)\, p(h_u \mid h_{ra}, h_{la}) \\
&\quad p(h_l \mid h_{rl}, h_{ll})\, p(h_{th}, th \mid h_u, hl)\,.
\end{aligned}
$$

The graphical model corresponding to this joint distribution is illustrated in Fig. 4.5a, where the graph $G$ specifies the dependencies between the latent states. Since $H$ is unobserved, Expectation Maximization can be used to estimate the model parameters from a set of training poses. Given that $G$ is a Directed Acyclic Graph we can compute all required expectations efficiently with dynamic programming (Huang, 2008). Once we have learned the parameters of the model we define our compression function to be:

$$
\phi(X^L) = \arg\max_{H}\; p\left(X^L, H\right)\;,
$$

and our decompression function to be:

$$
\phi^{-1}(H) = \arg\max_{X^L}\; p\left(X^L, H\right)\;.
$$

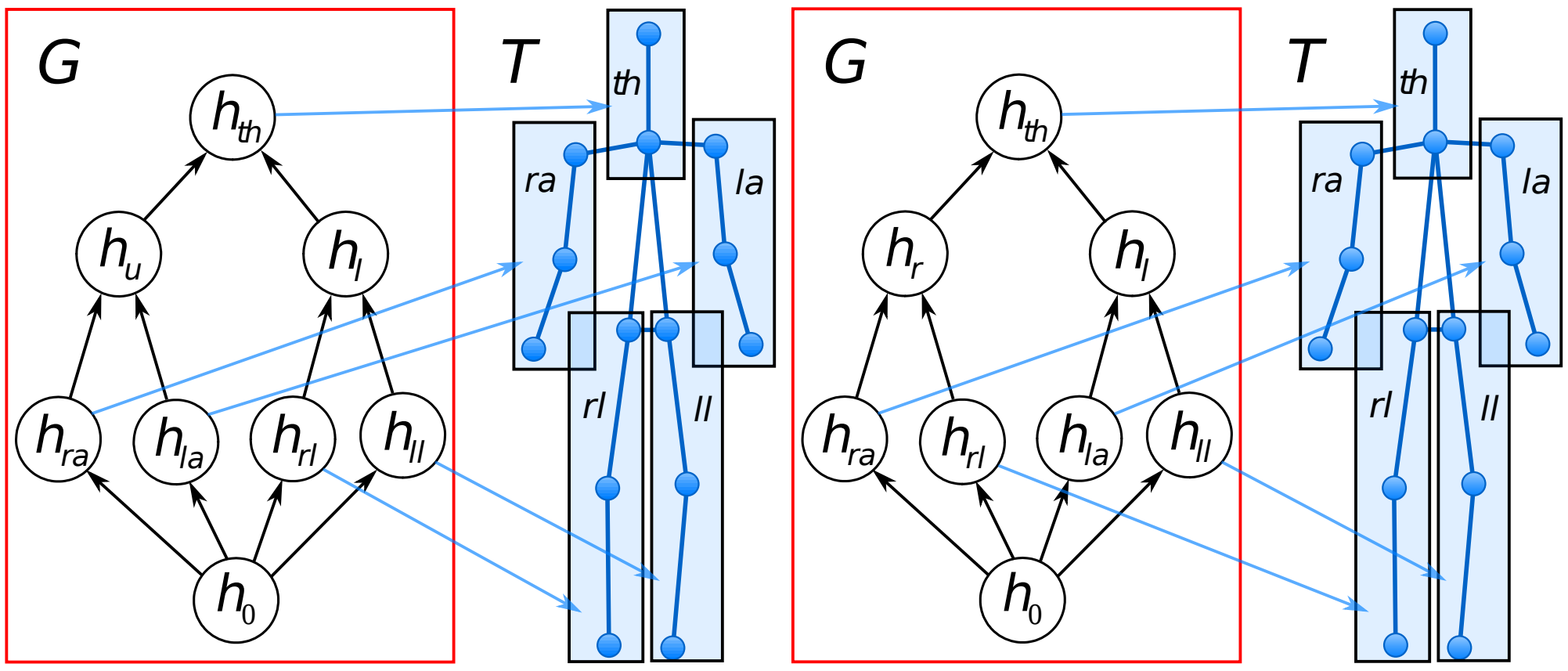

**(a)** UL grouping **(b)** RL grouping

**Figure 4.5: Directed Acyclic Graph models for 3D human pose.** We use a probabilistic graphical model with latent variables $G$ to represent the possible 3D human pose $T$ from a large set of discrete poses. Latent variables can either be mapped to the conjoint motion of various parts or be used as internal states containing internal structure of the pose. We show two different type of body part groupings: (a) Upper and Lower body grouping, and (b) Right and Left side grouping.

| States | 1 | 2 | 3 | 4 | 5 | **6** | 7 |
|---|---|---|---|---|---|---|---|
| **UL** | 123 | 108 | 87 | 64 | 61 | **51** | 46 |
| RL | 123 | 107 | 84 | 68 | 77 | 49 | 49 |

**Table 4.2: Influence of the number of latent states.** We evaluate them using the average reconstruction error (in mm). We compare the upper-lower grouping (UL) to a right-left grouping (RL), which can be seen to perform roughly the same. The values used are highlighted in bold.

Note that the decompression function is not technically speaking the true inverse of $\phi(X^L)$, clearly no such inverse exists since $\phi(X^L)$ is many to one. However, we can regard $\phi^{-1}(H)$ as a "probabilistic inverse" that returns the most probable pre-image of $H$. Our compression function maps points in $\mathcal{K}^5$ to points in $\mathcal{H}^8$. For example, when $k$=300 and $n$=9 we reduce the search space size from $10^{11}$ to $10^6$.

## Results

One of the major concerns when building the DAG is what structure the graph should have. This is an open problem and in general is left as a design decision. We experiment with using two different groupings: grouping right and left sides of the body (RL), and grouping upper and lower body (UL). An overview of both groupings can be seen in Fig. 4.5. Another hyperparameter is the number of latent states for each node. In order to minimize the number of hyperparameters we use the same number of states for all nodes. We experiment on the HumanEVA dataset (Sigal et al., 2010b). We evaluate

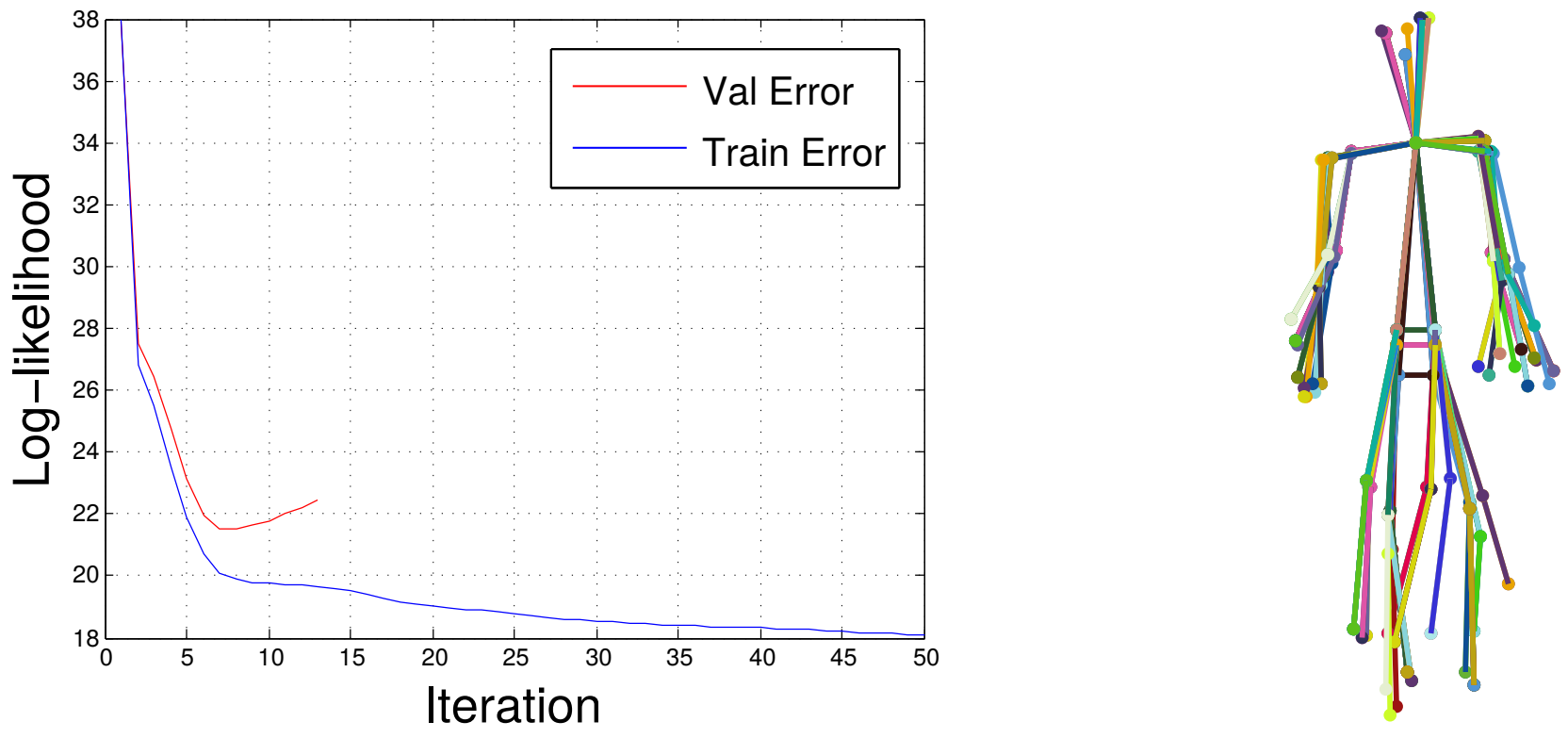

**(a)** Training of the model **(b)** Samples from the model using the method by compressing and decompressing the testing poses to obtain the reconstruction error $\epsilon_{rec}$ defined as

$$\epsilon_{rec}(X^L) = X^L - \phi^{-1}(\phi(X^L)) \ . \tag{4.6}$$

**Figure 4.6: Directed Acylcic Graph (DAG) model training and sampling.** We train the DAG model for the training action with 6 latent states and UL grouping. (a) Log-likelihood of the model for the training and validation splits. (b) Samples from the trained model. As we have trained with all three subjects simultaneously we see three different body sizes among the samples.

We show results for both groupings and number of latent states in Table 4.2. We can see that in this case both groupings perform nearly the same. While it is likely the graphical structure chosen is not optimal, the architecture does not seem to influence to heavily in the results. On the other hand the number of latent states is very important to be able to accurately model human pose. However, adding more latent states increases the size of the model, causing it to overfit. For our experiments we chose 6 latent states with a Upper and Lower body grouping (UL) that seems to provide a reasonable mix of good performance without having too many degrees of freedom.

The results of training the model can be seen in Fig. 4.6. We can see with 6 latent states per node the model overfits fairly quickly. We rely on a validation split in order to avoid the overfitting. Samples taken from the model can be seen in Fig. 4.6b. As we train with all 3 subjects simultaneously, three different body sizes can be identified. Most of the variability is, as expected of the walking action, located at the limb extremities.

The advantage of this model with respect to the linear latent model is clear: we are actually able to model and sample from $p(\mathbf{x})$ through the latent space. Therefore we do not need to perform any additional selection based on anthropomorphism nor other criterion. However, the model still has several disadvantages. Namely we are modelling the pose directly in 3D coordinates. This means that limb lengths may vary for different individuals, i.e., it is not an individual agnostic representation. Additionally the discretization is also not exempt of issues. It creates artefacts in the 3D pose when sampling in addition to increase the number of hyperparameters of the model. Despite this, we believe it is a step forward with respect to using linear latent models for the task of 3D pose generation.

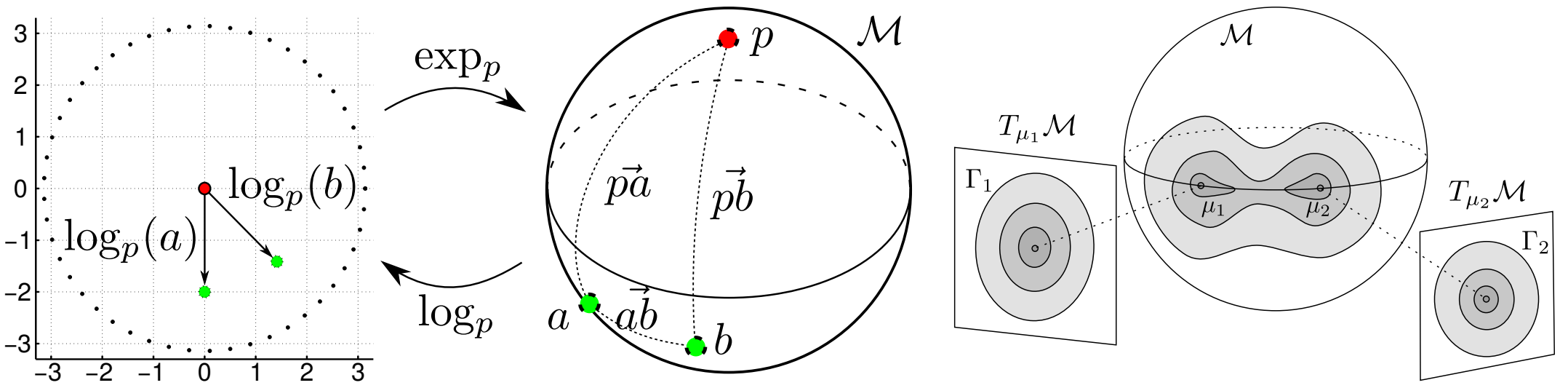

**(a)** $SO(2)$ Geodesics **(b)** Geodesic Finite Mixture Model

**Figure 4.7: Geodesics and Mixture Models.** (a) Representation of geodesics on the $S^2$ manifold. The tangent space ensures that $\|\log_o(x)\|$ is the true geodesic distance of $\vec{ox}$. However, $\|\log_o(a) - \log_o(b)\|$ is not the geodesic distance of $\vec{ab}$. (b) Illustration of the proposed mixture model approach. Each mixture component has its own tangent space, ensuring the consistency of the model while minimizing accuracy loss.

## 4.5 Geodesic Finite Mixture Models

In order to deal with problems that arose when using both the linear latent model and the Directed Acyclic Graph (DAG) model, we investigated into using Riemannian manifolds for representing the pose and modelling directly on the manifold. This allows us to obtain a individual-agnostic model in the continuous domain that can scale well to large datasets. More importantly, this model is consistent with the human pose manifold, i.e., it will only generate poses that have no limb deformation.

It is well known that human motion can be represented as points on a Riemannian manifold (Brubaker et al., 2012; Sommer et al., 2010). We consider the task of obtaining a probabilistic generative model of the poses, that is, approximate the Probability Density Function (PDF) of points on this manifold for a potentially large dataset. As we consider the situation in which the data lies on a *known* Riemannian manifold. We are able to create an efficient completely data-driven algorithm consistent with the manifold, i.e., an algorithm that yields a PDF defined exclusively on the manifold.

A standard procedure to operate on a manifold is to use the logarithmic map to project the data points onto the tangent space of the mean point on the manifold (Fletcher et al., 2004; Sommer et al., 2010; Huckemann et al., 2010). After this linearization, Euclidean statistics are computed and projected back to the manifold using the exponential map. This process is iteratively repeated until convergence of the computed statistics. Unfortunately, while this approximation is effective to model data with a reduced extent, it is prone to fail when dealing with data that covers wide regions of the manifold.

In the proposed finite mixture model, we overcome this limitation by simultaneously considering multiple tangent spaces, distributed along the whole manifold as seen in Fig. 4.7. We draw inspiration on the unsupervised algorithm from (Figueiredo and Jain, 2002), which given data lying in an Euclidean space, automatically computes the number of model components that minimize a message length cost. By representing each component as a distribution on the tangent space at its corresponding mean on the manifold, we are able to generalize the algorithm to Riemannian manifolds and at the same time mitigate the accuracy loss produced when using a single tangent space.

Furthermore, since our model is *semi-parametric*, we can handle an arbitrarily large number of samples. This is in contrast to existing *non-parametric* approaches (Pelletier, 2005) whose complexity grows with the training set size.

In this section we will show that our manifold-based finite mixture model can be effectively used both for non-linear data modeling and regression. In addition, we will show that the modeling of articulated structures can greatly benefit from our approach, allowing for instance, hallucinating subsets of missing joints of the human body without any specific training.

## Method

We next describe our approach, starting with some basic notions on Riemannian geometry and statistics on manifolds. We then integrate these tools in a mixture modeling algorithm to build consistent generative models.

### Manifolds, Geodesics and Tangent Spaces

Manifolds arise naturally in many real-world problems. One of the more well-known is the manifold representing spatial rotations. For example, when studying human motion, it is a common practice to use the spatial rotations of the different body parts to obtain a subject-agnostic representation of the whole body pose. This is usually done with angle representations that have an inherent periodicity and thus are not a vector space. By considering the Riemannian manifold of spatial rotations it is possible to use tangent spaces as a local vector space representation, and use powerful statistical tools based on Euclidean metrics. For an in depth description of Riemannian manifolds we refer the reader to (Boothby, 2003).

Geodesic distances, which we shall denote as $d(\cdot, \cdot)$, are the shortest distance along the manifold between two arbitrary points. This distance is generally not equivalent to an Euclidean one. The tangent space is a local vector space representation where the Euclidean distances between the origin and arbitrary points correspond to the geodesic distances on the manifold. Yet, as seen in Fig. 4.7a, this correspondence does not hold for two arbitrary points. A point $\mathbf{v} \in \mathcal{M}$ on the tangent space $T_{\mathbf{c}}\mathcal{M}$ at $\mathbf{c} \in \mathcal{M}$ can be mapped to the manifold $\mathcal{M}$ and back to the $T_{\mathbf{c}}\mathcal{M}$ by using the exponential and logarithmic maps respectively:

$$\exp_p : \begin{array}{ccc} T_{\mathbf{c}}\mathcal{M} & \longrightarrow & \mathcal{M} \\ \mathbf{v} & \longmapsto & \exp_{\mathbf{c}}(\mathbf{v}) = \mathbf{x} \end{array} \tag{4.7}$$

$$\log_p : \begin{array}{ccc} \mathcal{M} & \longrightarrow & T_{\mathbf{c}}\mathcal{M} \\ \mathbf{x} & \longmapsto & \log_{\mathbf{c}}(\mathbf{x}) = \mathbf{v} \end{array} \tag{4.8}$$

In general there is no closed-form of the $\exp_{\mathbf{c}}$ and $\log_{\mathbf{c}}$ maps for an arbitrary manifold. There are, though, approximations for computing them in Riemannian manifolds (Dedieu and Nowicki, 2005; Sommer et al., 2009). Additionally, efficient closed-form solutions exist for certain manifolds (Said et al., 2007). An interesting mapping for us will be the one between the unit sphere $S^2$ and its tangent space $T_{\mathbf{c}}S^2$. Let $\mathbf{x} = (x_1, x_2, x_3)^\top$, $\mathbf{y} = (y_1, y_2, y_3)^\top$ be two unit spoke directions in $S^2$ and $\mathbf{v} = (v_1, v_2)^\top$ a point in $T_{\mathbf{c}}S^2$. The $\exp_{\mathbf{c}}$ and $\log_{\mathbf{c}}$ maps are in this case:

$$\exp_{\mathbf{c}}(\mathbf{v}) = \mathbf{R}_{\mathbf{c}}^{-1}\left(v_1\frac{\sin\|\mathbf{v}\|}{\|\mathbf{v}\|},\ v_2\frac{\sin\|\mathbf{v}\|}{\|\mathbf{v}\|},\ \cos\|\mathbf{v}\|\right) \tag{4.9}$$

$$\log_p(\mathbf{x}) = \left(y_1\frac{\boldsymbol{\theta}}{\sin\boldsymbol{\theta}},\ y_2\frac{\boldsymbol{\theta}}{\sin\boldsymbol{\theta}}\right) \tag{4.10}$$

where $\mathbf{R}_{\mathbf{c}}$ is the rotation of $\mathbf{c}$ to the north pole, $\|\mathbf{v}\| = (v_1^2 + v_2^2)^{\frac{1}{2}}$, $\mathbf{y} = \mathbf{R}_{\mathbf{c}}\mathbf{x}$ and $\boldsymbol{\theta} = \arccos(y_3)$.

### Statistics on Tangent Spaces

While it is possible to define distributions on manifolds (Pennec, 2006), we shall focus on approximating Gaussian PDFs using the tangent space. Since it is a vector space, we can compute statistics that, by definition of the tangent space, are consistent with the manifold. For instance, the mean of $N$ points $\mathbf{x}_i$ on a manifold can be calculated as (Karcher, 1977)

$$\boldsymbol{\mu} = \arg\min_{\mathbf{c}} \sum_{i=1}^{N} d\left(\mathbf{x}_i,\ \mathbf{c}\right)^2 \ . \tag{4.11}$$

This is optimized iteratively using the $\exp_{\mathbf{c}}$ and $\log_{\mathbf{c}}$ maps,

$$\boldsymbol{\mu}(t+1) = \exp_{\boldsymbol{\mu}(t)}\left(\frac{\delta}{N}\sum_{i=1}^{N}\log_{\boldsymbol{\mu}(t)}\left(\mathbf{x}_i\right)\right) \ , \tag{4.12}$$

until $\|\boldsymbol{\mu}(t+1) - \boldsymbol{\mu}(t)\| < \epsilon$ for some threshold $\epsilon$, with $\delta$ being the step size parameter. Given the mean, it is then possible to estimate the covariance on the tangent space:

$$\boldsymbol{\Sigma} = \frac{1}{N}\sum_{i=1}^{N}\log_{\boldsymbol{\mu}}(\mathbf{x}_i)\log_{\boldsymbol{\mu}}(\mathbf{x}_i)^\top \ . \tag{4.13}$$

Knowing the mean value and the concentration matrix ($\boldsymbol{\Gamma} = \boldsymbol{\Sigma}^{-1}$) we can write the distribution that maximizes entropy on the tangent space as a normal distribution centered on the point $\boldsymbol{\mu} \in \mathcal{M}$, corresponding to the origin ($\boldsymbol{\nu} = 0$) in the tangent space:

$$\mathcal{N}_{\boldsymbol{\mu}}(\boldsymbol{\nu},\ \boldsymbol{\Sigma}^{-1}) = a\exp\left(-\frac{\log_{\boldsymbol{\mu}}(\mathbf{x})^\top\boldsymbol{\Sigma}^{-1}\log_{\boldsymbol{\mu}}(\mathbf{x})}{2}\right) \tag{4.14}$$

where $a$ is normalization term that ensures the normal distribution over the tangent space integrates to unity. If $T_{\boldsymbol{\mu}}\mathcal{M} \equiv \mathbb{R}^D$ this term simplifies to

$$a^{-1} = \sqrt{(2\pi)^D\det(\boldsymbol{\Sigma})} \ . \tag{4.15}$$

As the normalization term is dependent on the tangent space, it is not always easy to obtain. However, it is possible to approximate this normalization factor based on maximum entropy criterion computed directly on the manifold (Pennec, 2006). Since this situation is very unusual, it is not further explored in this thesis.

### Unsupervised Finite Mixture Modeling

Our approach for fitting the mixture model relies on (Figueiredo and Jain, 2002), a variant of the EM algorithm that uses the Minimum Message Length criterion (MML) to estimate the number of clusters and their parameters in an unsupervised manner.

Given an input dataset, this algorithm starts by randomly initializing a large number of mixtures. During the Maximization (M) step, a MML criterion is used to annihilate components that are not well supported by the data. In addition, upon EM convergence, the least probable mixture component is also forcibly annihilated and the algorithm continues until a minimum number of components is reached. (Figueiredo and Jain, 2002) is designed to work with data in an Euclidean space. To use it in Riemannian manifolds, we modify the M-step as follows.

Each mixture component is defined by its mean $\boldsymbol{\mu}_k$ and concentration matrix $\boldsymbol{\Gamma}_k = \boldsymbol{\Sigma}_k^{-1}$ as a normal distribution on its own tangent space $T_{\boldsymbol{\mu}_k}\mathcal{M}$:

$$p(\mathbf{x}|\boldsymbol{\theta}_k) \approx \mathcal{N}_{\boldsymbol{\mu}_k}\left(0,\ \boldsymbol{\Sigma}_k^{-1}\right) \tag{4.16}$$

with $\boldsymbol{\theta}_k = (\boldsymbol{\mu}_k, \boldsymbol{\Sigma}_k)$. Remember that the mean $\boldsymbol{\mu}_k$ is defined on the manifold $\mathcal{M}$, while the concentration matrix $\boldsymbol{\Gamma}_k$ is defined on the tangent space $T_{\boldsymbol{\mu}_k}\mathcal{M}$ at the mean $\boldsymbol{\nu}_k = 0$.

Also note that the dimensionality of the space embedding the manifold is larger than the actual dimension of the manifold, which in its turn is equal to the dimension of the tangent space. That is, $\dim(\text{Embedding}(\mathcal{M})) > \dim(T_{\mathbf{c}}\mathcal{M}) = \dim(\mathcal{M}) = D$. This dimensionality determines the total number of parameters $D_{\boldsymbol{\theta}}$ specifying each component, and, as we will explain below, plays an important role during component annihilation process. For full covariance matrices it can be easily found that $D_{\boldsymbol{\theta}} = D + D(D+1)/2$.

We next describe how the EM algorithm is extended from Euclidean to Riemmanian manifolds. Specifically, let us assume that $K$ components survived after iteration $t-1$. Then, in the E-step we compute the *responsibility* that each component $k$ takes for every sample $\mathbf{x}_i$:

$$w_k^{(i)} = \frac{\alpha_k(t-1)p(\mathbf{x}_i|\boldsymbol{\theta}_k(t-1))}{\sum_{k=1}^{K}\alpha_k(t-1)p(\mathbf{x}_i|\boldsymbol{\theta}_k(t-1))}\ , \tag{4.17}$$

for $k = 1, \ldots, K$ and $i = 1, \ldots, N$, and where $\alpha_k(t-1)$ are the relative weights of each component $k$.

In the M-step we update the weight $\alpha_k$, the mean $\boldsymbol{\mu}_k$ and covariance $\boldsymbol{\Sigma}_k$ for each of the components as follows:

$$\alpha_k(t) = \frac{1}{N}\sum_{i}^{N} w_k^{(i)} = \frac{w_k}{N} \tag{4.18}$$

$$\boldsymbol{\mu}_k(t) = \arg\min_{\mathbf{c}} \sum_{i=1}^{N} d\left(\frac{N}{w_k} w_k^{(i)}\mathbf{x}^{(i)},\ \mathbf{c}\right)^2 \tag{4.19}$$

$$\boldsymbol{\Sigma}_k(t) = \frac{1}{w_k}\sum_{i=1}^{N}\left(\log_{\boldsymbol{\mu}_k(t)}(\mathbf{x}^{(i)})\right)\left(\log_{\boldsymbol{\mu}_k(t)}(\mathbf{x}^{(i)})\right)^{\top} w_k^{(i)} \tag{4.20}$$

After each M-step, we follow the same annihilation criterion as in (Figueiredo and Jain, 2002), and eliminate those components whose accumulated responsibility $w_k$ is below a $D_{\boldsymbol{\theta}}/2$ threshold. A score for the remaining components based on the Minimum Message

Length is then computed. This EM process is repeated until the convergence of the score or until reaching a minimum number of components $K_{min}$. If this number is not reached, the component with the least responsibility is eliminated (even if it is larger than $D_{\boldsymbol{\theta}}/2$) and the EM process is repeated. Finally, the configuration with minimum score is retained (see (Figueiredo and Jain, 2002) for details), yielding a resulting distribution with the form

$$p(x|\boldsymbol{\theta}) = \sum_{k=1}^{K} \alpha_k p(\mathbf{x}|\boldsymbol{\theta}_k) \ . \tag{4.21}$$

To use the generative model in regression tasks, we need to split the mix into two components, $\mathbf{x} = (\mathbf{x}_A, \mathbf{x}_B)$ with $\boldsymbol{\mu}_k = (\boldsymbol{\mu}_{k,A}, \boldsymbol{\mu}_{k,B})$ and $\boldsymbol{\Gamma}_k = \begin{bmatrix} \boldsymbol{\Gamma}_{k,A} & \boldsymbol{\Gamma}_{k,AB} \\ \boldsymbol{\Gamma}_{k,BA} & \boldsymbol{\Gamma}_{k,B} \end{bmatrix}$. The regression function can be written as:

$$p(\mathbf{x}_A|\mathbf{x}_B, \boldsymbol{\theta}) = \frac{p(\mathbf{x}_A, \mathbf{x}_B|\boldsymbol{\theta})}{p(\mathbf{x}_B|\boldsymbol{\theta}_B)} = \frac{\sum_{k=1}^{K} \alpha_k p(\mathbf{x}_B|\boldsymbol{\theta}_{k,B}) p(\mathbf{x}_A|\mathbf{x}_B, \boldsymbol{\theta}_k)}{\sum_{k=1}^{K} \alpha_k p(\mathbf{x}_B|\boldsymbol{\theta}_{k,B})} \ . \tag{4.22}$$

Observe that this is a new mixture model $p(\mathbf{x}_A|\mathbf{x}_B, \boldsymbol{\theta}) = \sum_{k=1}^{K} \pi_k p(\mathbf{x}_A|\mathbf{x}_B, \boldsymbol{\theta}_k)$, with weights:

$$\pi_k = \frac{\alpha_k p(\mathbf{x}_B|\boldsymbol{\theta}_{k,B})}{\sum_{j=1}^{K} \alpha_j p(\mathbf{x}_B|\boldsymbol{\theta}_{j,B})} \ , \tag{4.23}$$

and $p(\mathbf{x}_A|\mathbf{x}_B, \boldsymbol{\theta}_k) = \mathcal{N}_{\boldsymbol{\mu}_{k,A|B}}(\boldsymbol{\nu}_{k,A|B}, \boldsymbol{\Gamma}_{k,A|B})$, where the mean and concentration matrix can be found to be:

$$\boldsymbol{\nu}_{k,A|B} = \boldsymbol{\Gamma}_{k,AB} \boldsymbol{\Gamma}_{k,B}^{-1} \log_{\boldsymbol{\mu}_k,B}(\mathbf{x}_B), \tag{4.24}$$

$$\boldsymbol{\Gamma}_{k,A|B} = \boldsymbol{\Gamma}_{k,A} - \boldsymbol{\Gamma}_{k,AB} \boldsymbol{\Gamma}_{k,B}^{-1} \boldsymbol{\Gamma}_{k,BA} \ . \tag{4.25}$$

### Results

We evaluated our approach on the Human3.6M Dataset (Ionescu et al., 2011, 2014) which has many subjects performing different activities. As is common practice (Ionescu et al., 2014), we model the poses in an individual-agnostic way by normalizing all the limbs by their segments and only modeling the rotation. The rotation of each joint is represented on the $S^2$ manifold which then are arranged in a tree structure (Tournier et al., 2009; Sommer et al., 2010). We note that other manifolds do exist such as one defined by using forward kinematics (Hauberg et al., 2012), however, this approach does not have closed form solutions of the $\exp_{\mathbf{c}}$ and $\log_{\mathbf{c}}$ operators. A simplified model of the human with 15 joints is used because the dimensionality of the model greatly increases the amount of the required data. This results in a 24-dimensional manifold, represented by a block-diagonal covariance matrix with 46 non-zero elements.

We divide the dataset into its five base scenarios: $S_1$, upper body movements; $S_2$, full body upright variations; $S_3$, walking variations; $S_4$, variations while seated on a chair; and $S_5$, sitting down on the floor. We train on subjects 1, 5, 6, 7, 8, 9 and 11 for each scenario using only the first of the actions in each scenario and test on the second one. The algorithm uses 2000 initial clusters and $\epsilon_s = 0.999$ for all scenarios.

We evaluate on the task of hallucinating limbs, i.e., probabilistically recovering missing parts $\mathbf{x}_A$ from the rest of the body $\mathbf{x}_B$. This is done by obtaining the expectation

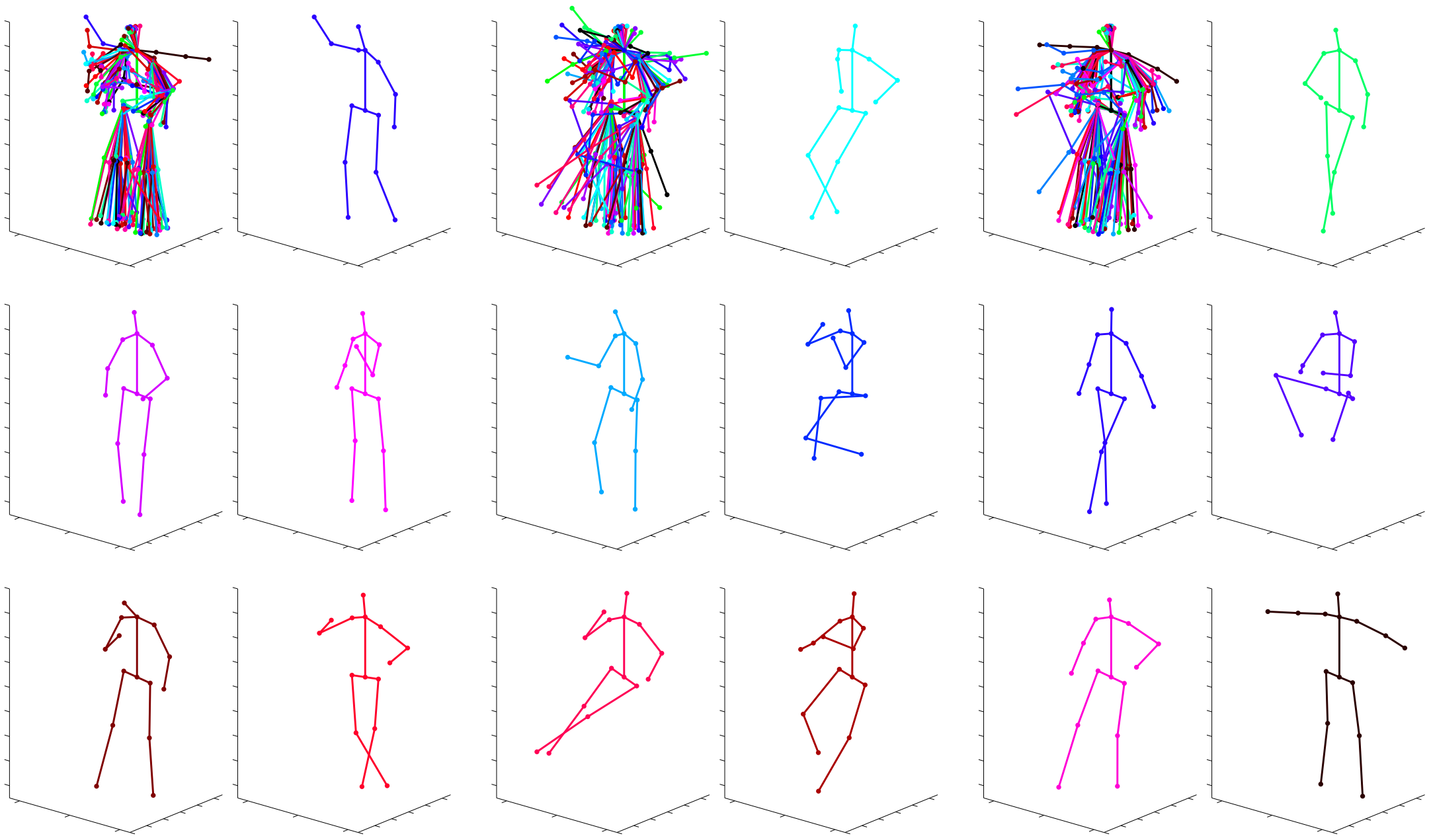

**(a)** Scenario 1 **(b)** Scenario 2 **(c)** Scenario 3

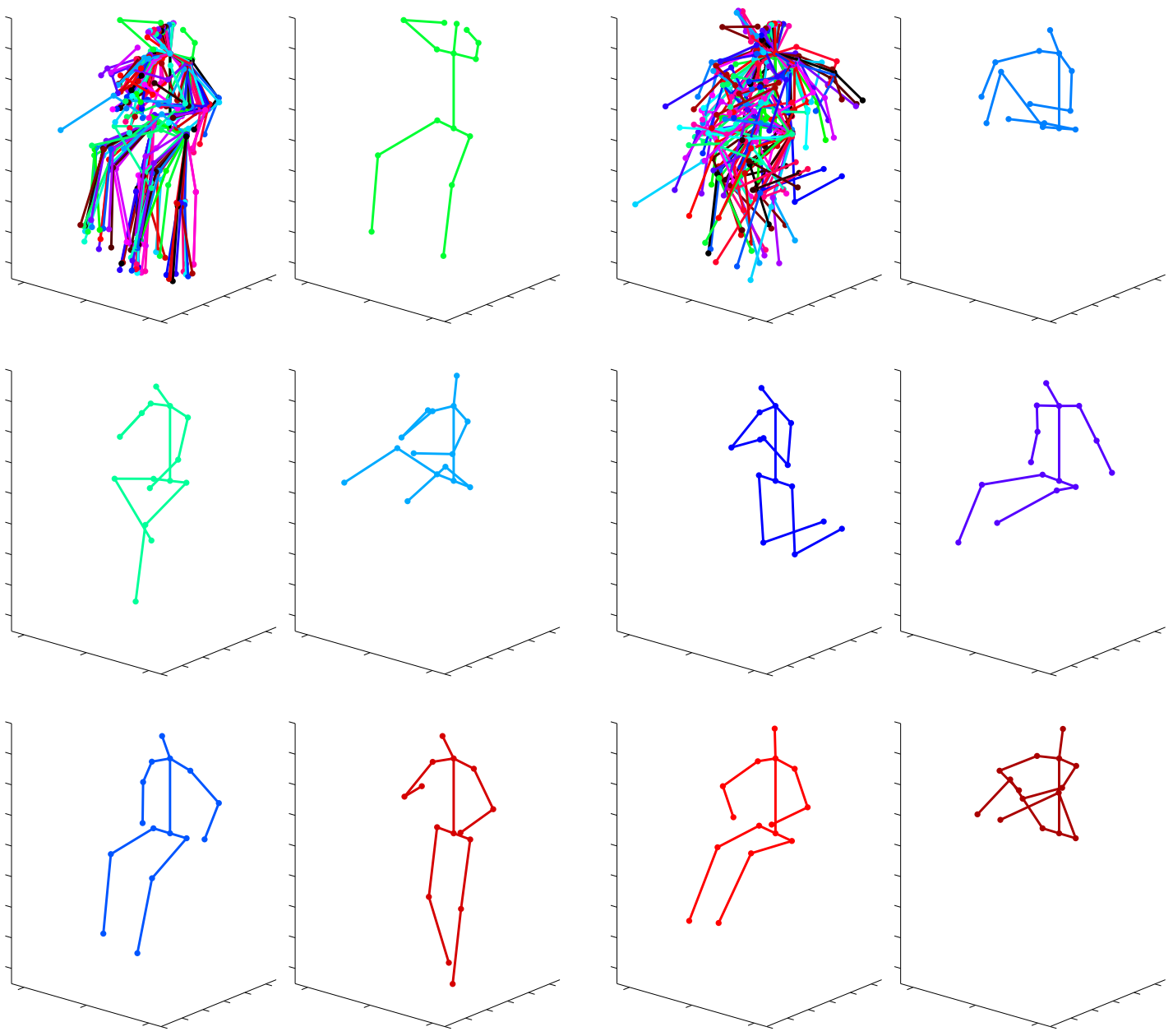

**(d)** Scenario 4 **(e)** Scenario 5

**Figure 4.8: Sampling from the Geodesic Finite Mixture Model**. We sample from the different mixtures trained on each of the scenarios. For visualization purposes, we perform 3D reconstruction using the limb lengths of subject 1. Note that the mixtures themselves represent poses and are independent of limb lengths. For each scenario, in the top left 30 samples are presented along with 5 particular examples. We can see there is a lot of diversity that our model is able to capture.

| | Scenario 1 | | | Scenario 2 | | | Scenario 3 | | |
|---|---|---|---|---|---|---|---|---|---|
| | **MGJE** | **MJE** | **MLLE** | **MGJE** | **MJE** | **MLLE** | **MGJE** | **MJE** | **MLLE** |
| **GFMM** | 0.446 | 105.8 | 0.0 | 0.468 | 110.1 | 0.0 | 0.349 | 81.7 | 0.0 |
| **vMF** | 0.481 | 114.5 | 0.0 | 0.568 | 134.8 | 0.0 | 0.470 | 110.2 | 0.0 |
| **1-TM** | 0.522 | 123.0 | 0.0 | 0.640 | 148.7 | 0.0 | 0.535 | 124.9 | 0.0 |
| **GMM** | 1.111 | 103.1 | 19.0 | 1.167 | 106.6 | 27.5 | 1.152 | 77.6 | 11.3 |

| | Scenario 4 | | | Scenario 5 | | |
|---|---|---|---|---|---|---|
| | **MGJE** | **MJE** | **MLLE** | **MGJE** | **MJE** | **MLLE** |
| **GFMM** | 0.458 | 108.2 | 0.0 | 0.597 | 135.7 | 0.0 |
| **vMF** | 0.496 | 118.0 | 0.0 | 0.698 | 162.3 | 0.0 |
| **1-TM** | 0.548 | 130.2 | 0.0 | 0.765 | 175.1 | 0.0 |
| **GMM** | 1.272 | 101.0 | 14.2 | 1.401 | 127.3 | 24.8 |

**Table 4.3: Comparing the Geodesic Finite Mixture Model (GFMM) with other approaches.** We compare against von Mises Fisher distributions (vMF), a single tangent space (1-TM) and an Euclidean Gaussian Mixture Model (GMM) in the limb reconstruction task. We show results using three different metrics: Mean Geodesic Joint Error (MGJE), Mean Joint Error (MJE), and Mean Limb Length Error (MLLE).

of the posterior $p(\mathbf{x}_A|\mathbf{x}_B, \boldsymbol{\theta})$ using a combination of the conditional means (Eq. (4.24)) at each tangent space with

$$\mathbf{x}_A = \arg\min_{\mathbf{c}} \sum_{k=1}^{K} d\left(\pi_k \boldsymbol{\nu}_{k,A|B},\ \mathbf{c}\right)^2 \tag{4.26}$$

We compare our method with von Mises Fisher distributions (vMF), a single tangent space (1-TM) and a Gaussian Mixture Model in Euclidean space (GMM). To compensate the fact that our approach is individual-agnostic, was trained with the GMM with same individual that was being tested. We define three metrics: Mean Geodesic Joint Error (MGJE), that accounts for the rotation error; Mean Joint Error (MJE), representing the Euclidean distance with the ground truth in mm; and Mean Limb Length Error (MLLE), which is the amount of limb length deformation in mm. All four models are trained using exactly the same parameters for all scenarios. The results are summarized in Table 4.3. Observe that while the MJE is similar for both our method and GMM, our approach has roughly one third of the MGJE error. Additionally, as our model is coherent with the manifold, there is no limb deformation (MLLE). Our model also consistently outperforms the vMF and 1-TM approaches, which once again shows that multiple tangent spaces with Gaussians can better capture the underlying distribution. The 1-TM approach has the worst MJE error but we can see it has a MGJE much closer to our model than the GMM. We also outperform the vMF model. We show four specific examples of the estimated distributions for our model in Fig. 4.9.

In summary we have proposed a novel data-driven approach for modeling the probability density function of data located on a Riemannian manifold. By using a mixture

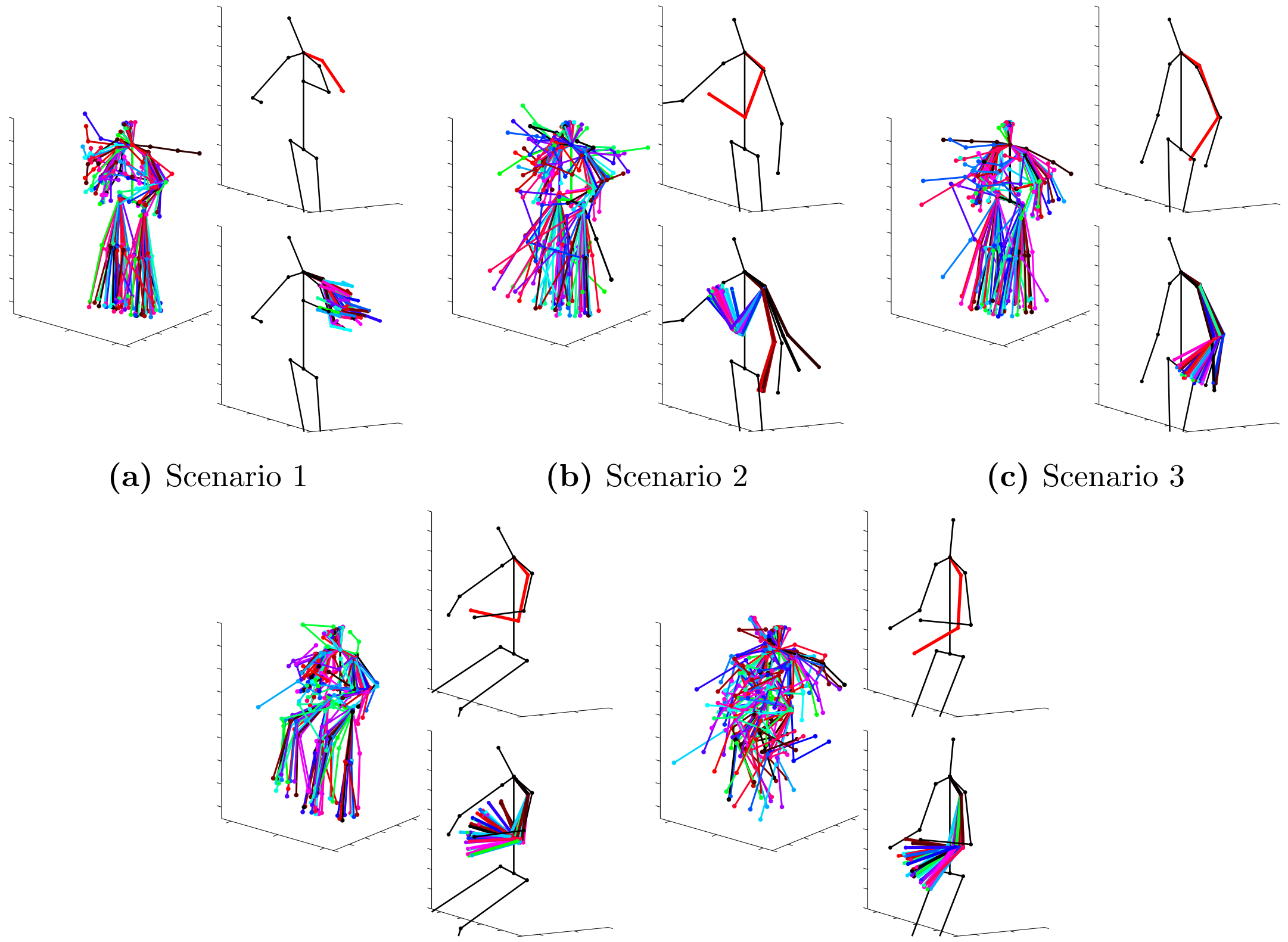

**(a)** Scenario 1 **(b)** Scenario 2 **(c)** Scenario 3

**(d)** Scenario 4 **(e)** Scenario 5

**Figure 4.9: Human pose regression with the GFMM.** We use the trained mixtures to perform regression on the left arm (3 limbs) of subjects on the Human3.6M dataset. On the left we plot 30 random samples from the full mixture. On the top-right we plot 30 random samples from the mixture of the conditional distribution corresponding to a particular frames. On the bottom-right the mean of the mixture is shown in red. The black lines indicate the input sample. As shown in Scenario 4, the distribution can be multimodal, however, one of the modes is very close to the true limb position.

of distributions, each with its own tangent space, we are able to ensure the consistency of the model while avoiding most of the linearization error caused by using one single tangent space. The approach has been experimentally validated on various synthetic examples that highlight their ability to both correctly approximate manifold distributions and discover the underlying data structure. Furthermore, the approach has been tested on a large and complex dataset, where it is shown to outperform the traditionally used Euclidean Gaussian Mixture Model, von Mises distributions and using a single tangent space in a regression task.

## 4.6 Summary

In this chapter we have presented three generative models of varying complexity and expressiveness that are suitable for 3D human pose. The linear latent model finds a linear transformation of the data that maximizes the covariance of the training data.

The Directed Acyclic Graph (DAG) model on the other hand defines a joint distribution over pose and a set of latent variables $p(X^L, H)$, and calculates expectations with dynamic programming. Finally the Geodesic Finite Mixture Model (GFMM) directly represents the data $p(\mathbf{x})$ with a Gaussian Mixture Model in which each mixture is defined on a separate local linear representation of the pose manifold.

Each have different strengths and weaknesses making them adequate for different tasks. The linear latent model is very simple and fast to calculate, however, it does not necessarily produce realistic poses. The DAG model is able to model the poses efficiently, but it relies on using discrete states for the poses. The GFMM on the other hand is a fully continuous generative model that is very efficient to sample on, but it relies on having a large amount of data to be able to train good models. In general, the optimal model for a particular framework or problem will depend heavily on many different constraints such as time complexity, or required accuracy. The approaches we have discussed in this chapter have very different properties that allow them to cover most if not all usage cases.

In the next chapter we will show full end-to-end systems that are able to exploit these different generative models to predict 3D human poses from single images.

# Chapter 5

# 3D Human Pose Estimation

Estimating the 3D human pose using a single image is a severely under-constrained problem, as many different body poses may have very similar image projections. In this chapter we shall explain two different methods for approaching this problem. The first one consists of estimating the 2D human pose, and then propagating the uncertainty of the 2D estimation to 3D, where the hypothesis can be disambiguated based on anthropomorphism to obtain a single solution. The second approach is based on jointly estimating the 2D and 3D pose using a combination of 2D discriminative body part detectors with a strong 3D generative model. By iteratively generating hypothesis and then ranking them based on how well they match the detector responses, we are able to find a single 2D and 3D human pose estimation.

## 5.1 Introduction

Computer vision has always had a focus on problems centered around humans and their environment. One of the more traditional computer vision problems has been that of detecting humans in static images (Dalal and Triggs, 2005). This problem was then extended to also estimating the 2D human pose from these images in which the Pictorial Structures (Felzenszwalb and Huttenlocher, 2005) and its variations (Ramanan, 2006; Andriluka et al., 2009; Yang and Ramanan, 2011; Andriluka et al., 2012) played a fundamental role. However, as humans are not 2D structures, many ambiguities can not be resolved with pure 2D models, leading to the advent of the 3D human pose estimation problem. In order to simplify the problem, most approaches either tackle the 3D human pose tracking problem (Lawrence and Moore, 2007; Urtasun et al., 2006; Zhao et al., 2011; Andriluka et al., 2010), the multiview 3D pose estimation problem (Bo and Sminchisescu, 2010; Burenius et al., 2013; Kazemi et al., 2013) (shown in Fig. 5.1) or a combination of both (Daubney and Xie, 2011; Yao et al., 2012a). On the other hand the single image monocular problem has only recently gained popularity.

In order to disambiguate the problem, one common approach is to assume that an underlying deformation model is available. Linear models (Balan et al., 2007) or sophisticated dimensionality reduction methods have been used for this purpose (Lawrence and Moore, 2007; Sminchisescu and Jepson, 2004; Urtasun et al., 2006). Alternatively, other techniques have focused on learning the mapping from 2D image observations to 3D poses (Okada and Soatto, 2008; Rogez et al., 2008; Sigal et al., 2009). In any event, most of these generative and discriminative approaches rely on the fact that 2D features, such as edges, silhouettes or joints may be easily obtained from the image.

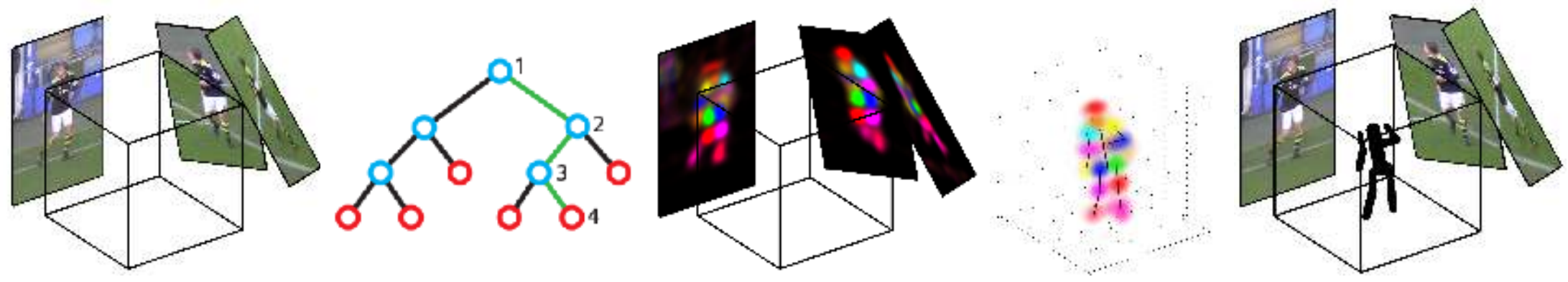

**Figure 5.1: Example of a multiview 3D pose estimation approach.** In this approach a random forest is first used to classify each pixel in each image as belonging to a part or the background. The results are then back-projected to a 3D volume. Corresponding mirror symmetric parts are found across views by using latent variables. Finally, a part-based model is used to estimate the 3D pose Figure reproduced from (Kazemi et al., 2013).

In this thesis, we get rid of the strong assumption that data association may be easily achieved, and propose two novel approaches for estimating the 3D pose of a person from a single image acquired with a calibrated but potentially moving camera. For this purpose we use a HOG-based discriminative appearance model trained independently on the PARSE dataset for 2D human pose estimation (Yang and Ramanan, 2011). By training the 2D appearance model independently, we are able to avoid overfitting the model to the 3D human pose estimation datasets that have very low appearance variability. Poor generalization is a general trait of models with appearance trained exclusively on the 3D dataset (Sminchisescu and Jepson, 2004).

The first approach we present consists of initially using a 2D human pose estimation algorithm (Yang and Ramanan, 2011) to obtain a noisy estimation of the 2D pose. Then by combining a linear latent pose model with a linear camera projection model, we are able to project the noisy 2D pose estimation into the latent space with a closed-form formulation to obtain a large set of pose hypothesis. As the linear latent model is not a probabilistic model, we rely on an anthropomorphism classifier to disambiguate the resulting poses to obtain a final solution. By exploiting linear formulations and fast classifiers, the algorithm is extremely efficient with the runtime dominated by the 2D human pose estimation. The downside of this approach is the fact it is a pipelined approach, i.e., if the 2D pose estimation is incorrect, the 3D pose estimation will also be incorrect.

Our second approach relaxes the hypothesis that we have a 2D human pose estimation and instead relies directly on a bank of discriminative independent 2D body part detectors. We use a Bayesian formulation very similar to the pictorial structures model which consists of the aforementioned 2D body part detectors in conjunction with a strong 3D pose generative model. In particular we use a Directed Acyclic Graph (DAG) to model the pose. By alternatively sampling from the generative model and then evaluating using the discriminative model we are able to jointly estimate the 2D and 3D pose.

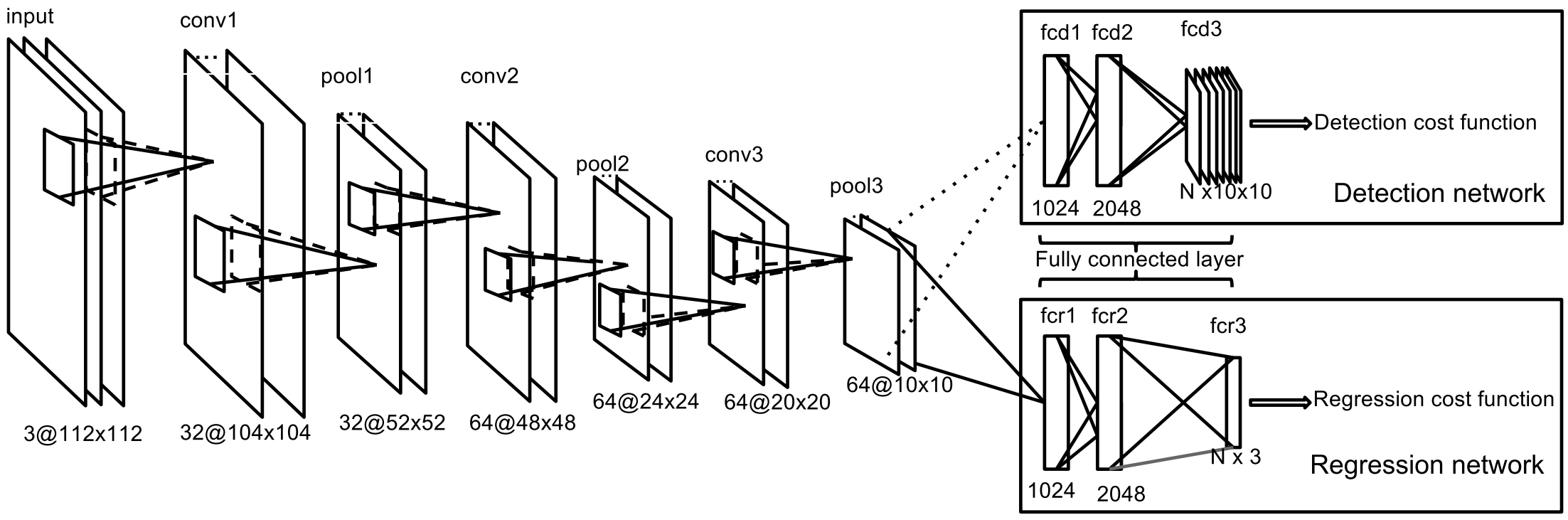

**Figure 5.2: Convolutional Neural Network for 3D Pose Estimation.** The network is initially pre-trained for the 2D human detection task using layers $fcd1$, $fcd2$, and $fcd3$. Afterwards training is continued for the 3D human pose estimation task, which is treated as a regression problem and uses layers $fcr1$, $fcr2$, and $fcr3$. Figure reproduced from (Li and Chan, 2014).

We evaluate both of our approaches numerically on the HumanEva dataset (Sigal et al., 2010b) and qualitatively on the TUD Stadtmitte sequence (Andriluka et al., 2010). Results for both approaches are competitive with the state-of-the-art despite our relaxation of restrictions.

## 5.2 Related work

Without using prior information, monocular 3D human pose estimation is known to be an ill-posed problem. In order to be disambiguate between the possible solutions, many methods to favor the most likely shapes have been proposed.

One of the most straightforward approaches consists of modeling the pose deformations as linear combinations of modes learned from training data (Balan et al., 2007). Since linear models are prone to fail in the presence of non-linear deformations, more accurate dimensionality reduction approaches based on spectral embedding (Sminchisescu and Jepson, 2004), Gaussian Mixtures (Howe et al., 1999) or Gaussian Processes (Urtasun et al., 2006; Lawrence and Moore, 2007; Zhao et al., 2011) have been proposed. However, these approaches rely on good initializations, and therefore, they are typically used in a tracking context. Other approaches follow a discriminative strategy and use learning algorithms such as support vector machines, mixtures of experts or random forest to directly learn the mappings from image evidence to the 3D pose space (Agarwal and Triggs, 2006; Okada and Soatto, 2008; Rogez et al., 2008; Sigal et al., 2009).

Most of the aforementioned solutions, though, oversimplify the 2D feature extraction problem, and typically rely on background subtraction approaches or on the fact that image evidence, such as edges or silhouettes, may be easily obtained from an image, or even assume known 2D (Salzmann and R.Urtasun, 2010; Ramakrishna et al., 2012).

With regard to the problem of directly predicting 2D poses on images, we find that one of the most successful methods is the pictorial structure model (Felzenszwalb and Huttenlocher, 2005) (later extended to the deformable parts model (Felzenszwalb et al., 2008)), which represents objects as a collection of parts in a deformable configuration and allows for efficient inference. Modern approaches detect each individual part using strong detectors (Andriluka et al., 2009; Singh et al., 2010; Tian and Sclaroff, 2010; Yang and Ramanan, 2011) in order to obtain good 2D pose estimations. The deformable parts model has also been extended to use 3D models for 3D viewpoint estimation of rigid objects (Pepik et al., 2012).

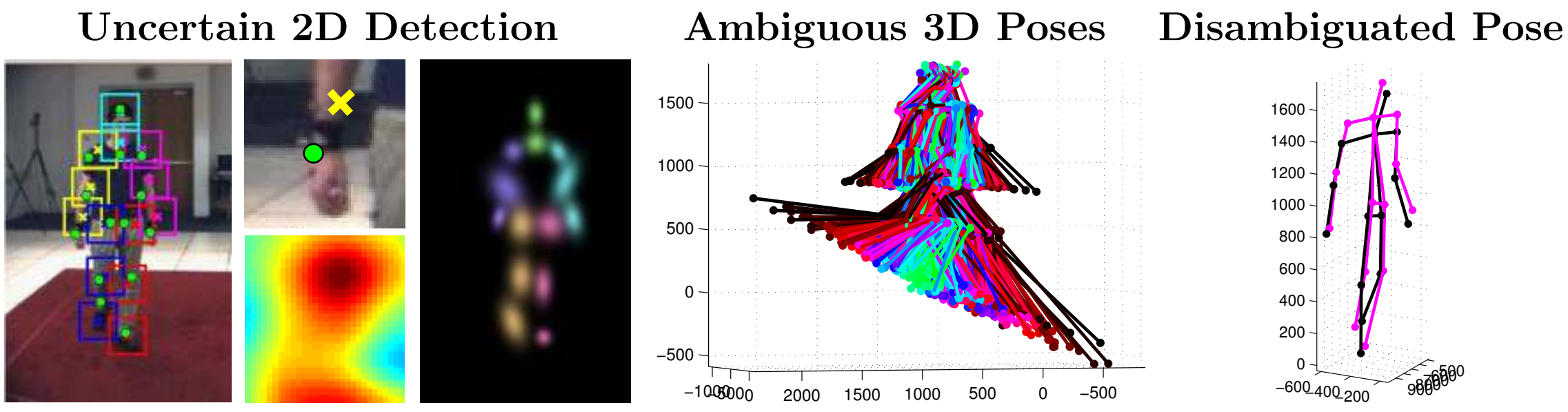

**Figure 5.3: 3D human pose estimation from noisy observations.** The left image shows the bounding box results of a body part detector and green dots indicate the true position of the joints. Note, in the middle, how the bounding box centers do not match the joint positions. Using the heat map scores of the classifier we represent the output of the 2D detector by Gaussian distributions, as shown on the right. Using the distribution of all the joints we sample the solution space and propose an initial set of ambiguous poses. By simultaneously imposing geometric and kinematic constraints that ensure the anthropomorphism, we are able to pick an accurate 3D pose (shown in magenta on the right) very similar to the ground truth (black).

Recently there has been an explosion in deep network approaches in the computer vision community that have made their way to 3D human pose estimation from single images (Li and Chan, 2014) using the architecture shown in Fig. 5.2. However, in general most works still focus on the 2D human pose estimation problem (Jain et al., 2014a; Toshev and Szegedy, 2014; Jain et al., 2014b). Currently 3D human pose estimation using deep networks suffers from an overfitting to the unnatural images of the training dataset. That is, in order to obtain a 3D ground truth, the subjects wear special clothing in conjunction with markers used in the 3D capture system. Appearance models learnt on these images do not generalize well to natural images. In contrast we propose using 2D body part detectors trained independently on natural images, which allow our appearance model to generalize much better.

## 5.3 3D Pose Estimation from Noisy 2D Observations

In order to robustly retrieve 3D human poses we proposed a new approach in which noisy observations are modeled as Gaussian distributions in the image plane and propagated forward to the shape space as shown in Fig. 5.3. This yields tight bounds on the solution space, which we explore using a probabilistic sampling strategy that guarantees the satisfaction of both geometric and anthropomorphic constraints. To favor efficiency, the exploration is performed hierarchically, starting from relatively lax and computationally efficient constraints up to highly restrictive and costly ones, until one single shape sample is retained. Our methodology outperforms approaches that optimize using only geometric constraints.

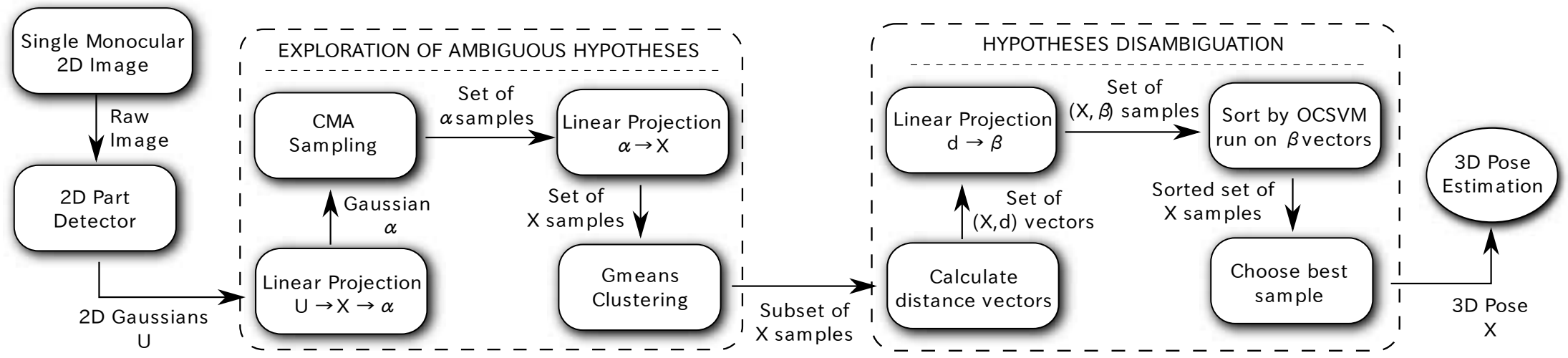

**Figure 5.4: Flowchart of the method for obtaining 3D human pose from single images.**

Drawing inspiration from (Andriluka et al., 2010; Sigal and Black, 2006) we propose retrieving 3D poses from the 2D body part positions estimated by state-of-the-art detectors (Andriluka et al., 2009; Felzenszwalb et al., 2008; Singh et al., 2010; Tian and Sclaroff, 2010; Yang and Ramanan, 2011). Although these detectors require a much reduced number of training samples, as they individually train each of the parts, they have shown impressive results in a wide range of challenging scenarios. However, their solutions have an associated uncertainty which, combined with the inherent ambiguity of the single view 3D detection, may lead to large errors in the estimated 3D shape. This is addressed in (Sigal and Black, 2006) by restricting the method to highly controlled settings, and in (Andriluka et al., 2010) by imposing temporal consistency. Other approaches (Guan et al., 2009; Salzmann and R.Urtasun, 2010; Taylor, 2000) guarantee the single frame solution, but simplify the 2D detection process by either manually clicking the position of the 2D joints or directly using the ground truth values obtained from motion capture systems.

In contrast, the approach we propose naturally deals with the uncertain observations of off-the-shelf body part detectors by modeling the position of each body part using a Gaussian distribution that is propagated to the shape space. This sets bounds on the solution search space, which we exhaustively explore to seek for the 3D pose configuration that best satisfies geometric (reprojection and length) and kinematic (anthropomorphic) constraints. To the best of our knowledge, (Daubney and Xie, 2011) is the only approach that has previously considered noisy observations, but only those related to the root node and not to all the nodes, as we do. In addition, the mentioned work imposes temporal constraints, while we are able to estimate the 3D pose using one single frame.

## Method

Figure 5.4 outlines our approach, which can be split into three major parts: 2D part detection, stochastic exploration of ambiguous hypotheses and disambiguation. The 2D body part estimation is based on the state-of-the-art detector (Yang and Ramanan, 2011) which is adapted to our usage by obtaining information from the classifier heatmaps to provide local 2D Gaussian inputs. Following (Moreno-Noguer et al., 2010), this uncertainty is propagated from the image plane to the shape space, thus reducing the size of the search space. We then use stochastic sampling to efficiently explore this region and propose a set of hypotheses that satisfy both reprojection and length constraints. This set of hypotheses must then be disambiguated by using some additional criteria. We show that only minimizing the reprojection and length errors does not generally give the best results and propose a new method based on coordinate-free geometry to help disambiguate while ensuring anthropomorphic-like shapes.

### 2D Body Part Detection

For body part detection we used (Yang and Ramanan, 2011) which learns a mixture-of-parts tree model encoding both co-occurrence and spatial relations. Each part is modeled as a mixture of HOG-based filters that account for the different appearances the part can take due to, for example, viewpoint change or deformation. Since the parts model is a tree, inference can be efficiently done using dynamic programming, even for a significant number of parts. The output of the detector is a bounding box for each body part, which we convert to a Gaussian distribution by computing the covariance matrix of the classification scores within the box. This is done because the method we propose below to estimate the 3D pose takes as input probability distributions.

### Estimating Ambiguous Solutions

The Gaussian distributions of each body part will be propagated to the shape space and used to propose a set of 3D hypotheses that both reproject correctly onto the image and retain the inter-joint distances of training shapes. However, due to the errors in the estimation of the 2D part location, there is no guarantee that minimizing these errors will yield the best pose estimate. We will show that this requires applying additional anthropomorphic constraints.

The approach we use to propagate the error and propose ambiguous solutions is inspired in (Moreno-Noguer et al., 2010), originally applied to non-rigid surface recovery. However, note that dealing with 3D human poses has an additional degree of complexity, because most joints can only be linked to two other joints. In contrast, when dealing with triangulated surfaces, each node is typically linked to six nodes. Therefore, the set of feasible human body configurations is much larger than the set of surface configurations. This will require using more sophisticated machinery such as integrating kinematic constraints within the process.

We start out by assuming the matrix $\mathbf{A}$ of internal camera parameters to be known and that the 3D points of the $n_v$ joints are expressed in the camera reference frame. We can then write the projection constraints of 3D points $\mathbf{x} = [\mathbf{p}_1^{\mathrm{T}}, \cdots, \mathbf{p}_{n_v}^{\mathrm{T}}]$ to their 2D correspondences $\mathbf{u} = [\mathbf{u}_1^{\mathrm{T}}, \cdots, \mathbf{u}_{n_v}^{\mathrm{T}}]_i$ as:

$$w_i \begin{bmatrix} \mathbf{u}_i \\ 1 \end{bmatrix} = \mathbf{A}\mathbf{p}_i = \begin{bmatrix} \mathbf{A}_{2\times 3} \\ \mathbf{a}_3^{\mathrm{T}} \end{bmatrix} \mathbf{p}_i \ , \tag{5.1}$$

where $w_i$ is a projective scalar, $\mathbf{A}_{2\times 3}$ are the first two rows of $\mathbf{A}$, and $\mathbf{a}_3^{\mathrm{T}}$ is the last one. Since from the last row we have $w_i = \mathbf{a}_3^{\mathrm{T}}\mathbf{p}_i$, we can write:

$$(\mathbf{u}_i \mathbf{a}_3^{\mathrm{T}} - \mathbf{A}_{2\times 3})\mathbf{p}_i = 0 \ . \tag{5.2}$$

For each 3D-to-2D correspondence we have 2 linear constraints on $\mathbf{p}_i$. As we have $n_v$ correspondences, one for each joint, the $2n_v$ constraints can be written as a linear system

$$\mathbf{M}\mathbf{x} = 0 \; , \tag{5.3}$$

where $\mathbf{M}$ is a $2n_v \times 3n_v$ matrix obtained from the known values of $\mathbf{u}$, and $\mathbf{A}$.

We also assume we have trained a linear latent model with $n_m$ dimensions and thus have:

$$\mathbf{x} = \mathbf{x}_0 + \mathbf{Q}\boldsymbol{\alpha} \; , \tag{5.4}$$

where $\mathbf{x}_0$ is the mean shape, $\mathbf{Q}$ is the $n_v \times n_m$ set of basis of the linear latent model, and $\boldsymbol{\alpha}$ are the $n_m$ weights in the latent space.

We can combine both the linear projection model from Eq.(5.3) and the linear latent model from Eq.(5.4) to obtain

$$\mathbf{M}\mathbf{Q}\boldsymbol{\alpha} + \mathbf{M}\mathbf{x}_0 = 0 \; , \tag{5.5}$$

in which any set of weights $\boldsymbol{\alpha}$ that is a solution will project at the right set of 2D correspondences $\mathbf{u}$.

We must now propagate the 2D Gaussian distributions found on the camera plane to the $\boldsymbol{\alpha}$ latent space weights. Following (Moreno-Noguer et al., 2010), the mean of this subspace can be computed as the least-squares solution of Eq. (5.5),

$$\boldsymbol{\mu}_\alpha = (\mathbf{B}^\mathrm{T}\mathbf{B})^{-1}\mathbf{B}^\mathrm{T}\mathbf{b} \; , \tag{5.6}$$

where $\mathbf{B} = \mathbf{M}\mathbf{Q}$ is a $2n_v \times n_m$ matrix and $\mathbf{b} = -\mathbf{M}\mathbf{x}_0$ is a $2n_v$ vector. The components of $\mathbf{B}$ and $\mathbf{b}$ are linear functions of the known parameters $\mathbf{u}_i$, $\mathbf{Q}$ and $\mathbf{A}$. The same can be done for the $2n_v \times 2n_v$ covariance matrix $\boldsymbol{\Sigma}_u$ built using the covariances $\boldsymbol{\Sigma}_{\mathbf{u}_i}$ of each body part. Its propagation yields a $n_m \times n_m$ matrix $\boldsymbol{\Sigma}_\alpha$ on the modal weights space,

$$\boldsymbol{\Sigma}_\alpha = \mathbf{J}_B \boldsymbol{\Sigma}_u \mathbf{J}_B^\mathrm{T} \; , \tag{5.7}$$

where $\mathbf{J}_B$ is the $n_m \times 2n_v$ Jacobian of $(\mathbf{B}^\mathrm{T}\mathbf{B})^{-1}\mathbf{B}^\mathrm{T}\mathbf{b}$.

The Gaussian distribution $\mathcal{N}(\boldsymbol{\mu}_\alpha, \boldsymbol{\Sigma}_\alpha)$ represents a region of the shape space containing 3D poses that will most likely project close to the detected 2D joint positions $\mathbf{u}_i$. We will now sample this region and propose a representative set of hypotheses. Note however, that the mean $\boldsymbol{\mu}_\alpha$ computed in Eq. (5.6) is unreliable, as it is computed from the $\mathbf{u}_i$'s which are not necessarily the true means of the distributions. We therefore do not draw all samples at once. Instead, we propose an evolution strategy in which we draw successive batches by sampling from a multivariate Gaussian whose mean and covariance are iteratively updated using the Covariance Matrix Adaptation (CMA) algorithm (Hansen, 2006) so as to simultaneously minimize reprojection and length errors.

More specifically, at iteration $k$ we draw $n_s$ random samples $\{\tilde{\boldsymbol{\alpha}}_i^k\}_{i=1}^{n_s}$ from the distribution $\mathcal{N}(\boldsymbol{\mu}_{\boldsymbol{\alpha}}^k, \mathcal{M}^2\boldsymbol{\Sigma}_{\boldsymbol{\alpha}}^k)$, where $\mathcal{M}$ is a constant that guarantees a certain confidence level (we set $\mathcal{M} = 4$ in all experiments). Each sample $\tilde{\boldsymbol{\alpha}}_i^k$ is assigned a weight $\pi_i^k$ proportional to $\varepsilon_{lr} = \varepsilon_\mathrm{l} \cdot \varepsilon_\mathrm{r}$ , the product of the length and reprojection errors:

$$\varepsilon_\mathrm{l} \;=\; \sum_{i,j\in\mathcal{N}} \left\| \tilde{l}_{ij} - l_{ij}^{\mathrm{train}} \right\| \sigma_{ij}^{-1}, \tag{5.8}$$

$$\varepsilon_\mathrm{r} \;=\; \sum_{i}^{n_v} \sqrt{(\tilde{\mathbf{u}}_i - \mathbf{u}_i)^\mathrm{T} \boldsymbol{\Sigma}_{\mathbf{u}_i}^{-1} (\tilde{\mathbf{u}}_i - \mathbf{u}_i)}, \tag{5.9}$$

where $l_{ij}^{\text{train}}$ is the mean distance in all training samples between the $i$-th and $j$-th joints, $\sigma_{ij}$ is the standard deviation, $\tilde{l}_{ij}$ is the length between joints $i$ and $j$ in the sample $\tilde{\boldsymbol{\alpha}}_i^k$, and the $\tilde{\mathbf{u}}_i$'s are their corresponding 2D projections.

Given the weights $\pi_i^k$ for all samples, we then update the mean and covariance of the distribution following the CMA strategy. The mean vector $\boldsymbol{\mu}_{\boldsymbol{\alpha}}^{k+1}$ is estimated as a weighted average of the samples. The update of the covariance matrix $\boldsymbol{\Sigma}_{\boldsymbol{\alpha}}^{k+1}$ consists of three terms: a scaled covariance matrix from the preceding step, a covariance matrix that estimates the variances of the best sampling points in the current generation, and a covariance that exploits information of the correlation between the current and previous generations. For further details, we refer the reader to (Hansen, 2006).

After each iteration a subset of the samples with smaller weights is retained and progressively accumulated for additional analysis. Note that instead of trying to optimize the error function, we use the error function with the CMA optimizer as a way to explore the solution space. When a specific number of samples ($10^4$ in practice) has been obtained, the problem then becomes how to disambiguate them to find one that represents an anthropomorphic pose.

As the detector input does not provide information on the orientation of the subject, we consider the possibility of the pose facing both directions by swapping the detected parts representing the left and right side of the body. This leads to two different distributions which we can then sample from.

Figure 5.5 shows an example of how the solution space is explored. Note that although the CMA algorithm converges relatively far from the optimal solution with minimal reconstruction error, some of the samples accumulated through the exploration process are good approximations. This is the key difference between using a plain CMA, which just seeks for one single solution, and our approach, that accumulates all samples and subsequently uses more stringent –although computationally more expensive– constraints to disambiguate.

After exploring the solution space, we have obtained a large number of samples that represent possible poses that have both low reprojection and length errors. However, since many of these samples are very similar, we reduce their number using a Gaussian-means clustering algorithm (Hamerly and Elkan, 2003). As shown in Fig. 5.6, we then consider the medoid of each cluster to be the candidate ambiguous shape. With this procedure, we can effectively reduce the number of samples from $10^4$ to around $10^2$.

### Hypotheses Disambiguation

The set of ambiguous shapes has been obtained by imposing relatively simple but computationally efficient constraints based on reprojection and length errors. In this section we will describe more discriminative criteria based on the kinematics of the anthropomorphic pose to further disambiguate them until obtaining a single solution.

For this purpose, we will first propose using a *coordinate-free kinematic representation* of the candidate shapes, based on the Euclidean Distance Matrix. Given the 3D position of the $n_v$ joints, we define the $n_v \times n_v$ matrix $\mathbf{D}$ such that, $d_{ij} = \|\mathbf{p}_i - \mathbf{p}_j\|$. It can be shown that this representation is unique for a given configuration. In addition, as it is a symmetric matrix with zero entries at the diagonal, it can be compactly represented by the $n_v(n_v-1)/2$ vector

$$\mathbf{d}_{\text{Kin}} = [d_{12}, \cdots, d_{1n_v}, d_{23}, d_{24}, \cdots, d_{(n_v-1)n_v}]^{\text{T}}. \tag{5.10}$$

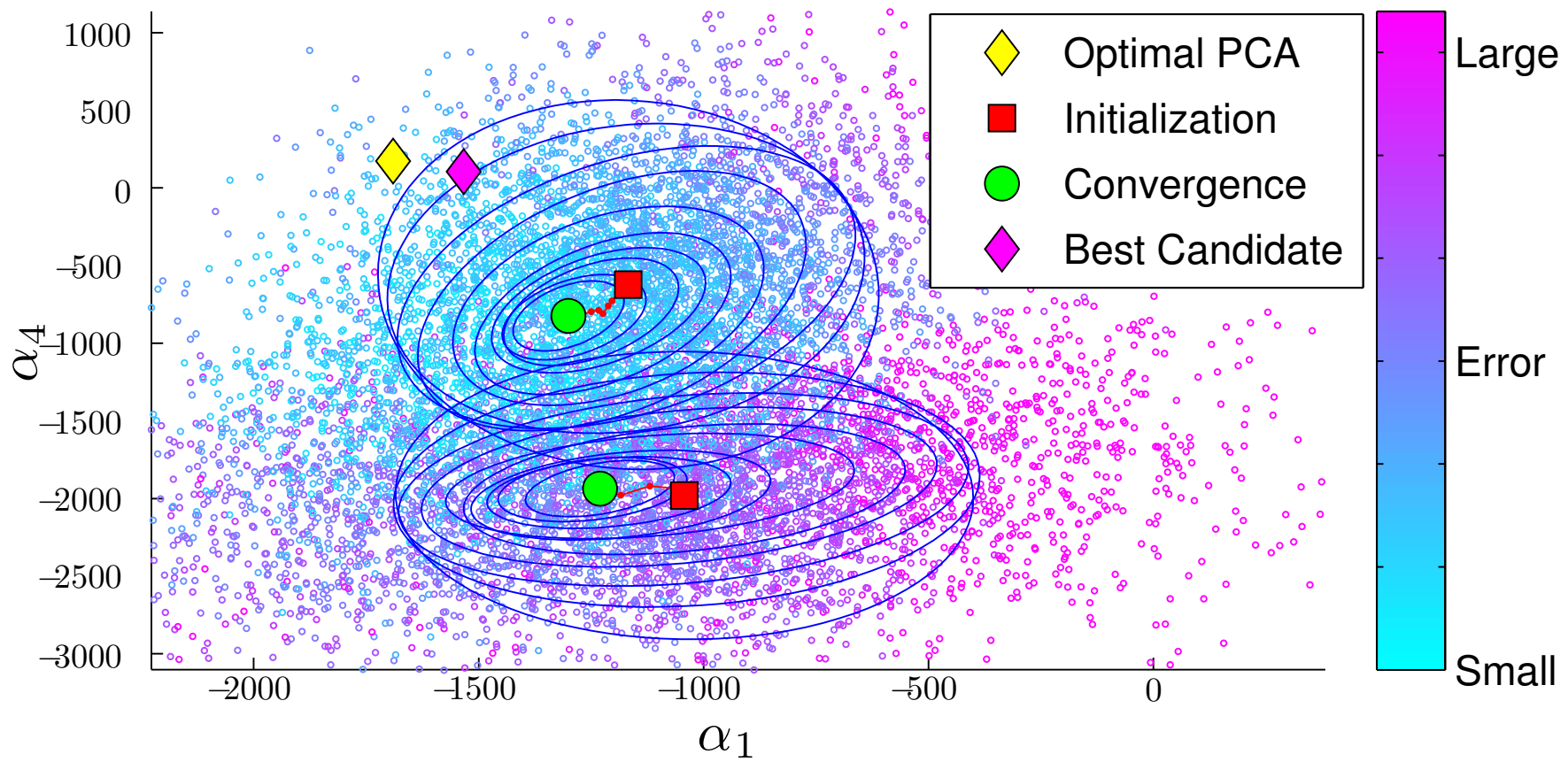

**Figure 5.5: Exploration of the solution space.** The figure plots the distribution of samples on the modal weights space and how the covariance matrix is progressively updated using the CMA algorithm. The two distributions represent both hypotheses of the directions the pose can be facing. In addition, the graph depicts the initial and the final configurations obtained with the CMA, and an optimal solution computed by directly projecting the ground-truth pose onto the PCA modes. The *Best Candidate* corresponds to the solution estimated by our approach. Note that although the CMA does not converge close to the optimal solution, some of the samples accumulated through the process lie very close, and thus, are potentially good solutions.

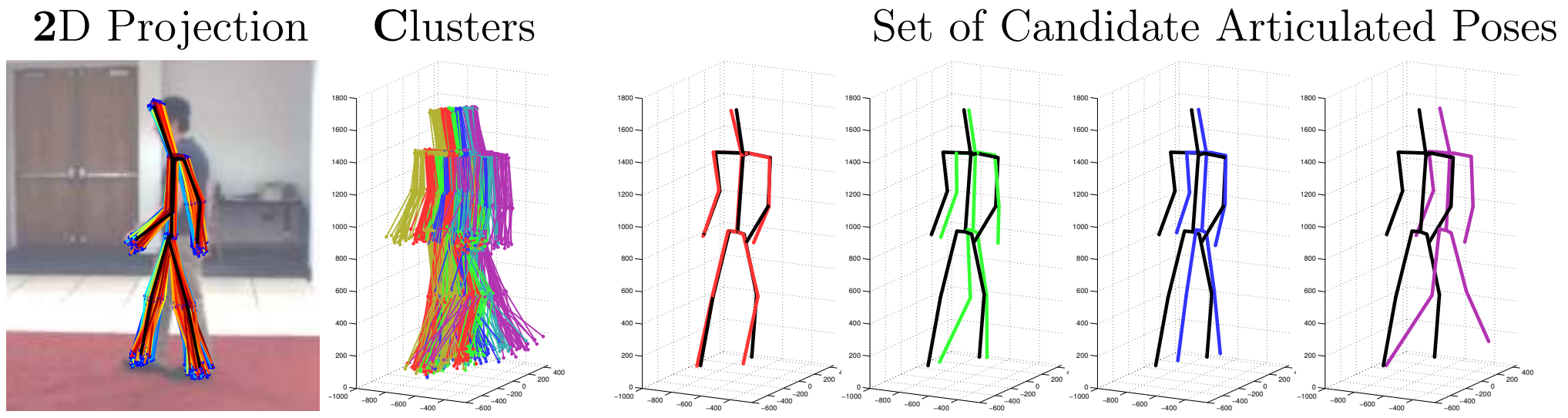

**Figure 5.6: Exploring the space of articulated shapes.** The first two plots represent the 2D projection and 3D view of the shape samples we generate. The colour of the 3D samples indicates the cluster to which they belong. The four graphs on the right represent the medoids of the clusters, which are taken to be the final set of ambiguous candidate shapes.

Given this unique representation of the pose kinematics, we then propose the treatment of the anthropomorphism as a regression problem. Specifically, we want to be able to calculate how different a 3D pose is from a set of training poses. We deal with this problem by using a one-class Support Vector Machine (OCSVM). The scores computed with this classifier can then be used to distinguish between clusters to determine the most anthropomorphic one.

In order to be able to properly determine the degree of anthropomorphism, and given that we have a limited amount of training data, we need to reduce the size of our pose representation and avoid the curse of dimensionality. For this purpose, we will use again PCA, and we will not directly train the classifier on the whole Euclidean distance vector, but with a linear projection $\boldsymbol{\beta}$ of it.

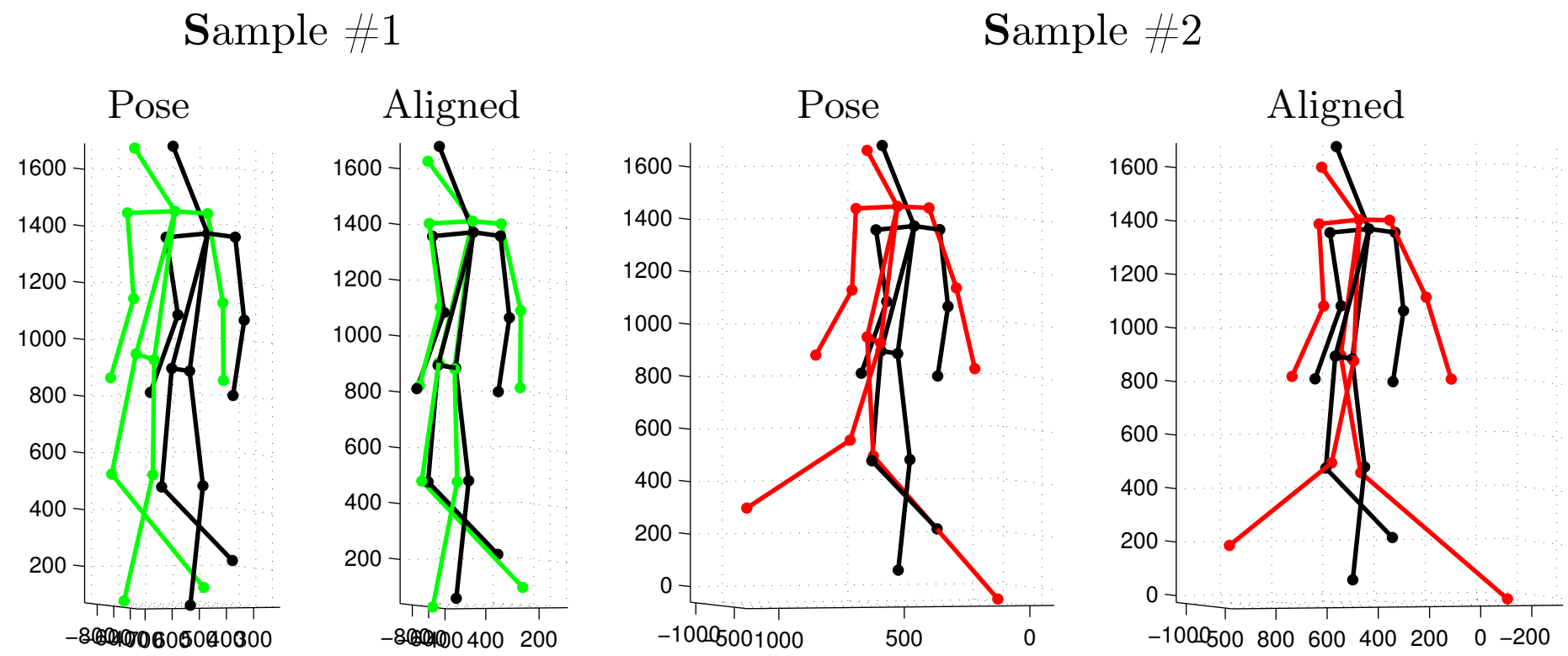

| | **Sample #1** | **Sample #2** |
|---|---|---|
| Error Value ($\varepsilon_{lr}$) | 6.883 | 6.885 |
| SVM Output | 2.8e-04 | -7.6e-03 |
| Reconst. Err. (mm) | 199.9 | 214.9 |
| Aligned Err. (mm) | 56.4 | 167.7 |

**Figure 5.7: Choosing human-like hypothesis via anthropomorphism.** The anthropomorphism factor obtained from the OCSVM can be used to choose more human-like models. In the top figures, the black lines represent the ground truth while the coloured lines represent the different poses. Note that although Sample #1 is far more human-like than Sample #2, both the error given by $\varepsilon_{lr}$ (consisting of reprojection error and limb length errors) and the reconstruction error are almost the same. In contrast, the output of the SVM (+1: anthropomorphic; -1:non-anthropomorphic) indicates that Sample #1 resembles more a human-like pose. A good way to validate anthropomorphism is by aligning the pose to the ground truth and measuring the reconstruction error after alignment.

One important thing to note is that the projection of the distance vectors $\mathbf{d}_{\mathrm{Kin}}$ to the subspace $\boldsymbol{\beta}$ implies a loss of information that can lead to non-anthropomorphic forms being projected close to anthropomorphic forms. In order to account for this effect, it is important to remove the clusters with the worst error value $\varepsilon_{lr}$. As shown in Fig. 5.7, this increases the likelihood that the results returned by the OCSVM correspond to an anthropomorphic form.

## Results

We evaluated the algorithm on two different datasets: the HumanEva dataset (Sigal et al., 2010a), which provides ground truth, and the TUD Stadtmitte sequence (Andriluka et al., 2010), which is a challenging urban environment with multiple people,

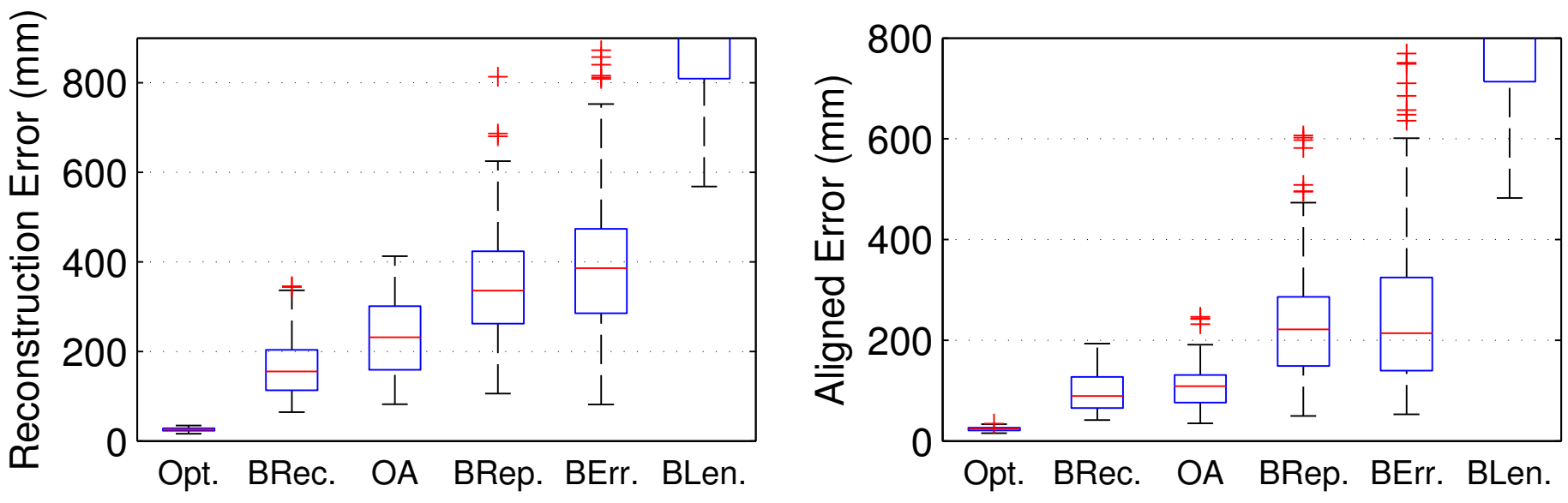

**Figure 5.8: Comparison of different error functions for 3D human pose estimation.** *Left:* Reconstruction errors on the HumanEva dataset for the sequence "S2 walk". *Right:* Same errors after rigid alignment of the shapes with the ground truth poses. It is more representative of the anthropomorphism of the pose compared to the plain reconstruction error, which only considers the distance between the joints of the retrieved pose and the ground truth. See text for a detailed description of the labels.

but without ground truth for a quantitative evaluation.

### Evaluation on the HumanEva dataset

We quantitatively evaluated the performance of our method, using the *walking* and *jogging* actions of the HumanEva dataset. For training the PCA and SVM, we used the motion captured data, independently for each action, for subjects "S1", "S2" and "S3", and used the "validation" sequences for testing.

Fig. 5.8 shows the distribution of the results on the "S2 walk" sequence. In Fig. 5.8-Left we plot the average reconstruction error of our approach (*OA*), and compare it against the reconstruction error of *Opt*: the best approximation we could achieve using PCA; *BRec*: sample with minimum reconstruction error among all samples generated during the exploration process; *BRep*: the sample with minimum reprojection error; *BLen*: the sample with minimum length error; and *BErr*: the sample that minimizes $\varepsilon_{lr}$ (consisting of reprojection error and limb length errors). Note that neither minimizing the reprojection error, the length error nor $\varepsilon_{lr}$ guarantees retrieving a good solution. We address this by also maximizing the similarity with anthropomorphic shapes. By doing this, the mean error per joint of the shapes we retrieve is around 230mm. Yet, most of this error is due to slight depth offsets which are hard to control due to the noise in the input data. In fact, if we perform a rigid alignment between these shapes and the ground truth ones, the error is reduced to about 100mm (Fig. 5.8-Right).

Finally, numeric results comparing with the state-of-the-art are given in Table 5.1. Note that this comparison is for guidance only, as different methods train and evaluate differently. For instance, although (Bo and Sminchisescu, 2010) yields significantly better results, it relies on strong assumptions, such as background subtraction, which both our approach and (Andriluka et al., 2010; Daubney and Xie, 2011) do not consider. Therefore, we believe that to truly position our approach, we should compare ourselves against (Andriluka et al., 2010; Daubney and Xie, 2011). In fact, the performance of all three methods is very similar, but we remind the reader that (Andriluka et al., 2010; Daubney and Xie, 2011) impose temporal consistency along the sequence, while we estimate the 3D pose using just one single image. A few sample images of the results we obtain are shown in Fig. 5.9.

| | Walking | | |
|---|---|---|---|
| | S1 | S2 | S3 |
| 3D Output | 99.6 (42.6) | 108.3 (42.3) | 127.4 (24.0) |
| 2D Input | 14.1 (7.5) | 19.1 (8.1) | 26.8 (8.0) |
| (Andriluka et al., 2010) | - | 107 (15) | - |
| (Daubney and Xie, 2011) | 89.3 | 108.7 | 113.5 |
| (Bo and Sminchisescu, 2010) | 38.2 (21.4) | 32.8 (23.1) | 40.2 (23.2) |
| | **Jogging** | | |
| | S1 | S2 | S3 |
| 3D Output | 109.2 (41.5) | 93.1 (41.1) | 115.8 (40.6) |
| 2D Input | 18.3 (6.3) | 18.1 (6.0) | 20.9 (6.1) |
| (Bo and Sminchisescu, 2010) | 42.0 (12.9) | 34.7 (16.6) | 46.4 (28.9) |

**Table 5.1: Comparison of results on the HumanEva dataset.** Comparing the results on the HumanEva dataset for the *walking* and *jogging* actions with all three subjects. All values are in mm with the standard deviation in parentheses if applicable. 2D values are in pixels. Absolute error is displayed for (Andriluka et al., 2010; Daubney and Xie, 2011), while our approach (OA) and (Bo and Sminchisescu, 2010) are relative error values. (Andriluka et al., 2010; Daubney and Xie, 2011) do not provide *jogging* data.

### Testing on Street Images

We have also used the TUD Stadtmitte sequence (Andriluka et al., 2010) to test the robustness of the algorithm. We consider the scenario with multiple people to detect. This sequence is especially challenging for 3D reconstruction as the camera has a long focal distance, which amplifies the propagation of the 2D errors to the 3D space.

Since we are dealing with real street images, walking pedestrian poses frequently do not match our limited training data: pedestrians may either carry an object or have their hands in their pockets, as seen in Fig. 5.10. Furthermore, the 2D body part detector generally fails to find the correct position of the hands (and consequently the arms) because of these occlusions. Despite these difficulties, our method is usually able to find the correct pose.

## 5.4 Joint 2D and 3D Pose Estimation

In this work, we get rid of the strong assumption that data association may be easily achieved, and propose a novel approach to jointly detect the 2D position and estimate the 3D pose of a person from one single image acquired with a calibrated but potentially moving camera. For this purpose we formulate a Bayesian approach combining a generative latent variable model that constrains the space of possible 3D body poses with a HOG-based discriminative model that constrains the 2D location of the body parts. The two models are simultaneously updated using an evolutionary strategy. In this manner 3D constraints are used to update image evidence while 2D observations are used to update the 3D pose. As shown in Fig. 5.11 these strong ties make it possible to accurately detect and estimate the 3D pose even when image evidence is very poor.

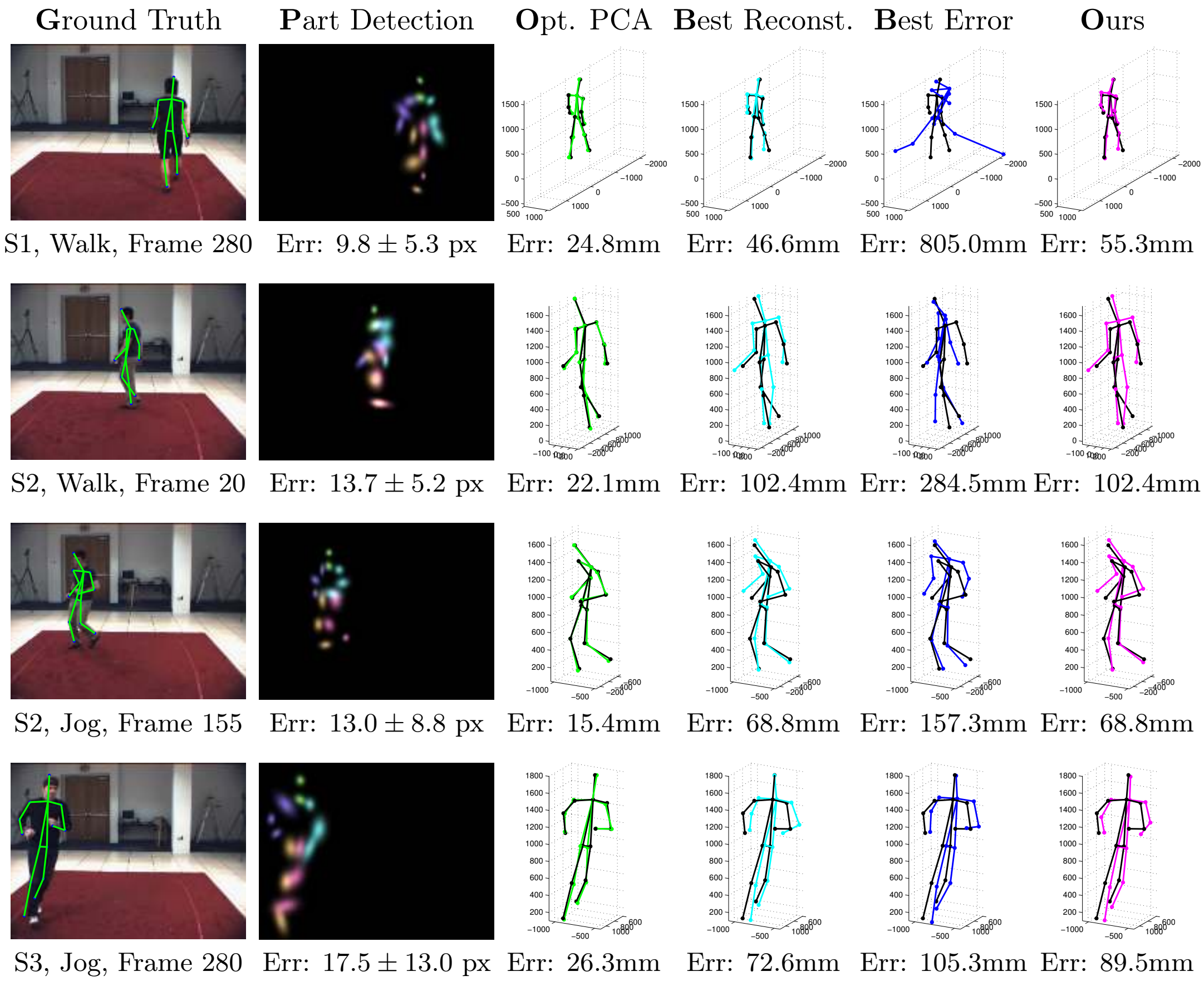

**Figure 5.9: 3D Human Pose Estimation Results.** *Leftmost two columns:* Raw image with 2D ground truth projection, and the 2D detection results with the associated average pixel distance from ground truth. *Rightmost four columns: Opt. PCA*: projection of the ground truth on the PCA; *Best Reconstruction*: the sample with lowest reconstruction error; *Best Error*: the sample with the lowest error $\varepsilon_{lr}$ (consisting of reprojection error and limb length errors); and *Ours*: the solution obtained. Below each solution we indicate the corresponding reconstruction error (in mm). Note that minimizing $\varepsilon_{lr}$ does not guarantee retrieving a good solution.

## Method

Figure 5.12 shows an overview of our model for simultaneous 2D people detection and 3D pose estimation. It consists of two main components: a 3D generative kinematic model, which generates pose hypotheses, and a discriminative part model, which weights the hypotheses based on image appearance. Drawing inspiration from the approach proposed in (Andriluka et al., 2009) for 2D articulated shapes, we represent this model using a Bayesian formulation.

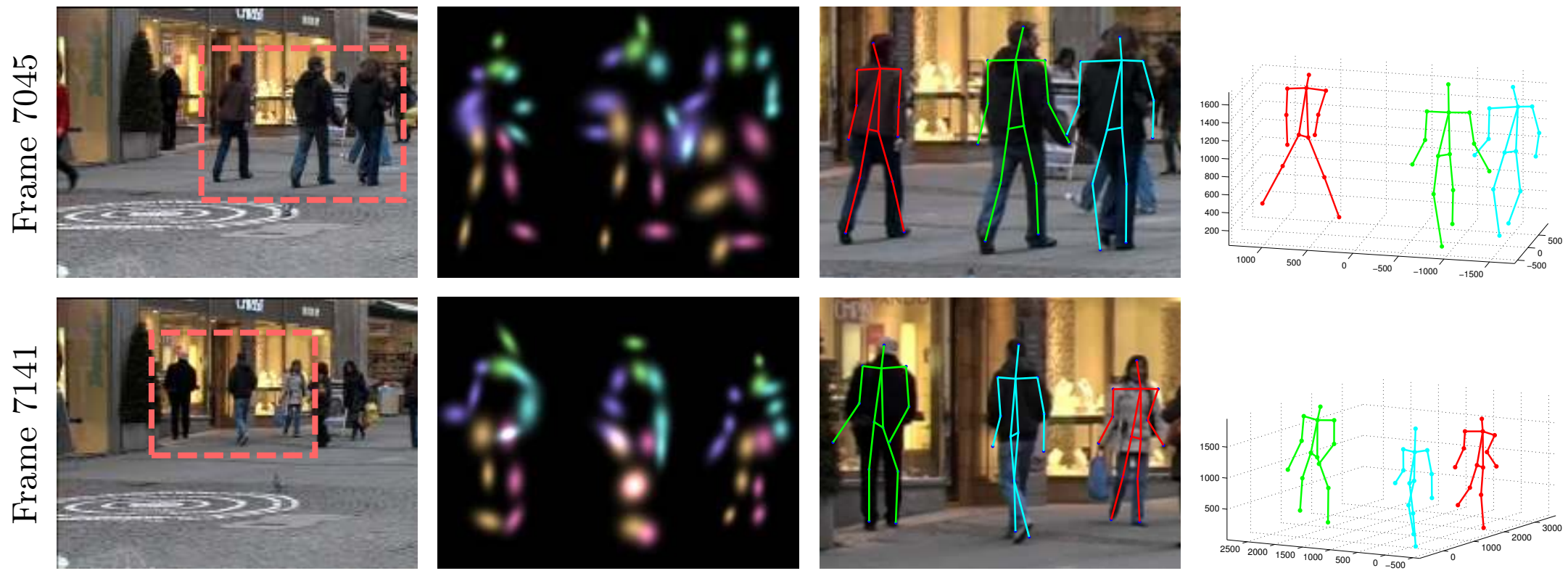

**Figure 5.10: Results on the TUD Stadtmitte sequence.** *Left three columns:* Raw input image, followed by the detected Gaussians and the reprojection of the estimated 3D pose on the scene. *Right column:* Estimated 3D pose.

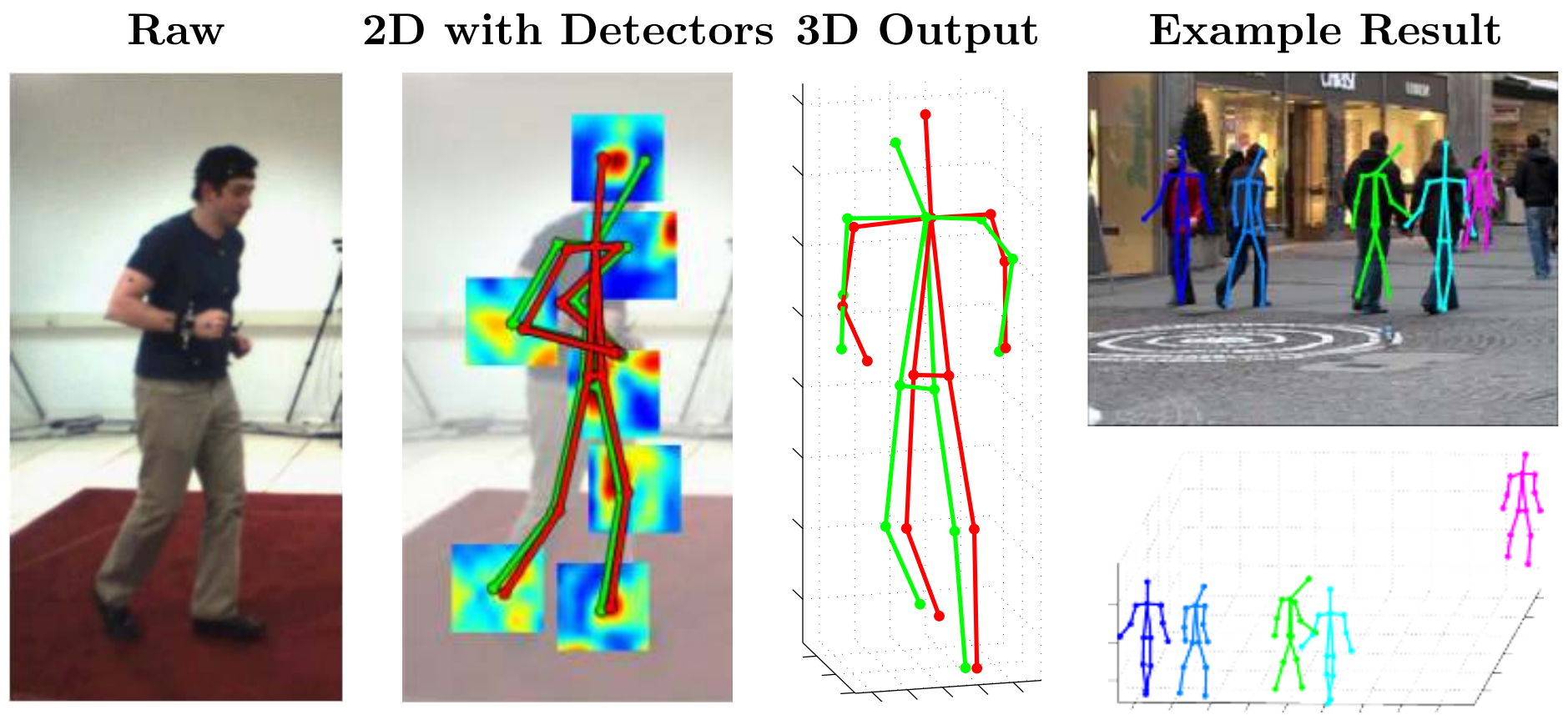

**Figure 5.11: Simultaneous estimation of 2D and 3D pose.** First column: Raw input image. Second column: Ground truth 2D pose (green) and the result of our approach (red). Additionally, we plot a few part detectors and their corresponding score, used to estimate 3D pose. Reddish areas represent regions with highest responses. Third column: 3D view of the resulting pose. Note that despite the detectors not being very precise, our generative model allows estimating a pose very close to the actual solution. Last Column: We show an example of a challenging scene with several pedestrians.

With this purpose, we represent 3D poses as a set of $N$ connected parts. Let $\mathbf{l} = \{\mathbf{l}_1, \dots, \mathbf{l}_N\}$ be their 2D configuration with $\mathbf{l}_i = (u_i, v_i, s_i)$. $(u_i, v_i)$ is the image position of the center of the part, and $s_i$ a scale parameter which will be defined below. In addition let $\mathbf{d} = \{\mathbf{d}_1, \dots, \mathbf{d}_N\}$ be the set of image evidence maps, i.e., the maps for every part detector at different scales and for the whole image. Assuming conditional independence of the evidence maps given $\mathbf{l}$, and that the part map $\mathbf{d}_i$ only depends on $\mathbf{l}_i$, the likelihood of the image evidence given a specific body configuration becomes:

$$p(\mathbf{d} \mid \mathbf{l}) = \prod_{i=1}^{N} p(\mathbf{d}_i \mid \mathbf{l}_i)\,. \tag{5.11}$$

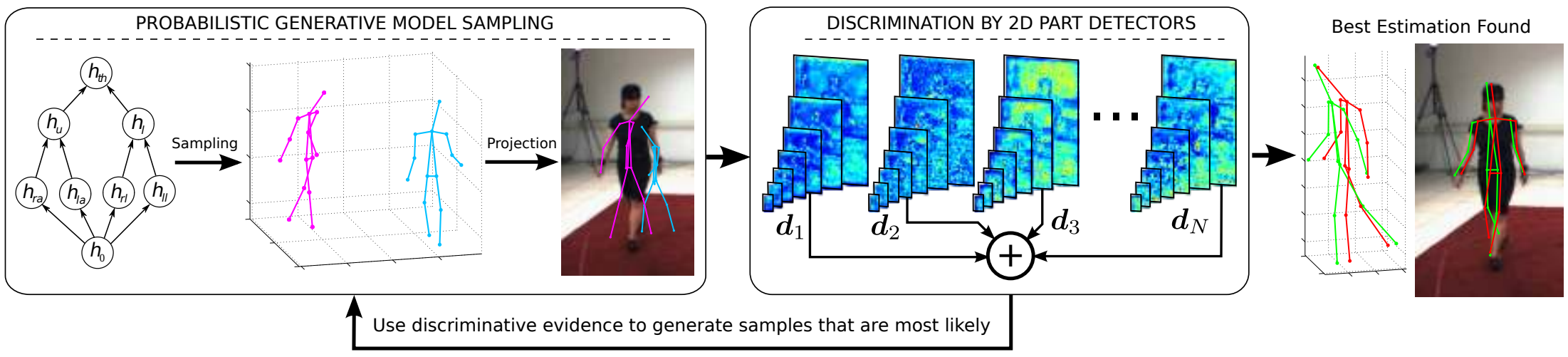

**Figure 5.12: Overview of our method for simultaneous 2D and 3D pose estimation.** Our approach consists of two main building blocks: a probabilistic generative model and a set of discriminative 2D part detectors. Our optimization framework simultaneously solves for both the 2D and 3D pose using an evolutionary strategy. A set of weighted samples are generated from the probabilistic generative model and are subsequently reweighted by the score given by the 2D part detectors. This process is repeated until convergence of the method. The rightmost figure shows results at convergence where the red shapes are the estimated poses and the green ones correspond to the ground truth.

In (Andriluka et al., 2009), Eq. (5.11) is further simplified under the assumption that the body configuration may be represented using a tree topology. This yields an additional efficiency gain, as it introduces independence constraints between branches, e.g., the left arm/leg does not depend on the right arm/leg. Yet, this causes the issue of the double counting, where the same arm/leg is considered to be both the left and right one. In (Andriluka et al., 2012) this is addressed by first solving an optimal tree and afterwards attempting to correct these artefacts using loopy belief propagation. Instead of using two stages, we directly represent our 3D model using a Directed Acyclic Graph, which enforces anthropomorphic constraints, and helps preventing the double counting problem.

Let $\mathbf{x} = \{\mathbf{p}_1, \dots, \mathbf{p}_N\}$ be the 3D model that projects on the 2D pose $\mathbf{l}$, where $\mathbf{p}_i = (x_i, y_i, z_i)$ is the 3D position of $i$-th part center. We write the posterior of $\mathbf{x}$ given the image evidence $\mathbf{d}$ by:

$$p(\mathbf{x} \mid \mathbf{d}) \propto \prod_{i=1}^{N} \left( p(\mathbf{d}_i \mid \mathbf{l}_i)\, p(\mathbf{l}_i \mid \mathbf{p}_i) \right) p(\mathbf{x})\,.$$

In order to handle the complexity of directly modeling $p(\mathbf{x})$, we propose approximating $\mathbf{x}$ through a generative model based on latent variables $\mathbf{h}$. This allows us to finally write the problem as:

$$p(\mathbf{x} \mid \mathbf{d}) \propto \underbrace{p(\mathbf{h})\, p(\mathbf{x} \mid \mathbf{h})}_{\text{generative}} \underbrace{\prod_{i=1}^{N} \left( p(\mathbf{d}_i \mid \mathbf{l}_i)\, p(\mathbf{l}_i \mid \mathbf{p}_i) \right)}_{\text{discriminative}}$$

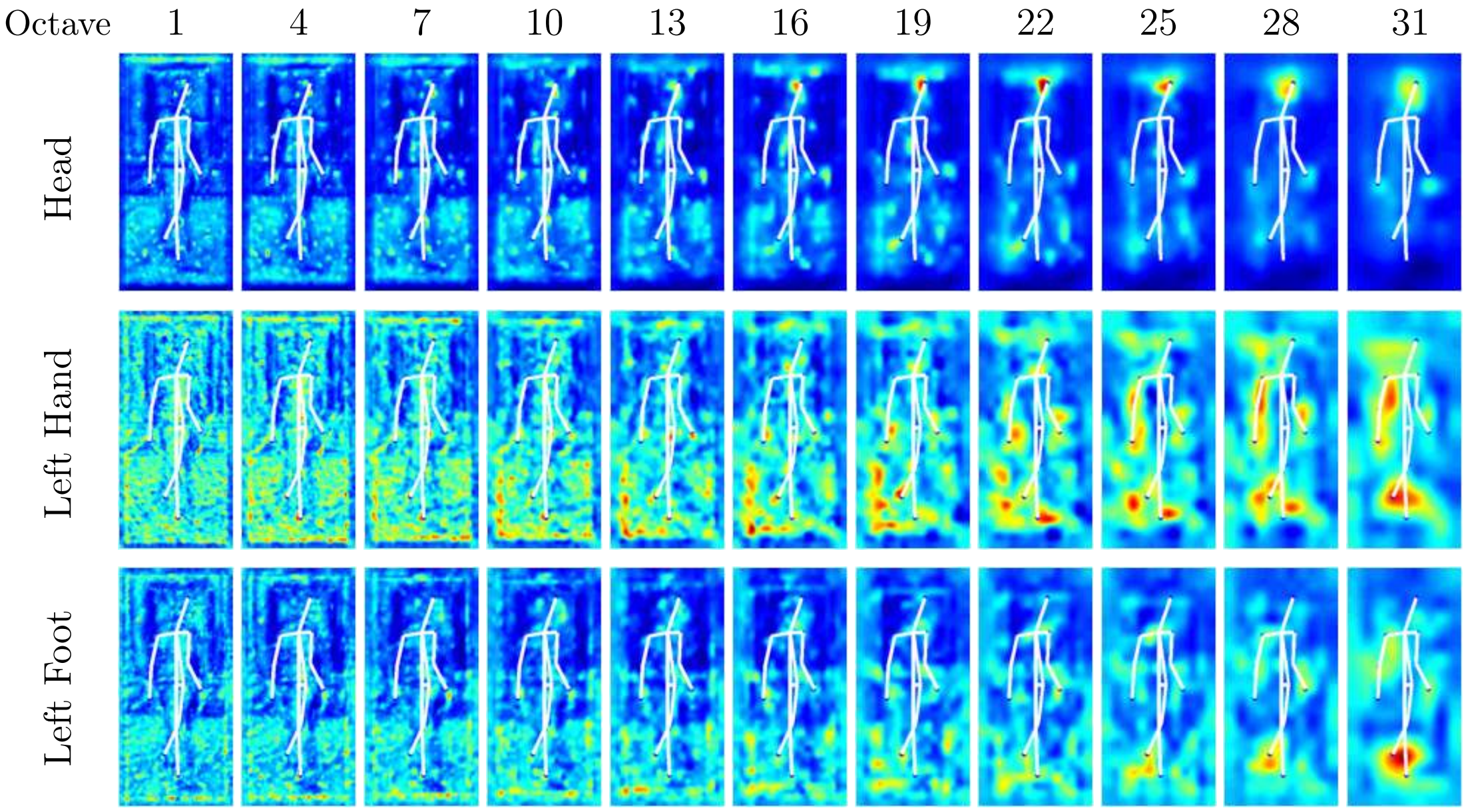

**Figure 5.13: 2D Part Detectors.** We visualize the response at different octaves for three different part detectors. We overlay the projection of the 3D ground truth in white to get an idea of the accuracy. The outputs are normalized for visualization purposes, with the dark blue and bright red areas corresponding to lower and higher responses respectively. Note that while some detectors, such as the head one in the first row, generally give good results, others do not, such as the left hand detector in the middle row. Our approach can handle these issues by combining these 2D part detectors with a generative model.

where the discriminative and generative components become clearly separated. We will next describe each of these components.

### Discriminative Detectors

Recent literature proposes two principal alternatives for discriminative detectors: the shape context descriptors built by applying boosting on the limbs (Andriluka et al., 2009), and the HOG template matching approach (Yang and Ramanan, 2011). For our purposes we have found the HOG-based template matching to be more adequate because it matches our joint-based 3D model better as we can place a detector at each joint part instead of having to infer the limb positions from the joints. In addition, the HOG template matching yields smoother responses, which is preferable when doing inference.

As mentioned above, each part $\mathbf{l}_i$ has an associated scale parameter $s_i$. This parameter is used to pick a specific scale among the evidence maps. Intuitively, if a part is far away, it should be evaluated by a detector at a small scale (high resolution). We therefore approximate the scale $s_i$ as:

$$s_i^{-1} = \alpha^{-1} \beta z_i \tag{5.12}$$

where $\alpha$ is the focal length of the camera and is used for normalization purposes. The parameter $\beta$ will be learned off-line and used during inference. Note that despite this parameter constant, the scale at which each part is evaluated is different as it depends on the $z_i$ coordinate.

Let $\mathbf{t} = \{\mathbf{t}_1, \dots, \mathbf{t}_N\}$ be the set of templates in the HOG space associated to each body part. These templates are provided in (Yang and Ramanan, 2011)[1]. Given a body part $\mathbf{l}_i$, its image evidence $\mathbf{d}_i$ is computed by evaluating the template $\mathbf{t}_i$ over the entire image for a range of scales $s_i$. Figure 5.13 illustrates the response of three part detectors at different scales. By interpreting each detector as a log-likelihood, the image evidence of a configuration $L$ can be computed as:

$$\log p\left(\mathbf{l} \mid \mathbf{d}\right) \approx \text{score}(\mathbf{l}) = \sum_{i=1}^{N} k_i \mathbf{d}_i(u_i, v_i, s_i) \ , \tag{5.13}$$

where $k_i$ is a weight associated to each detector, which is learned offline. It is meant to adapt the 2D detectors and 3D generative model due to the fact they were trained independently on different datasets.

Additionally, when evaluating a part detector at a point, we consider a small window from which we use the largest detector value in order to give additional noise tolerance to the detector. We find this necessary as small 2D errors can have large consequences in 3D positioning.

### Latent Generative Model

We use the Directed Acyclic Graph model explained in Section 4.4. This model discreticizes the poses to be able to learn the joint distribution of the 3D poses and a latent space by using Expectation Maximization. Given that the model does not contain loops, efficient inference can be done using dynamic programming. This allows the model to be used in a generative fashion in which the latent space is sampled and then the most likely 3D pose for the latent space configuration is efficiently estimated.

### Inference

The inference problem consists of computing:

$$<X^*> = \arg\max_{\mathbf{x}} \prod_{i=1}^{N} \left( p\left(\mathbf{d}_i \mid \mathbf{l}_i\right) p\left(\mathbf{l}_i \mid \mathbf{p}_i\right) \right) p\left(\mathbf{x} \mid \mathbf{h}\right) p\left(\mathbf{h}\right) \ . \tag{5.14}$$

We treat this as a global optimization problem where, given a set of 2D detections corresponding to the different parts from a single image, we optimize over both a rigid transformation and the latent states. Drawing inspiration in (Moreno-Noguer and Fua, 2013; Moreno-Noguer et al., 2010), we do this using a variation of the Covariance Matrix Adaptation Evolutionary Strategy (CMA-ES) (Hansen, 2006), which is a black box global optimizer that uses a Gaussian distribution in the search space to minimize

[1] Indeed, each of the part detectors provided in (Yang and Ramanan, 2011) is formed by several templates and we use their maximum score for each coordinate $(u, v, s)$. For ease of explanation we refer to them as a single template.

| | (Yang and Ramanan, 2011) | Ideal Detector | | | Our Approach | | |
|---|---|---|---|---|---|---|---|
| Err. | 2D | 2D | 3D | Pose | 2D | 3D | Pose |
| All | 21.7 | 11.0 | 106.6 | 51.6 | 19.5 | 237.3 | 55.3 |
| C1 | 19.5 | 11.1 | 113.8 | 52.3 | 18.9 | 239.1 | 55.2 |
| C2 | 22.9 | 11.1 | 109.7 | 51.2 | 19.6 | 245.8 | 55.4 |
| C3 | 22.8 | 10.8 | 96.2 | 51.2 | 20.0 | 227.1 | 55.4 |
| S1 | 21.8 | 10.2 | 96.8 | 63.4 | 19.9 | 277.2 | 69.3 |
| S2 | 21.8 | 10.8 | 108.0 | 44.8 | 18.6 | 206.6 | 46.8 |
| S3 | 21.6 | 12.3 | 119.0 | 43.7 | 20.1 | 221.4 | 46.6 |
| A1 | 20.9 | 10.7 | 106.0 | 56.2 | 19.3 | 254.4 | 60.3 |
| A2 | 22.7 | 11.3 | 107.2 | 46.6 | 19.7 | 219.0 | 50.0 |

**Table 5.2: Results on the HumanEva dataset.** We show results for the *walking* (A1) and *jogging* (A2) actions with all subjects (S1,S2,S3) and cameras (C1,C2,C3). We compare with the 2D error obtained using the 2D model from (Yang and Ramanan, 2011), based on the same part detectors we use. 2D, 3D and Pose Errors are defined in the text. Ideal detector corresponds to our approach using Gaussians with 20px covariance as 2D input instead of the 2D detections.

a function. In our case we perform:

$$\arg\max_{\mathbf{R},\mathbf{t},\mathbf{H}} \text{score}\left(\text{proj}_{\mathbf{R},\mathbf{t}}\left(\phi^{-1}\left(\mathbf{h}\right)\right)\right) + \log\left(p\left(\phi^{-1}\left(\mathbf{h}\right),\mathbf{h}\right)\right) \ ,$$

where $\text{proj}_{\mathbf{R},\mathbf{t}}(\cdot)$ is the result of applying a rotation $\mathbf{R}$, a translation $\mathbf{t}$ and projecting onto the image plane. We then take $\mathbf{x}^* = \mathbf{R}\phi^{-1}(\mathbf{h}^*) + \mathbf{t}$, that is, we obtain the most probable $\mathbf{x}^{L*}$ given $\mathbf{h}^*$ and perform a rigid transformation to the world coordinates to obtain $\mathbf{x}^*$.

## Results

We numerically evaluate our algorithm on the HumanEva benchmark (Sigal et al., 2010b), which provides 3D ground truth for various actions. In addition, we provide qualitative results on the TUD Stadtmitte sequence (Andriluka et al., 2010), a cluttered street sequence. For both cases, we compute the 2D observations using the detectors from (Yang and Ramanan, 2011) trained independently on the PARSE dataset (Ramanan, 2006).

We consider three error metrics: *2D error*, *3D error* and *3D pose error*. The 2D error measures the mean pixel difference between the 2D projection of the estimated 3D shape, and the ground truth. 3D error is the mean euclidean distance, in mm, with the ground truth, and the 3D pose error is the mean euclidean distance with the ground truth after performing a rigid alignment of the two shapes. This error is indicative of the local deformation error. We evaluate three times on every 5 images using all three cameras and all three subjects for both the *walking* and *jogging* actions, for a total of 1318 unique images.

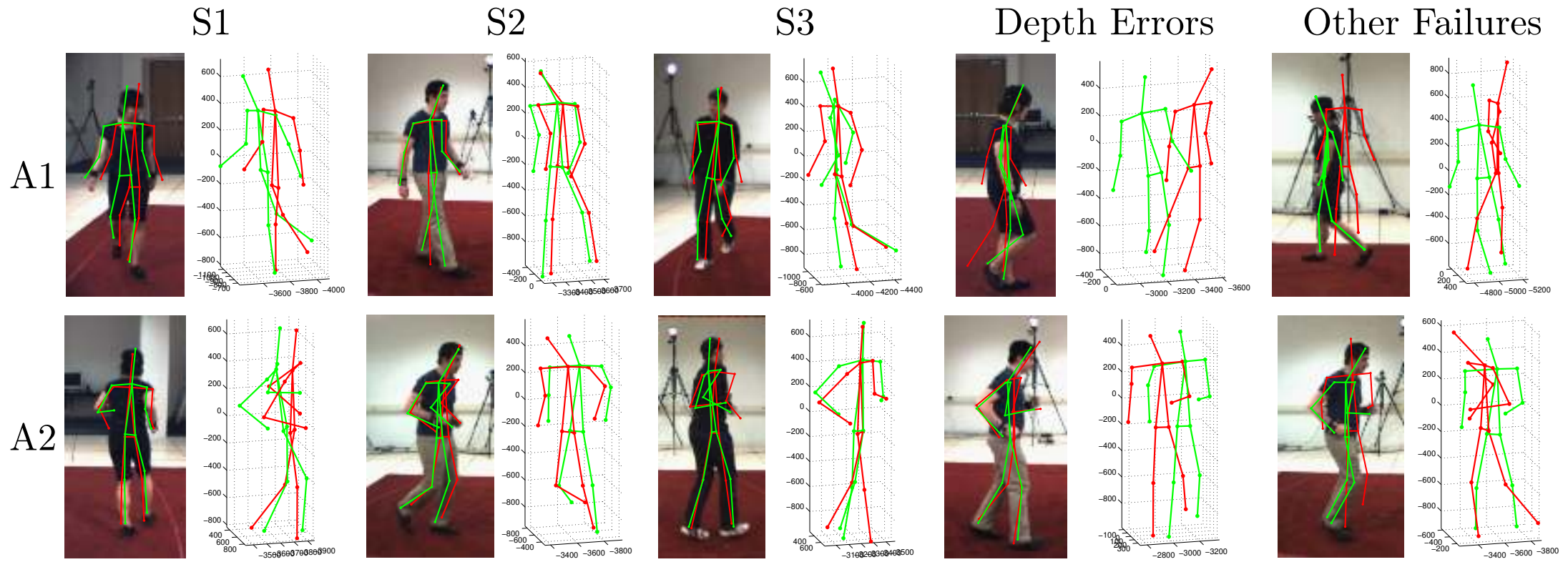

**Figure 5.14: Sample frames from the HumanEva dataset.** We show examples for both the *walking* (A1) and *jogging* (A2) actions. The first three columns correspond to successful 2D and 3D pose estimations for the three subjects (S1,S2,S3). The next two columns show typical failures cases of our algorithm. In the fourth column we see that occasionally we suffer from depth errors, where the 3D pose is correct but its depth is not. In the last column we plot other failures, mostly caused by very large errors of the 2D detector, due to mis-classifications or self-occlusions.

Table 5.2 summarizes the results of all experiments. We compare our approach using both Gaussians (20px Cov.) and the detector outputs as inputs. We see that using ideal detectors, even with large covariances, the absolute error is reduced to 45% of the full approach. An interesting result is that we outperform the 2D pose obtained by (Yang and Ramanan, 2011), using their own part detectors. This can likely be attributed to our joint 2D and 3D model. Nonetheless, although (Yang and Ramanan, 2011) is not an action specific approach as we are, this is still an interesting result as (Pepik et al., 2012) reports performance loss in 2D localization when using a 3D model. Figure 5.14 shows some specific examples. As expected, performance is better when there are fewer self-occlusions.

We also compare our results with (Andriluka et al., 2010; Bo and Sminchisescu, 2010; Daubney and Xie, 2011), and the approach presented in the previous section "From 2D Observations". This is just meant to be an indicative result, as the different methods are trained and evaluated differently. Table 5.3 summarizes the results using the pose error, corresponding to the "aligned error" mentioned in the previous section. The two algorithms that use temporal information (Andriluka et al., 2010; Daubney and Xie, 2011) are evaluated using absolute error. Moreover, (Daubney and Xie, 2011) uses two cameras, while the rest of the approaches are monocular. Due to our strong kinematic model we outperform all but (Bo and Sminchisescu, 2010). Yet, in this work the 2D detection step is relieved through background subtraction processes.

Finally, we present qualitative results on the TUD Stadtmitte sequence (Andriluka et al., 2010), which represents a challenging real-world scene with the presence of distracting clutter and occlusions. Some results can be seen in Fig. 5.15. Note that while the global pose seems generally correct, there are still some errors in the 3D pose due to the occlusions and to the fact that the walking style in the wild is largely different from that of the subjects of the HumanEva dataset used to train the generative model.

| | **Walking** (A1,C1) | | |
|---|---|---|---|
| | S1 | S2 | S3 |
| Joint 2D and 3D | 65.1 (17.4) | 48.6 (29.0) | 73.5 (21.4) |
| From 2D Observations | 99.6 (42.6) | 108.3 (42.3) | 127.4 (24.0) |
| (Daubney and Xie, 2011) | 89.3 | 108.7 | 113.5 |
| (Andriluka et al., 2010) | - | 107 (15) | - |
| (Bo and Sminchisescu, 2010) | 38.2 (21.4) | 32.8 (23.1) | 40.2 (23.2) |
| | **Jogging** (A2,C1) | | |
| | S1 | S2 | S3 |
| Joint 2D and 3D | 74.2 (22.3) | 46.6 (24.7) | 32.2 (17.5) |
| From 2D Observations | 109.2 (41.5) | 93.1 (41.1) | 115.8 (40.6) |
| (Bo and Sminchisescu, 2010) | 42.0 (12.9) | 34.7 (16.6) | 46.4 (28.9) |

**Table 5.3: Comparison against state-of-the-art approaches.** We present results for both the *walking* and *jogging* actions for all three subjects and camera C1. *From 2D Observations* refers to the first approach presented in this chapter.

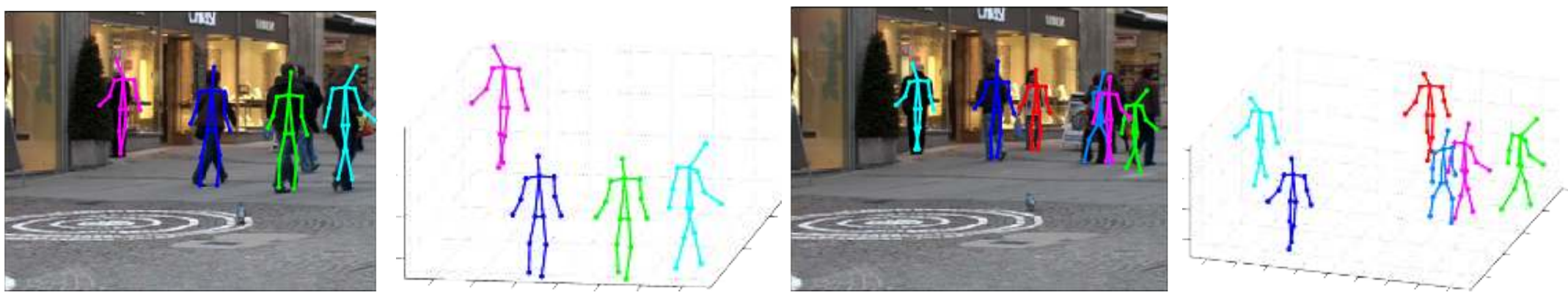

**Figure 5.15: Two sample frames from the TUD Stadtmitte sequence (Andriluka et al., 2010).**

## 5.5 Summary

We have presented two different approaches for 3D human pose estimation from single monocular images. This is an extremely challenging problem as it is severely underconstrained. Despite the difficulty we have been able to obtain very promising results with both approaches, each with a different set of strengths and weaknesses.

In the first approach a 2D pose estimation model is used to initially obtain a noisy observation that can be used for 3D pose estimation. The method consists of a linear projection model that constrains the possible 3D positions given the 2D observations in combination with a linear latent model for the 3D pose. By fusing both linear models, we are able to obtain a linear formulation of the camera constraints on the latent space. This allows a noisy 2D pose estimation to be projected to the 3D pose space in which a hypergaussian is obtained. Afterwards anthropomorphism is used as a criteria to discern the most realistic pose obtained from sampling from the hypotheses. The advantage of this formulation is in its simplicity, which allows it to be very computationally efficient, limited only by the speed of the 2D pose estimation model. However, by being a pipelined approach, the quality of the 3D output is limited by that of the 2D input.

For our second approach we employ 2D discriminative detectors directly as the model input and then perform 2D and 3D pose estimation jointly. This is done by extending the Bayesian formulation of pictorial structures to 3D and reformulating it as a hybrid generative-discriminative model. For the generative model we use a Directed Acyclic Graph, which allows probabilistic representing 3D human poses through a discrete latent space. The discriminative model consists of reweighting 2D detectors for the 3D human pose estimation task. We treat the inference as a global optimization problem and solve it iteratively using state-of-the-art techniques. By performing 2D and 3D pose estimation jointly, the 3D is able to correct errors caused by occlusions in the 2D detections.

Our findings with both approaches show that it is indeed possible to obtain good 3D human pose estimation results despite the complexity of the problem. This is done by leveraging the prior knowledge of human motion with generative models. Stronger priors allow much more accurate pose estimation by eliminating non-realistic pose hypotheses. We are also able to see that performing 2D and 3D pose estimation jointly is beneficial to performance. By delaying the pose estimation decision to the end and leveraging all the available information, the noise in the 2D part detectors is minimized.

In the next chapter we will show that the output of pose estimators can be used to create strong features when performing higher order reasoning.

# Chapter 6

# Probabilistically Modelling Fashion

> The finest clothing made is a person's skin, but, of course, society demands something more than this.
>
> *Mark Twain*

In this chapter we will discuss higher level computer vision tasks in a fashion context. These high level task push the boundaries of what is possible with modern computer vision and machine learning, while being the contributions that are able to influence people in their daily life.

In particular we shall focus on two tasks: semantic segmentation of clothings in fashion photographs, and prediction of fashionability while providing fashion advice. The first task is based on a standard problem in computer vision, however, the application domain is vastly different. In particular the large number of classes with high intra-class variability in combination with the fine-grained level of annotations make it an extremely challenging problem. In our second task we tackle the novel problem of predicting fashionability of an individual in an image with some associated metadata. We also take it a step further and attempt to give advice to the user on which outfits would suit her/him better.

All the work in this chapter is a result of collaborating with Prof. Raquel Urtasun and Prof. Sanja Fidler, from the University of Toronto.

## 6.1 Introduction

Fashion has a tremendous impact on our society. Clothing typically reflects the person's social status and thus puts pressure on how to dress to fit a particular occasion. Its importance becomes even more pronounced due to online social sites like Facebook and Instagram where one's photographs are shared with the world. We also live in a technological era where a significant portion of the population looks for their dream partner on online dating sites. People want to look good; business or casual, elegant or sporty, sexy but not slutty, and of course trendy, particularly so when putting their picture online. This is reflected in the growing online retail sales, reaching 370 billion dollars in the US by 2017, and 191 billion euros in Europe (Forbes Magazine, 2013).

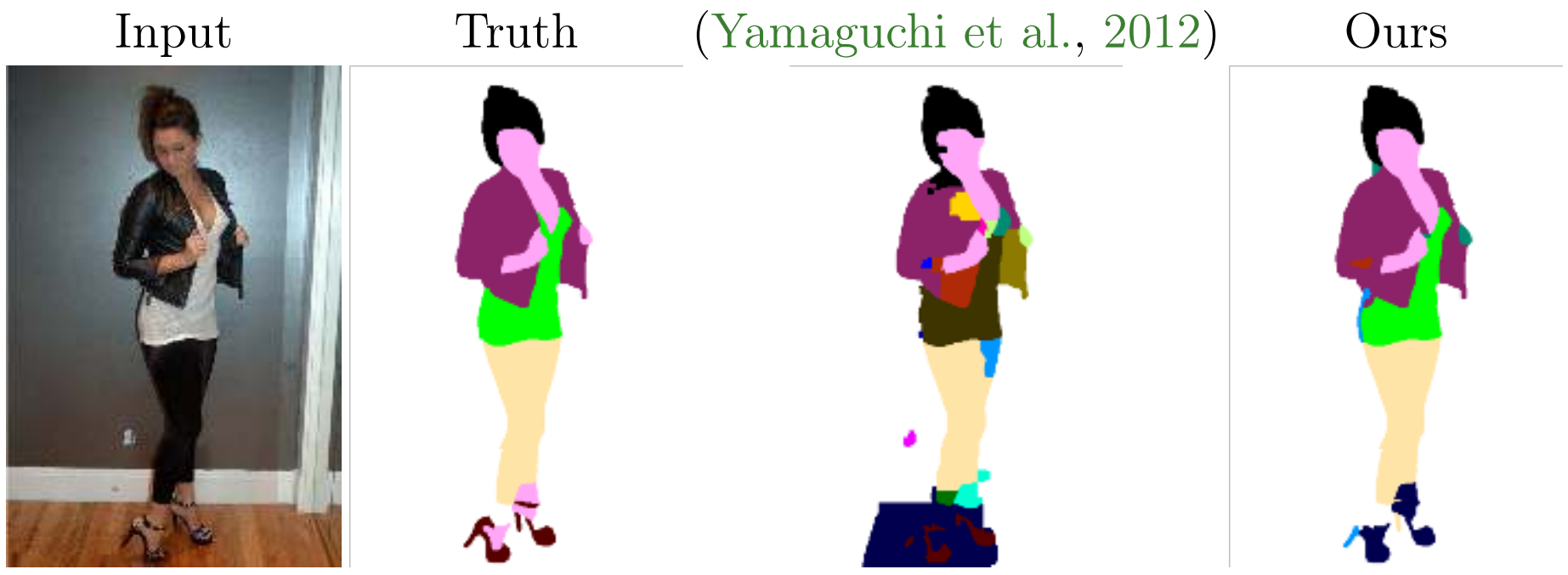

**Figure 6.1: Garment segmentation example.** We show our approach in a scenario where no a priori knowledge of which garments are worn is provided. We compare against state-of-the-art. Despite some mistakes, our result looks visually much more natural than the competing method.

Computer vision researchers have started to be interested in the subject due to the high impact of the application domain (Bossard et al., 2012; Bourdev et al., 2011; Chen et al., 2012; Gallagher and Chen, 2008; Hasan and Hogg, 2010; Jammalamadaka et al., 2013; Liu et al., 2012b; Wang and Ai, 2011). The main focus has been to infer clothing from photographs. This can enable a variety of applications such as virtual garments in online shopping. Being able to automatically parse clothing is also key in order to conduct large-scale sociological studies related to family income or urban groups (Murillo et al., 2012; Song et al., 2011).

In the context of fashion, Yamaguchi et al. (Yamaguchi et al., 2012), created *Fashionista*, a dataset of images and clothing segmentation labels. Great performance was obtained when the system was given information about which garment classes, but not their location, are present for each test image. Unfortunately, the performance of the state-of-the-art methods (Yamaguchi et al., 2012; Jammalamadaka et al., 2013) is rather poor when this kind of information is not provided at test time. This has been very recently partially addressed in (Yamaguchi et al., 2013) by utilizing over 300,000 weakly labeled images, where the weak annotations are in the form of image-level tags. In this chapter, we show an approach which outperforms the state-of-the-art significantly without requiring these additional annotations, by exploiting the specific domain of the task: clothing a person. An example of our result can be seen in Fig. 6.1.

In this chapter we also present a second task in which our goal is to predict how fashionable a person looks on a particular photograph. The fashionability is affected by the garments the subject is wearing, but also by a large number of other factors such as how appealing the scene behind the person is, how the image was taken, how visually appealing the person is, her/his age, etc. The garment itself being fashionable is also not a perfect indicator of someone's fashionability as people typically also judge how well the garments align with someone's "look", body characteristics, or even personality.

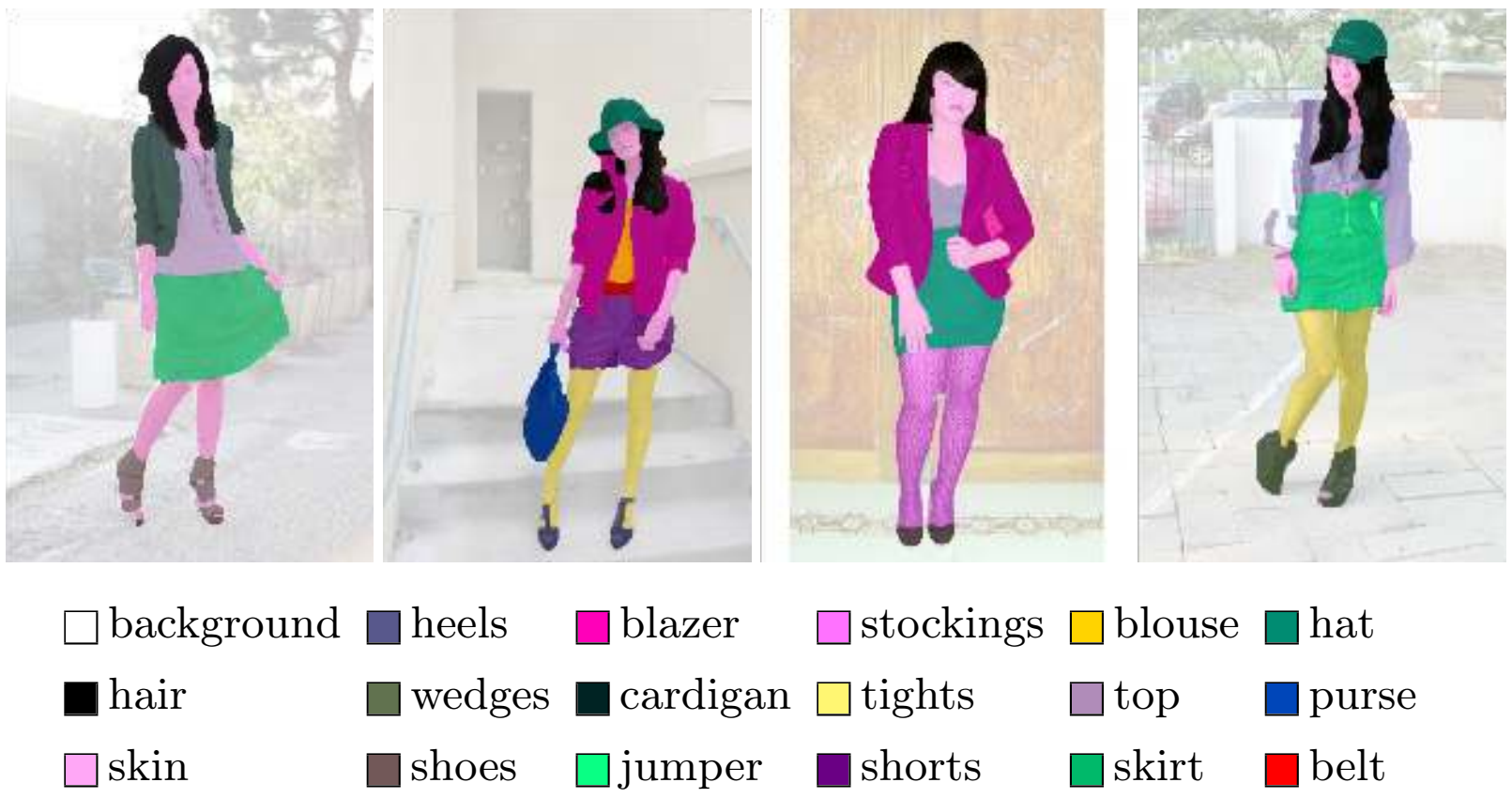

**Figure 6.2: Fine-grained annotations of the Fashionista dataset** (Yamaguchi et al., 2012). Many of the different classes are very difficult to distinguish even for humans. Observe the subtle differences between some classes such as footwear (heels, wedges, and shoes), blazer and cardigan, or stockings and tights. We also point out that this dataset has been annotated via superpixels, and thus the ground truth contains errors when superpixels do not align with the actual garments. We have not modified the ground truth segmentation in any way.

## 6.2 Segmentation of Garments in Fashion Images

The complexity of the task of human semantic segmentation comes from the inherent variability of pose and cloth appearances, the presence of self-occlusions as well as the potentially large number of classes. Consider for example Fig. 6.2: an autonomous system needs to distinguish between blazers and cardigans, stockings and tights, and heels, wedges and shoes, where the intra-class variability is fundamentally much larger than the inter-class variability. This fine-grained categorization is difficult to resolve even for humans who are not familiar with the fashion industry. The problem is further aggravated by the power law distribution of classes, as certain categories have very few examples. Thus, extra-care has to be taken into account to not over-predict the classes that are very likely to appear in each image, e.g., skin, hair.

In this section we address some of these challenges and formulate the problem as the one of inference in a Conditional Random Field (CRF), which takes into account the complex dependencies between clothing and human pose. Specifically, we develop a rich set of potentials which encode the person's global appearance and shape to perform figure/ground segmentation, shape and location likelihoods for each garment, which we call *clothelets*, and long-range similarity between segments to encourage, for example, T-Shirt pixels on the body to agree with the T-shirt pixels on the person's arm. We further exploit the fact the people are symmetric and dress as such as well by introducing symmetry-based potentials between different limbs. We also use a variety of different local features encoding cloth appearance as well as local shape of the person's parts. We demonstrate the effectiveness of our approach on the Fashionista dataset (Yamaguchi et al., 2012) and show that our approach significantly outperforms the existing state-of-the-art.

## Related Work

There has been a growing interest in recognizing outfits and clothing from still images. One of the first approaches on the subject was Chen et al. (Chen et al., 2006), which manually built a composite clothing model, that was then matched to input images. This has led to more recent applications for learning semantic clothing attributes (Chen et al., 2012), which are in turn used for describing and recognizing the identity of individuals (Bourdev et al., 2011; Gallagher and Chen, 2008), their style (Bossard et al., 2012), and performing sociological studies such as predicting the occupation (Song et al., 2011) or urban tribes (Murillo et al., 2012). Other tasks like outfit recommendations (Liu et al., 2012a) have also been investigated. However, in general, these approaches do not perform accurate segmentation of clothing, which is the goal of our approach. Instead, they rely on more coarse features such as bounding boxes and focus on producing generic outputs based on the presence/absence of a specific type of outfit. It is likely that the performance of such systems would improve if accurate clothing segmentation would be possible.

Recent advances in 2D pose estimation (Bourdev and Malik, 2009; Yang and Ramanan, 2011) have enabled a more advanced segmentation of humans (Dong et al., 2013). However, most approaches have focused on figure/ground labeling (Ladicky et al., 2013; Wang and Koller, 2011). Additionally, pose information has been used as a feature in clothing related tasks such as finding similar worn outfits in the context of online shopping (Liu et al., 2012b).

Segmentation and classification of garments has been addressed in the restrictive case in which the labels are known beforehand (Yamaguchi et al., 2012). The original paper tackled this problem in the context of fashion photographs which depicted one person typically in an upright pose. This scenario has also been extended to the case where more than one individual can be present in the image (Jammalamadaka et al., 2013). In order to perform the segmentation, conditional random fields are used with potentials linking clothing and pose. However, the performance of these approaches drops significantly when no information about the outfit is known a priori (i.e., no tags are provided at test time). The paper doll approach (Yamaguchi et al., 2013) uses over 300,000 weakly labeled training images and a small set of fully labeled examples in order to enrich the model of (Yamaguchi et al., 2012) with a prior over image labels. As we will show in the experimental evaluation, our method can handle this scenario without having to resort to additional training images. Furthermore, it consistently outperforms (Yamaguchi et al., 2012, 2013).

CRFs have been very successful in semantic segmentation tasks. Most approaches combine detection and segmentation by using detectors as additional image evidence (Yao et al., 2012b; Fidler et al., 2013). Co-occurrence potentials have been employed to enforce consistency among region labels (Ladicky et al., 2010). Part-based detectors have also been aligned to image contours to aid in object segmentation (Brox et al., 2011). All these strategies have been applied to generic segmentation problems, where one is interested in segmenting classes such as car, sky or trees. Pixel-wise labeling of clothing is, however, a much more concrete task, where strong domain specific information, such as 2D body pose, can be used to reduce ambiguities.

| Type | Name | Description |
|---|---|---|
| unary | Simple features ($\phi_{i,j}^{simple}(y_i)$) | Assortment of simple features. |
| unary | Object mask ($\phi_{i,j}^{obj}(y_i)$) | Figure/ground segmentation ask. |
| unary | Clothelets ($\phi_{i,j}^{cloth}(y_i)$) | Pose-conditioned garment likelihood masks. |
| unary | Ranking ($\phi_{i,j}^{o2p}(y_i)$) | Rich set of region ranking features. |
| unary | Bias ($\phi_{j}^{bias}(y_i)$ and $\phi_{p,j}^{bias}(l_p)$) | Class biases. |
| pairwise | Similarity ($\phi_{m,n}^{simil}(y_m, y_n)$) | Similarity between superpixels. |
| pairwise | Compatibility ($\phi_{i,p}^{comp}(y_i, l_p)$) | Edges between limb segments and superpixels. |

**Table 6.1: Overview of the different types of potentials used in the proposed CRF model.**

## Clothing a Person

Our proposed Conditional Random Field (CRF) takes into account complex dependencies that exist between garments and human pose. We obtain pose by employing a 2D articulated model by Yang et al. (Yang and Ramanan, 2011) which predicts the main keypoints such as head, shoulders, knees, etc. As (Yamaguchi et al., 2012), we will exploit these keypoints to bias the clothing labeling in a plausible way (e.g., a hat is typically on the head and not the feet). To manage the complexity of the segmentation problem we represent each input image with a small number of superpixels (P.Arbelaez et al., 2011). Our CRF contains a variable encoding the garment class (including background) for each superpixel. We also add limb variables which encode the garment associated with a limb in the human body and correspond to edges in the 2D articulated model. We use the limb variables to propagate information while being computationally efficient.

Our CRF contains a rich set of potentials which exploit the domain of the task. We use the person's global appearance and shape to perform figure/ground segmentation in order to narrow down the scope of cloth labeling. We further use shape and location likelihoods for each garment, which we call *clothelets*. We exploit the fact that people are symmetric and typically dress as such by forming long-range consistency potentials between detected symmetric keypoints of the human pose. We finally also use a variety of different features that encode appearance as well as local shape of superpixels.

We will now discuss the different potentials in the model. All our potentials belong to the exponential function family, however, for notation simplicity we will write the logarithm of the potentials and not make explicit the dependency with the observed variables.

### Pose-aware Model

Given an input image represented with superpixels, our goal is to assign a clothing label (or background) to each of them.

More formally, let $y_i \in \{1, \cdots, C\}$ be the class associated with the $i$-th superpixel, and let $l_p$ be the $p$-th limb segment defined by the edges in the articulated body model. Each limb $l_p$ is assumed to belong to one class, $l_p \in \{1, \cdots, C\}$. To encode body symmetries in an efficient manner, we share limb variables between the left and right part of the human body, e.g., the left and the right leg share the same limb variables. We propose several domain inspired potentials, the overview of which is presented in Table 6.1. We emphasize that the weights associated with each potential in our CRF will be learned using structure prediction. We now explain each potential in more detail.

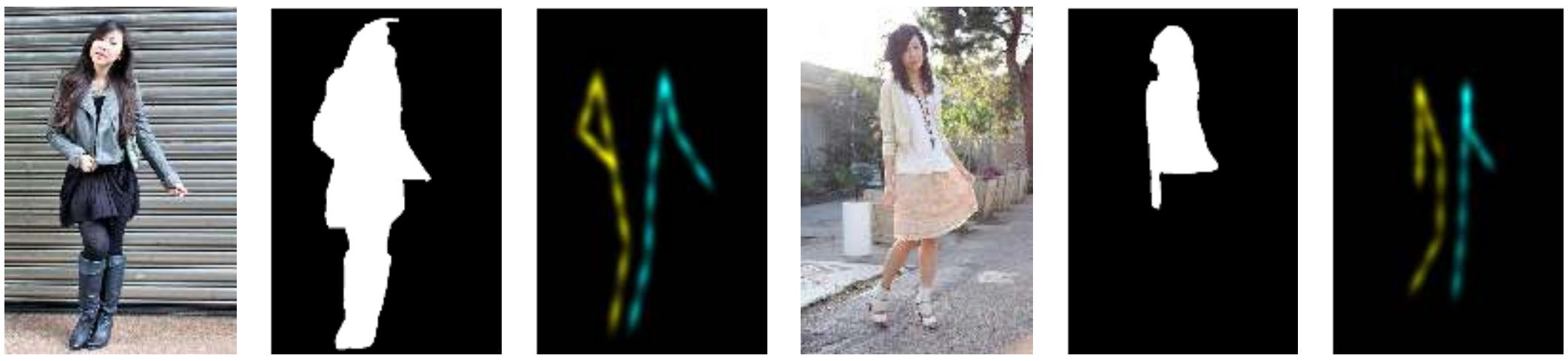

**Figure 6.3: Visualization of object masks (Carreira and Sminchisescu, 2012) and limbs (obtained via (Yang and Ramanan, 2011)).** Note that CPMC typically generates high quality results, e.g. the one in the left image, but can also completely miss large parts of the body as shown in the image on the right.

**Simple Features:** Following (Yamaguchi et al., 2012), we concatenate a diverse set of simple local features and train a logistic regression classifier for each class. In particular, we use colour features, normalized histograms of RGB and CIE L*a*b* colour; texture features, Gabor filter responses; and location features: both normalized 2D image coordinates and pose-relative coordinates. The output of the logistic functions are then used as unary features in the CRF. This results in a unary potential with as many dimensions as classes:

$$\phi_{i,j}^{simple}(y_i) = \begin{cases} w_j^{simple}\sigma_j^{simple}(f_i), & \text{if } y_i = j \\ 0, & \text{otherwise} \end{cases}, \tag{6.1}$$

where $w_j^{simple}$ is the weight for class $j$, $\sigma_j^{simple}(f_i)$ is the score of the classifier for class $j$, and $f_i$ is the concatenation of all the features for superpixel $i$. Notice we have used $C$ different unary potentials, one for each class. By doing this, we allow the weights of a variety of potentials and classes be jointly learned within the model.

**Figure/Ground Segmentation:** To facilitate clothing parsing we additionally compute how likely it is that each superpixel belongs to a person. We do this by computing a set of bottom-up region proposals using the Constrained Parametric Min-Cuts (CPMC) approach (Carreira and Sminchisescu, 2012), which repetitively solves a figure/ground energy minimization problem with different parameters and seeds via parametric min-cuts. We take top $K$ (we set $K = 100$) regions per image and use Order 2 Pooling (O2P) (Carreira et al., 2012) to score each region into figure/ground (person-vs-background). Since we know that there is a person in each image, we take at least the top scoring segment per image, no matter its score. For images with multiple high scoring segments, we take the union of all segments with scores higher than a learned threshold (Carreira et al., 2012). We define a unary potential to encourage the superpixels that lie inside the foreground mask to take any of the clothing labels (and not

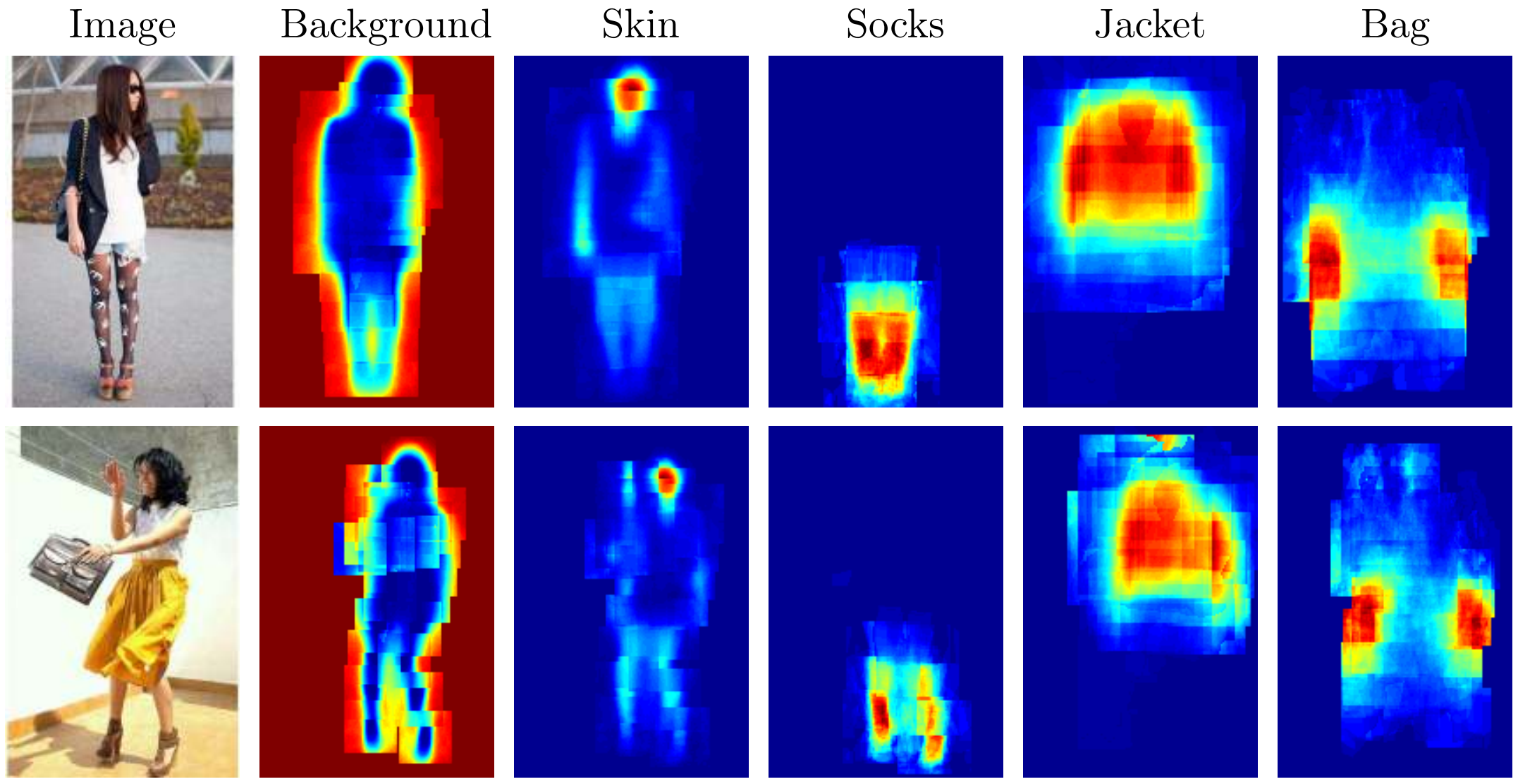

**Figure 6.4: Visualization of clothelets.** We show several clothelets for two different input images.

background):

$$\phi^{obj}_{i,j}(y_i) = \begin{cases} w^{cmpc}_1 \sigma^{cpmc} \cdot |\neg M_{fg} \cap S_i|/|S_i|, & \text{if } y_i = 1 \\ w^{cpmc}_j \sigma^{cpmc} \cdot |M_{fg} \cap S_i|/|S_i|, & \text{otherwise} \end{cases}, \quad (6.2)$$

where $y_i = 1$ encodes the background class, $w^{cpmc}_j$ is the weight for class $j$, $\sigma^{cpmc}$ is the score of the foreground region, $S_i$, $M_{fg}$ are binary masks defining the superpixel and foreground, respectively, and $\neg M_{fg}$ is a mask of all pixels not in foreground. Fig. 6.3 shows examples of masks obtained by (Carreira et al., 2012). Note that while in some cases it produces very accurate results, in others, it performs poorly. These inaccurate masks are compensated by other potentials.

**Clothelets:** Our next potential exploits the statistical dependency between the location on the human body and garment type. Its goal is to make use of the fact that e.g. jeans typically cover the legs and not the head. We compute a likelihood of each garment appearing in a particular relative location of the human pose. In particular, for each training example we take a region around the location of each joint (and limb), the size of which corresponds to the size of the joint part template encoded in (Yang and Ramanan, 2011). We average the GT segmentation masks for each class across the training examples. In order to capture garment classes that stray away from the pose, we use boxes that are larger than the part templates in (Yang and Ramanan, 2011). At test time, the masks for each class are overlaid relative to the inferred pose and normalized by the number of non-zero elements. Areas with no information are assigned to the background class. The potential is then defined as

$$\phi^{cloth}_{i,j}(y_i) = \begin{cases} w^{cloth}_j (\text{clothelet}^j_i \cdot S_i)/|S_i|, & \text{if } y_i = j \\ 0, & \text{otherwise} \end{cases}, \quad (6.3)$$

where $w_j^{cloth}$ is the weight for class $j$, $\text{clothelet}_i^j$ is the clothelet for the $j$-th class, and $\cdot$ is the dot product. Fig. 6.4 depicts clothelets for a few sample classes. Note that these likelihoods are conditioned on the pose.

**Shape Features:** This potential uses a set of rich features that exploit both the shape and local appearance of garments. In particular, we use eSIFT and eMSIFT proposed by (Carreira et al., 2012) for region description. Given a region, both descriptors extract SIFT inside the region and enrich it with the relative location and scale within the region. Second-order pooling is used to define the final region descriptor. eSIFT and eMSIFT differ slightly in how the descriptors are pooled, eSIFT pools over both the region and background of the region, while eMSIFT pools over the region alone.

While (Carreira et al., 2012) defines the features over full object proposals, here we compute them over each superpixel. As such, they capture more local shape of the part/limb and local appearance of the garment. We train a logistic classifier for each type of feature and class and use the output as our potential:

$$\phi_{i,j}^{o2p}(y_i) = \begin{cases} w_j^{o2p} \sigma_j^{o2p}(r_i), & \text{if } y_i = j \\ 0, & \text{otherwise} \end{cases}, \tag{6.4}$$

with $w_j^{o2p}$ is the weight for class $j$, $\sigma_j^{o2p}(r_i)$ the classifier score for class $j$, and $r_i$ the feature vector for superpixel $i$.

**Bias:** We use a simple bias for the background to encode the fact that it is the class that appears more frequently. Learning a weight for this bias is equivalent to learning a threshold for the foreground, however within the full model. Thus:

$$\phi^{bias}(y_i) = \begin{cases} w^{bias}, & \text{if } y_i = 1 \\ 0, & \text{otherwise} \end{cases}, \tag{6.5}$$

where $w^{bias}$ is the single bias weight.

**Similarity:** In CRFs, neighboring superpixels are typically connected via a (contrast-sensitive) Potts model encouraging smoothness of the labels. For clothing parsing, we want these connections to act on a longer range. That is, a jacket is typically split in multiple disconnected segments due to a T-shirt, tie, and/or a bag. Our goal is to encourage superpixels that are similar in appearance to agree on the label, even though they may not be neighbors in the image.

We follow (Uijlings et al., 2013) and use size similarity, fit similarity that measures how well two superpixels fit each other; and colour and texture similarity, with the total of 12 similarity features between each pair of superpixels. We then train a logistic regression to predict if two superpixels should have the same label or not. In order to avoid setting connections on the background, we only connect superpixels that overlap with the bounding box of the 2D pose detection. Note that connecting all pairs of similar superpixels would slow down inference considerably. To alleviate this problem, we compute the minimum spanning tree using the similarity matrix and use the top 10 edges to connect 10 pairs of superpixels in each image. We form a pairwise potential between each connected pair:

$$\phi^{simil}_{m,n}(y_m, y_n) = \begin{cases} w^{simil}_i \sigma^{simil}_{m,n}, & \text{if } y_m = y_n = i \\ 0, & \text{otherwise} \end{cases}, \tag{6.6}$$

where $w^{simil}_i$ is the weight for class $i$, and $\sigma^{simil}_{m,n}$ is the output of the similarity classifier.

**Limb Segment Bias:** We use a per-class bias on each limb segment to capture a location specific bias, e.g., hat only appears in the head:

$$\phi^{bias}_{p,j}(l_p) = \begin{cases} w^{bias}_{p,j}, & \text{if } l_p = j \\ 0, & \text{otherwise} \end{cases}, \tag{6.7}$$

where $w^{bias}_{p,j}$ is a weight for part limb $p$ and class $j$ that allows us to compute which classes are more frequent in each limb.

**Compatibility Segmentation-Limbs:** We define potentials connecting limb segments with nearby superpixels encouraging them to agree in their labels. Towards this goal, we first define a Gaussian mask centered between two joints. More formally, for two consecutive joints with coordinates $J_a = (u_a, v_a)$ and $J_b = (u_b, v_b)$, we define the mask based on the following Normal distribution:

$$M(J_a, J_b) = \mathcal{N}\left(\frac{J_a + J_b}{2}, \; R \begin{pmatrix} q_1 \|J_a - J_b\| & 0 \\ 0 & q_2 \end{pmatrix} R^{\mathrm{T}}\right), \tag{6.8}$$

where $R$ is a 2D rotation matrix with an angle $\arctan(\frac{u_a - u_b}{v_a - v_b})$, and $q_1$ and $q_2$ are two hyperparameters controlling the spread of the mask longitudinally and transversely, respectively. In order to additionally encode symmetry, a single limb segment is used for both left and right sides of the body, e.g., left and right shins will share the same keypoint. The strength of the connection is based on the overlap between the superpixels and the Gaussian mask:

$$\phi^{comp}_{i,p}(y_i, l_p) = \begin{cases} w^{comp}_{j,p} M(J_a, J_b) \cdot S_i, & \text{if } y_i \neq 1 \text{ and } y_i = k_p = j \\ 0, & \text{otherwise} \end{cases}, \tag{6.9}$$

where $w^{comp}_{i,p}$ is the weight for part $p$ and class $j$. For computational efficiency, edges with connection strengths below a threshold are not set in the model. Some examples of the limb segment masks are shown in Fig. 6.3. We can see the masks fit the body tightly to avoid overlapping with background superpixels.

**Full Model:** We define the energy of the full model to be the sum of three types of energies encoding unary and pairwise potentials that depend on the superpixel labeling, as well as an energy term linking the limb segments and the superpixels:

$$E(\mathbf{y}, l) = E_{unary}(\mathbf{y}) + E_{similarity}(\mathbf{y}) + E_{limbs}(\mathbf{y}, l) \,. \tag{6.10}$$

This energy is maximized during inference. The unary terms are formed by the concatenation of appearance features, figure/ground segmentation, clothelets, shape features

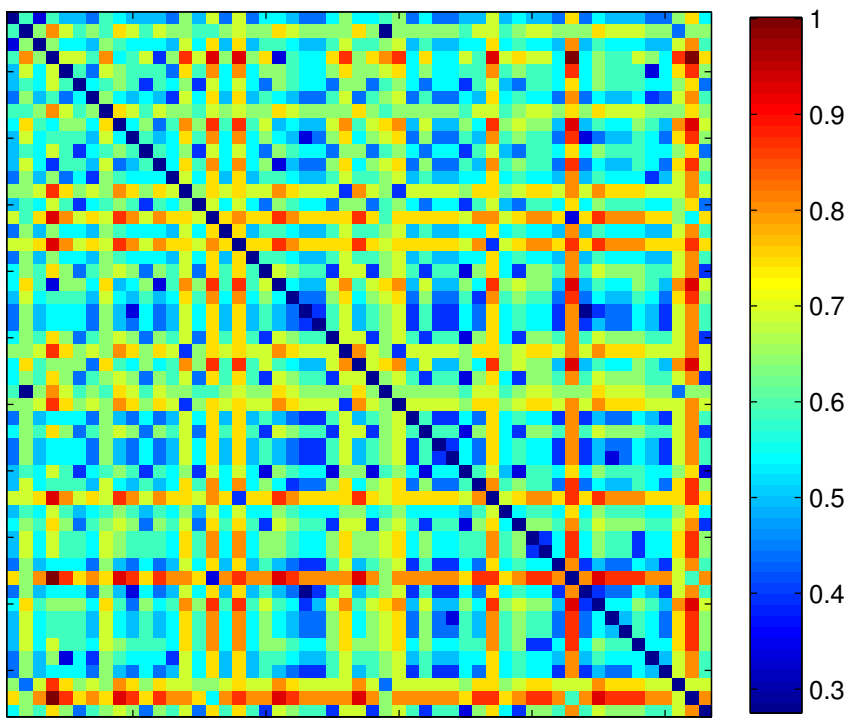

| Clothing classes | | | Similarity ($d_{\text{lch}}$) |
|---|---|---|---|
| jumper | vs | sweater | 1.000 |
| jumper | vs | sweatshirt | 0.697 |
| jumper | vs | top | 0.557 |
| jumper | vs | stockings | 0.507 |
| jumper | vs | heels | 0.465 |
| jumper | vs | sunglasses | 0.341 |
| jumper | vs | watch | 0.317 |

**Figure 6.5: Wordnet-based similarity function.** We use the inverse of the Leacock-Chodorow Similarity between the classes in the Fashionista dataset to weigh our loss function. We display the similarity matrix between all the classes on the left. Some individual values of the similarity between the jumper class and several other classes can be seen on the right.

and background bias for a total of $K = (1 + 5C)$ features

$$E_{unary}(\mathbf{y}) = \sum_{i=1}^{N} \sum_{j=1}^{K} \phi_{i,j}^{unary}(y_i) \ , \tag{6.11}$$

where $N$ is the number of superpixels. The pairwise features encode the similarity between different pairs of superpixels as we describe above

$$E_{similarity}(\mathbf{y}) = \sum_{(m,n)\in\text{pairs}} \phi_{m,n}^{simil}(y_m, y_n) \ . \tag{6.12}$$

The limb-superpixel compatibility term is defined as

$$E_{limbs}(\mathbf{y}, l) = \sum_{p=1}^{M} \left( \sum_{j=1}^{C} \left( \phi_{p,j}^{bias}(l_p) + \sum_{i=1}^{N} \phi^{comp}(y_i, l_p) \right) \right) \ , \tag{6.13}$$

for a total of $(M + C)$ features, with $M$ the number of limb segments.

### Learning and Inference

Our model is a multi-label CRF which contains cycles and thus inference is NP-hard. We use a message passing algorithm, distributed convex belief propagation (Schwing et al., 2011) to perform inference. It belongs to the set of LP-relaxation approaches, and has convergence guarantees. This is not the case in other message passing algorithms such as loopy-BP.

To learn the weights, we use the primal-dual method of (Hazan and Urtasun, 2010) (we use the implementation of (Schwing et al., 2012)), shown to be more efficient than other structure prediction learning algorithms.

As loss-function, we use the semantic similarity between the different classes in order to penalize mistakes between unrelated classes more than similar ones. We do this via Wordnet (Miller, 1995), which is a large lexical database in which sets of cognitive synonyms (synsets) are interlinked by means of semantic and lexical relationships. We can unambiguously identify each of the classes with a single synset, and then proceed to calculate similarity scores between these synsets that represent the semantic similarity between the classes, in order to penalize mistakes with dissimilar classes more.

| | 29 Classes | | 56 Classes | | |
|---|---|---|---|---|---|
| Method | Y2012 | Ours | Y2012 | Y2013 | Ours |
| Jaccard index | 12.32 | **20.52** | 7.22 | 9.22 | **12.28** |

**Table 6.2: Comparison against the state-of-the-art.** We compare on two different datasets: Fashionista v0.2 with 56 classes and Fashionista v0.3 with 29 classes, against Y2012 (Yamaguchi et al., 2012) and Y2013 (Yamaguchi et al., 2013).

In particular, we choose the corpus-independent Leacock-Chodorow Similarity score. This score takes into account the shortest path length $p$ between both synsets and the maximum depth of the taxonomy $d$ at which they occur. It is defined as the relationship $-\log(p/2d)$. A visualization of the dissimilarity between all the classes in the dataset can be seen in Fig. 6.5. We therefore define the loss-function as:

$$\Delta^y(y_i, y_i^*) = \begin{cases} 0, & \text{if } y_i = y_i^* \\ d_{\text{lch}}(y_i, y_i^*), & \text{otherwise} \end{cases} \tag{6.14}$$

with $d_{\text{lch}}(\cdot,\cdot)$ being the inverse Leacock-Chodorow Similarity score between both classes. For the limb segments we use a 0-1 loss:

$$\Delta^k(k_i, k_i^*) = \begin{cases} 0, & \text{if } k_i = k_i^* \\ 1, & \text{otherwise} \end{cases} \tag{6.15}$$

### Experimental Evaluation

We evaluate our approach on both a the Fashionista dataset v0.3 (Yamaguchi et al., 2012), and the setting of (Yamaguchi et al., 2013) with the Fashionista dataset v0.2. Both datasets are taken from chictopia.com in which a single person appears wearing a diverse set of garments. The dataset provides both annotated superpixels as well as 2D pose annotations. A set of evaluation metrics and the full source code of approaches (Yamaguchi et al., 2012, 2013) are provided. Version 0.2 has 685 images and v0.3 has 700 images. Note that v0.3 is not a superset of v0.2.

We have modified the Fashionista v0.3 dataset in two ways. First we have compressed the original 54 classes into 29. This is due to the fact that many classes that appear have very few occurrences. In fact, in the original dataset, 13 classes have 10 or fewer examples and 6 classes have 3 or fewer instances. This means that when performing a random split of the samples into training and test subsets, there is a high probability that some classes will only appear in one of the subsets. We therefore compress the classes by considering both semantic similarity and the number of instances.

| | null | hair | skin | glasses | accessories | misc garments | footwear | shoes | socks | boots | purse | bag | hat | dress | upper body | shirt | sweater | top | jacket | vest | blouse | jeans | leggings | pants | shorts | skirt | stockings | tights | belt |
|---|---|---|---|---|---|---|---|---|---|---|---|---|---|---|---|---|---|---|---|---|---|---|---|---|---|---|---|---|---|
| null | .97 | .00 | .00 | 0.0 | 0.0 | 0.0 | .00 | .00 | .00 | .00 | 0.0 | .00 | 0.0 | .00 | .00 | 0.0 | 0.0 | 0.0 | .00 | 0.0 | 0.0 | .00 | 0.0 | .00 | 0.0 | 0.0 | 0.0 | 0.0 | 0.0 |
| hair | .03 | .67 | .05 | .00 | .00 | .01 | .00 | .00 | 0.0 | 0.0 | .00 | .00 | .08 | .00 | .05 | .00 | .00 | .00 | .04 | .02 | .01 | 0.0 | .00 | .00 | 0.0 | 0.0 | 0.0 | .00 | 0.0 |
| skin | .04 | .02 | .82 | .00 | .00 | .00 | .02 | .01 | .00 | .00 | .00 | .00 | .00 | .01 | .02 | .01 | .00 | .00 | .00 | 0.0 | .00 | .00 | 0.0 | .00 | 0.0 | .00 | .01 | .01 | 0.0 |
| glasses | .01 | .14 | .14 | .63 | .00 | .01 | 0.0 | 0.0 | 0.0 | 0.0 | 0.0 | 0.0 | .04 | 0.0 | .00 | 0.0 | 0.0 | 0.0 | .02 | 0.0 | 0.0 | 0.0 | 0.0 | 0.0 | 0.0 | 0.0 | 0.0 | 0.0 | .01 |
| accessories | .10 | .02 | .15 | .01 | .14 | .07 | .01 | .00 | .05 | .06 | .02 | .03 | .02 | .12 | .04 | .02 | .00 | .05 | .03 | .00 | .00 | 0.0 | 0.0 | .01 | 0.0 | 0.0 | 0.0 | 0.0 | .04 |
| misc garments | .01 | .15 | .04 | .01 | .00 | .06 | .00 | .01 | 0.0 | .00 | .00 | .02 | 0.0 | .07 | .10 | .02 | .04 | .07 | .16 | .07 | .02 | .01 | .03 | .04 | 0.0 | .00 | .00 | .00 | .03 |
| footwear | .12 | 0.0 | .04 | 0.0 | 0.0 | .00 | .43 | .30 | .04 | .03 | 0.0 | .00 | 0.0 | 0.0 | 0.0 | 0.0 | 0.0 | 0.0 | 0.0 | 0.0 | 0.0 | .00 | 0.0 | .01 | 0.0 | 0.0 | .01 | .01 | 0.0 |
| shoes | .10 | .00 | .05 | 0.0 | 0.0 | 0.0 | .19 | .35 | .07 | .11 | .02 | .02 | 0.0 | 0.0 | 0.0 | 0.0 | 0.0 | 0.0 | 0.0 | .00 | 0.0 | .01 | .02 | .01 | 0.0 | 0.0 | .03 | .01 | 0.0 |
| socks | .01 | 0.0 | .04 | 0.0 | 0.0 | 0.0 | .02 | .07 | .66 | .09 | 0.0 | 0.0 | 0.0 | 0.0 | 0.0 | 0.0 | 0.0 | 0.0 | 0.0 | 0.0 | 0.0 | .00 | 0.0 | .00 | 0.0 | 0.0 | .05 | .04 | 0.0 |
| boots | .08 | .00 | .03 | 0.0 | 0.0 | 0.0 | .08 | .15 | .09 | .35 | 0.0 | .01 | 0.0 | 0.0 | 0.0 | 0.0 | 0.0 | 0.0 | 0.0 | 0.0 | 0.0 | .08 | .06 | .01 | 0.0 | .01 | .02 | .02 | 0.0 |
| purse | .17 | .00 | .11 | 0.0 | 0.0 | .01 | .01 | 0.0 | .00 | .03 | .10 | .31 | 0.0 | .05 | .05 | .04 | 0.0 | .01 | .05 | 0.0 | .00 | 0.0 | 0.0 | 0.0 | .01 | .02 | 0.0 | .01 | .01 |
| bag | .14 | .00 | .07 | 0.0 | .00 | .01 | .01 | .01 | .01 | .03 | .11 | .27 | 0.0 | .05 | .04 | .01 | .02 | .00 | .02 | .03 | .00 | .02 | .00 | .02 | .01 | .08 | .02 | .01 | .00 |
| hat | .13 | .11 | .03 | .00 | 0.0 | 0.0 | 0.0 | 0.0 | 0.0 | 0.0 | 0.0 | 0.0 | .70 | 0.0 | 0.0 | 0.0 | .00 | 0.0 | .00 | .01 | 0.0 | 0.0 | 0.0 | 0.0 | 0.0 | 0.0 | 0.0 | 0.0 | .00 |
| dress | .01 | .01 | .03 | 0.0 | .01 | .03 | .00 | .00 | 0.0 | .00 | .04 | .05 | 0.0 | .26 | .05 | .07 | .04 | .03 | .05 | .04 | .02 | .01 | .01 | .05 | .01 | .17 | .00 | .00 | .01 |
| upper body | .01 | .02 | .05 | 0.0 | .00 | .02 | .00 | .00 | .00 | .01 | .03 | .01 | .01 | .04 | .37 | .02 | .05 | .02 | .11 | .05 | .04 | .02 | .00 | .09 | .00 | .01 | .00 | 0.0 | .00 |
| shirt | .02 | .02 | .04 | 0.0 | .00 | .04 | .00 | .00 | .00 | .00 | .00 | .00 | 0.0 | .09 | .08 | .27 | .04 | .12 | .11 | .01 | .05 | .00 | 0.0 | .02 | .03 | .01 | 0.0 | .01 | .03 |
| sweater | .02 | .01 | .05 | 0.0 | .00 | .03 | .01 | .01 | .01 | 0.0 | 0.0 | .00 | .00 | .13 | .20 | .10 | .07 | .04 | .04 | .01 | .05 | .00 | 0.0 | .16 | .01 | .03 | 0.0 | .00 | .01 |
| top | .01 | .02 | .08 | 0.0 | .00 | .07 | .00 | 0.0 | .02 | 0.0 | .01 | .01 | .00 | .08 | .06 | .15 | .07 | .19 | .07 | .03 | .06 | 0.0 | .03 | 0.0 | .03 | .01 | 0.0 | 0.0 | .02 |
| jacket | .03 | .02 | .02 | 0.0 | .00 | .03 | .00 | .00 | 0.0 | .00 | .01 | .01 | .00 | .03 | .29 | .05 | .03 | .03 | .31 | .07 | .01 | 0.0 | .00 | .02 | .00 | 0.0 | 0.0 | 0.0 | .00 |
| vest | .01 | .01 | .04 | 0.0 | .00 | .04 | 0.0 | .00 | 0.0 | .00 | .02 | .02 | .00 | .16 | .11 | .18 | .02 | .04 | .23 | .08 | .00 | 0.0 | .01 | 0.0 | .00 | .00 | 0.0 | .01 | .01 |
| blouse | .03 | .03 | .02 | 0.0 | .01 | .05 | .00 | 0.0 | .00 | .00 | .02 | .00 | 0.0 | .14 | .11 | .13 | .05 | .15 | .09 | .00 | .09 | .01 | 0.0 | 0.0 | .00 | .03 | 0.0 | 0.0 | .01 |
| jeans | .01 | 0.0 | .01 | 0.0 | .00 | 0.0 | .00 | .01 | .04 | .03 | .00 | .02 | 0.0 | .05 | .01 | .01 | .00 | .00 | .00 | 0.0 | 0.0 | .37 | .15 | .24 | .03 | 0.0 | .02 | .01 | .01 |
| leggings | .08 | 0.0 | .03 | 0.0 | 0.0 | 0.0 | 0.0 | .01 | .29 | .02 | .02 | .03 | 0.0 | .02 | 0.0 | .00 | 0.0 | 0.0 | 0.0 | 0.0 | 0.0 | .12 | .17 | .09 | .00 | 0.0 | .05 | .07 | 0.0 |
| pants | .01 | 0.0 | .08 | 0.0 | 0.0 | .00 | .01 | .00 | .04 | .01 | .00 | .02 | .04 | .01 | .00 | .02 | 0.0 | 0.0 | 0.0 | .00 | 0.0 | .25 | .10 | .27 | .00 | .02 | .01 | .09 | .00 |
| shorts | .00 | 0.0 | .02 | .01 | 0.0 | 0.0 | .00 | 0.0 | 0.0 | 0.0 | 0.0 | .02 | .00 | .22 | .02 | .03 | .00 | .03 | .00 | .00 | 0.0 | .02 | .02 | .02 | .47 | .10 | 0.0 | .00 | .02 |
| skirt | .04 | 0.0 | .05 | 0.0 | 0.0 | .01 | .00 | .01 | 0.0 | .05 | .02 | .05 | 0.0 | .09 | .03 | .04 | .05 | .01 | .02 | .00 | .01 | .03 | .08 | .06 | .09 | .20 | 0.0 | .03 | .02 |
| stockings | .01 | 0.0 | .09 | 0.0 | 0.0 | .00 | .00 | .01 | .18 | .16 | 0.0 | .01 | 0.0 | .06 | 0.0 | 0.0 | 0.0 | 0.0 | 0.0 | 0.0 | 0.0 | .03 | .06 | .07 | 0.0 | .00 | .17 | .14 | .01 |
| tights | .09 | 0.0 | .13 | 0.0 | 0.0 | 0.0 | .00 | .00 | .21 | .05 | .00 | .01 | 0.0 | 0.0 | 0.0 | 0.0 | 0.0 | 0.0 | 0.0 | 0.0 | 0.0 | .04 | .07 | .16 | 0.0 | .00 | .10 | .12 | 0.0 |
| belt | .02 | .01 | .03 | 0.0 | .01 | .00 | .01 | 0.0 | .00 | 0.0 | .02 | .01 | 0.0 | .11 | .02 | .01 | .02 | .04 | .02 | 0.0 | .02 | .00 | .00 | .02 | .00 | .02 | 0.0 | 0.0 | .60 |

**Figure 6.6: Confusion matrix for our approach on the Fashionista v0.3 dataset.**

For evaluation on Fashionista v0.3 we consider a random 50-50 train-test split. As previously stated, we do not have information about which classes are present in the scene. We employ the publicly available code of (Yamaguchi et al., 2012) as the baseline. We evaluate on Fashionista v0.2 according to the methodology in (Yamaguchi et al., 2013). This consists of a split with 456 images for training and 229 images for testing. Note that (Yamaguchi et al., 2013) uses 339,797 additional weakly labeled images from the Paper doll dataset for training, which we do not use.

Following PASCAL VOC, we report the average class intersection over union (Jaccard index). This metric is the most similar to human perception as it considers all true positives, true negatives and false positives. It is nowadays a standard measure to evaluate segmentation and detection (Everingham et al., 2010; Geiger et al., 2013; Deng et al., 2009).

**Comparison to State-of-the-Art:** We compare our approach against (Yamaguchi et al., 2012, 2013). The approach of (Yamaguchi et al., 2012) uses a CRF with very simple features. We adapt the code to run in the setting in which the labels that appear in the image are not known a priori. Note also that (Yamaguchi et al., 2013) uses a look-up approach on a separate dataset to parse the query images. The results of the comparison can be seen in Table 6.2. Note that our approach consistently outperforms both competing methods on both datasets, even though (Yamaguchi et al., 2013) uses 339,797 additional images for training. We roughly obtain a 60% relative improvement on Jaccard index metric with respect to (Yamaguchi et al., 2012) and a 30% improvement over (Yamaguchi et al., 2013). The full confusion matrix of our method can be seen in Fig. 6.6. We can identify several classes that have large appearance variation and similar positions that get easily confused, such as Footwear with Shoes and Jeans with Pants.

| Method | CPMC | Y2012 | Clothelets | Y2013 | Ours |
|---|---|---|---|---|---|
| Pixel Accuracy | - | 77.98 | 77.09 | 84.68 | **84.88** |
| Person/Bck. Accuracy | 85.39 | 93.79 | 94.77 | 95.79 | **97.37** |

**Table 6.3: Evaluation on foreground segmentation task.** We compare CPMC (Carreira et al., 2012), Y2012 (Yamaguchi et al., 2012), clothelets, Y2013 (Yamaguchi et al., 2013) and our approach for segmenting the foreground on the Fashionista v0.2 dataset.

| Method | | 29 Classes | 56 Classes |
|---|---|---|---|
| Y2012 | Estimated | 12.32 | 7.22 |
| | GT Pose | 12.39 | 7.41 |
| | No Pose | 10.54 | 5.22 |
| Ours | Estimated | 20.52 | 12.28 |
| | GT Pose | 21.01 | 12.46 |
| | No Pose | 16.56 | 9.64 |

**Table 6.4: Influence of pose.** We compare against Y2012 (Yamaguchi et al., 2012) in three different scenarios: estimated 2D pose, ground truth 2D pose and no pose information at all.

**Foreground Segmentation:** We also evaluate person-background segmentation results. Note that the binary segmentation in our model is obtained by putting all foreground garment classes to the person class. In Table 6.3, we show results for both pixel accuracy considering all the different classes, and the two class case of foreground/background segmentation accuracy. We see that the best results are obtained by the approaches reasoning jointly about the person and clothing. Even a simple model that uses only clothelets outperform the object mask by nearly 10% despite simply being a unary. Our approach outperforms the baseline CPMC (Carreira et al., 2012) by 12%, and achieves a 4% over (Yamaguchi et al., 2012) and 2% over (Yamaguchi et al., 2013).

**Pose Influence:** We next investigate the importance of having an accurate pose estimate. Towards this goal, we analyze three different scenarios. In the first one, the pose is estimated by (Yang and Ramanan, 2011). The second case uses the ground-truth pose, while the last one does not use pose information at all. As shown in Table 6.4, the poses in this dataset are not very complex as performance does not increase greatly when using ground truth instead of estimated pose. However, without pose information, performance drops 20%. This shows that our model is truly pose-aware.

A breakdown of the effect of pose on all the classes is shown in Fig. 6.7. Some classes like hat, belt or boots benefit greatly from pose information while others like shorts, tights or skin do not really change.

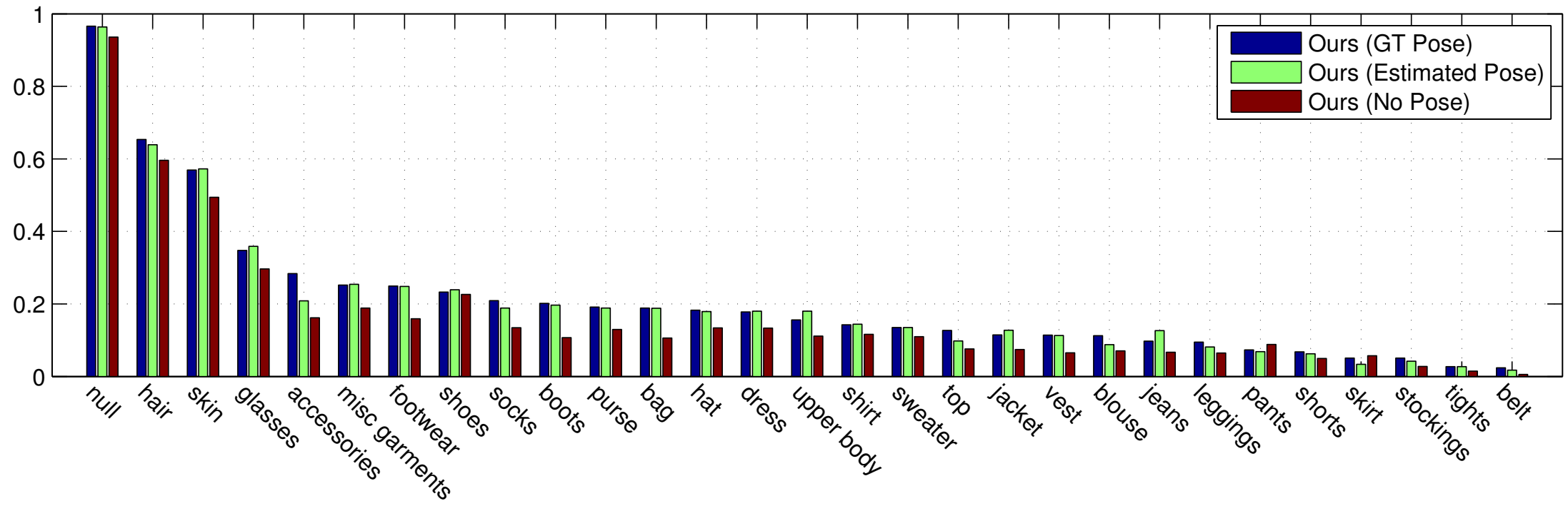

**Figure 6.7: Per class results for our model.** We evaluate with Jaccard index metric on Fashionista v0.3.

| Threshold | 0.16 | 0.10 | 0.05 |
|---|---|---|---|
| Mean superpixels/image | 50 | 120 | 290 |
| Jaccard index | 69.44 | 83.07 | 100 |

**Table 6.5: Oracle performance for different superpixel levels.** We show results for the Fashionista v0.3 dataset.

| Method | Simple features | Clothelets | eSIFT | eMSIFT |
|---|---|---|---|---|
| 29 Classes | 13.80 | 8.91 | 16.65 | 13.65 |
| 56 Classes | 7.93 | 3.02 | 9.29 | 7.80 |

**Table 6.6: Different results using only unary potentials in our model.** Both eSIFT and eMSIFT are ranking features from (Carreira et al., 2012).

**Oracle Performance:** Unlike (Yamaguchi et al., 2012), we do not use the fine level superpixels, but instead use coarser superpixels to speed up learning and inference. Table 6.5 shows that using coarser superpixels lowers the maximum achievable performance. However, by having larger areas, the local features become more discriminative. We also note that the dataset (Yamaguchi et al., 2012) was annotated by labeling the finer superpixels. As some superpixels do not follow boundaries well, the ground truth contains a large number of errors. We did not correct those, and stuck with the original annotations.

**Importance of the Features:** We also evaluate the influence of every potential in our model in Table 6.6. The eSIFT features obtain the best results under the Jaccard index metric. The high performance of eSIFT can be explained by the fact that it also takes into account the super pixel's background, thus capturing local context of garments. This feature alone surpasses the simple features from (Yamaguchi et al., 2012) despite that it does not use pose information. By combining all the features we are able to improve the results greatly. We show some qualitative examples of the different feature activations in Fig. 6.8. We also evaluate the model in a leave-one-out fashion. That is, for each unary we evaluate the rest of the unaries in the model without it. Results are shown in Table 6.7.

| Method | Jaccard index |
|---|---|
| Full Model | 12.28 |
| No similarity ($\phi^{simil}_{m,n}(y_m, y_n)$) | 11.64 |
| No limb segments ($\phi^{comp}_{i,p}(y_i, l_p)$) | 12.24 |
| No simple features ($\phi^{simple}_{i,j}(y_i)$) | 10.07 |
| No clothelets ($\phi^{cloth}_{i,j}(y_i)$) | 11.94 |
| No object mask ($\phi^{obj}_{i,j}(y_i)$) | 10.02 |
| No eSIFT ($\phi^{o2p(eSIFT)}_{i,j}(y_i)$) | 10.70 |
| No eMSIFT ($\phi^{o2p(eMSIFT)}_{i,j}(y_i)$) | 12.25 |

**Table 6.7: Importance of the different potentials in the full model.** We evaluate the model after removing individually all the different potentials to evaluate their usefulness. We consider the 56 class setting.

**Qualitative Results:** We show qualitative results for both our approach and the current state-of-the-art in Fig. 6.9. We can see a visible improvement over (Yamaguchi et al., 2012), especially on person/background classification due to the strength of the proposed clothelets and person masks which in combination give strong cues on person segmentation. There are also several failure cases of our algorithm. One of the main failure cases is a breakdown of the superpixels caused by clothing texture. An excess of texture leads to an oversegmentation where the individual superpixels are no longer discriminative enough to individually identify (Fig. 6.9-bottom-left), while too much similarity with the background leads to very large superpixels that mix foreground and background. Additionally it can be seen that failures in pose detection can lead to missed limbs (Fig. 6.9-bottom-right).

**Computation Time:** Our full model takes several hours to train and roughly 20 minutes to evaluate on the full test set (excluding feature computation), on a single machine. On the same machine (Yamaguchi et al., 2012) takes more than twice the time for inference. Additionally, (Yamaguchi et al., 2012) uses grid-search for training, which does not scale to a large amount of weights, that as we have shown, are able to provide an increase in performance. Even with only 2 weights, (Yamaguchi et al., 2012) is several times slower to train than our model. Furthermore, (Yamaguchi et al., 2013) reports a training time of several days in a distributed environment.

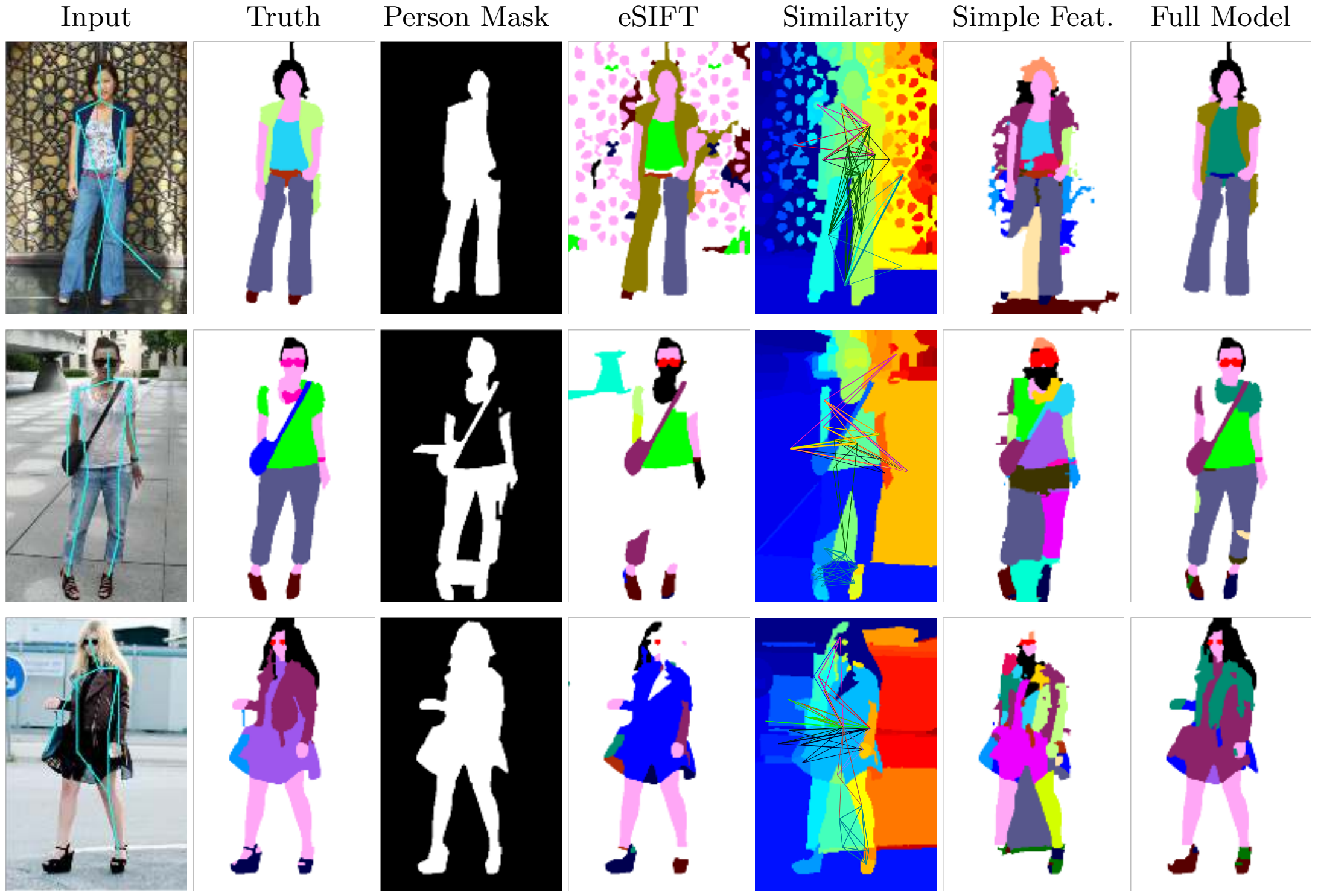

**Figure 6.8: Visualization of different features.** We show various feature activations for example images. For similarity we display the connections between the superpixels. While both the simple features from (Yamaguchi et al., 2012) and eSIFT provide decent segmentation results, they have poorly defined boundaries. These are corrected via person masks (Carreira et al., 2012) and clothelets. Further corrections are obtained by pairwise potentials such as symmetry and similarity. These results highlight the importance of combining complementary features. For class colours we refer to Fig. 6.9.

## 6.3 Modelling Fashionability

In this section we focus on predicting how fashionable a person looks on a particular photograph. Our aim here is to give a rich feedback to the user: not only whether the photograph is appealing or not, but also to make suggestions of what clothing or even the scenery the user could change in order to improve her/his look, as illustrated in Fig. 6.10. We parametrize the problem with a Conditional Random Field that jointly reasons about several important fashionability factors: the type of outfit and garments, the type of user, the setting/scenery of the photograph, and fashionability of the user's photograph. Our model exploits several domain-inspired features, such as beauty, age and mood inferred from the image, the scene type of the photograph, and if available, meta-data in the form of where the user is from, how many online followers she/he has, the sentiment of comments by other users, etc.

Since no dataset with such data exists, we created our own from online resources. We collected 144,169 posts from the fashion website chictopia.com to create our *Fashion144k* dataset. In a post, a user publishes a photograph of her/himself wearing a new outfit, typically with a visually appealing scenery behind the user. Each post also contains text in the form of descriptions and garment tags, as well as other users' comments. It also contains votes or "likes" which we use as a proxy for fashionability. We refer the reader to Fig. 6.11 for an illustration of a post.

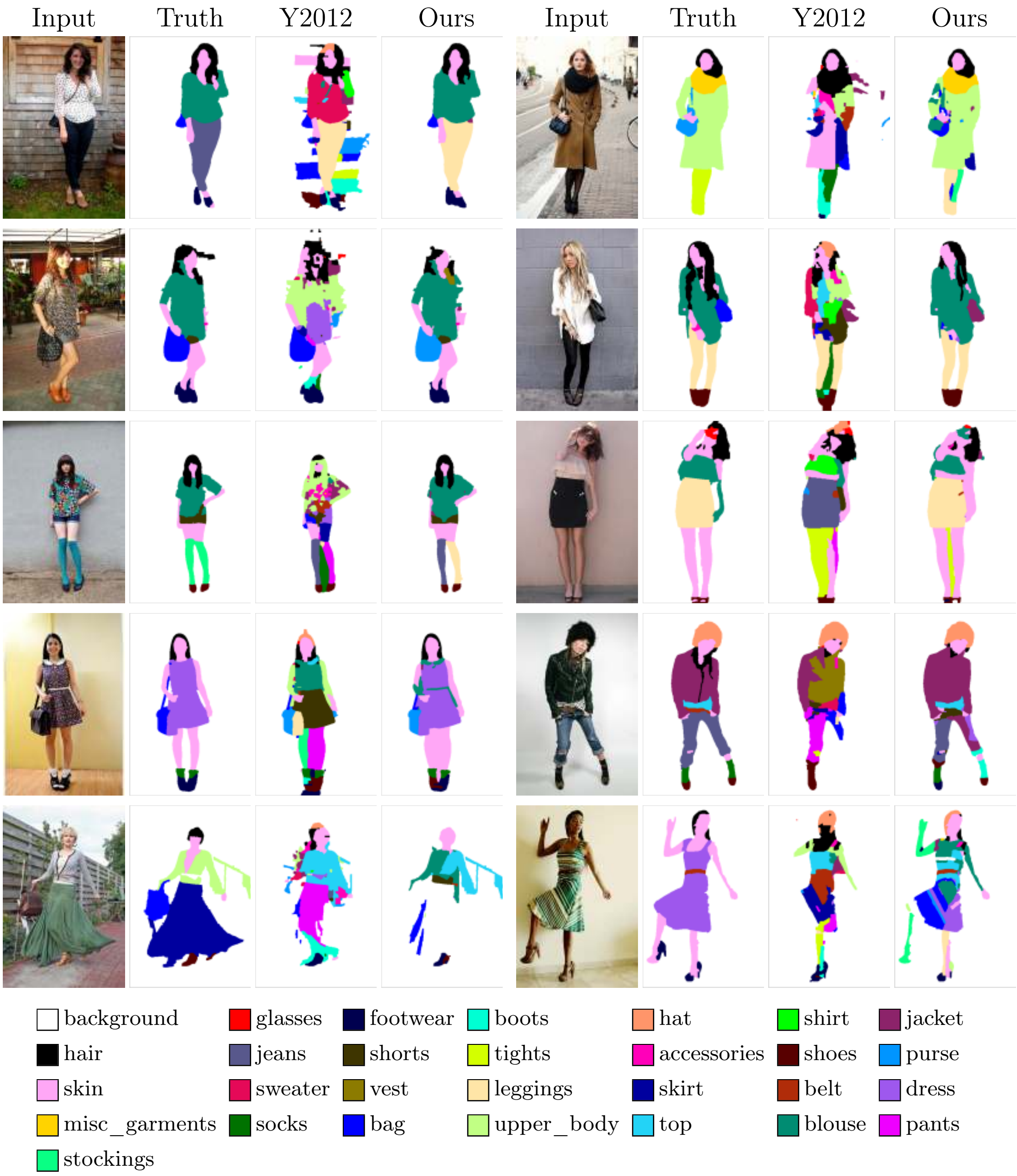

**Figure 6.9: Qualitative clothing segmentation results.** Results on Fashionista v0.2 with 29 classes, comparing our approach with the state-of-the-art Y2012 (Yamaguchi et al., 2012). In the top four rows we show good results obtained by our model. In the bottom row we show failure cases that are in general caused by 2D pose estimation failure, superpixel failures or chain failures of too many potentials.

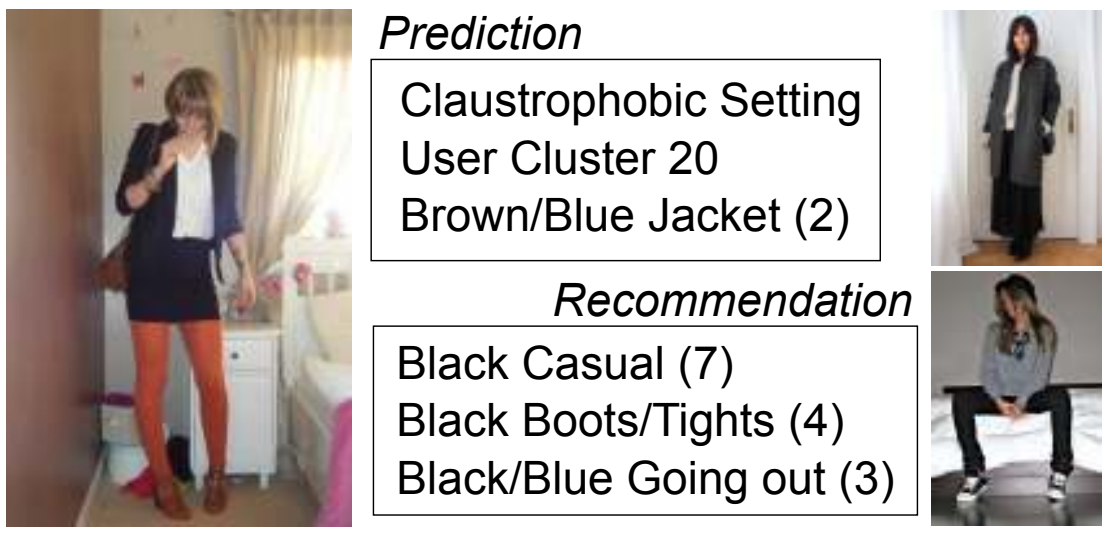

**Figure 6.10: Example of an outfit recommendation.** We show how our model is able to provide outfit recommendations for hte post on the left. In this case the user is wearing what we have identified as "Brown/Blue Jacket". This photograph obtains a score of 2 out of 10 in fashionability. Additionally the user is classified as belonging to cluster 20 and took a picture in the "Claustrophobic" setting. If the user were to wear a "Black Casual" outfit as seen on the right, our model predicts she would improve her fashionability to 7 out of 10. This prediction is conditioned on the user, setting and other factors allowing the recommendations to be tailored to each particular user.

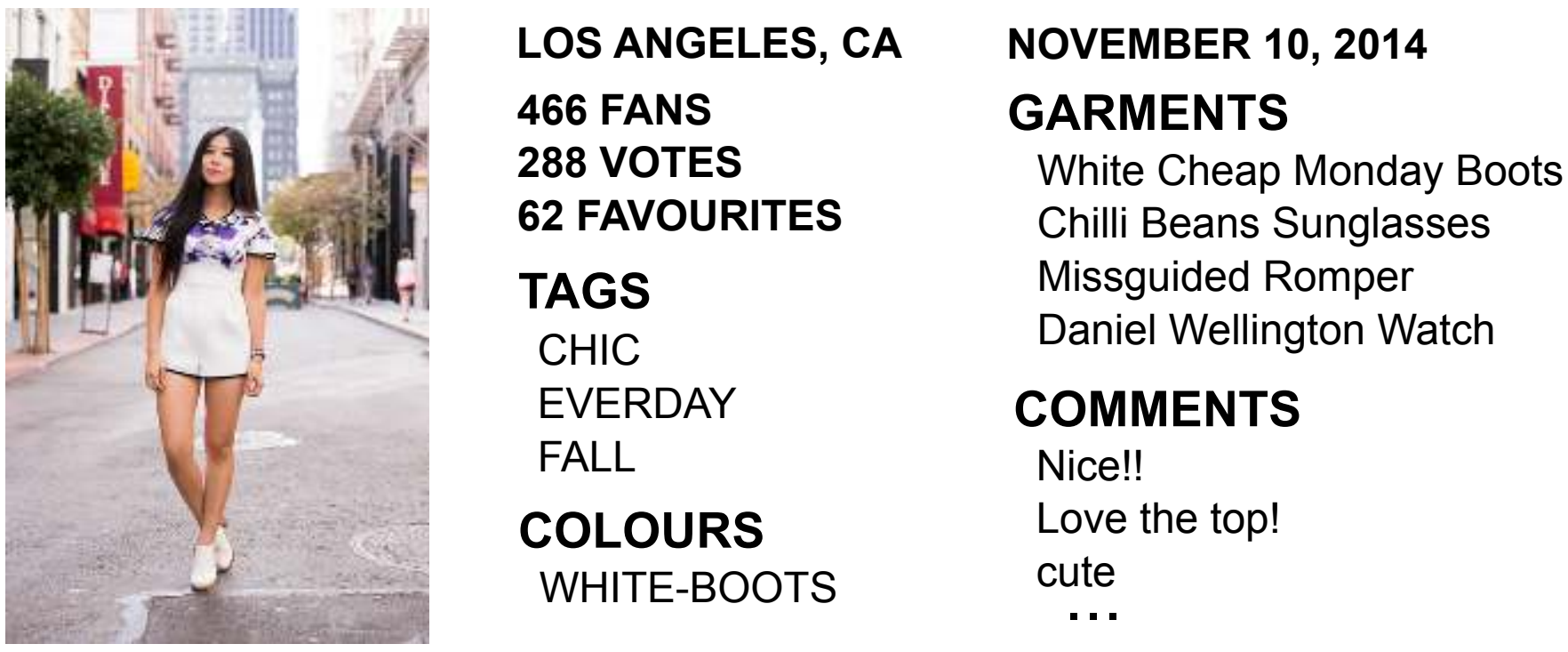

**Figure 6.11: Anatomy of a post from the Fashion144k dataset.** It consists always of at least a single image with additional metadata that can take the form of tags, list of nouns and adjectives, discrete values or arbitrary text.

As another contribution, we provide a detailed analysis of the data, in terms of fashionability scores across the world and the types of outfits people in different parts of the world wear. We also analyze outfit trends through the last six years of posts spanned by our dataset. Such analysis is important for the users, as they can adapt to the trends in "real-time" as well as to the fashion industry which can adapt their new designs based on the popularity of garments types in different social and age groups.

### Related Work

Fashion has a high impact on our everyday lives. This also shows in the growing interest in clothing-related applications in the vision community. Early work focused on manually building composite clothing models to match to images (Chen et al., 2006). In (Jammalamadaka et al., 2013; Simo-Serra et al., 2014a; Yamaguchi et al., 2013, 2012;

Yang et al., 2014), the main focus was on clothing parsing in terms of a diverse set of garment types. Most of these works follow frameworks for generic segmentation with additional pose-informed potentials. They showed that clothing segmentation is a very challenging problem with the state-of-the-art capping at 12% accuracy (Simo-Serra et al., 2014a).

More related to our line of work are recent applications such as learning semantic clothing attributes (Chen et al., 2012), identifying people based on their outfits, predicting occupation (Song et al., 2011) and urban tribes (Murillo et al., 2012), outfit recommendations (Liu et al., 2012a), and predicting outfit styles (Kiapour et al., 2014). Most of these approaches address very specific problems with fully annotated data. In contrast, the model we propose is more general, allowing to reason about several properties of one's photo: the aesthetics of clothing, the scenery, the type of clothing the person is wearing, and the overall fashionability of the photograph. We do not require any annotated data, as all necessary information is extracted by automatically mining a social website.

Our work is also related to the recent approaches that aim at modeling the human perception of beauty. In (Dhar et al., 2011; Gygli et al., 2013; Isola et al., 2014; Khosla et al., 2014) the authors addressed the question of what makes an image memorable, interesting or popular. This line of work mines large image datasets in order to correlate visual cues to popularity scores (defined as e.g., the number of times a Flickr image is viewed), or "interestingness" scores acquired from physiological studies. In our work, we tackle the problem of predicting fashionability. We also go a step further from previous work by also identifying the high-level semantic properties that cause a particular aesthetics score, which can then be communicated back to the user to improve her/his look. The closest to our work is (Khosla et al., 2013) which is able to infer whether a face is memorable or not, and modify it such that it becomes. The approach is however very different from ours, both in the domain and in formulation.

### Fashion144k Dataset

We collected a novel dataset that consists of 144,169 user posts from a clothing-oriented social website chictopia.com. In a post, a user publishes one to six photographs of her/himself wearing a new outfit. Generally each photograph shows a different angle of the user or zooms in on different garments. Users sometimes also add a description of the outfit, and/or tags of the types and colours of the garments they are wearing. Not all users make this information available, and even if they do, the tags are usually not complete, i.e. not all garments are tagged. Users typically also reveal their geographic location, which, according to our analysis, is an important factor on how fashionability is being perceived by the visitors of the post. Other users can then view these posts, leave comments and suggestions, give a "like" vote, tag the post as a "favorite", or become a "follower" of the user. There are no "dislike" votes making the data challenging to work with from the learning perspective. An example of a post can be seen in Fig. 6.11.

We parsed all the useful information of each post to create Fashion144k. The oldest entry in our dataset dates to March 2nd in 2008, the first post to the chictopia website. The last crawled post is May 22nd 2014. We refer the reader to Table 6.8 for detailed statistics of the dataset. We can see the large diversity in meta-data. Perhaps expected, the website is dominated by female users (only 5% are male).

| Property | Total | Per User | Per Post |
|---|---|---|---|
| posts | 144169 | 10.09 (30.48) | - |
| users | 14287 | - | - |
| locations | 3443 | - | - |
| males | 5% | - | - |
| fans | - | 116.80 (1309.29) | 1226.60 (3769.97) |
| comments | - | 14.15 (15.43) | 20.09 (27.51) |
| votes | - | 106.08 (108.34) | 150.76 (129.78) |
| favourites | - | 18.49 (22.04) | 27.01 (27.81) |
| photos | 277537 | 1.73 (1.00) | 1.93 (1.24) |
| tags | 13192 | 3.43 (0.75) | 3.66 (1.12) |
| colours | 3337 | 2.06 (1.82) | 2.28 (2.06) |
| garments | - | 3.14 (1.57) | 3.22 (1.72) |

**Table 6.8: Statistics of the Fashion144k dataset.** We compute various statistics of our dataset. We show global statistics, statistics for the different users, and averages for all posts.

Measuring Fashionability of a Post. Whether a person on a photograph is truly fashionable is probably best decided by fashion experts. It is also to some extent a matter of personal taste, and probably even depends on the nationality and the gender of the viewer. fashionability. In particular, we base our measure of interest on each post's number of votes, analogous to "likes" on other websites. The main issue with votes is the strong correlation with the time when the post was published. Since the number of users fluctuate, so does the number of votes. Furthermore, in the first months or a year since the website was created, the number of users (voters) was significantly lower than in the recent years.

As the number of votes follows a power-law distribution, we use the logarithm for a more robust measure. We additionally try to eliminate the temporal dependency by calculating histograms of the votes for each month, and fit a Gaussian distribution to it. We then bin the distribution such that the expected number of posts for each bin is the same. By doing this we are able to eliminate almost all time dependency and obtain a quasi-equal distribution of classes, which we use as our fashionability measure, ranging from 1 (not fashionable) to 10 (very fashionable). We can see an overview of how we obtain the fashionability metric in Fig. 6.12.

The dataset is very rich in information. In order to gain more insights we perform some data mining. Figure 6.13 shows the number of posts and fashionability scores mapped to the globe via the user's geographic information. Table 6.9 reveals some of the most trendy cities in the world, according to chictopia users and our measure.

| City Name | Posts | Fashionability |
|---:|---|---|
| Manila | 4269 | 6.627 |
| Los Angeles | 8275 | 6.265 |
| Melbourne | 1092 | 6.176 |
| Montreal | 1129 | 6.144 |
| Paris | 2118 | 6.070 |
| Amsterdam | 1111 | 6.059 |
| Barcelona | 1292 | 5.845 |
| Toronto | 1471 | 5.765 |
| Bucharest | 1385 | 5.667 |
| New York | 4984 | 5.514 |
| London | 3655 | 5.444 |
| San Francisco | 2880 | 5.392 |
| Madrid | 1747 | 5.371 |
| Vancouver | 1468 | 5.266 |
| Jakarta | 1156 | 4.398 |

**Table 6.9: Mean fashionability of cities with at least 1000 posts.** We compute the mean fashionability of all the posts from the different cities in the dataset. We then show the most fashionable cities with at least 1000 posts.

| Feature | Dim. | Description |
|---:|---:|:---:|
| Fans | 1 | Number of user's fans. |
| $\Delta$T | 1 | Time between post creation and download. |
| Comments | 5 | Sentiment analysis (Socher et al., 2013) of comments. |
| Location | 266 | Distance from location clusters (Simo-Serra et al., 2014b). |
| Personal | 21 | Face recognition attributes. |
| Style | 20 | Style of the photography (Karayev et al., 2014). |
| Scene | 397 | Output of scene classifier trained on (Xiao et al., 2010). |
| Tags | 209 | Bag-of-words with post tags. |
| Colours | 604 | Bag-of-words with colour tags. |
| Singles | 121 | Bag-of-words with split colour tags. |
| Garments | 1352 | Bag-of-words with garment tags. |

**Table 6.10: Overview of the different features used.**

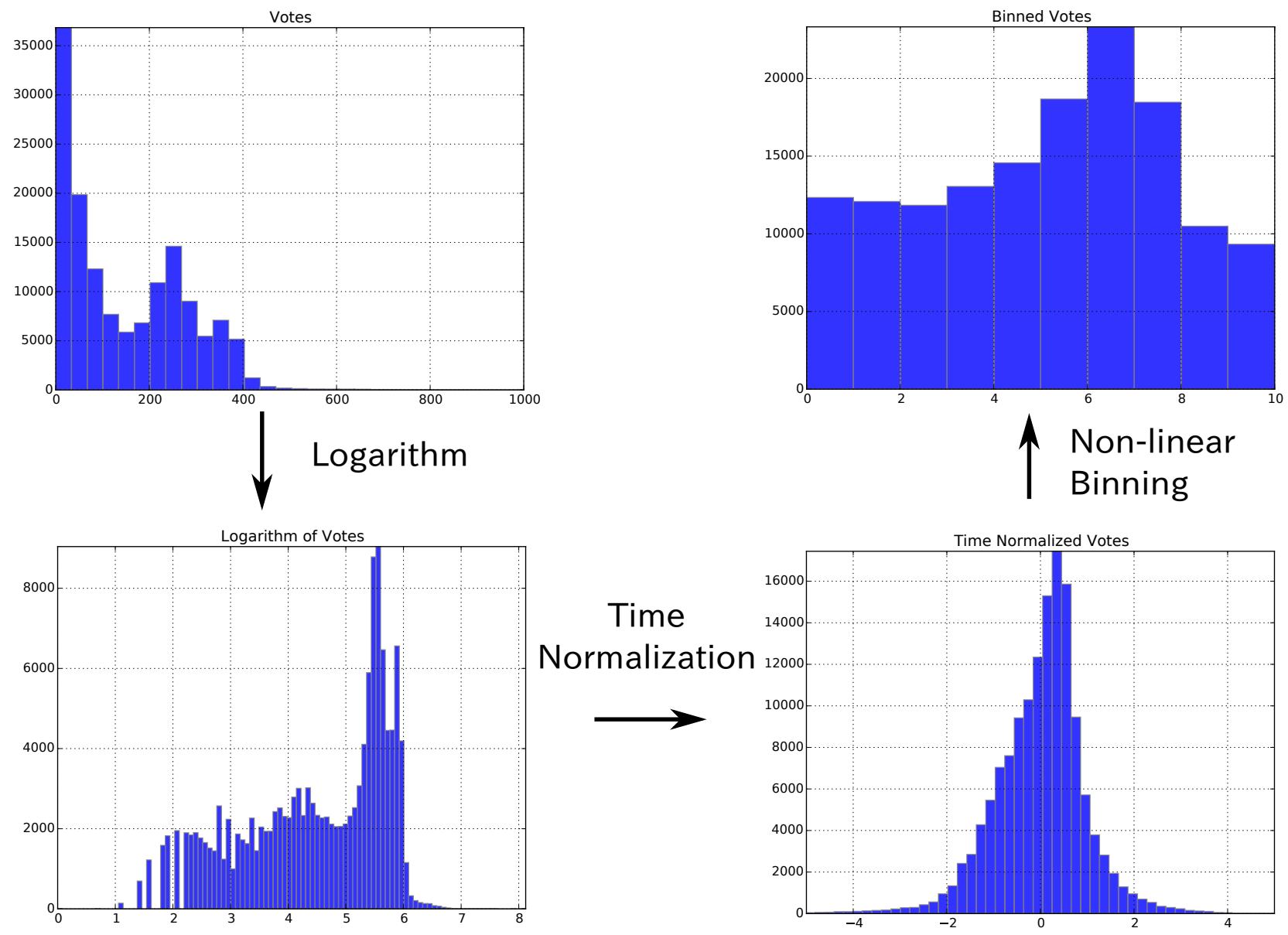

**Figure 6.12: Normalization of votes.** We objectively model fashionability using the votes. In order to remove the temporal component of the votes we perform a non-linear binning on the logarithm of the votes for a specific month. We note that it is not a flat distribution as we might have hoped. This is because we make an assumption that the votes have a Gaussian distribution, which is not entirely true.

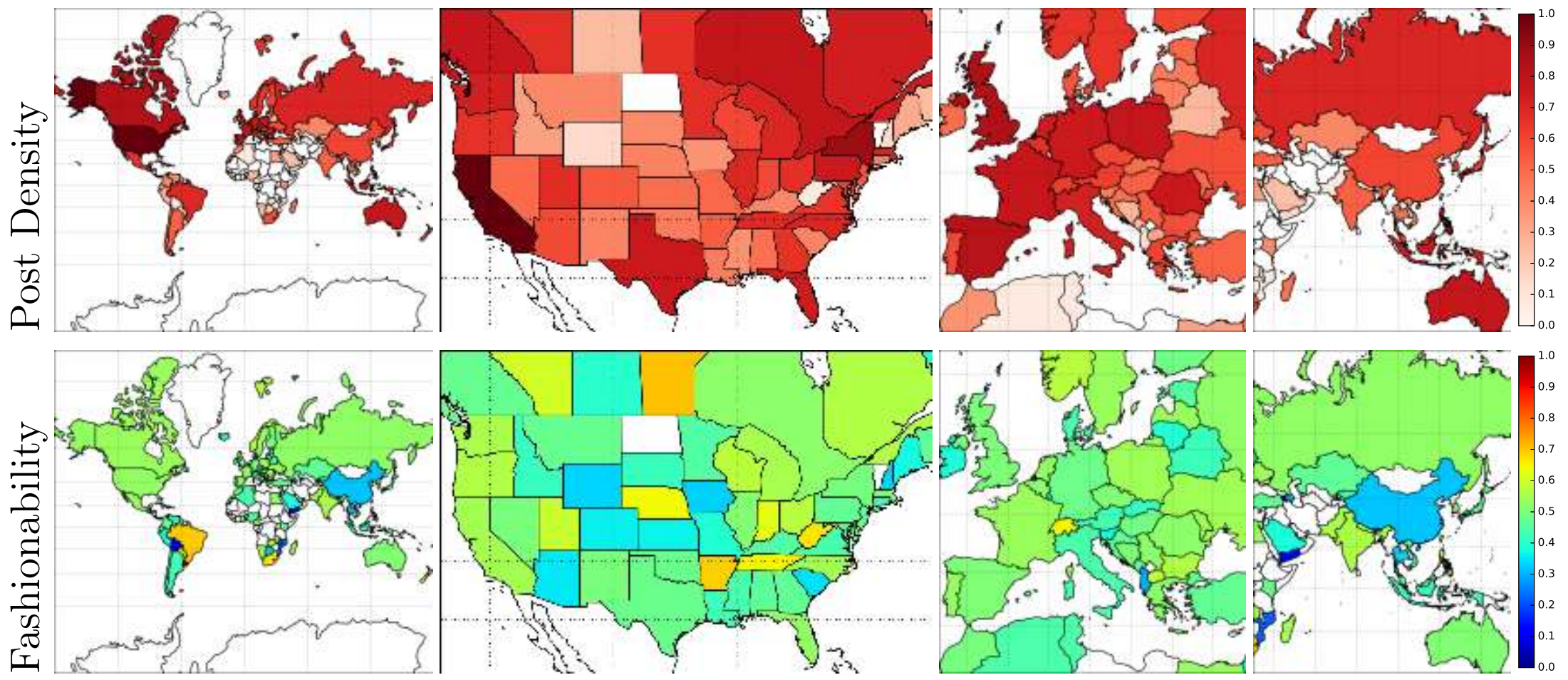

**Figure 6.13: Visualization of the density of posts and fashionability by country.** In the top row we show the logarithm of the number of posts per country as a representation of post density. In the bottom row we show the mean fashionability of the posts for all the countries.

## Discovering Fashion from Weak Data

Our objective is not only to be able to predict fashionability of a given post, but we want to create a model that can understand fashion at a higher level. For this purpose we make use of a Conditional Random Field (CRF) to learn the different outfits, types of people and settings. Settings can be interpreted as where the post is located, both at a scenic and geographic level. Our potentials make use of deep networks over a wide variety of features exploiting Fashion144k images and meta-data to produce accurate predictions of how fashionable a post is.

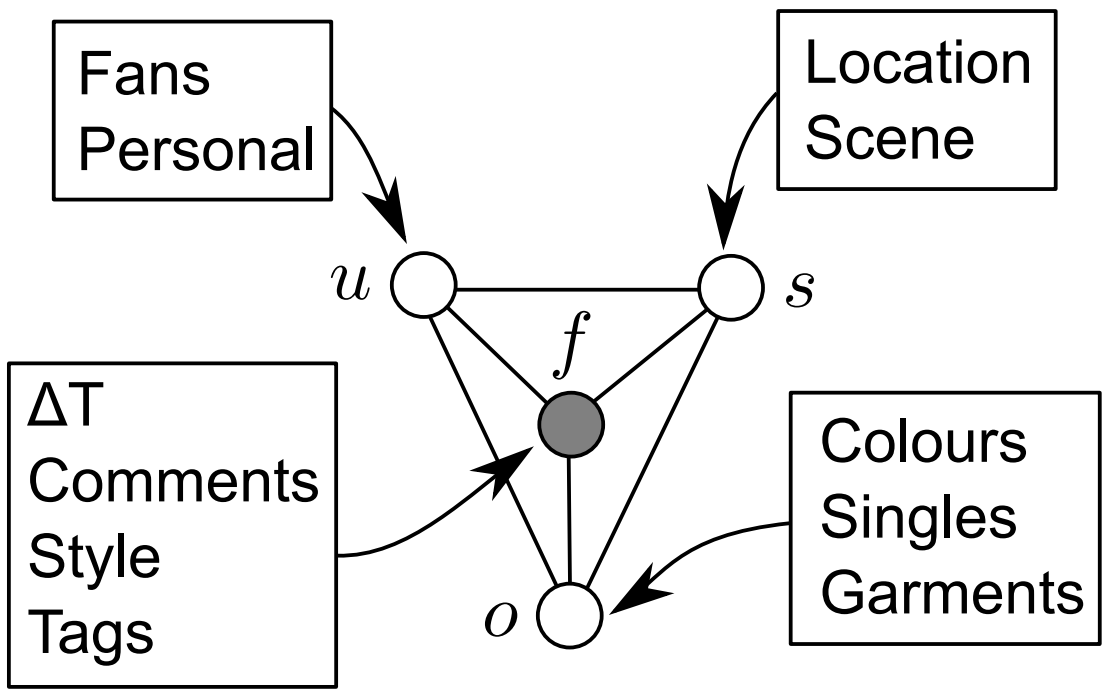

**Figure 6.14: An overview of the CRF model and the features used by each of the nodes.**

More formally, let $u \in \{1, \cdots, N_U\}$ be a random variable capturing the type of user, $o \in \{1, \cdots, N_O\}$ the type of outfit, and $s \in \{1, \cdots, N_S\}$ the setting. Further, we denote $f \in \{1, \cdots, 10\}$ as the fashionability of a post $\mathbf{x}$. We represent the energy of the CRF as a sum of energies encoding unaries for each variable as well as non-parametric pairwise potentials which reflect the correlations between the different random variables. We thus define

$$\begin{aligned} E(u, o, s, f) = & E_{user}(u) + E_{out}(o) + E_{set}(s) + E_{fash}(f) \\ & + E_{np}^{uf}(u, f) + E_{np}^{of}(o, f) + E_{np}^{sf}(s, f) \\ & + E_{np}^{uo}(u, o) + E_{np}^{so}(s, o) + E_{np}^{us}(u, s) \end{aligned} \tag{6.16}$$

We refer the reader to Fig. 6.14 for an illustration of the graphical model. We now define the potentials in more detail.

**User:** We compute user specific features encoding the logarithm of the number of fans that the particular user has as well as the output of a pre-trained neural network-based face detector enhanced to predict additional face-related attributes. In particular, we use rekognition.com which computes attributes such as ethnicity, emotions, age, beauty, etc. We run this detector on all the images of each post and only keep the features for the image with the highest score. We then compute our unary potentials as the output of a small neural network with two hidden layers that takes as input the user's high dimensional features and produces an 8D feature map $\phi_u(x)$. We refer the reader to Fig. 6.15 for an illustration. Our user unary potentials are then defined as

$$E_{user}(u = i, \mathbf{x}) = \mathbf{w}_{u,i}^T \phi_u(\mathbf{x})$$

with $\mathbf{x}$ all the information included in the post. Note that we share the features and learn a different weight for each user latent state.

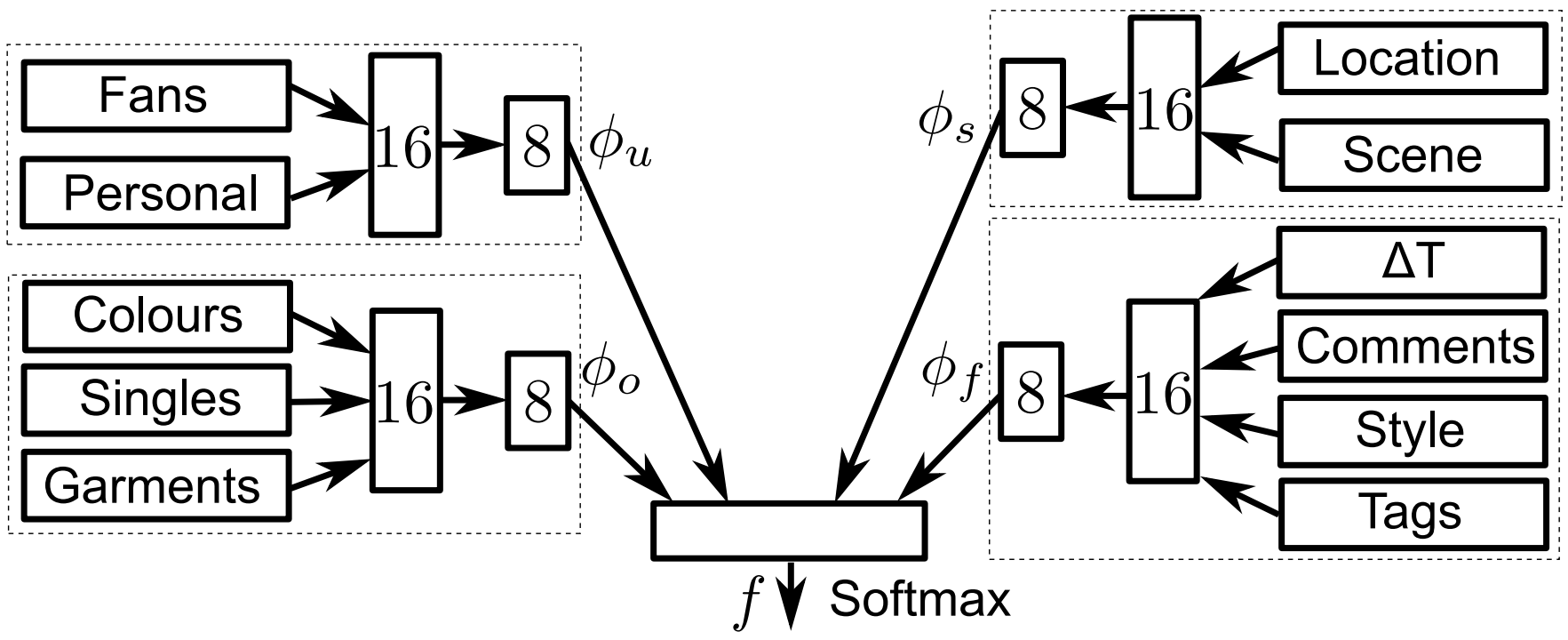

**Figure 6.15: Illustration of the type of deep network architecture used to learn features.** We can see that it consists of four network joined together by a softmax layer. The output of the different networks $\phi_f$, $\phi_o$, $\phi_u$, and $\phi_s$ are then used as features for the CRF.

**Outfit:** We use a bag-of-words approach on the “garments” and “colours” meta-data provided in each post. Our dictionary is composed of all words that appear at least 50 times in the training set. This results in 1352 and 604 words respectively and thus our representation is very sparse. Additionally we split the colour from the garment in the “colours” feature, e.g., red-dress becomes red and dress, and also perform bag-of-words on this new feature. We then compute our unary potentials as the output of a small neural network with two hidden layers that takes as input the outfit high dimensional features and produces an 8D feature map $\phi_o(\mathbf{x})$. We refer the reader to Fig. 6.15 for an illustration. Our outfit unary potentials are then defined as

$$E_{out}(o = i, \mathbf{x}) = \mathbf{w}_{o,i}^T \phi_o(\mathbf{x})$$

with $\mathbf{x}$ all the information included in the post. Note that as with the users we share the features and learn a different weight for each outfit latent state.

**Setting:** We try to capture the setting of each post by using both a pre-trained scene classifier and the user-provided location. For the scene classifier we have trained a multi-layer perceptron with a single 1024 unit hidden layer and softmax layer on the SUN Dataset (Xiao et al., 2010). We randomly use 70% of the 130,519 images as the training set, 10% as the validation set and 20% as the test set. We use the Caffe pre-trained network (Jia, 2013) to obtain features for each image which we then use to learn to identify each of the 397 classes in the dataset, corresponding to scenes such as “art_studio”, “vineyard” or “ski_slope”. The output of the 397D softmax layer is used as a feature along with the location. As the location is written in plain text, we first look up the latitude and longitude. We project all these values on the unit sphere and add some small Gaussian noise to account for the fact that many users will write more generic locations such as “Los Angeles" instead of the real address. We then perform unsupervised clustering using geodesic distances (Simo-Serra et al., 2014b) and use the geodesic distance from each cluster center as a feature. We finally compute our unary potentials as the output of a small neural network with two hidden layers that takes as input the settings high dimensional features and produce an 8D feature map $\phi_s(\mathbf{x})$.

Our outfit unary potentials are then defined as

$$E_{set}(s = i, \mathbf{x}) = \mathbf{w}_{s,i}^T \phi_s(\mathbf{x})$$

with $\mathbf{x}$ all the information included in the post. Note that as with the users and outfits we share the features and learn a different weight for each settings latent state.

**Fashion:** We use the time between the creation of the post and when the post was crawled as a feature, as well as bag-of-words on the "tags". To incorporate the reviews, we parse the comments with the sentiment-analysis model of (Socher et al., 2013). This model attempts to predict how positive a review is on a 1-5 scale (1 is extremely negative, 5 is extremely positive). We used a pre-trained model that was trained on the rotten tomatoes dataset. We run the model on all the comments and sum the scores for each post. We also extract features using the style classifier proposed in (Karayev et al., 2014) that is pre-trained on the Flickr80k dataset to detect 20 different image styles such as "Noir", "Sunny", "Macro" or "Minimal". This captures the fact that a good photography style is correlated with the fashionability score. We then compute our unary potentials as the output of a small neural network with two hidden layers that takes as input the settings high dimensional features and produce an 8D feature map $\phi_f(\mathbf{x})$. Our outfit unary potentials are then defined as

$$E_{fash}(f = i, \mathbf{x}) = \mathbf{w}_{f,i}^T \phi_f(\mathbf{x})$$

Once more, we shared the features and learn separate weights for each fashionability score.

**Correlations:** We use a non-parametric function for each pairwise and let the CRF learn the correlations. Thus

$$E_{np}^{uf}(u = i, f = j) = w_{i,j}^{uf}$$

Similarly for the other pairwise potentials.

### Learning and Inference

We learn our model using a two step approach: we first jointly train the deep networks that are used for feature extraction to predict fashionability as shown in Fig 6.15, and estimate the initial latent states using clustering. We then learn the CRF model (2430 weights) using the primal-dual method of (Hazan and Urtasun, 2010). In particular, we use the implementation of (Schwing et al., 2012). As task loss we use the $L_1$ norm for fashionability, and encourage the latent states to match the initial clustering. We perform inference using message passing (Schwing et al., 2011).

## Experimental Evaluation

We perform a detailed quantitative evaluation on the 10-class fashionability prediction task. Furthermore, we provide a qualitative evaluation on other high level tasks such as visualizing changes in trends and outfit recommendations.

| Attribute | Corr. |
|---|---|
| Economy class | -0.137 |
| Income class | -0.111 |
| log(GDP) | 0.258 |
| log(Population) | 0.231 |

| Attribute | Corr. |
|---|---|
| age | -0.025 |
| beauty | 0.066 |
| eye_closed | 0.022 |
| gender | -0.037 |
| smile | -0.023 |
| asian | 0.024 |
| calm | 0.023 |
| happy | -0.024 |
| sad | 0.023 |

**Table 6.11: Effect of various attributes on the fashionability.** Economy and Income class refer to a 1-7 scale in which 1 corresponds to most developed or rich country while 7 refers to least developed or poor country. For the face recognition features we only show those with absolute values above 0.02. In all cases we show the Pearson Coefficients.

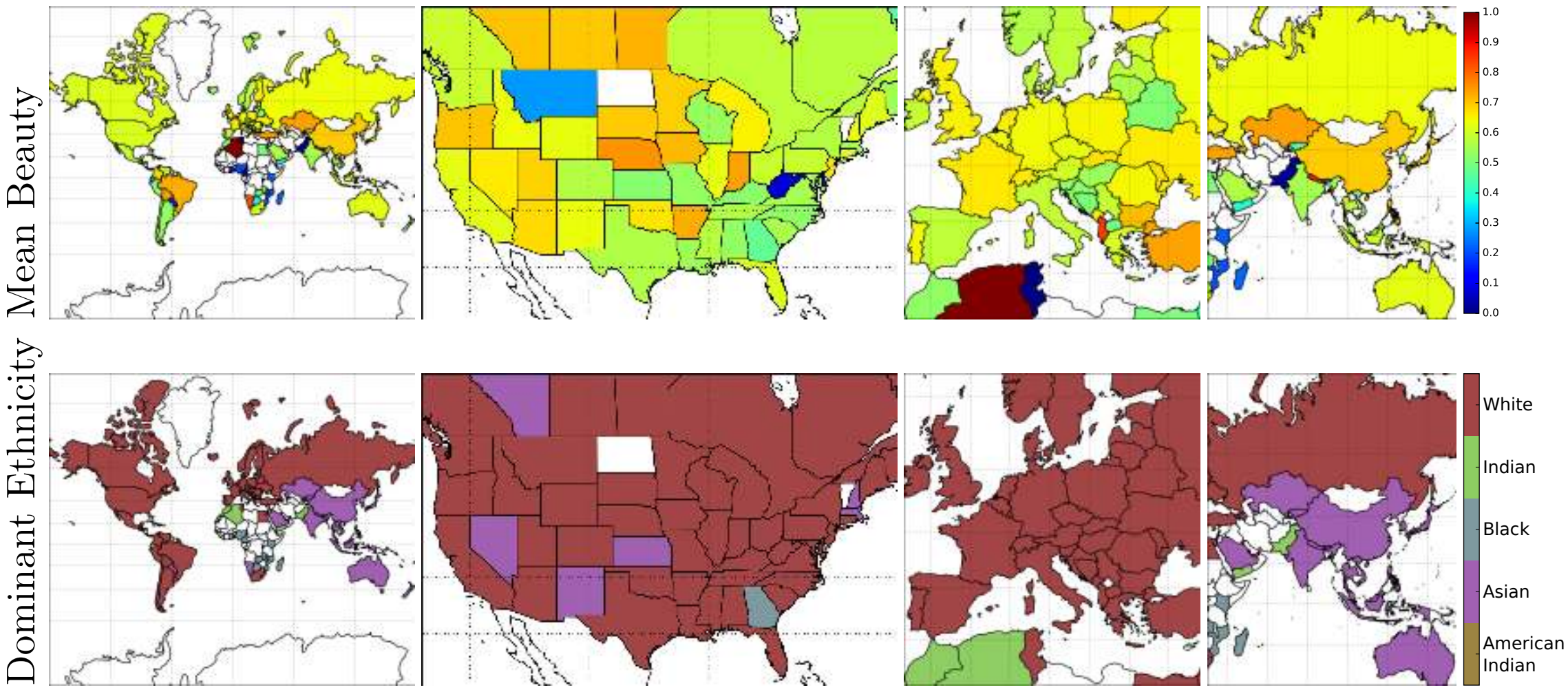

**Figure 6.16: Visualization of mean beauty and dominant ethnicity by country.** We average the beauty and ethnicity scores of all posts and show results for all the countries.

| Model | Acc. | Pre. | Rec. | IOU | $L_1$ |
|---|---|---|---|---|---|
| CRF | 29.27 | 30.42 | 28.69 | 17.36 | 1.46 |
| Deep Net | 30.42 | 31.11 | 30.26 | 18.41 | 1.45 |
| Log. Reg. | 23.92 | 22.54 | 22.99 | 12.55 | 1.91 |
| Baseline | 16.28 | - | 10.00 | 1.63 | 2.32 |
| Random | 9.69 | 9.69 | 9.69 | 4.99 | 3.17 |

**Table 6.12: Fashionability prediction results.** We show results for classification for random, a baseline that predicts only the dominant class, a standard logistic regression on our features, the deep network used to obtain features for the CRF and the final CRF model. We show accuracy, precision, recall, intersection over union (IOU), and $L_1$ norm as different metrics for performance.

### Correlations

We first analyze the correlation between fashionability and various features in our model. We consider the effect of the country on fashionability: in particular, we look the effect of economy, income class, Gross Domestic Product (GDP) and population. Results are in Table 6.11-left. A strong relationship is clear: poorer countries score lower in fashionability than the richer, sadly a not very surprising result.

We also show face-related correlations in Table 6.11-right. Interestingly, but not surprising, younger and more beautiful users are considered more fashionable. Additionally, we show the mean estimated beauty and dominant inferred ethnicity on the world map in Fig. 6.16. Brazil dominates the Americas in beauty, France dominates Spain, and Turkey dominates in Europe. In Asia, Kazakhstan scores highest, followed by China. There are also some high peaks which may be due to a very low number of posts in a country. The ethnicity classifier also seems to work pretty well, as generally the estimation matches the ethnicity of the country.

### Predicting Fashionability

We use 60% of the dataset as a train set, 10% as a validation, and 30% as test, and evaluate our model for the fashionability prediction task. Results of various model instantiations are reported in Table 6.12. While the deep net obtains slightly better results than our CRF, the model we propose is very useful as it simultaneously identifies the type of user, setting and outfit of each post. Additionally, as we show later, the CRF model allows performing much more flexible tasks such as outfit recommendation or visualization of trends. Since the classification metrics such as accuracy, precision, recall, and intersection over union (IOU) do not capture the relationship between the different fashionability levels, we also report the $L_1$ norm between the ground truth and the predicted label. In this case both the CRF and the deep net obtain virtually the same performance.

Furthermore, we show qualitative examples of true positives, false positives, true negatives and false positives in Fig. 6.17. Note that while we are only visualizing images, there is a lot of meta-data associated to each image.

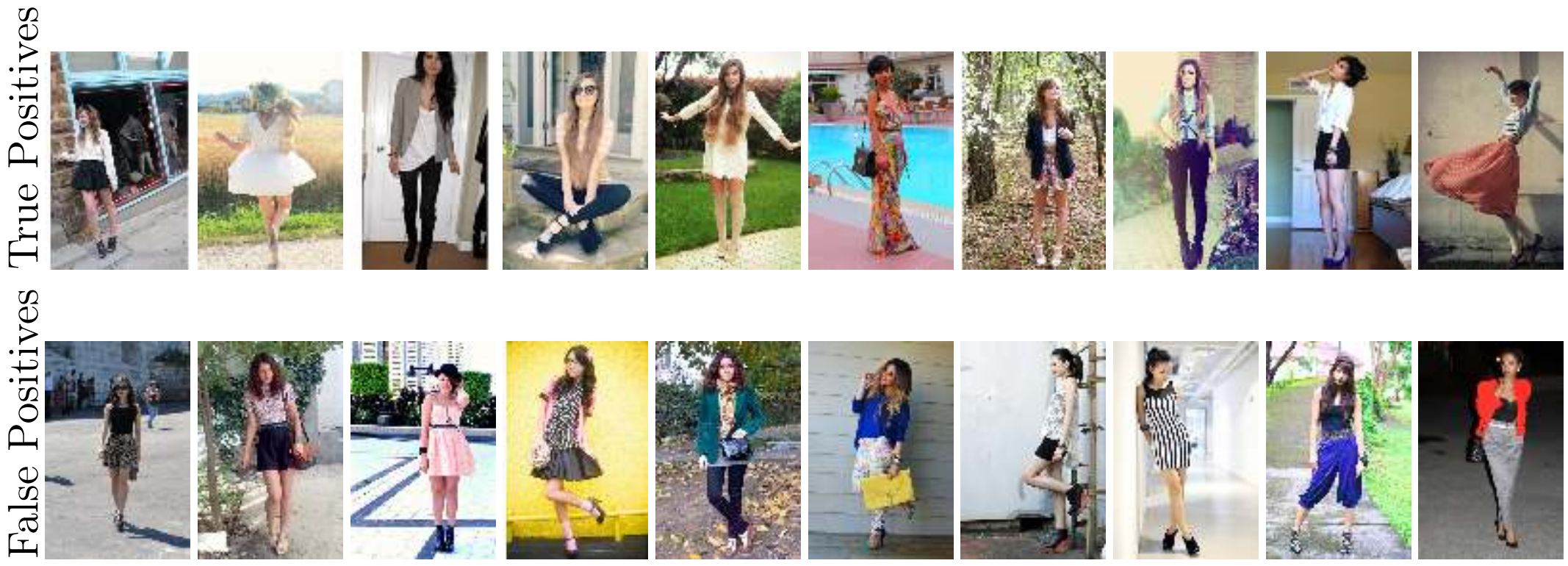

**Figure 6.17: True and false positive fashionability prediction examples.** We show some examples of the classification task obtained with our CRF model.

| Feature | Single feature | Leave one out |
|---|---|---|
| Baseline | 16.3 | 23.9 |
| Comments | 19.7 | 21.6 |
| Tags | 17.4 | 23.7 |
| $\Delta$T | 17.2 | 23.4 |
| Style | 16.3 | 23.4 |
| Location | 16.9 | 23.3 |
| Scene | 16.1 | 23.3 |
| Fans | 18.9 | 23.2 |
| Personal | 16.3 | 23.1 |
| Colours | 15.9 | 23.0 |
| Singles | 17.2 | 22.8 |
| Garments | 16.2 | 22.7 |

**Table 6.13: Evaluation of features for predicting fashionability.** We evaluate the different features for the fashionability prediction task using logistic regression. We show two cases: performance of individual features, and performance with all but one feature, which we call leave one out.

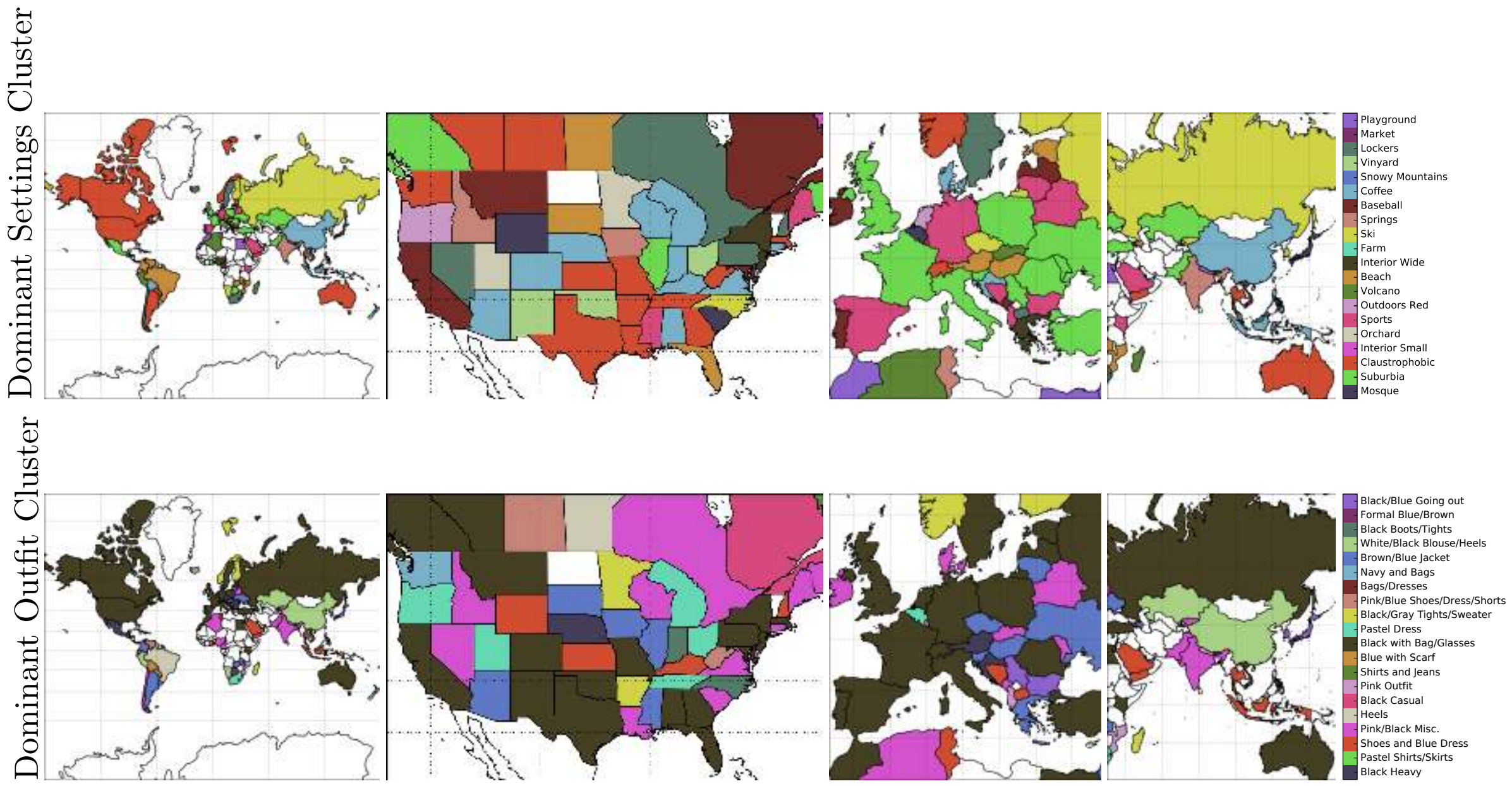

**Figure 6.18: Visualization of the dominant latent clusters.** We show results for the settings and outfit nodes in our CRF by country.

In order to analyze the individual contribution of each of the features, we show their individual prediction power as well as how much performance is lost when a feature is removed. The individual performances of the various features are shown in the second column of Table 6.13. We can see that in general the performance is very low. Several features even perform under the baseline (Personal, Scene, and Colours). The strongest features are Comments and Fans, which, however, are still not a very strong indicator of fashionability as one would expect. In the leave one out case shown in the third column, removing any feature causes a drop in performance. This means that some features are not strong individually, but carry complementary information to other features and thus still contribute to the whole. In this case we see that the most important feature is once again Comments, likely caused by the fact that most users that comment positively on a post also give it a vote.

### Identifying Latent States

In order to help interpreting the results we manually attempt to give semantic meaning to the different latent states discovered by our model. While some states are harder to assign a meaning due to the large amount of data variation, other states like, e.g., the settings states corresponding to "Ski" and "Coffee" have a clear semantic meaning. A visualization of the location of some of the latent states can be seen in Fig. 6.18.

By visualizing the pairwise weights between the fashionability node and the different nodes we can also identify the "trendiness" of different states (Fig. 6.19). For example, the settings state 1 corresponding to "Mosque" is clearly not fashionable while the state 2 and 3 corresponding to "Suburbia" and "Claustrophobic", respectively, have positive gradients indicating they are fashionable settings.

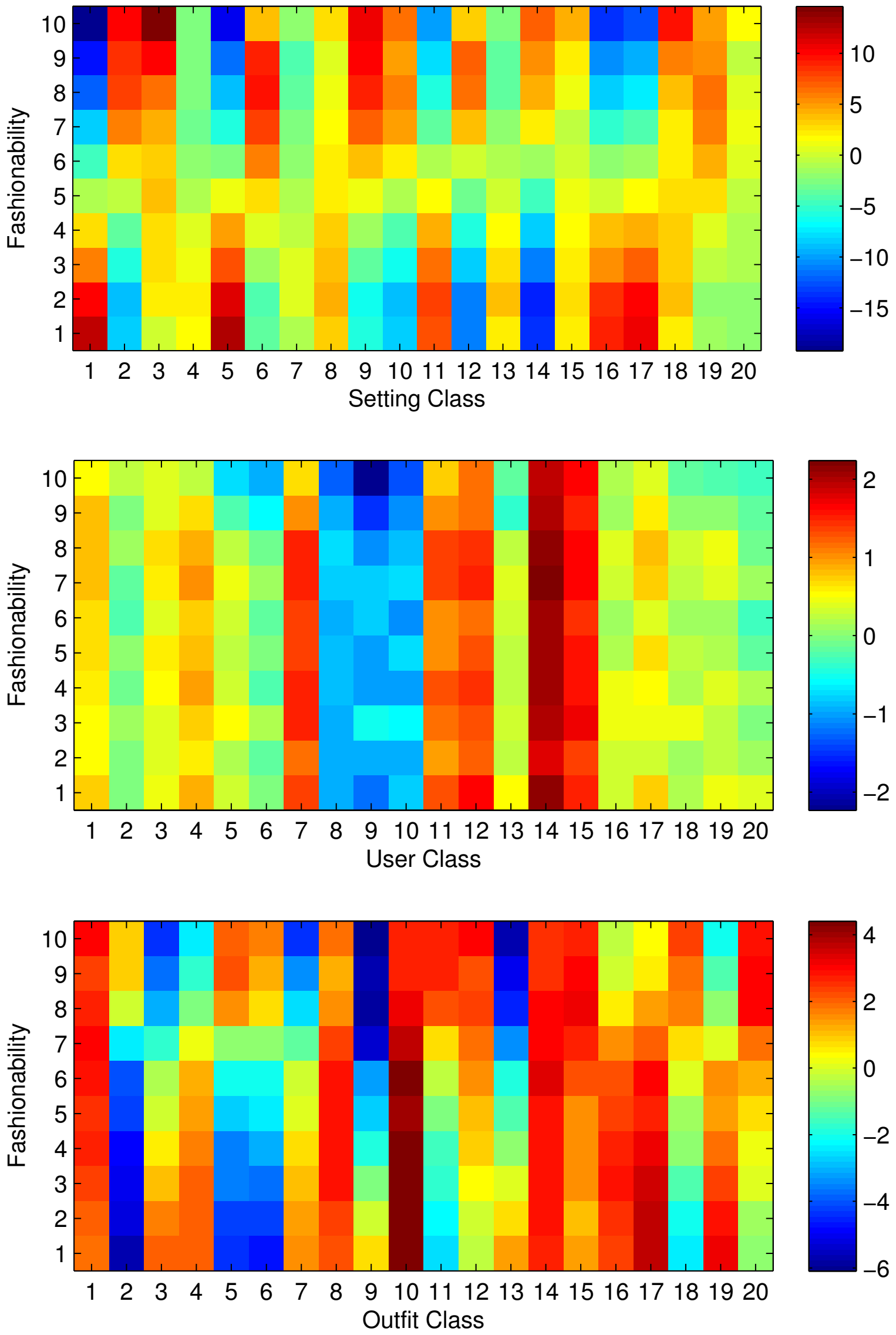

**Figure 6.19: Visualizing pairwise potentials between nodes in the CRF.** By looking at the pairwise between fashionability node and different states of other variables we are able to distinguish between fashionable and non-fashionable outfits and settings.

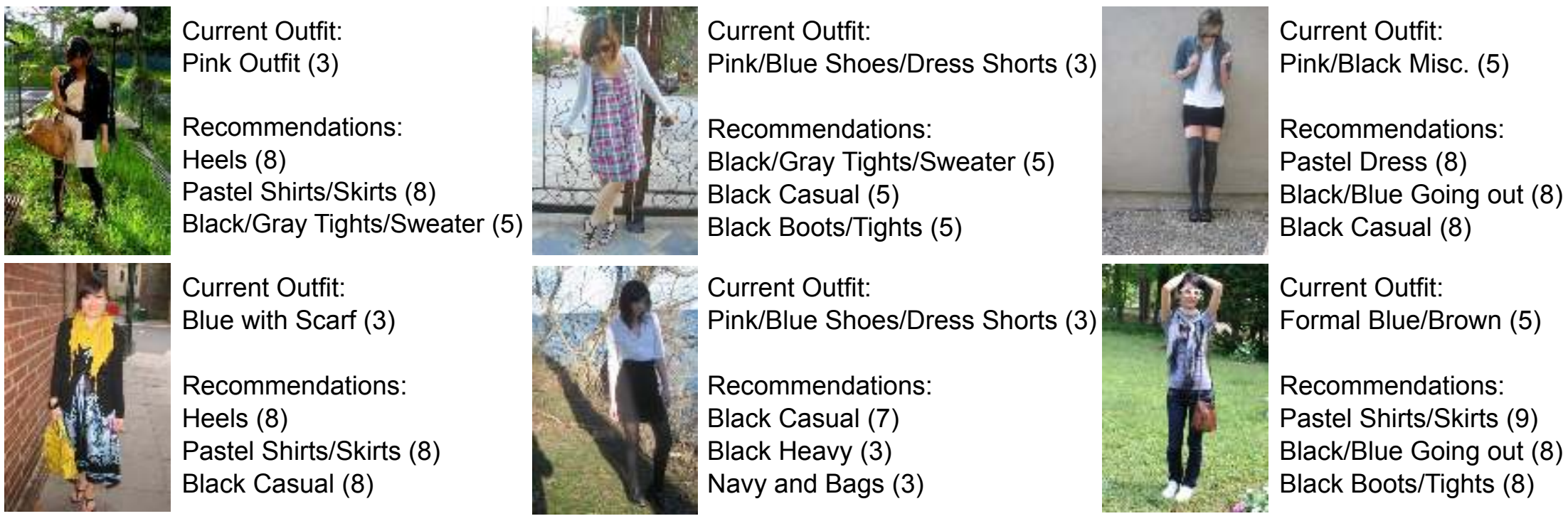

**Figure 6.20: Examples of recommendations provided for our model.** In parenthesis we see the fashionability of the user as predicted by our model.

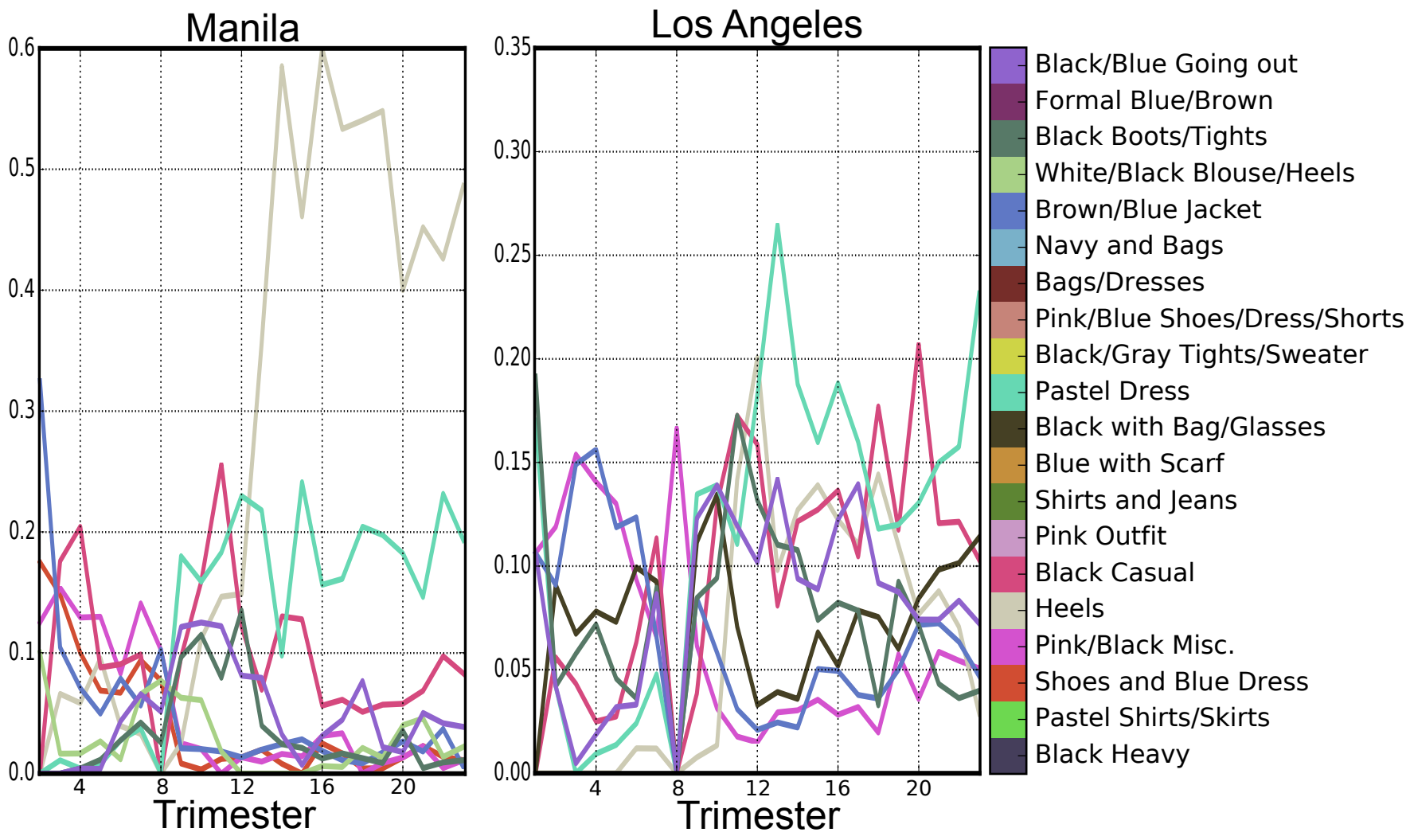

**Figure 6.21: Visualization of the temporal evolution of the different trends in Manila and Los Angeles.** The less significant clusters have been manually removed to decrease clutter.

### Outfit Recommendation

An exciting property of our model is that it can be used for outfit recommendation. In this case, we take a post as an input and estimate the outfit that maximizes the fashionability while keeping the other variables fixed. In other words, we are predicting what the user should be wearing in order to maximize her/his look instead of their current outfit. We show some examples in Fig. 6.20. This is just one example of the flexibility of our model. Other tasks such what is the least fitting outfit, what is the best place to go to with the current outfit, or what types of users this outfit fits the most, can also be done with the same model.

**Estimation Fashion Trends**

By incorporating temporal information we can try to visualize the changes in trends for a given location. In particular we look at the trendiest cites in the dataset, that is Manila and Los Angeles, as per Table 6.9. We visualize these results in Fig. 6.21. For Manila, one can see that while until the 8th trimester, outfits like “Pastel Skirts/Shirts” and “Black with Bag/Glasses” are popular, after the 12th trimester there is a boom of “Heels” and “Pastel Dress”. Los Angeles follows a roughly similar trend. For LA however, before the 8th trimester, “Brown/Blue Jacket” and “Pink/Black Misc” are popular, while afterwards “Black Casual” is also fairly popular. We’d like to note that in the 8th trimester there appears to have been an issue with the chictopia website, causing very few posts to be published, and as a consequence, results in unstable outfit predictions.

## 6.4 Summary

In this chapter we have dealt with understanding images in the challenging context of fashion. This has only recently become the focus of research and is proving to require many innovations in order to have competitive results. It is an exciting topic that can prove to be invaluable to users in an era in which one’s personal image is critical to be successful in society: from applying to universities to finding a job, and even to succeed in one’s love life.

First we have tackled the challenging problem of clothing parsing in fashion images. We have shown that our approach is able to obtain a significant improvement over the state-of-the-art in the challenging Fashionista dataset by exploiting appearance, figure/ground segmentation, shape and location priors for each garment as well as similarity between segments and symmetries between different human body parts. Despite these promising results, we believe much can still be done to improve. For example, one of the most occurring mistakes are missing glasses or other small garments. We believe a multi-resolution approach is needed to handle the diversity of garment classes.

We have also presented a novel task of predicting fashionability of users photographs. In order to evaluate this task we have collected a large-scale dataset by crawling the largest fashion-oriented social website. Our proposed Conditional Random Fields model is able to reason about settings, users and their fashionability. We are able to predict the visual aesthetics related to fashion, which can also be used to analyze fashion trends in the world or individual cities, and potentially different age groups and outfit styles. Furthermore our model can also be used for outfit recommendation. This is an important first step to building more complex and powerful models that will be able to understand fashion, trends, and users a whole in order to improve the experience of users in the modern day society. We will make the dataset and code public in hopes that this will inspire other researchers to tackle this challenging task.

It is important to note that to obtain our competitive results in these challenging tasks, many of the techniques and approaches developed in other chapter of this thesis have played a fundamental role. In order to perform high level tasks such as predicting fashionability it is essential to have a good grasp on mid and low level cues. It is no longer sufficient to focus on a very niche problem with a narrow understanding of one’s field.

# Chapter 7

# Conclusions

In this thesis we have analyzed the problem of understanding human-centric images at various levels. Our work is a push towards more ambitious goals such as estimating fashionability and proposing better outfits which we presented in Chapter 6. In order to be able to tackle these higher level tasks, we have presented a swath of techniques from low level image features to full 3D pose estimation algorithms. We finally showed the importance of using strong low and mid level cues in high level tasks.

To recapitulate the contributions:

1. We have proposed and evaluated two different approaches for local feature description. On one hand, we have used heat diffusion to robustly describe image patches undergoing deformation and illumination changes. On the other hand, we have shown that it is not only possible to learn descriptors using convolutional networks, but also extremely beneficial when done with high levels of mining denoted "fracking".
2. For 3D human pose estimation and other problems we have investigated several generative models. In particular, we presented two novel models for representing 3D human pose. One is based on a Directed Acyclic Graph with a discrete representation of the poses, and is used to learn a mapping between a latent space and the pose space. The other models the pose as data on a Riemannian manifold using a Gaussian mixture with local approximations.
3. We have also presented two frameworks for 3D human pose estimation from single monocular images. We show how it is possible to use noisy 2D observations to obtain more reliable 3D pose estimations. Additionally, we demonstrate the benefits of integrating the 2D and 3D pose estimation into a single end-to-end system that performs both tasks simultaneously.
4. Finally, we show more higher level applications in the challenging context of fashion. In particular, we show results with a Conditional Random Fields model using very strong mid level cues for the complex task of semantic segmentation of clothing. We also propose the novel high level task of predicting fashionability from images. We show it is possible to design models that are able to interpret the image and provide recommendations on how to improve one's fashionableness.

All these works share the same context, that is, they are designed for usage in human-centric images. While the lower level approaches such as descriptors and generative models do also generalize to many other problems, the 3D human pose estimation approaches and the fashion understanding developments are specific to human-centric images.

Additionally, we would like to point out we have made most of the code and datasets of this thesis publicly available[1]. We believe this is important for the reproducibility of our results and for the benefit of the community in general. We acknowledge that there are still parts of this thesis not made publicly available, however, we have the intention of releasing more in the near future.

## 7.1 Future Work

We have set the basis for many different tools that we have not yet fully exploited in this thesis. The most direct line of work is to attempt to take more advantage of our low and mid level cues in other tasks. In particular our descriptors have not been tested in full applications yet, although they have been thoroughly evaluated on real images.

For more tangible directions of work, we can refer to our deep learning approach for feature point descriptors. In particular, while we were able to obtain great improvements, there are still many open questions. For example, is it possible to learn robustness to different types of properties, such as rotation or scale, independently? What exactly is the effect of the training data on a pre-learned network? Is it possible to exploit other datasets in order to obtain better generalizations? It is also not clear whether the best performing network's hyperparameters are optimal. This requires a much more high level large-scale analysis and in particular the creation of a more complete large dataset for feature point matching.

Our generative model based on clustering on Riemannian manifolds also has shown great promise, but has not been used in a real 3D pose estimation framework. As the strength of this model holds on the fact it can quickly estimate distributions for missing limbs, if it were feasible to integrate it with an occlusion estimator, it would be possible to handle occlusions in an elegant manner. Furthermore, as we show in (Simo-Serra et al., 2015c), it is possible to use the same framework for tracking. Therefore another logical extension of the work presented in this thesis would be to tackle the problem of 3D human pose tracking. This would require extending the Bayesian formulation of our simultaneous 2D and 3D human pose estimation to integrate the temporal component.

Finally, we have only started to explore the possibilities in the world of fashion. We have presented a novel large dataset called *Fashion144k* for the prediction of fashionability, but have only started approaching this challenging problem. As shown by our preliminary results, this is a very complex problem which requires development of much stronger features and cues, such as our clothing segmentation algorithm, in order to be able to perform reliable predictions. Additionally, due to the large scale of the dataset, it is imperative for these feature extraction algorithms to be fast and scale well to the size of the dataset. For this purpose, we think it is fundamental to try to develop new approaches as well as explore the existing literature to see what cues we can exploit to build models that can truly understand fashion.

[1] http://www.iri.upc.edu/people/esimo/

## 7.2 Trends in Computer Vision and Machine Learning

In the elaboration of this thesis, we have exploited many existing tools from the fields of computer vision and machine learning. As both these fields are moving extremely fast, we would like to briefly mention some major changes we have experienced during the course of this thesis and give insights as to which direction we believe it is heading.

Probably the strongest change we have been able to experience has been the irruption of deep learning in computer vision. At the beginning of the thesis (2011), these tools were not widely available nor very relevant, however, since the publication of (Krizhevsky et al., 2012) we have seen a large influx of incredible results that have been overturning most well established techniques (Girshick et al., 2014; Simonyan and Zisserman, 2014). While the pure quantitative results of these approaches have been unbeatable in most tasks, a number of concerned voices have appeared questioning whether or not we fully understand them. This has led to the current trend of not only obtaining great results with these systems, but also understanding why we are able to obtain these great results (Zeiler and Fergus, 2014; Chatfield et al., 2014).

Competing with alternative approaches against deep learning in many problems such as classification seems like a lost cause. Not only because it seems like no other algorithms can compete with them on certain problems, but additionally the resources some of the next generation of networks require to train are out of the reach of all except the largest research laboratories. The unrelenting increase of available data and the infrastructure needed to process them are a concern for the competitiveness of small research centers. However, despite the enormous results on classifications, and more recently detection, problems, we believe there are still holes that can't be filled with pure deep learning approaches. For example structured prediction problems in which the output is subject to a set of constraints are very complicated to model with neural networks.

While results in problems with more tradition have been greatly affected by the deep learning tsunami, we have also seen an emergence of very interesting new problems in computer vision. Related to this thesis we have seen the first dataset for segmentation in fashion images in 2012 (Yamaguchi et al., 2012). In tune with this, we proposed our own dataset for predicting fashionionability and presented the novel task of giving fashion advice. On a similar line of research it is also worth to mention the work of (Pirsiavash et al., 2014), which attempts to assess different actions performed by a user, such as pool diving or figure skating, and give advice on how to improve the performance of the action. Another interesting and recent line of research seeks to find out what makes photographs of faces memorable and attempts to modify them to become even more so (Khosla et al., 2013). These are all very high level tasks that were unthinkable just a few years back. We believe we are at a point of inflection and things that were science fiction just a decade back will start to become more and more commonplace.

Another heavily questioned aspect is currently the publication model in computer vision and machine learning. There is already an antecedent of a major change in the machine learning literature when forty members resigned from the Editorial Board of the Machine Learning Journal (MLJ), and moved to support the newly created Journal of Machine Learning Research (JMLR) (Jordan, 2001). This was a move away from restricted access pay-per-view to a free-access model, which we personally believe has greatly benefited the community. Along the same lines, during the Computer Vision and Pattern Recognition (CVPR) 2013 conference in Portland, Oregon, moving to an open access model was also decided by voting (PAMI-TC, 2013). This is the unavoidable consequence of the distribution abilities of the internet and the widespread practice of researchers hosting their papers on their website to encourage diffusion.

In 2013 we additionally saw the birth of a completely new publication model. The new deep learning oriented International Conference on Learning Representations (ICLR) had its first edition. Instead of using the well established double-blind review process with a rebuttal period, they opted for single blind reviewing process, which was common in machine learning over a decade ago. This may seem like a step backwards, however, this is the consequence of moving to an open review model [2]. In this conference papers are put online during submission time and the entire reviewing process is made transparent, i.e., reviews and responses are public and shown in realtime. Additionally, editing of the paper after the initial submission is actively encouraged. This drastically changes the publication model and is once again the inevitable result of a new widespread practice: publishing preliminary work on the arXiv website [3] when submitting to a top-tier conference such as CVPR. While it is not clear if this new model is more fair than double-blind reviewing, it provides an interesting alternative for a time when it is hard to enforce the anonymousness of the submitting authors.

Finally, very recently we have seen the results of a most interesting experiment. The denoted "NIPS Experiment" (Price, 2014) consisted of forming two independent program committees and assigning a random 10% of submitted papers to both committees. The ambitious objective of this feat was to quantify the amount of randomness in acceptance of papers. Anyone who has submitted to a research conference is familiar with this randomness. A poll of submitters indicated that most agreed that the divergence would be found to be around 30%. During the conference in December 2014 the results were published and it was found that an astounding 57% of accepted papers would have been rejected by the other reviewing team. The implications of this result are not yet understood.

As we can see, not only has there been a landslide in the techniques used for computer vision with the irruption of deep learning, there have also been many changes as the current publishing model is being questioned. We believe that there are still more important changes to come in the short term, however, they seem to be going in the right direction and will be beneficial to the future of our research.

[2] http://www.iclr.cc/doku.php?id=pubmodel
[3] http://arxiv.org/

# Bibliography

A. Yao, J. Gall, L. V. G. and Urtasun, R. (2011). Learning probabilistic non-linear latent variable models for tracking complex activities. In *Neural Information Processing Systems*.

Achanta, R., Shaji, A., Smith, K., Lucchi, A., Fua, P., and Susstrunk, S. (2012). Slic superpixels compared to state-of-the-art superpixel methods. *IEEE Transactions Pattern Analylis and Machine Intelligence*, 34(11):2274–2282.

Agarwal, A. and Triggs, B. (2006). Recovering 3d human pose from monocular images. *IEEE Transactions Pattern Analylis and Machine Intelligence*, 28(1):44–58.

Andriluka, M., Roth, S., and Schiele, B. (2009). Pictorial Structures Revisited: People Detection and Articulated Pose Estimation. In *IEEE Conference on Computer Vision and Pattern Recognition*.

Andriluka, M., Roth, S., and Schiele, B. (2010). Monocular 3D Pose Estimation and Tracking by Detection. In *IEEE Conference on Computer Vision and Pattern Recognition*.

Andriluka, M., Roth, S., and Schiele, B. (2012). Discriminative Appearance Models for Pictorial Structures. *International Journal of Computer Vision*, 99(3).

Anguelov, D., Srinivasan, P., Koller, D., Thrun, S., Rodgers, J., and Davis, J. (2005). Scape: shape completion and animation of people. *ACM Transactions on Graphics*, 24(3):408–416.

Balan, A. O., Sigal, L., Black, M. J., Davis, J. E., and Haussecker, H. W. (2007). Detailed Human Shape and Pose from Images. In *IEEE Conference on Computer Vision and Pattern Recognition*.

Bay, H., Tuytelaars, T., and Gool, L. V. (2006). SURF: Speeded up robust features. In *European Conference on Computer Vision*, pages 404–417.

Belongie, S., Malik, J., and Puzicha, J. (2002). Shape matching and object recognition using shape contexts. *IEEE Transactions Pattern Analylis and Machine Intelligence*, 24(4):509–522.

Berg, A., Berg, T., and Malik, J. (2005). Shape matching and object recognition using low distortion correspondences. In *IEEE Conference on Computer Vision and Pattern Recognition*, volume 1, pages 26–33.

Bo, L. and Sminchisescu, C. (2010). Twin Gaussian Processes for Structured Prediction. *International Journal of Computer Vision*, 87:28–52.

Boothby, W. M. (2003). *An Introduction to Differentiable Manifolds and Riemannian Geometry. Revised 2nd Ed.* Academic, New York.

Bossard, L., Dantone, M., Leistner, C., Wengert, C., Quack, T., and Gool, L. V. (2012). Apparel classifcation with style. In *Asian Conference on Computer Vision*.

Bourdev, L., Maji, S., and Malik, J. (2011). Describing people: A poselet-based approach to attribute classification. In *International Conference on Computer Vision*.

Bourdev, L. and Malik, J. (2009). Poselets: Body part detectors trained using 3d human pose annotations. In *International Conference on Computer Vision*.

Boykov, Y., Veksler, O., and Zabih, R. (2001). Fast approximate energy minimization via graph cuts. *IEEE Transactions Pattern Analylis and Machine Intelligence*, 23(11):1222–1239.

Bromley, J., Guyon, I., Lecun, Y., Säckinger, E., and Shah, R. (1994). Signature verification using a "siamese" time delay neural network. In *Neural Information Processing Systems*.

Bronstein, A., Bronstein, M., Bruckstein, A., and Kimmel, R. (2007). Analysis of two-dimensional non-rigid shapes. *International Journal of Computer Vision*, 78(1):67–88.

Bronstein, M. and Kokkinos, I. (2010). Scale-invariant heat kernel signatures for non-rigid shape recognition. In *IEEE Conference on Computer Vision and Pattern Recognition*, pages 1704–1711.

Brown, M., Hua, G., and Winder, S. (2011). Discriminative learning of local image descriptors. *IEEE Transactions Pattern Analylis and Machine Intelligence*, 33(1):43–57.

Brox, T., Bourdev, L., Maji, S., and Malik, J. (2011). Object segmentation by alignment of poselet activations to image contours. In *IEEE Conference on Computer Vision and Pattern Recognition*.

Brubaker, M. A., Salzmann, M., and Urtasun, R. (2012). A Family of MCMC Methods on Implicitly Defined Manifolds. *Journal of Machine Learning Research - Proceedings Track*, 22:161–172.

Burenius, M., Sullivan, J., and Carlsson, S. (2013). 3d pictorial structures for multiple view articulated pose estimation. In *IEEE Conference on Computer Vision and Pattern Recognition*.

C. Strecha, A. Bronstein, M. B. and Fua, P. (2012). LDAHash: Improved matching with smaller descriptors. *IEEE Transactions Pattern Analylis and Machine Intelligence*, 34(1).

Cai, H., Mikolajczyk, K., and Matas, J. (2011). Learning linear discriminant projections for dimensionality reduction of image descriptors. *IEEE Transactions Pattern Analylis and Machine Intelligence*, 33(2):338–352.

Calonder, M., Lepetit, V., Ozuysa, M., Trzcinski, T., Strecha, C., and Fua, P. (2012). BRIEF: Computing a local binary descriptor very fast. *IEEE Transactions Pattern Analylis and Machine Intelligence*, 34(7):1281–1298.

Carreira, J., Caseiroa, R., Batista, J., and Sminchisescu, C. (2012). Semantic segmentation with second-order pooling. In *European Conference on Computer Vision*.

Carreira, J. and Sminchisescu, C. (2012). CPMC: Automatic Object Segmentation Using Constrained Parametric Min-Cuts. *IEEE Transactions Pattern Analylis and Machine Intelligence*, 34(7):1312–1328.

Chang, C.-C. and Lin, C.-J. (2011). LIBSVM: A library for support vector machines. *ACM Transactions on Intelligent Systems and Technology*, 2:27:1–27:27.

Chatfield, K., Simonyan, K., Vedaldi, A., and Zisserman, A. (2014). Return of the devil in the details: Delving deep into convolutional nets. *CoRR*, abs/1405.3531.

Chavel, I. (1984). *Eigenvalues in Riemannian Geometry*. London Academic Press.

Chen, H., Gallagher, A., and Girod, B. (2012). Describing clothing by semantic attributes. In *European Conference on Computer Vision*.

Chen, H., Xu, Z. J., Liu, Z. Q., and Zhu, S. C. (2006). Composite templates for cloth modeling and sketching. In *IEEE Conference on Computer Vision and Pattern Recognition*.

Cheng, H., Liu, Z., Zheng, N., and Yang, J. (2008). A deformable local image descriptor. In *IEEE Conference on Computer Vision and Pattern Recognition*.

Cho, M., Lee, J., and Lee, K. (2009). Feature correspondence and deformable object matching via agglomerative correspondence clustering. In *International Conference on Computer Vision*, pages 1280–1287.

Collobert, R., Kavukcuoglu, K., and Farabet, C. (2011). Torch7: A matlab-like environment for machine learning. In *BigLearn, NIPS Workshop*.

Culurciello, E., Jin, J., Dundar, A., and Bates, J. (2013). An analysis of the connections between layers of deep neural networks. *CoRR*, abs/1306.0152.

Dalal, N. and Triggs, B. (2005). Histograms of oriented gradients for human detection. In *IEEE Conference on Computer Vision and Pattern Recognition*.

Daubney, B. and Xie, X. (2011). Tracking 3D Human Pose with Large Root Node Uncertainty. In *IEEE Conference on Computer Vision and Pattern Recognition*.

Davis, B. C., Bullitt, E., Fletcher, P. T., and Joshi, S. (2007). Population Shape Regression from Random Design Data. In *International Conference on Computer Vision*.

Davis, J. and Goadrich, M. (2006). The relationship between PR and ROC curves. In *International Conference in Machine Learning*.

de Goes, F., Goldenstein, S., and Velho, L. (2008). A hierarchical segmentation of articulated bodies. In *Proceedings of the Symposium on Geometry Processing*, pages 1349–1356.

Dedieu, J.-P. and Nowicki, D. (2005). Symplectic methods for the approximation of the exponential map and the newton iteration on riemannian submanifolds. *Journal of Complexity*, 21(4):487 – 501.

Deng, J., Dong, W., Socher, R., jia Li, L., Li, K., and Fei-fei, L. (2009). Imagenet: A large-scale hierarchical image database. In *IEEE Conference on Computer Vision and Pattern Recognition*.

Dhar, S., Ordonez, V., and Berg, T. (2011). High level describable attributes for predicting aesthetics and interestingness. In *IEEE Conference on Computer Vision and Pattern Recognition*.

Donahue, J., Jia, Y., Vinyals, O., Hoffman, J., Zhang, N., Tzeng, E., and Darrell, T. (2013). Decaf: A deep convolutional activation feature for generic visual recognition. *CoRR*, abs/1310.1531.

Dong, J., Chen, Q., Xia, W., Huang, Z., and Yan, S. (2013). A deformable mixture parsing model with parselets. In *International Conference on Computer Vision*.

Everingham, M., Van Gool, L., Williams, C. K. I., Winn, J., and Zisserman, A. (2010). The pascal visual object classes (voc) challenge. *International Journal of Computer Vision*, 88(2):303–338.

Fan, B., Wu, F., and Hu, Z. (2012). Rotationally invariant descriptors using intensity order pooling. *IEEE Transactions Pattern Analylis and Machine Intelligence*, 34(10):2031–2045.

Fan, R.-E., Chang, K.-W., Hsieh, C.-J., Wang, X.-R., and Lin, C.-J. (2008). LIBLINEAR: A library for large linear classification. *Journal of Machine Learning Research*, 9:1871–1874.

Farhadi, A., Endres, I., Hoiem, D., and Forsyth, D. (2009). Describing objects by their attributes. In *IEEE Conference on Computer Vision and Pattern Recognition*.

Felzenszwalb, P., McAllester, D., and Ramanan, D. (2008). A Discriminatively Trained, Multiscale, Deformable Part Model. In *IEEE Conference on Computer Vision and Pattern Recognition*.

Felzenszwalb, P. F., Girshick, R. B., McAllester, D., and Ramanan, D. (2010). Object Detection with Discriminatively Trained Part Based Models. *IEEE Transactions Pattern Analylis and Machine Intelligence*, 32(9):1627–1645.

Felzenszwalb, P. F. and Huttenlocher, D. P. (2005). Pictorial Structures for Object Recognition. *International Journal of Computer Vision*, 61:55–79.

Fidler, S., Sharma, A., and Urtasun, R. (2013). A sentence is worth a thousand pixels. In *IEEE Conference on Computer Vision and Pattern Recognition*.

Figueiredo, M. and Jain, A. (2002). Unsupervised Learning of Finite Mixture Models. *IEEE Transactions Pattern Analylis and Machine Intelligence*, 24(3):381–396.

Fletcher, P., Lu, C., Pizer, S., and Joshi, S. (2004). Principal Geodesic Analysis for the Study of Nonlinear Statistics of Shape. *IEEE Transactions on Medical Imaging*, 23(8):995–1005.

Forbes Magazine (2013). US Online Retail Sales To Reach $370B By 2017; €191B in Europe. http://www.forbes.com. [Online; accessed 14-March-2013].

Gallagher, A. C. and Chen, T. (2008). Clothing cosegmentation for recognizing people. In *IEEE Conference on Computer Vision and Pattern Recognition*.

Geiger, A., Lenz, P., Stiller, C., and Urtasun, R. (2013). Vision meets robotics: The kitti dataset. *International Journal of Robotics Research*.

Girshick, R., Donahue, J., Darrell, T., and Malik, J. (2014). Rich feature hierarchies for accurate object detection and semantic segmentation. In *IEEE Conference on Computer Vision and Pattern Recognition*.

Gębal, K., Bærentzen, J. A., Aanæs, H., and Larsen, R. (2009). Shape analysis using the auto diffusion function. In *Proceedings of the Symposium on Geometry Processing*, pages 1405–1413.

Gong, Y., Lazebnik, S., Gordo, A., and Perronnin, F. (2012). Iterative quantization: A procrustean approach to learning binary codes for large-scale image retrieval. In *IEEE Transactions Pattern Analylis and Machine Intelligence*.

Guan, P., Weiss, A., Balan, A., and M.J.Black (2009). Estimating Human Shape and Pose from a Single Image. In *International Conference on Computer Vision*.

Gupta, R. and Mittal, A. (2007). Illumination and Affine-Invariant Point Matching using an Ordinal Approach. In *International Conference on Computer Vision*.

Gupta, R. and Mittal, A. (2008). Smd: A locally stable monotonic change invariant feature descriptor. In *European Conference on Computer Vision*, pages 265–277.

Gupta, R., Patil, H., and Mittal, A. (2010). Robust order-based methods for feature description. In *IEEE Conference on Computer Vision and Pattern Recognition*.

Gygli, M., Grabner, H., Riemenschneider, H., Nater, F., and Gool, L. (2013). The interestingness of images. In *International Conference on Computer Vision*.

Hamerly, G. and Elkan, C. (2003). Learning the k in k-means. In *Neural Information Processing Systems*.

Hansen, N. (2006). The CMA Evolution Strategy: a Comparing Review. In *Towards a new evolutionary computation. Advances on estimation of distribution algorithms*, pages 75–102. Springer.

Hasan, B. and Hogg, D. (2010). Segmentation using deformable spatial priors with application to clothing. In *British Machine Vision Conference*.

Hasler, N., Stoll, C., Sunkel, M., Rosenhahn, B., and Seidel, H.-P. (2009). A statistical model of human pose and body shape. *Computer Graphics Forum*, 28(2):337–346.

Hauberg, S., Sommer, S., and Pedersen, K. S. (2012). Natural metrics and least-committed priors for articulated tracking. *Image and Vision Computing*, 30(6):453–461.

Hazan, T. and Urtasun, R. (2010). A primal-dual message-passing algorithm for approximated large scale structured prediction. In *Neural Information Processing Systems*.

Heikkilä, M., Pietikäinen, M., and Schmid, C. (2009). Description of interest regions with local binary patterns. *Pattern Recognition*, 42(3):425–436.

Hou, X. and Zhang, L. (2007). Saliency detection: A spectral residual approach. In *IEEE Conference on Computer Vision and Pattern Recognition*.

Howe, N. R., Leventon, M. E., and Freeman, W. T. (1999). Bayesian Reconstruction of 3D Human Motion from Single-Camera Video. In *Neural Information Processing Systems*, pages 820–826. MIT Press.

Huang, L. (2008). Advanced dynamic programming in semiring and hypergraph frameworks. In *Coling 2008: Advanced Dynamic Programming in Computational Linguistics: Theory, Algorithms and Applications - Tutorial notes*, pages 1–18, Manchester, UK. Coling 2008 Organizing Committee.

Huckemann, S., Hotz, T., and Munk, A. (2010). Intrinsic Shape analysis: Geodesic PCA for Riemannian Manifolds Modulo Isometric Lie Group Actions. *Statistica Sinica*, 20:1–100.

Ionescu, C., Li, F., and Sminchisescu, C. (2011). Latent Structured Models for Human Pose Estimation. In *International Conference on Computer Vision*.

Ionescu, C., Papava, D., Olaru, V., and Sminchisescu, C. (2014). Human3.6m: Large scale datasets and predictive methods for 3d human sensing in natural environments. *IEEE Transactions Pattern Analylis and Machine Intelligence*.

Isola, P., Xiao, J., Parikh, D., Torralba, A., and Oliva, A. (2014). What makes a photograph memorable? *IEEE Transactions Pattern Analylis and Machine Intelligence*, 36(7):1469–1482.

Itti, L. and Koch, C. (2000). A saliency-based search mechanism for overt and covert shifts of visual attention. *Vision Research*, 40:1489–1506.

Jahrer, M., Grabner, M., and Bischof, H. (2008). Learned local descriptors for recognition and matching. In *Computer Vision Winter Workshop*.

Jain, A., Tompson, J., Andriluka, M., Taylor, G. W., and Bregler, C. (2014a). Learning human pose estimation features with convolutional networks. In *International Conference on Learning Representations*.

Jain, A., Tompson, J., Lecun, Y., and Bregler, C. (2014b). Modeep: A deep learning framework using motion features for human pose estimation. In *Asian Conference on Computer Vision*.

Jammalamadaka, N., Minocha, A., Singh, D., and Jawahar, C. (2013). Parsing clothes in unrestricted images. In *British Machine Vision Conference*.

Jarrett, K., Kavukcuoglu, K., Ranzato, M., and LeCun, Y. (2009). What is the best multi-stage architecture for object recognition? In *International Conference on Computer Vision*.

Jia, Y. (2013). Caffe: An open source convolutional architecture for fast feature embedding. http://caffe.berkeleyvision.org/.

Jordan, M. (2001). Leading ML researchers issue statement of support for JMLR. http://www.jmlr.org/statement.html. [Online; accessed 3-February-2015].

Karayev, S., Hertzmann, A., Winnemoeller, H., Agarwala, A., and Darrell, T. (2014). Recognizing image style. In *British Machine Vision Conference*.

Karcher, H. (1977). Riemannian center of mass and mollifier smoothing. *Communications on Pure and Applied Mathematics*, 30(5):509–541.

Kazemi, V., Burenius, M., Azizpour, H., and Sullivan, J. (2013). Multiview body part recognition with random forests. In *British Machine Vision Conference*.

Ke, Y. and Sukthankar, R. (2004). PCA-SIFT: a more distinctive representation for local image descriptors. In *IEEE Conference on Computer Vision and Pattern Recognition*, volume 2, pages 506–513.

Khosla, A., Bainbridge, W. A., Torralba, A., and Oliva, A. (2013). Modifying the memorability of face photographs. In *International Conference on Computer Vision*.

Khosla, A., Sarma, A. D., and Hamid, R. (2014). What makes an image popular? In *International World Wide Web Conference*.

Kiapour, M. H., Yamaguchi, K., Berg, A. C., and Berg, T. L. (2014). Hipster wars: Discovering elements of fashion styles. In *European Conference on Computer Vision*.

Kokkinos, I., Bronstein, M., and Yuille, A. (2012). Dense Scale Invariant Descriptors for Images and Surfaces. Research Report RR-7914, INRIA.

Kovnatsky, A., Bronstein, M., Bronstein, A., and Kimmel, R. (2011). Photometric heat kernel signatures. In *Scale Space and Variational Methods in Computer Vision*, pages 616–627.

Krizhevsky, A., Sutskever, I., and Hinton, G. E. (2012). Imagenet classification with deep convolutional neural networks. In *Neural Information Processing Systems*.

Krähenbühl, P. and Koltun, V. (2011). Efficient inference in fully connected crfs with gaussian edge potentials. In *Neural Information Processing Systems*.

Ladicky, L., Russell, C., Kohli, P., and Torr, P. H. S. (2010). Graph cut based inference with co-occurrence statistics. In *European Conference on Computer Vision*.

Ladicky, L., Torr, P. H. S., and Zisserman, A. (2013). Human pose estimation using a joint pixel-wise and part-wise formulation. In *IEEE Conference on Computer Vision and Pattern Recognition*.

Lafferty, J. D., McCallum, A., and Pereira, F. C. N. (2001). Conditional random fields: Probabilistic models for segmenting and labeling sequence data. In *International Conference in Machine Learning*.

Lawrence, N. D. (2005). Probabilistic Non-linear Principal Component Analysis with Gaussian Process Latent Variable Models. *Journal of Machine Learning Research*, 6:1783–1816.

Lawrence, N. D. and Moore, A. J. (2007). Hierarchical Gaussian Process Latent Variable Models. In *International Conference in Machine Learning*, pages 481–488, New York, NY, USA. ACM.

Lecun, Y., Bottou, L., Bengio, Y., and Haffner, P. (1998). Gradient-based learning applied to document recognition. *Proceedings of the IEEE*, 86(11):2278–2324.

Leordeanu, M. and Hebert, M. (2005). A spectral technique for correspondence problems using pairwise constraints. In *International Conference on Computer Vision*, pages 1482–1489.

Lévy, B. (2006). Laplace-Beltrami Eigenfunctions: Towards an Algorithm that Understands Geometry. In *IEEE International Conference on Shape Modeling and Applications*.

Li, R., Tian, T.-P., Sclaroff, S., and Yang, M.-H. (2010). 3d human motion tracking with a coordinated mixture ofăfactor analyzers. *International Journal of Computer Vision*, 87(1-2):170–190.

Li, S. and Chan, A. B. (2014). 3d human pose estimation from monocular images with deep convolutional neural network. In *Asian Conference on Computer Vision*.

Lin, D., Fidler, S., and Urtasun, R. (2013). Holistic scene understanding for 3d object detection with rgbd cameras. In *International Conference on Computer Vision*.

Ling, H. and Jacobs, D. (2005). Deformation invariant image matching. In *International Conference on Computer Vision*, pages 1466–1473.

Ling, H. and Jacobs, D. (2007). Shape classification using the inner-distance. *IEEE Transactions Pattern Analylis and Machine Intelligence*, 29(2):286–299.

Ling, H., Yang, X., and Latecki, L. (2010). Balancing deformability and discriminability for shape matching. In *European Conference on Computer Vision*.

Liu, S., Feng, J., Song, Z., Zhang, T., Lu, H., Changsheng, X., and Yan, S. (2012a). Hi, magic closet, tell me what to wear! In *ACM International Conference on Multimedia*.

Liu, S., Song, Z., Liu, G., Xu, C., Lu, H., and Yan, S. (2012b). Street-toshop: Cross-scenario clothing retrieval via parts alignment and auxiliary set. In *IEEE Conference on Computer Vision and Pattern Recognition*.

Lowe, D. G. (2004). Distinctive image features from scale-invariant keypoints. *International Journal of Computer Vision*, 60(2):91–110.

Mikolajczyk, K. and Schmid, C. (2005). A performance evaluation of local descriptors. *IEEE Transactions Pattern Analylis and Machine Intelligence*, 10(27):1615–1630.

Miller, G. A. (1995). Wordnet: a lexical database for english. *Communications of the ACM*, 38(11):39–41.

Mobahi, H., Collobert, R., and Weston, J. (2009). Deep learning from temporal coherence in video. In *International Conference in Machine Learning*.

Morel, J. and Yu, G. (2009). ASIFT: A new framework for fully affine invariant image comparison. *SIAM Journal on Imaging Sciences*, 2(2):438–469.

Moreno-Noguer, F. (2011). Deformation and illumination invariant feature point descriptor. In *IEEE Conference on Computer Vision and Pattern Recognition*, pages 1593–1600.

Moreno-Noguer, F. and Fua, P. (2013). Stochastic exploration of ambiguities for non-rigid shape recovery. *IEEE Transactions Pattern Analylis and Machine Intelligence*, 35(2):463–475.

Moreno-Noguer, F., M.Salzmann, Lepetit, V., and Fua, P. (2009). Capturing 3d stretchable surfaces from single images in closed form. In *IEEE Conference on Computer Vision and Pattern Recognition*, pages 1842–1849.

Moreno-Noguer, F. and Porta, J. (2011). Probabilistic simultaneous pose and non-rigid shape recovery. In *IEEE Conference on Computer Vision and Pattern Recognition*, pages 1289 –1296.

Moreno-Noguer, F., Porta, J., and Fua, P. (2010). Exploring ambiguities for monocular non-rigid shape estimation. In *European Conference on Computer Vision*, pages 370–383.

Murillo, A. C., Kwak, I. S., Bourdev, L., Kriegman, D., and Belongie, S. (2012). Urban tribes: Analyzing group photos from a social perspective. In *CVPR Workshops*.

Okada, R. and Soatto, S. (2008). Relevant Feature Selection for Human Pose Estimation and Localization in Cluttered Images. In *European Conference on Computer Vision*, pages 434–445, Berlin, Heidelberg. Springer-Verlag.

Oliva, A. and Torralba, A. (2001). Modeling the shape of the scene: A holistic representation of the spatial envelope. *International Journal of Computer Vision*, 42(3):145–175.

Osendorfer, C., Bayer, J., Urban, S., and van der Smagt, P. (2013). Convolutional neural networks learn compact local image descriptors. In *International Conference on Neural Information Processing*.

PAMI-TC (2013). The July 2013 edition of the PAMI-TC newsletter. http://www.computer.org/web/tcpami/july-2013. [Online; accessed 3-February-2015].

P.Arbelaez, M.Maire, C.Fowlkes, and J.Malik (2011). Contour detection and hierarchical image segmentation. In *IEEE Transactions Pattern Analylis and Machine Intelligence*.

Pelletier, B. (2005). Kernel Density Estimation on Riemannian Manifolds. *Statistics & Probability Letters*, 73(3):297 – 304.

Pennec, X. (2006). Intrinsic Statistics on Riemannian Manifolds: Basic Tools for Geometric Measurements. *Journal of Mathematical Imaging and Vision*, 25(1):127–154.

Pennec, X., Fillard, P., and Ayache, N. (2006). A Riemannian framework for tensor computing. *International Journal of Computer Vision*, 66(1):41–66.

Pepik, B., Stark, M., Gehler, P., and Schiele, B. (2012). Teaching 3D Geometry to Deformable Part Models. In *IEEE Conference on Computer Vision and Pattern Recognition*, Providence, RI, USA. accepted as oral.

Philbin, J., Isard, M., Sivic, J., and Zisserman, A. (2010). Descriptor learning for efficient retrieval. In *European Conference on Computer Vision*, pages 677–691.

Pinkall, U. and Polthier, K. (1993). Computing discrete minimal surfaces and their conjugates. *Experimental Mathematics*, 2(1):15–36.

Pirsiavash, H., Vondrick, C., and Torralba, A. (2014). Assessing the quality of actions. In *European Conference on Computer Vision*.

Price, E. (2014). The NIPS Experiment. http://blog.mrtz.org/2014/12/15/the-nips-experiment.html. [Online; accessed 3-February-2015].

Quiñonero-candela, J., Rasmussen, C. E., and Herbrich, R. (2005). A Unifying View of Sparse Approximate Gaussian Process Regression. *Journal of Machine Learning Research*, 6:1939–1959.

Ramakrishna, V., Kanade, T., and Sheikh, Y. A. (2012). Reconstructing 3D Human Pose from 2D Image Landmarks. In *European Conference on Computer Vision*.

Ramanan, D. (2006). Learning to parse images of articulated bodies. In *NIPS*, volume 19, page 1129.

Raviv, D., Bronstein, M. M., Sochen, N., Bronstein, A. M., and Kimmel, R. (2011). Affine-invariant diffusion geometry for the analysis of deformable 3d shapes. In *IEEE Conference on Computer Vision and Pattern Recognition*.

Reuter, M., Wolter, F., and Peinecke, N. (2006). Laplace-beltrami spectra as 'shape-dna' of surfaces and solids. *Computer Aided Design*, 38(4):342–366.

Rogez, G., Rihan, J., Ramalingam, S., Orrite, C., and Torr, P. (2008). Randomized Trees for Human Pose Detection. In *IEEE Conference on Computer Vision and Pattern Recognition*.

Rubio, A., Villamizar, M., Ferraz, L., nate Sńchez, A. P., Ramisa, A., Simo-Serra, E., Moreno-Noguer, F., and Sanfeliu, A. (2015). Efficient Pose Estimation for Complex 3D Models. In *International Conference on Robotics and Automation*.

Rustamov, R. (2007). Laplace-beltrami eigenfunctions for deformation invariant shape representation. In *Eurographics Symposium on Geometry Processing*, pages 225–233.

Said, S., Courtry, N., Bihan, N. L., and Sangwine, S. (2007). Exact Principal Geodesic Analysis for data on $SO(3)$. In *European Signal Processing Conference*.

Salzmann, M. and R.Urtasun (2010). Combining Discriminative and Generative Methods for 3D Deformable Surface and Articulated Pose Reconstruction. In *IEEE Conference on Computer Vision and Pattern Recognition*, pages 647–654.

Sanchez, J., Ostlund, J., Fua, P., and Moreno-Noguer, F. (2010). Simultaneous pose, correspondence and non-rigid shape. In *IEEE Conference on Computer Vision and Pattern Recognition*, pages 1189–1196.

Schwing, A., Hazan, T., Pollefeys, M., and Urtasun, R. (2011). Distributed message passing for large scale graphical models. In *IEEE Conference on Computer Vision and Pattern Recognition*.

Schwing, A. G., Hazan, T., Pollefeys, M., and Urtasun, R. (2012). Efficient structured prediction with latent variables for general graphical models. In *International Conference in Machine Learning*.

Sermanet, P., Chintala, S., and LeCun, Y. (2012). Convolutional neural networks applied to house numbers digit classification. In *International Conference on Pattern Recognition*.

Serradell, E., Glowacki, P., Kybic, J., Moreno-Noguer, F., and Fua, P. (2012). Robust non-rigid registration of 2d and 3d graphs. In *IEEE Conference on Computer Vision and Pattern Recognition*, pages 996–1003.

Shi, L., Yu, Y., and Feng, N. B. W.-W. (2006). A fast multigrid algorithm for mesh deformation. *ACM SIGGRAPH*, 25(3):1108–1117.

Sigal, L., Balan, A., and Black, M. (2010a). Humaneva: Synchronized video and motion capture dataset and baseline algorithm for evaluation of articulated humanămotion. *International Journal of Computer Vision*, 87(1-2):4–27.

Sigal, L., Balan, A. O., and Black, M. J. (2010b). HumanEva: Synchronized Video and Motion Capture Dataset and Baseline Algorithm for Evaluation of Articulated Human Motion. *International Journal of Computer Vision*, 87(1-2):4–27.

Sigal, L., Bhatia, S., Roth, S., Black, M., and Isard, M. (2004). Tracking loose-limbed people. In *IEEE Conference on Computer Vision and Pattern Recognition*.

Sigal, L. and Black, M. J. (2006). Predicting 3D People from 2D Pictures. In *Conference on Articulated Motion and Deformable Objects*, pages 185–195.

Sigal, L., Isard, M., Haussecker, H. W., and Black, M. J. (2012). Loose-limbed People: Estimating 3D Human Pose and Motion Using Non-parametric Belief Propagation. *International Journal of Computer Vision*, 98(1):15–48.

Sigal, L., Memisevic, R., and Fleet, D. (2009). Shared kernel information embedding for discriminative inference. In *IEEE Conference on Computer Vision and Pattern Recognition*, pages 2852–2859. IEEE.

Simo-Serra, E., Fidler, S., Moreno-Noguer, F., and Urtasun, R. (2014a). A High Performance CRF Model for Clothes Parsing. In *Asian Conference on Computer Vision*.

Simo-Serra, E., Fidler, S., Moreno-Noguer, F., and Urtasun, R. (2015a). Neuroaesthetics in Fashion: Modeling the Perception of Fashionability. In *IEEE Conference on Computer Vision and Pattern Recognition*.

Simo-Serra, E., Quattoni, A., Torras, C., and Moreno-Noguer, F. (2013). A Joint Model for 2D and 3D Pose Estimation from a Single Image. In *IEEE Conference on Computer Vision and Pattern Recognition*.

Simo-Serra, E., Ramisa, A., Alenyà, G., Torras, C., and Moreno-Noguer, F. (2012). Single Image 3D Human Pose Estimation from Noisy Observations. In *IEEE Conference on Computer Vision and Pattern Recognition*.

Simo-Serra, E., Torras, C., and Moreno-Noguer, F. (2014b). Geodesic Finite Mixture Models. In *British Machine Vision Conference*.

Simo-Serra, E., Torras, C., and Moreno-Noguer, F. (2015b). DaLI: Deformation and Light Invariant Descriptor. *International Journal of Computer Vision*, 1:1–1.

Simo-Serra, E., Torras, C., and Moreno-Noguer, F. (2015c). Lie Algebra-Based Kinematic Prior for 3D Human Pose Tracking. In *International Conference on Machine Vision and Applications*.

Simo-Serra, E., Trulls, E., Ferraz, L., Kokkinos, I., and Noguer, F. M. (2014c). Fracking Deep Convolutional Image Descriptors. *CoRR*, abs/1412.6537.

Simonyan, K., Vedaldi, A., and Zisserman, A. (2014). Learning local feature descriptors using convex optimisation. *IEEE Transactions Pattern Analylis and Machine Intelligence*, 36(8):1573–1585.

Simonyan, K. and Zisserman, A. (2014). Very deep convolutional networks for large-scale image recognition. *CoRR*, abs/1409.1556.

Singh, V. K., Nevatia, R., and Huang, C. (2010). Efficient Inference with Multiple Heterogeneous Part Detectors for Human Pose Estimation. In *European Conference on Computer Vision*, pages 314–327.

Sminchisescu, C. and Jepson, A. (2004). Generative Modeling for Continuous Non-Linearly Embedded Visual Inference. In *International Conference in Machine Learning*, pages 96–.

Sochen, N., Kimmel, R., and Malladi, R. (1998). A general framework for low level vision. *IEEE Transactions on Image Processing*, 7(3):310–318.

Socher, R., Perelygin, A., Wu, J., Chuang, J., Manning, C. D., Ng, A. Y., and Potts, C. (2013). Recursive deep models for semantic compositionality over a sentiment treebank. In *Conference on Empirical Methods on Natural Language Processing*.

Sommer, S., Lauze, F., Hauberg, S., and Nielsen, M. (2010). Manifold Valued Statistics, Exact Principal Geodesic Analysis and the Effect of Linear Approximations. In *European Conference on Computer Vision*.

Sommer, S., Tatu, A., Chen, C., Jurgensen, D., De Bruijne, M., Loog, M., Nielsen, M., and Lauze, F. (2009). Bicycle Chain Shape Models. In *IEEE Conference on Computer Vision and Pattern Recognition*.

Song, Z., Wang, M., s. Hua, X., and Yan, S. (2011). Predicting occupation via human clothing and contexts. In *International Conference on Computer Vision*.

Sun, J., Ovsjanikov, M., and Guibas, L. (2009). A concise and provably informative multi-scale signature based on heat diffusion. In *Eurographics Symposium on Geometry Processing*, pages 1383–1392.

Szegedy, C., Toshev, A., and Erhan, D. (2013). Deep neural networks for object detection. In *Neural Information Processing Systems*.

Tang, F., Lim, S. H., Chang, N., and Tao, H. (2009). A novel feature descriptor invariant to complex brightness changes. In *IEEE Conference on Computer Vision and Pattern Recognition*, pages 2631–2638.

Taylor, C. (2000). Reconstruction of articulated objects from point correspondences in a single uncalibrated image. *CVIU*, 80:349–363.

Taylor, G., Sigal, L., Fleet, D., and Hinton, G. (2010). Dynamical binary latent variable models for 3d human pose tracking. In *IEEE Conference on Computer Vision and Pattern Recognition*.

Tenenbaum, J. B., Silva, V., and Langford, J. C. (2000). A Global Geometric Framework for Nonlinear Dimensionality Reduction. *Science*, 290(5500):2319–2323.

Tian, T.-P. and Sclaroff, S. (2010). Fast Globally Optimal 2D Human Detection with Loopy Graph Models. In *IEEE Conference on Computer Vision and Pattern Recognition*.

Todt, E. and Torras, C. (2004). Detecting salient cues through illumination-invariant color ratios. *Robotics and Autonomous Systems*, 48(2):111–130.

Tola, E., Lepetit, V., and Fua, P. (2010). Daisy: An efficient dense descriptor applied to wide-baseline stereo. *IEEE Transactions Pattern Analylis and Machine Intelligence*, 32(5):815–830.

Torresani, L., Kolmogorov, V., and Rother, C. (2008). Feature correspondence via graph matching: Models and global optimization. In *European Conference on Computer Vision*, volume 2, pages 596–609.

Toshev, A. and Szegedy, C. (2014). Deeppose: Human pose estimation via deep neural networks. In *IEEE Conference on Computer Vision and Pattern Recognition*.

Tournier, M., Wu, X., Courty, N., Arnaud, E., and Revéret, L. (2009). Motion compression using principal geodesics analysis. In *Proceedings of Eurographics*.

Trulls, E., Kokkinos, I., Sanfeliu, A., and Moreno-Noguer, F. (2013). Dense segmentation-aware descriptors. In *IEEE Conference on Computer Vision and Pattern Recognition*.

Trulls, E., Tsogkas, S., Kokkinos, I., Sanfeliu, A., and Moreno-Noguer, F. (2014). Segmentation-aware deformable part models. In *IEEE Conference on Computer Vision and Pattern Recognition*.

Trzcinski, T., Christoudias, .and Lepetit, V., and Fua, P. (2012). Learning image descriptors with the boosting-trick. In *Neural Information Processing Systems*.

Uijlings, J. R. R., van de Sande, K. E. A., Gevers, T., and Smeulders, A. W. M. (2013). Selective search for object recognition. *International Journal of Computer Vision*, 104(2):154–171.

Urtasun, R., Fleet, D. J., and Fua, P. (2006). 3D People Tracking with Gaussian Process Dynamical Models. In *IEEE Conference on Computer Vision and Pattern Recognition*.

Urtasun, R., Fleet, D. J., and Lawrence, N. D. (2007). Modeling human locomotion with topologically constrained latent variable models. In *Proceedings of the 2nd Conference on Human Motion: Understanding, Modeling, Capture and Animation*.

Vaxman, A., Ben-Chen, M., and Gotsman, C. (2010). A multi-resolution approach to heat kernels on discrete surfaces. *ACM SIGGRAPH*, 29(4):121.

Vedaldi, A. and Fulkerson, B. (2008). VLFeat: An open and portable library of computer vision algorithms. http://www.vlfeat.org/.

Vedaldi, A. and Soatto, S. (2005). Features for recognition: Viewpoint invariance for non-planar scenes. In *International Conference on Computer Vision*, pages 1474–1481.

Wang, H. and Koller, D. (2011). Multi-level inference by relaxed dual decomposition for human pose segmentation. In *IEEE Conference on Computer Vision and Pattern Recognition*.

Wang, J., Fleet, D., and Hertzmann, A. (2005). Gaussian process dynamical models. In *Neural Information Processing Systems*.

Wang, J., Fleet, D., and Hertzmann, A. (2008). Gaussian process dynamical models for human motion. *IEEE Transactions Pattern Analylis and Machine Intelligence*, 30(2):283–298.

Wang, N. and Ai, H. (2011). Who blocks who: Simultaneous clothing segmentation for grouping images. In *International Conference on Computer Vision*.

Wang, Z., Fan, B., and Wu, F. (2011). Local intensity order pattern for feature description. In *International Conference on Computer Vision*, pages 603 –610.

Wesseling, P. (2004). *An Introduction to Multigrid Methods.* John Wiley & Sons.

Winder, S., Hua, G., and Brown, M. (2009). Picking the best daisy. In *IEEE Conference on Computer Vision and Pattern Recognition.*

Xiao, J., Hays, J., Ehinger, K. A., Oliva, A., and Torralba, A. (2010). Sun database: Large-scale scene recognition from abbey to zoo. In *IEEE Conference on Computer Vision and Pattern Recognition.*

Yamaguchi, K., Kiapour, M. H., and Berg, T. L. (2013). Paper doll parsing: Retrieving similar styles to parse clothing items. In *International Conference on Computer Vision.*

Yamaguchi, K., Kiapour, M. H., Ortiz, L. E., and Berg, T. L. (2012). Parsing clothing in fashion photographs. In *IEEE Conference on Computer Vision and Pattern Recognition.*

Yang, W., Luo, P., and Lin, L. (2014). Clothing co-parsing by joint image segmentation and labeling. In *IEEE Conference on Computer Vision and Pattern Recognition.*

Yang, Y. and Ramanan, D. (2011). Articulated Pose Estimation with Flexible Mixtures-of-Parts. In *IEEE Conference on Computer Vision and Pattern Recognition.*

Yanowitz, S. and Bruckstein., A. (1989). A new method for image segmentation. *Computer Vision, Graphics, and Image Processing*, 46(1):82–95.

Yao, A., Gall, J., and Van Gool, L. (2012a). Coupled action recognition and pose estimation from multiple views. *International Journal of Computer Vision*, 100(1):16–37.

Yao, J., Fidler, S., and Urtasun, R. (2012b). Describing the scene as a whole: Joint object detection, scene classification and semantic segmentation. In *IEEE Conference on Computer Vision and Pattern Recognition.*

Yezzi, A. (1998). Modified curvature motion for image smoothing and enhancement. *IEEE Transactions on Image Processing*, 7(3):345–352.

Zeiler, M. and Fergus, R. (2014). Visualizing and understanding convolutional networks. In *European Conference on Computer Vision.*

Zhao, X., Fu, Y., and Liu, Y. (2011). Human Motion Tracking by Temporal-Spatial Local Gaussian Process Experts. *IEEE Transactions on Image Processing*, 20(4):1141–1151.